\documentclass[phd,tocprelim]{cornell}
\let\ifpdf\relax
\usepackage{amsmath,amsfonts,amssymb,amsthm} 
\usepackage{mathtools} 
\usepackage{graphicx,float} 
\usepackage{algorithm, algpseudocode, algorithmicx} 
\usepackage{bm} 
\usepackage{bbm} 
\usepackage[hidelinks]{hyperref} 
\hypersetup{
colorlinks = true,
urlcolor = blue,
linkcolor = red,
citecolor = green
}
\usepackage{enumitem} 
\usepackage{grffile} 

\DeclareMathOperator{\sech}{sech}
\newcommand{\mathe}{\mathbb{E}}
\newcommand{\norm}[1]{\left\lVert\,#1\,\right\rVert}
\newcommand{\thinnorm}[1]{\left\lVert#1\right\rVert}
\newcommand{\shortnorm}[1]{\big\lVert\,#1\,\big\rVert}

\DeclareMathOperator*{\argmin}{arg\,min}
\newcommand*{\citealt}[1]{\cite{#1}}
\newcommand*{\citep}[1]{(\cite{#1})}

\newtheorem{assumption}{Assumption}
\newtheorem{theorem}{Theorem}
\newtheorem{proposition}{Proposition}
\newtheorem{definition}{Definition}
\newtheorem{corollary}{Corollary}
\newtheorem{lemma}{Lemma}
\newtheorem{condition}{Condition}
\newtheorem{prf}{Proof}

\numberwithin{assumption}{chapter} 
\numberwithin{theorem}{chapter} 
\numberwithin{proposition}{chapter} 
\numberwithin{definition}{chapter} 
\numberwithin{corollary}{chapter} 
\numberwithin{lemma}{chapter} 
\numberwithin{condition}{chapter} 
\numberwithin{prf}{chapter} 
\usepackage{changepage} 

\title {Gradient Estimation and Variance Reduction in Stochastic and Deterministic Models}
\author {Ronan Keane}
\conferraldate {May}{2022}
\degreefield {Systems Science and Engineering}
\copyrightholder{Ronan Keane}
\copyrightyear{2022}

\begin{document}

\maketitle
\makecopyright

\begin{abstract}
\subsubsection{Abstract}
It seems that in the current age, computers, computation, and data have an increasingly important role to play in scientific research and discovery. This is reflected in part by the rise of machine learning and artificial intelligence, which have become great areas of interest not just for computer science but also for many other fields of study. More generally, there have been trends moving towards the use of bigger, more complex and higher capacity models. It also seems that stochastic models, and stochastic variants of existing deterministic models, have become important research directions in various fields. For all of these types of models, gradient-based optimization remains as the dominant paradigm for model fitting, control, and more. This dissertation considers unconstrained, nonlinear optimization problems, with a focus on the \textit{gradient} itself, that key quantity which enables the solution of such problems. 

In chapter 1, we introduce the notion of reverse differentiation, a term which describes the body of techniques which enables the efficient computation of gradients. We cover relevant techniques both in the deterministic and stochastic cases. We present a new framework for calculating the gradient of problems which involve both deterministic and stochastic elements. The resulting gradient estimator can be applied in virtually any situation (including many where automatic differentiation alone fails due to the fact that it does not give gradient terms due to score functions). In chapter 2, we analyze the properties of the gradient estimator, with a focus on those properties which are typically assumed in convergence proofs of optimization algorithms. That chapter attempts to bridge some of the gap between what is assumed in a mathematical optimization proof, and what we need to prove to get a convergence result for a specific model/problem formulation. Chapter 3 gives various examples of applying our new gradient estimator. We further explore the idea of working with piecewise continuous models, that is, models with distinct branches and if statements which define what specific branch to use. We also discuss model elements that cause problems in gradient-based optimization, and how to reformulate a model to avoid such issues. Lastly, chapter 4 presents a new optimal baseline for use in the variance reduction of gradient estimators involving score functions. We forsee that methodology as becoming a key part of gradient estimation, as the presence of score functions is a key feature in our gradient estimator. In somewhat of a departure from the previous chapters, chapters 5 and 6 present two studies in transportation, one of the core emerging application areas which have motivated this dissertation. 
\end{abstract}

\begin{biosketch}
Ronan Keane is a current PhD Candidate in Systems Science and Engineering at Cornell University. Prior to starting his PhD, Ronan received an MS in Applied Math from the University of Washington. His research focuses on traffic flow theory, mathematical modeling, and optimization. Within optimization, his interests lie in nonlinear optimization, stochastic optimization/gradient estimation, and machine learning. Ronan's dissertation research developed new methods for formulating and solving nonlinear optimization problems involving stochastic and/or piecewise continuous models. He is actively developing havsim (Human and Autonomous Vehicle SIMulation, calibration, and optimization), a software package for solving optimization and control problems involving traffic models. In 2020 he was a fellow at the Institute for Pure and Applied Mathematics (IPAM) as part of their program 'Mathematical Challenges and Opportunities for Autonomous Vehicles'. 
\end{biosketch}

\begin{acknowledgements}
I want to acknowledge and thank my advisor, H. Oliver Gao for all his help and support over the last five years. I also want to thank my special committee members, Samitha Samaranayake, Jamol Pender, and Richard Rand.
 
\noindent This dissertation is dedicated to my parents. 
\end{acknowledgements}

\contentspage

\normalspacing \setcounter{page}{1} \pagenumbering{arabic}
\pagestyle{cornell} \addtolength{\parskip}{0.5\baselineskip}

\chapter{Gradient Estimation and Reverse Differentiation} 
\section{Introduction}
The first three chapters of this thesis are mainly concerned with solving unconstrained nonlinear optimization problems using gradient-based optimization. In such a problem, we have some nonlinear function, $F$, which depends on parameters $\theta$ (where $F$ is a scalar and $\theta$ a vector). The goal is to find the $\theta$ which will minimize the value of $F$. There are a multitude of different problems and applications which can be formulated and solved as unconstrained nonlinear optimization problems. Most notably, at least in recent years, is machine learning and specifically deep neural networks (also known as deep learning).

To solve such problems, we will require a way to compute the \textbf{gradient}, which we denote as $\dfrac{\partial F}{\partial \theta}$. The gradient is defined simply as the vector of partial derivatives
\begin{align*} 
\dfrac{\partial F(\theta)}{\partial \theta} = \begin{bmatrix} \dfrac{\partial F(\theta)}{\partial \theta_1} & \dfrac{\partial F(\theta)}{\partial \theta_2} & \ldots & \dfrac{\partial F(\theta)}{\partial \theta_m} \end{bmatrix}
\end{align*}
where $\theta$ is a vector of length $m$ with components $\theta_1, \ldots, \theta_m$. If we can compute the gradient, then there are a huge number of widely available algorithms/software which can be used to minimize $F$. Thus, our concern in this chapter will be how to compute the gradient for a large class of $F$. 

Reverse differentiation (RD) is a family of techniques which efficiently calculate the gradient of a scalar function. By using RD, it is possible to compute $\dfrac{\partial F}{\partial \theta}$ in approximately the same amount of time it takes to calculate $F$, regardless of how complicated $F$ is or how many components are in $\theta$. This makes RD a powerful tool which enables the solution of highly nonlinear problems involving complex models with many parameters. 

Examples of reverse differentiation include backpropagation, reverse-mode automatic differentiation, vector-Jacobian products, and the adjoint method. All of these methods have time complexities which do not depend on the number of parameters $m$. In contrast, a technique such as finite differences scales linearally with $m$. Reverse differentiation can also be applied in cases when calculating the gradient by hand is impossible or simply too complicated to be practical. There are a plethora of tools, known as automatic differentiation software, which can be used to implement RD. The machine learning community in particular has embraced RD as a core principle and been responsible for the creation and maintenance of many of these tools, such as \texttt{tensorflow} \cite{tensorflow}. In section \ref{sec-1-2} we will explain how RD works and explain how to implement it in python.

Gradient estimation refers to the case where the function $F$ is a random variable. In such a situation, we seek to minimize the expected value of $F$, i.e. $\mathe[F]$. We assume that it is impossible (or simply far too computationally expensive) to actually compute $\mathe[F]$ or $\dfrac{\partial \mathe[F]}{\partial \theta}$. Instead we seek to compute a gradient estimator, denoted $\hat g$, such that $\mathe[\hat g] = \dfrac{\partial \mathe[F]}{\partial \theta}$. Note that $\hat g$ is a vector of random variables. Reverse differentiation in this context would mean that $\hat g$ can be sampled in approximately the same amount of time that it takes to sample $F$. We will introduce the tools used in gradient estimation in section \ref{sec-1-3} and unify it with reverse differentiation in section \ref{sec-1-4}.

\section{Gradients of Deterministic Functions} \label{sec-1-2}
\subsection{vector-Jacobian products and automatic differentiation}
Jacobian-vector products and vector-Jacobian products are the key pillars of automatic differentiation software \cite{autodiff-ML-survey, griewank-walther}. For some function $F(x) : \mathbb{R}^m \rightarrow \mathbb{R}^k$, the vector-Jacobian product is defined as 
\begin{align*} 
\text{vjp}(u, F, x) := u^\top\dfrac{\partial F}{\partial x}
\end{align*}
for some constant vector $u \in \mathbb{R}^k$. ($u$ is a column vector, we use the so-called numerator layout where $\partial F/ \partial x$ is $\mathbb{R}^{k \times m}$. Due to \cite{griewank-walther}, we have that the time complexity of calculating a $\text{vjp}$ is
\begin{align*} 
TIME(F, \text{vjp}(u, F, x)) \leq 4 \, TIME(F).
\end{align*}
That is to say, the time complexity of the $\text{vjp}$ is bounded by a constant factor of 4 times the complexity of $F$. 

This constant time complexity (with respect to the underlying function being differentiated) is what makes automatic differentiation powerful. In particular, consider some scalar function $f(\theta): \mathbb{R}^m \rightarrow \mathbb{R}$. The gradient $\partial f / \partial \theta$ can be efficiently calculated by $\text{vjp}(1, f, \theta)$ (where $1$ is a scalar). This corresponds to the most common use case of automatic differentiation. The fact that $TIME(\text{vjp}(1, f, \theta)) \leq 4 TIME(f)$ is the precise justification and meaning of the broad statement that scalar functions can be efficiently differentiated. 

It is beyond the scope of this document to explain the precise meaning/definition of the time complexity measure $TIME$, or to explain how a vector-Jacobian product is implemented at a low level in computer code. For those questions, \cite{griewank-walther} already provides an authoritative source. 

Related to vector-Jacobian products is the Jacobian-vector product (jvp), defined as
\begin{align*} 
\text{jvp}(F, x, v) := \dfrac{\partial F}{\partial x}v
\end{align*}
 for a constant vector $v \in \mathbb{R}^m$. The Jacobian-vector product also has a constant time complexity in the sense that $TIME(F, \text{jvp}(F, x, v)) \leq (5/2) TIME(F)$. Compared to vjp, the jvp does not arise as often, so it is not as commonly used. Note that jvp and vjp are not just trivial transposes of each other; $\text{jvp}(F, x, v)$ is a vector of dimension $k$, whereas $\text{vjp}(u, F, x)$ is a vector of dimension $m$.
 
\subsubsection{How to implement a vector-Jacobian product in tensorflow}
The following code snippet explains how to calculate $\text{vjp}(u, F, x)$ in python using tensorflow. The function $F$ takes as input the tensor $x$. The shape of $x$ is arbitrary, and the output of $F$ can also have any shape. The tensor $u$ must have the same shape as the output of $F$. The resulting vector-Jacobian product will have the same shape as $x$.
\begin{algorithm}[h]
\caption{Reverse Pass} \label{vjp-psuedocode}
\begin{algorithmic}
\State \texttt{import tensorflow as tf}
\State \texttt{with tf.GradientTape() as g:}
\State \quad \texttt{g.watch(x)}
\State \quad \texttt{output = F(x)}
\State $u^\top \dfrac{\partial F}{\partial x}$ = \texttt{g.gradient(output, x, output\_gradients = u)}
\end{algorithmic}
\end{algorithm}


\subsection{The adjoint method}
Consider the optimization problem
\begin{align*}
\underset{\theta}{\min} & \quad  f(x_1, \ldots, x_n)       \stepcounter{equation}\tag{\theequation}\label{eqn1}\\
\text{s.t.} & \quad x_{i} = h_i(x_{i-1}, \theta)  \quad i = 1, \ldots, n 
\end{align*}
We call $h_i$ the model, which has parameters $\theta$ and generates the output $x_i \in \mathbb{R}^{k_i}$. We can interpret the model in several different ways depending on the problem. $x_i$ can be the current state at time $i$, which is iteratively updated by the model (e.g. a time varying system where $x_i$ are generated through some time stepping procedure). We can view $h$ as an algorithm with multiple steps $h_i$, where each $x_i$ represents the algorithm variables during the step $i$. For a neural network, we could view each $h_i$ as the $i$\textsuperscript{th} layer, with the activations $x_i$. In any case, $f$ is the scalar objective function, which depends on at least one of the $x_i$. $x_0$ has no dependence on the parameters and is interpreted either as an input to the model/algorithm or as initial conditions. \\
Because the model is deterministic, one option to calculate the gradient $\partial f/ \partial \theta$ is to use the adjoint method \cite{Pontryagin}. Applying the adjoint method, (see appendix \ref{adjoint-method-tutorial} for more details) we arrive at
\begin{align*} 
\dfrac{\partial f}{\partial \theta} = \sum_{i=1}^{n}\lambda_i^\top \dfrac{\partial h_i}{\partial \theta}  \stepcounter{equation}\tag{\theequation}\label{2.2}
\end{align*}
where $\lambda_i$ are defined by the adjoint equations
\begin{align*} 
& \lambda_{n}^\top  = \dfrac{\partial f}{\partial x_n} \\
& \lambda_{i-1}^\top = \dfrac{\partial f}{\partial x_{i-1}} + \lambda_i^\top \dfrac{\partial h_i}{\partial x_{i-1}} = 0 \quad i = n, n-1, \ldots, 2 . \stepcounter{equation}\tag{\theequation}\label{2.3}
\end{align*}
Then to evaluate the gradient, one first performs the ``forward solve'' by computing $x_1 = h_1(x_0, \theta)$ through $x_n = h_n(x_{n-1}, \theta)$. Then the ``reverse solve'' \eqref{2.3} is computed starting from $\lambda_n$ first, and then $\lambda_1$ last. 

A key feature of the adjoint method is that it produces expressions which are easily evaluated by vector-Jacobian products.
\begin{align*} 
& \dfrac{\partial f}{\partial \theta} = \sum_{i=0}^{n-1}\text{vjp}(\lambda_i, h_i, \theta) \\
& \lambda_{n}^\top = \text{vjp}(1, f, x_n),\qquad \lambda_{i-1}^\top = \text{vjp}(\lambda_i, h_i, x_{i-1}) + \text{vjp}(1, f, x_{i-1}) \stepcounter{equation}\tag{\theequation}\label{adjoint-benefit}
\end{align*}
Assuming that the main computational cost in \eqref{eqn1} is evaluating the model $h$, the cost of evaluating the forward solve is $\sum_{i=1}^{n}TIME(h_i)$ and the cost of the reverse solve is $\leq 4\sum_{i=1}^{n}TIME(h_i)$. Note that to achieve the bound of 4, we simply concatenate $[x_{i-1}, \theta]$ so that only a single $\text{vjp}$ is required for each $h_i$.

Note that the motivation for presenting the adjoint method will be its use in section \ref{sec-1-4} to create a new, generic reverse differentiation algorithm which can be applied to stochastic models or mixed stochastic and deterministic models. If the goal is merely to calculate the gradient of \eqref{eqn1}, the best way to do so is most likely to rely on automatic differentiation software to compute $\text{vjp}(1, f, \theta)$.

\subsection{Reverse Differentiation}
We use the term reverse differentiation to describe methods such as backpropagation, (reverse mode) automatic differentiation, and the adjoint method which compute a forward and reverse pass in order to ``efficiently'' calculate a gradient. By efficient, we mean that the gradient can be evaluated with the same asymptotic time complexity as evaluating the objective. Reverse differentiation can be contrasted with techniques such as finite differences or simultaneous perturbation (SPSA) \cite{SPSA}. Finite differences requires at least $m+1$ forward solves ($\theta \in \mathbb{R}^m$), hence it is not efficient and becomes intractible as the number of parameters increases. SPSA is efficient, but produces gradient estimates with high bias \cite{adjointpaper}. 

It can be noted that various approaches may be used in order to achieve reverse differentiation. For example, to compute the gradient of \eqref{eqn1}, we could simply take $\text{vjp}(1, f, \theta)$. Another approach could be to use \eqref{2.3} to compute the $\lambda_i$ and then calculate the gradient as $\dfrac{\partial f}{\partial \theta} = \sum_{i=0}^{n-1}\text{vjp}(\lambda_i, h_i, \theta)$. A third possibility would be to not rely on vector-Jacobian products at all: we could calculate the relevant partial derivatives of $f$ and $h_i$ by hand and implement those functions directly (although this will perhaps be less efficient than a vector-Jacobian product). Other approaches are possible. There are also choices to be made in terms of what intermediate quantities are stored. Automatic differentiation software uses a ``tape'' which records all computations which take place. Of course, this recording comes with extra memory requirements. In certain problems, it is possible to drastically reduce the memory required by not storing any intermediates at all; rather than keeping $x_1, \ldots, x_n$ in memory during the forward pass, \cite{neural-ode} only keeps the current $x_i$ in memory so that one ends up with only $x_n$ after the forward pass is completed. Then, $x_{n-1}, \ldots, x_1$ are recomputed during the reverse pass. This can drastically reduce the memory usage, but it introduces extra numerical error (because the recomputed $x_1, \ldots, x_{n-1}$ are not exactly the same as their original values) and also adds extra computation time (due to having to evaluate $h_1, \ldots, h_{n-1}$ a second time). 

It's clear that various reverse differentiation approaches can differ in terms of their numerical accuracy, memory usage, and computation time. Regardless of the exact approach/methodology used, all reverse differentiation shares in common a) a time complexity which does not depend on the dimension of $\theta$ and b) a high level of numerical accuracy (ideally at the level of the machine epsilon). 

\section{Gradients of Functions of Random Variables} \label{sec-1-3}
Consider the function $y(x)$ where $x$ is a random variable. Here the goal is to derive an unbiased estimator for the gradient $\frac{\partial }{\partial \theta} \mathe_x[y(x)] $.
There are three possible methods for this: the pathwise derivative, the score function (also known as the likelihood ratio), and the weak derivative \cite{fu-gradient, monte-carlo-gradient}. Of these approaches, the weak derivative requires a seperate computation for each parameter. This makes weak derivatives incompatible with reverse differentiation, so we do not consider them. 

\subsection{Pathwise derivatives}
The pathwise derivative can also be called ``infinitestimal perturbation analysis'' or the ``reparametrization trick''/''reparametrization gradient''\cite{monte-carlo-gradient, fu-gradient}. If $x$ can be expressed as a deterministic function of some other random variable $z$, where $z$ does not depend on the parameters, then we can write 
\begin{align*}
\dfrac{\partial}{\partial \theta} \mathe_x [ y(x) ]  = \dfrac{\partial}{\partial \theta} \mathe_z \big[ \mathe_x[ y(x) | z] \big] =  \dfrac{\partial}{\partial \theta} \mathe_z [ y\big( x(\theta, z)\big)] = \mathe_z\big[\dfrac{\partial y}{\partial x}\dfrac{\partial x}{\partial \theta}\big]
\end{align*}
where the last equality assumes we can interchange the expectation and derivative. 
The key observation here is that $x(\theta, z)$ is viewed as if it was a deterministic function with input $z$, so that the gradient is simply given by the chain rule as normal. For this to work, $z$ must not depend on the parameters, and $x(\theta, z)$ must be differentiable with respect to $\theta$. 

Some common random variables admit pathwise derivatives. For example, if $x \sim \text{exponential}(\lambda)$, by using the inverse CDF of the exponential we obtain $x(\lambda, z) = -\log (z)/\lambda$ where $z \sim \text{Uniform}(0,1)$. If $x \sim \mathcal{N}(\theta_1, \theta_2^2)$, then $x(\theta, z) = \theta_1 + \theta_2z$ where $z \sim N(0,1)$. 

Discrete random variables do not have pathwise derivatives. However, for the sake of completeness, we point out that there have been estimators proposed such as the gumbel softmax and straight-through gumbel softmax estimators which give a pathwise derivative for a ``relaxed'' categorical distribution \cite{gumbel-softmax}. Those estimators are biased because the gumbel softmax distribution is an approximation to the categorical distribution (refer to \cite{rebar, relax} for further discussion on this topic). 

Because deriving a pathwise derivative estimator is the same as differentiating a deterministic function, automatic differentiation software can easily give pathwise derivatives. It should also be noted that pathwise derivatives typically have significantly lower variance than score functions \cite{variance-of-derivative-estimators, variance-reduction-reparameterization-trick}.

\subsection{Score Functions}
Gradients due to score functions can also be referred to as ``likelihood ratio'' estimators/gradients or the ``Reinforce'' gradient. Here we are required to have the probability density of $x$ (or probability mass function if $x$ is discrete), which we denote $p(x, \theta)$. The score function estimator draws samples of the gradient as 
\begin{align*} 
y(x) \dfrac{\partial \log p(x, \theta)}{\partial \theta}  . \stepcounter{equation}\tag{\theequation}\label{likelihood}
\end{align*}
 To derive this \cite{stochastic-computation-graphs}, write 
 \begin{align*} 
 & \dfrac{\partial}{\partial \theta} \mathe_x [ y(x)] = \dfrac{\partial}{\partial \theta} \int y(x) p(x, \theta)dx = \int  y(x) \dfrac{\partial p(x, \theta)}{\partial \theta} dx \\
 & = \int  y(x) \dfrac{\partial \log p(x, \theta)}{\partial \theta} p(x, \theta) dx = \mathe_x \bigg[ y(x) \dfrac{\partial \log p(x, \theta)}{\partial \theta} \bigg] .
 \end{align*}
Just like the pathwise derivative, the expectation and derivative must be able to be interchanged.

Compared to the pathwise derivative, score functions tend to be more general. In particular, neither $y$ or $x$ need to be differentiated: we can produce unbiased gradient estimators for discrete-valued distributions, and also for objective functions which are not continuous. 

\section{A General Recipe for Reverse Differentiation} \label{sec-1-4}
Consider calculating the gradient of the generic optimization problem
\begin{align*} 
\underset{\theta}{\min} & \quad  \mathe_{z_1^n,  y_1^n} [ f(x_1, \ldots, x_n) ] \stepcounter{equation}\tag{\theequation}\label{eqn6}\\ 
\text{s.t.} & \quad x_{i} = h_i(x_{i-1}, \theta, y_{i}, z_{i}) \  \quad i = 1, \ldots, n .
\end{align*}
The notation $\mathe_{z_1^n, y_1^n}[f(\cdot)]$ means the expectation is with respect to the joint distribution over $z_1, \ldots, z_n, y_1, \ldots, y_n$. The $z_1^n := \{z_i : i = 1, \ldots, n\}$ represent any random variables used in pathwise derivatives (so their distribution cannot depend on $\theta$ or any $x_i$). Any input random variables with distributions which do depend on $\theta$ or $x_i$ are denoted as $y_1^n := \{y_i : i = 1, \ldots, n\}$ and will be used with score functions. Each $y_{i}$ has the known conditional probability density $p_i(y_{i}\, | \, x_{i-1}, \theta, z_{i})$. Note that this simply means that $x_{i-1}$ and $z_i$ should be simulated first before it is possible to sample $y_i$. In this case, $p_i(y_{i}\, | \, x_{i-1}, \theta, z_{i})$ is straightforward to define, whereas the distribution $p_i(y_i \, | \, \theta)$ will typically be unknown.

We define $Z$ as the support of $z_1^n$ and $Y$ as the support of $y_1^n$. We will assume that $Y$ does not depend on either $z_1^n$ or $\theta$ (the former assumption is only to simplify notation).

For a given $y_i, z_i$, the output $x_i$ is given deterministically by the model $h_i$. After calculating $x_i$, we can then sample $y_{i+1}$ and the process iterates until all the output $x_1, \ldots, x_n$ has been generated. Note that the initial conditions/input $x_0$ is considered to be fixed. If the input needs to be random or depend on the parameters, then $x_1$ should be considered as the input and $x_0$ can be arbitrary. 

\textit{Note that we cannot simply apply automatic differentiation to calculate the gradient of \eqref{eqn6}}. Meaning, if we were to calculate $\text{vjp}(1, f, \theta)$, we would get an incorrect gradient. Our approach for deriving a correct gradient estimator is based on the use of the adjoint method.

 For the estimator to be unbiased, we need to be able to interchange derivative and expectation, which is an issue discussed in detail in chapter \ref{chapter-2}. For the estimator to merely be defined, we require the following differentiability condition. 

\begin{condition} \label{a1}
For any $z_1^n, y_1^n \in Z \times Y$, the partial derivatives 
\begin{align*}
\dfrac{\partial h_i}{\partial \theta}, \ \ \dfrac{\partial p_i(y_i | x_{i-1}, z_i )}{\partial \theta}, \ \ \dfrac{\partial f}{\partial x_i}
\end{align*}
exist for all $i = 1, \ldots, n$ and the partial derivatives
\begin{align*} 
\dfrac{\partial h_i}{\partial x_{i-1}}, \ \ \  \dfrac{\partial p_i(y_i | x_{i-1}, z_i)}{\partial x_{i-1}}
\end{align*}
exist for $i = 2, \ldots, n$. 
\end{condition}
All of our results still apply for a relaxed version of Condition \ref{a1}. For some given $\theta$, let $\mathcal{ND}(\theta) \subseteq Z\times Y$ be the set of $z_1^n, y_1^n$ such that the relevant partial derivatives exist when $z_1^n, y_1^n$ are restricted to $Z \times Y \setminus \mathcal{ND}(\theta)$. If $\mathcal{ND}(\theta)$ has measure zero, or equivalently if $\mathbb{P}(y_1^n, z_1^n \in \mathcal{ND}(\theta) )= 0$, we can simply exclude such values of $y_1^n, z_1^n$ from the expectation without changing the expected value. 

\begin{theorem} \label{thm1-1}
Let condition \ref{a1} hold. The gradient of \eqref{eqn6} is given by 
\begin{align*} 
\dfrac{\partial}{\partial \theta} \mathe_{z_1^n,  y_1^n} [f( x_1, \ldots, x_n)] = \mathe_{z_1^n,  y_1^n} \left[ \sum_{i=1}^{n} f(x_1, \ldots, x_n) \dfrac{\partial \log p_i(y_i \,|\, x_{i-1}, z_{i})}{\partial \theta} + \lambda_i^\top \dfrac{\partial h_i}{\partial \theta} \right] \stepcounter{equation}\tag{\theequation}\label{eqn7}.
\end{align*}
where it is additionally assumed the derivative and expectation can be interchanged. 
Any $z_1^n, y_1^n \in Z \times Y$ corresponds to the adjoint variables $\{ \lambda_i : i=1, \ldots, n\}$ defined by the deterministic equations
\begin{align*} 
& \lambda_{n}^\top =  \dfrac{\partial f}{\partial x_n}  \stepcounter{equation}\tag{\theequation}\label{8.0}\\ 
& \lambda_{i-1}^\top =  \dfrac{\partial f}{\partial x_{i-1}} + \lambda_i^\top \dfrac{\partial h_i}{\partial x_{i-1}} + f(x_1, \ldots, x_n) \dfrac{\partial \log p_i(y_i \,|\, x_{i-1}, z_{i})}{\partial x_{i-1}} \quad i = n, \ldots, 2 . \stepcounter{equation}\tag{\theequation}\label{8}
\end{align*}
\end{theorem}
Recall that the only scalar quantities are the loss function $f$ and probability densities $p_i(y_i \, | \, \cdot)$. All other variables/functions are vectors. We treat the derivative of a length $j$ vector with respect to a length $k$ vector as a $(j, k)$ matrix. Note as well that in our notation we will often write $f$, $h_i$ etc. ignoring the function arguments. 

\textit{Proof.} We will assume $y_i, z_i$ are continuous for all $i$; alternatively considering the discrete case is trivial. \\
Since $x_{i} - h_i = 0$ by definition, we have
\begin{align*} 
 & \dfrac{\partial}{\partial \theta} \mathe_{z_1^n,  y_1^n} \big [ f(x_1, \ldots,  x_n ) \big ] = \dfrac{\partial}{\partial \theta} \mathe_{z_1^n,  y_1^n} \big[ f(x_1, \ldots, x_n) + \sum_{i=1}^{n}\lambda_i^\top(x_{i} - h_i) \big] \\
 = & \dfrac{\partial}{\partial \theta}\underset{y_1^n, z_1^n \in Y \times Z}{\int} \bigg[  f(x_1, \ldots, x_n) + \sum_{i=1}^{n}\lambda_i^\top(x_{i} - h_i) \bigg] p(y_1\, | \, x_0, z_1)p(y_2\, | \, x_1, z_2)\ldots \\
 & \qquad \qquad \qquad p(y_n\, | \, x_{n-1}, z_n) \mu(z_1^n)dy_1dz_1\ldots dy_n dz_n
\end{align*}
where $\mu(z_1^n)$ is the joint probability density for $z_1, \ldots, z_n$. The adjoint variables $\lambda_i$ are as of yet undetermined; equality holds for any choice of $\lambda_1, \ldots, \lambda_n$. To condense notation define 
\begin{align*} 
& \Pi_1^{n} :=p(y_1\, | \, x_0, z_1)p(y_2\, | \, x_1, z_2)\ldots p(y_n\, | \, x_{n-1}, z_n)  \\
& \Pi_{-k} := \Pi_1^{k-1}\Pi_{k+1}^n .
\end{align*}
Continuing with these definitions and taking the derivative 
\begin{align*} 
= & \dfrac{\partial}{\partial \theta}\underset{y_1^n, z_1^n \in Y \times Z}{\int} \bigg[  f(x_1, \ldots, x_n) + \sum_{i=1}^{n}\lambda_i^\top(x_{i} - h_i) \bigg] \Pi_1^n \mu(z_1^n)dy_1dz_1\ldots dy_n dz_n \\
= &\underset{y_1^n, z_1^n \in Y \times Z}{\int} \bigg[ \bigg( \sum_{i=1}^{n}\dfrac{\partial f}{\partial x_i}\dfrac{\partial x_i}{\partial \theta} + \lambda_i^\top\Big(\dfrac{\partial x_i}{\partial \theta} - \dfrac{\partial h_i}{\partial x_{i-1}}\dfrac{\partial x_{i-1}}{\partial \theta}  - \dfrac{\partial h_i}{\partial \theta}\Big) \bigg) \Pi_1^n  \\
& \hphantom{\underset{y_1^n, z_1^n \in Y \times Z}{\int}} + \Big( f(x_1, \ldots, x_n) + \sum_{i=1}^{n}\lambda_i^\top(x_{i} - h_i) \Big)\sum_{j=1}^n \bigg(\dfrac{\partial p_j(y_j\,|\,x_{j-1}, z_j)}{\partial x_{j-1}}\dfrac{\partial x_{j-1}}{\partial \theta} +  \\
& \hphantom{\underset{y_1^n, z_1^n \in Y \times Z}{\int}} + \dfrac{\partial p_j(y_j\,|\,x_{j-1}, z_j)}{\partial \theta} \bigg)\Pi_{-j}  \bigg]\mu(z_1^n)  dy_1dz_1\ldots dy_n dz_n \\
= & \underset{y_1^n, z_1^n \in Y \times Z}{\int} \bigg[  \bigg( \sum_{i=1}^{n}\dfrac{\partial f}{\partial x_i}\dfrac{\partial x_i}{\partial \theta} + \lambda_i^\top\Big(\dfrac{\partial x_i}{\partial \theta} - \dfrac{\partial h_i}{\partial \theta}\Big) \bigg) + \sum_{i=1}^{n-1}-\lambda_{i+1}^\top\dfrac{\partial h_{i+1}}{\partial x_{i}}\dfrac{\partial x_{i}}{\partial \theta} \\
& \hphantom{\underset{y_1^n, z_1^n \in Y \times Z}{\int}} + f(x_1, \ldots, x_n)\sum_{j=1}^{n-1}\dfrac{\partial  \log p_{j+1}(y_{j+1}\,|\,x_{j}, z_{j+1})}{\partial x_{j}}\dfrac{\partial x_{j}}{\partial \theta}\\
& \hphantom{\underset{y_1^n, z_1^n \in Y \times Z}{\int}} + f(x_1, \ldots, x_n)\sum_{j=1}^n \dfrac{\partial \log p_j(y_j\,|\,x_{j-1}, z_j)}{\partial \theta}  \bigg] \Pi_{1}^n\mu(z_1^n)  dy_1dz_1\ldots dy_n dz_n .
\end{align*}
The last equality uses that $\partial x_0 / \partial \theta$ is zero since the input $x_0$ is assumed to be fixed, and also again uses that $x_i - h_i = 0$. Now collecting the $\partial x_i / \partial \theta $ terms:
\begin{adjustwidth}{-1.5cm}{0cm} \vspace{-1.3cm}
\begin{align*} 
= & \underset{y_1^n, z_1^n \in Y \times Z}{\int} \bigg( \dfrac{\partial f}{\partial x_n} + \lambda_n^T \bigg)\dfrac{\partial x_n}{\partial \theta} \Pi_1^n \mu(z_1^n)dy_1dz_1\ldots dy_ndz_n \\
 + & \underset{y_1^n, z_1^n \in Y \times Z}{\int}\sum_{i=1}^{n-1} \bigg( \dfrac{\partial f}{\partial x_i} + \lambda_{i}^\top - \lambda_{i+1}^\top \dfrac{\partial h_{i+1}}{\partial x_n} + f(\cdot) \dfrac{\partial \log p_{i+1}(y_{i+1}\, | \, \cdot)}{\partial x_i}\bigg)\dfrac{\partial x_i}{\partial \theta} \Pi_1^n \mu(z_1^n)dy_1dz_1\ldots dy_ndz_n \\
 + & \underset{y_1^n, z_1^n \in Y \times Z}{\int}\sum_{i=1}^n \bigg( f(\cdot)\dfrac{\partial \log p_i(y_i \, | \, x_{i-1}, z_i)}{\partial \theta} - \lambda_i^\top\dfrac{\partial h_i}{\partial \theta}\bigg) \Pi_1^n \mu(z_1^n)dy_1dz_1\ldots dy_ndz_n \stepcounter{equation}\tag{\theequation}\label{thm1proof1}.
\end{align*}
\end{adjustwidth}
We want to define $\lambda_1, \ldots, \lambda_n$ in order to satisfy the equations
\begin{adjustwidth}{-1.5cm}{0cm} \vspace{-1.3cm}
\begin{align*} 
& \underset{y_1^n, z_1^n \in Y \times Z}{\int} \bigg( \dfrac{\partial f}{\partial x_n} + \lambda_n^T \bigg)\dfrac{\partial x_n}{\partial \theta} \Pi_1^n \mu(z_1^n)dy_1dz_1\ldots dy_ndz_n  = 0 \\
& \underset{y_1^n, z_1^n \in Y \times Z}{\int}\sum_{i=1}^{n-1} \bigg( \dfrac{\partial f}{\partial x_i} + \lambda_{i}^\top - \lambda_{i+1}^\top \dfrac{\partial h_{i+1}}{\partial x_i} + f(\cdot) \dfrac{\partial \log p_{i+1}(y_{i+1}\, | \, \cdot)}{\partial x_i}\bigg)\dfrac{\partial x_i}{\partial \theta} \Pi_1^n \mu(z_1^n)dy_1dz_1\ldots dy_ndz_n = 0.
\end{align*}
\end{adjustwidth}
This follows immediately if any $z_1^n, y_1^n \in Z \times Y$ corresponds to the $\{\lambda_i : i = 1, \ldots, n\}$ defined by
\begin{align*} 
& \dfrac{\partial f}{\partial x_n} + \lambda_n^T = 0 \\
& \dfrac{\partial f}{\partial x_i} + \lambda_{i}^\top - \lambda_{i+1}^\top \dfrac{\partial h_{i+1}}{\partial x_i} + f(\cdot) \dfrac{\partial \log p_{i+1}(y_{i+1}\, | \, x_i, z_{i+1})}{\partial x_i} = 0 \quad i = 1, \ldots, n-1 \stepcounter{equation}\tag{\theequation}\label{thm1proof2}. 
\end{align*}
Thus from \eqref{thm1proof1}, \eqref{thm1proof2} we have 
\begin{align*} 
& \dfrac{\partial}{\partial \theta} \mathe_{z_1^n,  y_1^n} \big [ f(x_1, \ldots,  x_n ) \big ] \\ 
= & \underset{y_1^n, z_1^n \in Y \times Z}{\int}\sum_{i=1}^n \bigg( f(\cdot)\dfrac{\partial \log p_i(y_i \, | \, x_{i-1} z_i)}{\partial \theta} - \lambda_i^\top\dfrac{\partial h_i}{\partial \theta}\bigg) \Pi_1^n \mu(z_1^n)dy_1dz_1\ldots dy_ndz_n \\
= & \mathe_{z_1^n,  y_1^n}\bigg[ \sum_{i=1}^n f(x_1, \ldots, x_n)\dfrac{\partial \log p_i(y_i \, | \, x_{i-1} z_i)}{\partial \theta} - \lambda_i^\top\dfrac{\partial h_i}{\partial \theta}\bigg] .
\end{align*}
The statement of the theorem follows after using the substitution $\lambda_i^* = -\lambda_i$ to simplify the minus signs. $\qed$ \\

Theorem \ref{thm1-1} assumes a combination of a deterministic/pathwise derivatives and score functions. If none of the $h_i$ depend on $y_i$, then $\frac{\partial \log p(y_i)}{\partial \theta}, \frac{\partial \log p(y_{i})}{\partial x_{i-1}}$ can be taken as zero and \eqref{eqn7}\,-\,\eqref{8} reduces to be the same as the deterministic case \eqref{2.2}\,-\,\eqref{2.3} (in this case, automatic differentiation of $f$ with respect to $\theta$ gives the correct gradient). The opposite case is where none of the $h_i$ depend on $\theta$ so that $\frac{\partial h_i}{\partial \theta} = 0$. The estimator then reduces to a sum of score functions, $\sum_{i=1}^{n}f \frac{\partial \log p(y_{i})}{\partial \theta}$, and there is no need to calculate the adjoint variables.

In the remainder of the paper we use the notation $\hat g(\theta)$ to refer to the gradient estimator, which we define as a single sample of \eqref{eqn7}. For a given $y_1^n$, $z_1^n$, the corresponding (deterministic) gradient estimate is denoted as $\hat g(\theta, y_1^n, z_1^n)$.

Algorithm \ref{adjoint-alg} gives a method for efficiently sampling $\hat g(\theta)$ using only vector-Jacobian products. Algorithm \ref{tensorflow-psuedocode} also gives psuedocode for an efficient implementation of the reverse pass using tensorflow. As discussed previously, the computational cost of sampling $\hat g(\theta)$ is proportional to the cost of calculating $\{h_i(\cdot), p_i(y_i | \cdot)\}$. The cost does not scale with the dimension of $\theta$ or the dimension of the $x_i$'s.

\begin{algorithm}[h]
\caption{Gradient Estimation} \label{adjoint-alg}
\begin{algorithmic}
\State \textbf{Inputs:} \ $\theta$, $x_0$, model $\{h_i\}$ and conditional probability densities $\{p_i\}$
\For{$i$ in $1, \ldots, n$ }  \Comment{forward pass}
\State sample $z_{i}$ and sample $y_i$ with probability $p(y_i \,|\, x_{i-1}, \theta, z_{i})$
\State $x_{i} = h_i(x_{i-1}, \theta, y_i, z_{i})$
\State store $x_{i}, y_i, z_{i}$ in memory
\EndFor
\State $F = f(x_1, \ldots, x_n)$
\State $\lambda = 0$, $\hat g = 0$
\For{$i$ in $n, \ldots, 1$} \Comment{reverse pass}
\State $\lambda = \lambda + \text{vjp}(1, f, x_i)$
\State $\hat g =  \hat g + \text{vjp}(F, \log p_i(y_i\,|\, x_{i-1}, z_{i}), \theta) + \text{vjp}(\lambda, h_i, \theta)$
\State $\lambda = \text{vjp}(\lambda, h_i, x_{i-1}) + \text{vjp}(F, \log p_i(y_i\,|\, x_{i-1}, z_{i}), x_{i-1})$ 
\EndFor
\State \textbf{Return:} \ $F$, $\hat g$
\end{algorithmic}
\end{algorithm}
\subsubsection{Tensorflow psuedocode}
To reflect what we believe are the most common cases encountered in practice, we assume that the loss function is of the form $f(x_1, \ldots, x_n) = \sum_{i=1}^nf_i(x_i)$, i.e. the loss is \textit{summable}.
It is also assumed that there is a single function $h_i^*(x_{i-1}, y_i, z_i, \theta)$ which returns both $x_i$ as well as $\log p_i(y_i \, | \, x_{i-1}, \theta, z_i)$.
\begin{algorithm}[h]
\caption{Reverse Pass} \label{tensorflow-psuedocode}
\begin{algorithmic}
\State \textbf{Inputs:} objective value $F$, $\theta$, $x_0$, $\{x_i, y_i, z_i\}$ \Comment{Forward pass already completed}
\State $\lambda = 0$, $\hat g = 0$
\For{ $i$ in $n, n-1, \ldots, 1$}
\State  with \texttt{tf.GradientTape()} as \texttt{dh}:
\State \quad \texttt{dh.watch}($x_{i-1}$)
\State \quad $[x_i,  \, \log p_i(y_i \, | \, x_{i-1}, \theta, z_i)] = h_i^*(x_{i-1}, y_i, z_i, \theta)$
\State with \texttt{tf.GradientTape()} as \texttt{df}:
\State \quad \texttt{df.watch}($x_i$)
\State \quad f = $f_i(x_i)$
\State $\frac{\partial f}{\partial x_i}$ = \texttt{df.gradient}(f, $x_i$)
\State $\lambda = \lambda + \frac{\partial f}{\partial x_i}$
\State $\big[\lambda^T\frac{\partial h_i}{\partial x_{i-1}} + F \frac{\partial \log p_i(y_i \, | \, \cdot)}{\partial x_{i-1}}, \  \lambda^T\frac{\partial h_i}{\partial \theta} + F \frac{\partial \log p_i(y_i \, | \, \cdot)}{\partial \theta}\big]$ = \ldots
\State \qquad \texttt{dh.gradient}($[x_i,  \, \log p_i(y_i \, | \, x_{i-1}, \theta, z_i)]$, $[x_{i-1}, \theta]$, \ldots 
\State \qquad \hphantom{\texttt{dh.gradient}(}\texttt{output\_gradients} =[$\lambda$, $F$]) 
\State $\hat g$ = $\hat g + \lambda^T\frac{\partial h_i}{\partial \theta} + F \frac{\partial \log p_i(y_i \, | \, \cdot)}{\partial \theta}$
\State $\lambda = \lambda^T\frac{\partial h_i}{\partial x_{i-1}} + F\frac{\partial \log p_i(y_i \, | \, \cdot)}{\partial x_{i-1}}$
\EndFor 
\State \textbf{Return:} $\hat g$
\end{algorithmic}
\end{algorithm}
This psuedocode also works to calculate the gradient of a batch of forward solves. In that case, $x_i$, $F$, $\log p_i(y_i \, | \, \cdot)$ should all have an extra dimension in their first axis corresponding to the batch. Note that gradient tape is only used in the reverse pass, which makes this is a low memory implementation; if $h_i$, $p_i$ are neural networks, the activations of any hidden layers are not stored in memory. The tradeoff is the computation time equivalent of an extra forward pass.

\subsection{Extensions of Theorem 1.1}
\subsubsection{More complicated $h_i$}
One common situation is where $h_i$ may depend on past inputs beyond just $x_{i-1}, y_i, z_i$. Specifically, consider $x_i = h_i(x_{i-\tau}, x_{i-\tau+1}, \ldots, x_{i-1}, z_1, \ldots, z_i, y_1, \ldots, y_i)$ where $\tau>1$. Assume similarly that $y_i$ has the conditional probability distribution \\
$p_i(y_i \, | \, x_{i-\tau}, \ldots, x_{i-1}, z_1, \ldots, z_i, y_1, \ldots, y_{i-1})$. The gradient and $\lambda_n$ are still given by \eqref{eqn7}, \eqref{8.0}, with the other adjoint variables given by 
\begin{align*} 
& \lambda_{i-1}^\top =  \dfrac{\partial f}{\partial x_{i-1}} + \sum_{j=i}^{\max\{n, i+\tau-1\}}\left( \lambda_j^T \dfrac{\partial h_j}{\partial x_{i-1}} + f(x_1, \ldots, x_n) \dfrac{\partial \log p_j(y_j \,|\, \cdot)}{\partial x_{i-1}} \right) \quad i = n, \ldots, 2 . \stepcounter{equation}\tag{\theequation}\label{adjointv2}
\end{align*}
The proof follows precisely along the same lines of theorem \ref{thm1-1}. We see that if $h_j$ depends on $x_{i-1}$, it will contribute a $\text{vjp}(\lambda_j, h_j, x_{i-1})$ to $\lambda_{i-1}$. Likewise if $p_j$ depends on $x_{i-1}$, it contributes a $\text{vjp}(f, \log p_j(y_j\, | \, \cdot), x_{i-1})$ to $\lambda_{i-1}$. Note as well that $h_i$ or $p_i(y_i \, | \, \cdot)$ may depend on $z_1, \ldots, z_{i-1}$ and/or $y_1, \ldots, y_{i-1}$ with no change to the estimator.

In order to use algorithm \ref{tensorflow-psuedocode} in this case, one needs to maintain memory of the $\tau$ adjoint variables $\lambda_{i-\tau}, \ldots, \lambda_{i-1}$ at any given index $i$ during the reverse pass. One still does a single vector-Jacobian product with respect to $h_i^*$, i.e. 
\begin{align*} 
\text{vjp}([\lambda_i, \, F], h_i^*, [x_{i-\tau}, x_{i-\tau+1}, \ldots, x_{i-1}, \theta])
\end{align*}
and does the update $\lambda_j = \lambda_j +\lambda_i^\top\frac{\partial h_i}{\partial x_j} + F\frac{\partial \log p_i(y_i \, | \, \cdot)}{\partial x_{j}}$ for each $j = i-\tau, \ldots, i-1$. 

\subsubsection{Summable loss}
Another common situation is when the objective is of the form 
\begin{align*} 
f(x_1, \ldots, x_n) = \sum_{i=1}^n f_i(x_i). \stepcounter{equation}\tag{\theequation}\label{summable-loss}
\end{align*}
It can be shown (in chapter 4, section \ref{properties}) that in this case, we can weight the score functions as
\begin{align*} 
& f(x_1, \ldots, x_n)\dfrac{\partial \log p_i(y_i \, | \, \cdot)}{\partial \theta} \rightarrow \sum_{j=i}^nf_j(x_j)\dfrac{\partial \log p_i(y_i \, | \, \cdot)}{\partial \theta} \\
&f(x_1, \ldots, x_n)\dfrac{\partial \log p_i(y_i \, | \, \cdot)}{\partial x_{i-1}} \rightarrow \sum_{j=i}^nf_j(x_j)\dfrac{\partial \log p_i(y_i \, | \, \cdot)}{\partial x_{i-1}} .
\end{align*}
This modification will reduce the variance of the score function terms without adding any bias, provided that \eqref{summable-loss} holds.

\section{On the Use of Score Functions and Mixed Score Function / Pathwise Derivative Estimators} \label{sec-1-5}
Score functions can be used to construct unbiased gradient estimators in situations where issues with differentiability arise. We consider cases where a) $f$ is not continuous, b) $x_i$ is a mixture of continuous/discrete variables, and/or c) $h_i$ is a piecewise function where the switching condition may have explicit dependence on parameters. 
\subsection{When the objective is not continuous}
If $f$ is not continuous then $\partial f / \partial x_i$ does not exist and condition \ref{a1} is therefore violated. But suppose our model is such that given $x_{i-1}$, the distribution of $x_i$ is known with density $p_i(x_i \, | \, x_{i-1}, \theta, z_i)$. Then the model should be formulated as
\begin{align*} 
x_i = y_i \qquad y_i \text{ has density } p_i(y_i \, | \, x_{i-1}, \theta, z_i) \stepcounter{equation}\tag{\theequation}\label{eg-1}
\end{align*}
so that $h_i( x_{i-1}, \theta, y_i, z_i)$ is simply $y_i$. It follows that $f(x_1, \ldots, x_n) = f(y_1, \ldots, y_n)$, so the objective value is completely determined by $y_1^n$ and is constant with respect to $\theta$. The corresponding gradient estimator is just the sum of score functions
\begin{align*} 
\sum_{i=1}^n f(x_1, \ldots, x_n) \dfrac{\partial \log p_i(y_i \, | \, x_{i-1}, \theta)}{\partial \theta}
\end{align*}
which exists provided that both $\partial \log p_i(y_i \, | \, \cdot) / \partial \theta$ and $f$ exist: there are no continuity or differentiability requirements on $f$. 
\subsection{Combinations of continuous and discrete variables}
In various practical problems, one may deal with a simulation which contains both discrete and continuous variables. Our estimator gives a convenient way to deal with this situation, by using score functions to deal with the discrete variables. Let $x_i$ be the concatenation of $[x_i^{\rm cont},  \ x_i^{\rm disc}]$, where the superscript indicates the two parts are continuous and discrete respectively. Assume $x_i^{\rm cont}$ is given by a deterministic part or pathwise derivatives, and $x_i^{\rm disc}$ has a known density conditional on $x_{i-1}$. Then the model is formulated as 
\begin{align*} 
x_i = \begin{bmatrix} h_i^{\rm cont}(x_{i-1}, z_i, \theta) & y_i \end{bmatrix} \qquad y_i \text{ has density } p_i(y_i \, | \, x_{i-1}, \theta, z_i).  \stepcounter{equation}\tag{\theequation}\label{eg-2}
\end{align*} 
Again, the key observation here is that for a given $y_i$, $x_i^{\rm disc}$ is constant with respect to $\theta$. The resulting gradient estimator is 
\begin{align*} 
&\hat g(\theta) = \sum_{i=1}^{n} f(x_1, \ldots, x_n) \dfrac{\partial \log p_i(y_i \,|\, x_{i-1}, z_{i})}{\partial \theta} + \lambda_i^\top \dfrac{\partial h_i^{\rm cont}}{\partial \theta}, \qquad \lambda_{n}^\top =  \dfrac{\partial f}{\partial x_n^{\rm cont}} \\ 
& \lambda_{i-1}^\top =  \dfrac{\partial f}{\partial x_{i-1}^{\rm cont}} + \lambda_i^\top \dfrac{\partial h_i^{\rm cont}}{\partial x_{i-1}^{\rm cont}} + f(x_1, \ldots, x_n) \dfrac{\partial \log p_i(y_i \,|\, x_{i-1}, z_{i})}{\partial x_{i-1}^{\rm cont}} \quad i = n, \ldots, 2
\end{align*}
so one only needs to worry about differentiability with respect to $x_i^{\rm cont}$ and not $x_i^{\rm disc}$.
\subsection{Piecewise functions with parameter dependent switching conditions}
Consider a model of the form
\begin{align*} 
h_i(x_{i-1}, \theta, z_i) = \begin{cases} 
 h_i^1(x_{i-1}, \theta, z_i) & \text{ if }\gamma_i^1(x_{i-1}, \theta) > 0  \\ 
  h_i^2(x_{i-1}, \theta, z_i) & \text{otherwise}
  \end{cases}. \stepcounter{equation}\tag{\theequation}\label{eg-3-motivation}
\end{align*}
We say the model is piecewise because it has two seperate ``branches''  denoted as $h_i^1, \, h_i^2$, and the ``switching condition'' $\gamma_i^1$ controls which branch to use for any given input. The extension to having an arbitrary number of branches is clear. Such a form may arise, for example, in modeling human behavior or decision making (having multiple distinct behaviors/decisions), modeling a physical system with several distinct phases/modes, or in certain hierarchical models (e.g. \cite{hierarchical-networks}). One might worry about the possible discontinuities (in both $h_i$ and its Jacobians) which can occur when $\gamma_i^1 = 0$. But actually there is a much bigger and more fundamental issue: the switching condition $\gamma_i^1$ will never contribute to the gradient. Thus while it may be possible learn both $h_i^1$ and $h_i^2$ using gradient based optimization, the same is not true for $\gamma_i^1$. To address this type of model, we propose the canonical form
\begin{align*} 
h_i(x_{i-1}, \theta, y_i, z_i) = \begin{cases} 
 h_i^1(x_{i-1}, \theta, z_i) & \text{ if }y_i = 1 \\ 
  h_i^2(x_{i-1}, \theta, z_i) & \text{ if }y_i = 2
  \end{cases} \quad y_i \in \{1, 2\} \text{ with pmf } p_i(y_i \, | \, x_{i-1}, \theta, z_i). \stepcounter{equation}\tag{\theequation}\label{eg-3}
\end{align*}
Now which of the two branches the model is in is encoded by $y_i$, which takes on either a value of 1 or 2. This formulation actually fixes both of the problems in \eqref{eg-3-motivation}. The switching condition now is controlled by the pmf of $y_i$, denoted as $p_i(y_i \, | \, \cdot)$, and can be learned using gradient based optimization by applying the theorem \ref{thm1-1} estimator. Also, for a given $y_i$ we will be in a single branch irrespective of how $\theta$ is changed. Therefore we will only need $h_i^1$ and $h_i^2$ to both be differentiable with respect to $x_{i-1}$ and $\theta$, without needing to consider the previous issue of what happens when $\gamma_i^1 = 0$. Of course the caveat is that \eqref{eg-3} is not exactly the same as \eqref{eg-3-motivation}. In particular since the present framework assumes that the support of $y_i$ does not depend on $\theta$, there must always be some positive probability of choosing the other branch. But we can still define the pmf of $y_i$ such its distribution approaches \eqref{eg-3-motivation}, for example if $\mathbb{P}(y_i = 1 \, | \, x_{i-1}, \theta, z_i) := \mathbb{P}(\epsilon z_i < \gamma_i^1(x_{i-1}, \theta))$ where $z_i \sim \text{Normal}(0, 1)$. Then \eqref{eg-3} becomes \eqref{eg-3-motivation} in the limit $\epsilon \rightarrow 0$.

We find this aspect of our gradient estimator especially exciting as it gives a new way to deal with the rich space of piecewise functions. Moreover, it is remarkable that this approach not only fixes the issue with discontinuities in the objective/gradient, but also gives a rigorous way to learn the switching condition. If we imagine our model as a computer code, a continuous function is like a single code block, that executes from start to finish. A piecewise continuous function has if-else statements, corresponding to different choices/branches in the program. The canonical form \eqref{eg-3} gives a way to learn such piecewise continuous functions. This idea is explored further in chapter 3. We also show an example of learning a switching condition and recovering the deterministic behavior in section \ref{sec-3-3}.

\chapter{Convergence of Stochastic Gradient Descent} \label{chapter-2}
\section{Introduction}
In the previous chapter, we presented the idea of being able to efficiently calculate the gradient for an arbitrary unconstrained nonlinear optimization problem. In particular, the main contribution was being able to do this in the \textit{stochastic setting}, when the objective function depends on random variables, and those random variables may have distributions which depend on $\theta$. In this setting, the optimization problem will be solved using a stochastic gradient method, such as the well-known stochastic gradient descent (SGD). Our main goal in this chapter is to understand when we can expect a method such as SGD to actually converge to a stationary point. We first introduce some relevant theory of stochastic gradient methods. This will motivate the study of two main properties: unbiasedness of the gradient estimator, and Lipschitz continuity of $\mathe[f]$. We will apply our analyses to deep neural networks in section \ref{sec-2-4}. We also briefly mention the purely deterministic case in \ref{sec-2-5}. 
\section{Stochastic Gradient Methods} \label{SG methods} 
Stochastic gradient methods \cite{bottou-nocedal} represent a broad class of iterative nonlinear optimization algorithms which solve the problem
\begin{align*} 
\underset{\theta}{\min} \quad \mathe_{\omega}[ F(\theta, \omega) ] \stepcounter{equation}\tag{\theequation}\label{SG0}
\end{align*}
where $\omega$ is some random variable, and the expectation is assumed to be well defined. This problem is also studied under the keyword of stochastic approximation. A stochastic gradient method takes the form
\begin{align*} 
\theta_{k+1} = \theta_k - \alpha_k \xi(\theta_k) \stepcounter{equation}\tag{\theequation}\label{SG}
\end{align*}
where $\theta_k$ are the parameter values in iteration $k$, $\alpha_k$ is the stepsize (also known as learning rate) at iteration $k$, and $\xi(\theta_k)$ is the search direction. The search direction is computed based on sample(s) of the gradient, where we denote a single gradient estimate as $\hat g(\theta)$. In the simplest case, known as stochastic gradient descent (SGD), $\xi(\theta_k) = \hat g(\theta_k)$. 

An important question is establishing sufficient conditions for when we can expect SGD to give a satisfactory solution of \eqref{SG0}. We will consider two different sets of conditions, due to \cite{nemirovski} and \cite{bottou-nocedal} respectively. In both cases, we will require the following. Let $\theta \in \Theta$ for a compact set $\Theta$, and let $\mathe[F(\theta)]$ be differentiable for $\theta \in \Theta$. The second moment of the gradient estimator must be bounded, i.e. 
\begin{align*} 
\mathe[ \norm{\hat g(\theta)}^2 ] \leq M \quad \forall \theta \in \Theta \stepcounter{equation}\tag{\theequation}\label{nemirovski bound}
\end{align*}
for some constant $M$. We denote $\norm{\cdot}$ as the L2 (Euclidean) norm. Additionally, $\hat g$ must be unbiased so that $\mathe[\hat g] = \frac{\partial}{\partial \theta}\mathe[F]$. Assume $K$ iterations of SGD, with some constant stepsize $\alpha$. 

The guarentee of \cite{nemirovski} further requires that $\Theta$ is convex, and that $\mathe[F(\theta)]$ is convex on $\Theta$. Then,
\begin{align*} 
& \mathe [ F( \bar \theta) - F_{\rm inf} ]  \leq \dfrac{D_{\theta} + M K \alpha^2 }{2K \alpha}. \stepcounter{equation}\tag{\theequation}\label{nemirovski} \\
& \bar \theta := \frac{1}{K}\sum_{k=1}^K\theta_k \qquad D(\theta) := \underset{\theta \in \Theta}{\max} \norm{\theta - \theta_1}^2 .
\end{align*}
where $F_{\rm inf}$ is the optimal value of $\mathe[F(\theta)]$ on $\Theta$. In this result, note that the optimal stepsize $\alpha = \sqrt{D_{\theta}/(M K)}$ depends on $K$, so the number of iterations should ideally be fixed or known in advance. 

\cite{bottou-nocedal} don't require convexity. Instead, suppose that the \textit{expected} gradient is Lipschitz continuous:
\begin{align*} 
\norm{\dfrac{\partial}{\partial \theta} \mathe [ F(\theta^*) ] -  \dfrac{\partial}{\partial \theta} \mathe [ F(\theta^{**}) ] } \leq L \norm{\theta^* - \theta^{**}} \ \  \forall \  \theta^*, \theta^{**} \in \mathbb{R}^m .
\end{align*}
The resulting convergence guarentee is that 
\begin{align*} 
\mathe \left[\dfrac{1}{K} \sum_{k=1}^K \norm{\dfrac{\partial}{\partial \theta} \mathe[F(\theta_k)]}^2 \right] \leq \alpha L M + \dfrac{2(F(\theta_1) - F_{\rm inf})}{K \alpha} \stepcounter{equation}\tag{\theequation}\label{bottou}
\end{align*}
where $\alpha \leq 1/(L)$.

These results hold for SGD or mini-batch SGD. In practice, methods such as ADAM \cite{ADAM}, coordinate descent \cite{coordinate-descent} or SGD with momentum \cite{nesterov-momentum} are typically preferred over SGD; the discussion of these methods are outside the scope of this paper. However, we note that the conditions outlined here are quite standard, and at the bare minimum, the gradient estimator $\hat g$ should be unbiased and its second moment should be bounded. 

\section{Unbiasedness and Convergence of SGD} \label{analysis}
For some given problem, can we expect SGD to converge when using theorem \ref{thm1-1} as the gradient estimator? In section \ref{SG methods}, we saw three important properties which are commonly used in establishing convergence: unbiasedness of the gradient estimator, the second moment of the gradient estimator being bounded, and Lipschitz continuity of the expected gradient. In this section we will examine the connections between these three properties, and also establish procedures for formally showing that those properties hold. In particular, we will see that if it is possible to to show the unbiasedness of $\hat g(\theta)$, we will probably also be able to conclude that the second moment, $\mathe[\norm{\hat g(\theta)}^2]$, is bounded, and that the expected objective is at least locally Lipschitz continuous. In the next section we will apply our analysis to a deep feedforward neural network with $\tanh$ or ReLU activations, where the input to the network follows a normal distribution. 

First we will consider the question of unbiasedness. As mentioned earlier, unbiasedness of $\hat g$ is achieved when the derivative/expectation interchange in Theorem \ref{thm1-1} is justified by the dominated convergence theorem. Oftentimes, especially in machine learning literature, the interchange is simply assumed to be permissable. Here we are concerned with formally justifying this. This question has been studied previously in \cite{interchange-ecuyer}, \cite{glasserman} (ch. 7.2.2), and \cite{monte-carlo-optimization-glasserman}. \cite{glasserman} (ch. 7), \cite{unified-ecuyer}, and \cite{fu-gradient} also provide several examples. 

Our first approach for establishing unbiasedness relies on the mean value theorem due to \cite{dieudonne-mvt-thm} (Theorem 8.5.2). 
We consider a point $\theta_0$ in the parameter space, and want to give sufficient conditions for $\hat g(\theta_0)$ to be an unbiased estimator.
\begin{condition}\label{a2}
Let $\hat g(\theta_0, y_1^n, z_1^n)$ exist except for $z_1^n, y_1^n \in \mathcal{ND}(\theta_0)$ where $\mathcal{ND}(\theta_0)$ is a subset of $Z \times Y$ with measure 0. 
\end{condition}
\noindent Denote $e_k$ as the $k$\textsuperscript{th} standard basis vector.
\begin{condition}\label{a4}
For each $k = 1, 2, \ldots, \text{dim}(\theta)$ there exists a constant $\alpha_k$ such that for any $z_1^n, y_1^n \in Y \times Z  \setminus \mathcal{ND}(\theta_0)$, $\hat g(\bar \theta, y_1^n, z_1^n)$ exists for $\bar \theta \in [\theta_0, \theta_0 + \alpha_ke_k] \setminus \beta_k(z_1^n, y_1^n)$ where $\beta_k(z_1^n, y_1^n)$ is a countable set. 
\end{condition}
\noindent Define $x_{i}(\theta, z_1^i, y_1^i)$ as the corresponding deterministic value of $x_i$ given $\theta, y_1^i, z_1^i$, i.e.
\begin{align*} 
x_i(\theta, z_1^i, y_1^i) := h_{i}(x_{i-1}(\theta, z_1^{i-1},y_1^{i-1}), \theta, y_i, z_i).
\end{align*}
For $i=0$ we simply have $x_0(\theta, z_1^0, y_1^0) := x_0$. 
\begin{condition}\label{a3}
For any $k=1, \ldots, \text{dim}(\theta)$, $i = 1, \ldots, n$, $z_1^n, y_1^n \in Z \times Y$, let $h_i(\bar x_{i-1}, \bar \theta, z_i, y_i)$ and $p_i(y_{i} \, | \, \bar x_{i-1}, \bar \theta, z_i)$ be continuous for $\bar x_{i-1} \in \{x_{i-1}(\theta, z_1^{i-1}, y_1^{i-1}) : \theta \in [\theta_0, \theta_0+\alpha_ke_k]\}$, $\bar \theta \in [\theta_0, \theta_0 + \alpha_ke_k]$. For any $k=1, \ldots, \text{dim}(\theta)$, $z_1^n, y_1^n \in Z \times Y$, let $f(\bar x_1, \ldots, \bar x_n)$ be continuous for $\{\bar x_i \in \{x_{i}(\theta, z_1^{i}, y_1^{i}): \theta \in [\theta_0, \theta_0+\alpha_ke_k]\} : i = 1, \ldots, n\}$.
\end{condition}
Condition \ref{a2} ensures that $\mathe_{z_1^n, y_1^n}[\hat g(\theta_0)]$ exists. Condition \ref{a4} is a relatively unrestrictive differentiability condition. We can consider each component of $\theta$ seperately, one at a time. For the $k$\textsuperscript{th} component of $\theta$, we can allow a countable number of non-differentiable points. The set of non-differentiable points $\beta_k(z_1^n, y_1^n)$ can also depend on the random variables, so we just need to ensure that any particular realization of $z_1^n, y_1^n \in Z \times Y \setminus \mathcal{ND}(\theta_0)$ can only result in at most a countable number of non-differentiable points. Condition \ref{a3} is a continuity condition and can be considered to be the most restrictive condition in practice. The model, probability densities, and objective function must all be continuous with respect to both $\theta$ as well as the $x_i$'s. 

Denote $L(\theta, y_1^n, z_1^n)$ as the likelihood of sampling $y_1^n, z_1^n$ with parameters $\theta$, so 
\begin{align*} 
L(\theta, y_1^n, z_1^n) :=  \prod_{i=1}^n p_i(y_i | x_{i-1}(\theta, z_1^{i-1}, y_1^{i-1}), \theta, z_i) \mu(z_1^n)
\end{align*}
where $\mu(z_1^n)$ is the joint density for $z_1, \ldots, z_n$. To show unbiasedness we lastly need to verify the integrability condition in the following theorem.
\begin{theorem}\label{unbiased1}
Let conditions \ref{a2}, \ref{a4} and \ref{a3} hold. If for each $k = 1, \ldots, \text{dim}(\theta)$,
\begin{align*} 
 \underset{z_1^n, y_1^n \in (Z \times Y) \setminus \mathcal{ND}(\theta_0)}{\int} \underset{\bar \theta \in [\theta_0, \theta_0+\alpha_ke_k] \setminus \beta_k  }{\sup} \norm{\hat g(\bar \theta, y_1^n, z_1^n)} L(\bar \theta, y_1^n, z_1^n)dz_1^ndy_1^n < \infty \stepcounter{equation}\tag{\theequation}\label{equnbiased1}
\end{align*}
then $\hat g(\theta_0)$ is unbiased. 
\end{theorem}

\textit{Proof.} 
Consider the function
\begin{align*} 
F(\theta, y_1^n, z_1^n) := f(x_1(\theta, z_1^1, y_1^1), \ldots, x_n(\theta, z_1^n, y_1^n)) L(\theta, y_1^n, z_1^n)
\end{align*}
whose gradient is
\begin{adjustwidth}{-2cm}{0cm} \vspace{-1.3cm}
\begin{align*} 
\dfrac{\partial}{\partial \theta} F(\theta, y_1^n, z_1^n) = \dfrac{\partial}{\partial \theta} \Big(f(x_1, \ldots, x_n) + \sum_{i=1}^n\big(\lambda_i - h_i(x_{i-1}, \theta, y_i, z_i)\big) \Big)L(\theta, y_1^n, z_1^n) = \hat g(\theta, y_1^n, z_1^n)L(\theta, y_1^n, z_1^n).
\end{align*}
\end{adjustwidth}
By condition \ref{a3}, $F(\theta, y_1^n, z_1^n)$ is a continuous function of $\theta$ when $\theta \in [\theta_0, \theta_0 + \alpha_ke_k]$ for any value of $k$. Moreover by condition \ref{a4}, $\frac{\partial}{\partial \theta}F(\theta, y_1^n, z_1^n)$ exists for $\theta \in [\theta_0, \theta_0 + \alpha_ke_k] \setminus \beta_k$. By considering the $\mathbb{R}\rightarrow\mathbb{R}$ function $\mathcal{F}(\alpha_k) := F(\theta_0 + \alpha_ke_k, y_1^n, z_1^n)$, we can conclude by Dieudonne's mean value theorem \cite{dieudonne-mvt-thm} (Theorem 8.5.2) that
\begin{align*} 
& \left\lvert \dfrac{F(\theta + \alpha_k e_k, y_1^n, z_1^n) - F(\theta, y_1^n, z_1^n)}{\alpha_k} \right\rvert \leq \underset{\bar \theta \in [\theta_0, \theta_0 + \alpha_ke_k] \setminus \beta_k}{\sup} \big\lvert \hat g^k(\bar \theta, y_1^n, z_1^n) \big\rvert L(\bar \theta, y_1^n, z_1^n)  \\
\leq & \underset{\bar \theta \in [\theta_0, \theta_0 + \alpha_ke_k] \setminus \beta_k}{\sup} \norm{\hat g(\bar \theta, y_1^n, z_1^n)}L(\bar \theta, y_1^n, z_1^n). \stepcounter{equation}\tag{\theequation}\label{unbiased1proof1}
\end{align*}
where $\hat g^k$ is the $k$\textsuperscript{th} component of $\hat g$. Similarly let $\theta^k$ be the $k$\textsuperscript{th} component of $\theta$. 

To finish the proof,
\begin{align*} 
& \dfrac{\partial}{\partial \theta_0^k} \mathe_{z_1^n, y_1^n}\left[ f(x_1, \ldots, x_n) \right] = \underset{\alpha_k \rightarrow 0}{\lim} \underset{z_1^n, y_1^n \in (Z \times Y) \setminus \mathcal{ND}(\theta_0)}{\int} \dfrac{F(\theta_0 + \alpha_k e_k, y_1^n, z_1^n) - F(\theta_0, y_1^n, z_1^n)}{\alpha_k} dz_1^n dy_1^n \\
= &   \underset{z_1^n, y_1^n \in (Z \times Y) \setminus \mathcal{ND}(\theta_0)}{\int}\underset{\alpha_k \rightarrow 0}{\lim} \dfrac{F(\theta_0 + \alpha_k e_k, y_1^n, z_1^n) - F(\theta_0, y_1^n, z_1^n)}{\alpha_k} dz_1^n dy_1^n \\
 = & \underset{z_1^n, y_1^n \in (Z \times Y) \setminus \mathcal{ND}(\theta)}{\int}\hat g^k(\theta, y_1^n, z_1^n)  dz_1^n dy_1^n \\
 = & \mathe_{z_1^n, y_1^n}\big[\hat g^k(\theta)\big] .
\end{align*} 
The second line applies the dominated convergence theorem, as the the dominating function is established by \eqref{unbiased1proof1} and is assumed to be integrable per \eqref{equnbiased1}. As this argument holds for all $k$, it follows that $\frac{\partial}{\partial \theta}\mathe[f] = \mathe[\hat g(\theta)]$. \qed \\

Note that if $Y$ depends on $z_1^n$, the same argument as presented here applies. On the other hand, the case where $Y$ depends on $\theta$ (either directly or indirectly through the $x_i$'s) must be analyzed differently and can be considered in the future.

Equation \eqref{equnbiased1} shows the connection between proving the gradient is unbiased and proving that the second moment of the gradient is bounded. By recalling the definition of the second moment of $\hat g(\theta)$, 
\begin{align*} 
\underset{z_1^n, y_1^n \in (Z \times Y) \setminus \mathcal{ND}(\theta_0)}{\int} \norm{\hat g(\theta, y_1^n, z_1^n)}^2 L(\theta, y_1^n, z_1^n)dz_1^ndy_1^n 
\end{align*}
the connection is obvious. 

When applying theorem \ref{unbiased1}, it is clear how the conditions \ref{a2}, \ref{a4} and \ref{a3} can be verified. To assist with verifying the integrability condition \eqref{equnbiased1}, the next lemma establishes a generic bound on $\norm{\hat g}$. 
\begin{lemma}\label{integrability} The gradient estimator $\hat g$ satisfies the following inequality
\begin{align*} 
\norm{\hat g} \leq & \sum_{i=1}^n\norm{f(x_1, \ldots, x_n) \dfrac{\partial \log p_i(y_i \, | \, \cdot)}{\partial \theta}} + \sum_{i=1}^n \norm{\dfrac{\partial f}{\partial x_i}}\sum_{j=1}^i\prod_{k=1}^{i-j}\norm{\dfrac{\partial h_{i-k+1}}{\partial x_{i-k}}}\norm{\dfrac{\partial h_j}{\partial \theta}} \\
+ & \sum_{i=1}^{n-1} \norm{f(x_1, \ldots, x_n) \dfrac{\partial \log p_{i+1}(y_{i+1} \, | \, \cdot)}{\partial x_{i}}}\sum_{j=1}^{i}\prod_{k=1}^{i-j}\norm{\dfrac{\partial h_{i-k+1}}{\partial x_{i-k}}}\norm{\dfrac{\partial h_j}{\partial \theta}}. \stepcounter{equation}\tag{\theequation}\label{integrabilitylemma}
\end{align*} 
For a matrix $A$, $\norm{A}$ denotes the matrix norm induced by the Euclidean norm, i.e. $\norm{A} := \underset{v : \thinnorm{v} = 1}{\sup}\norm{Av}$. 
\end{lemma}
\textit{Proof.} 
\begin{align*} 
\norm{\hat g} & = \norm{\sum_{i=1}^n \Big(f(x_1, \ldots, x_n) \dfrac{\partial \log p_i(y_i \, | \, \cdot)}{\partial \theta} + \lambda_i^\top\dfrac{\partial h_i}{\partial \theta} \Big)} \\
& \leq \sum_{i=1}^n \norm{f(x_1, \ldots, x_n)\dfrac{\partial \log p_i(y_i \, | \, \cdot)}{\partial \theta}} + \sum_{i=1}^n \norm{\lambda_i} \norm{\dfrac{\partial h_i}{\partial \theta} }  .  \stepcounter{equation}\tag{\theequation}\label{lemma2eqn1}
\end{align*}
Now using \eqref{8} and assuming $i < n$, we have this inequality for $\norm{\lambda_i}$:
\begin{align*} 
& \norm{\lambda_i} \leq \norm{\dfrac{\partial f}{\partial x_i}} + \norm{f(\cdot) \dfrac{\partial \log p_{i+1}(y_{i+1}\, | \, \cdot)}{\partial x_i}} + \norm{\lambda_{i+1}}\norm{\dfrac{\partial h_{i+1}}{\partial x_i}} .
\end{align*}
Expanding out $\lambda_{i+1}$, and then $\lambda_{i+2}$ etc. in a recursive manner, we reach the inequality
\begin{align*} 
& \norm{\lambda_i} \leq \sum_{j=i}^n\norm{\dfrac{\partial f}{\partial x_j}}\prod_{k=1}^{j-i}\norm{\dfrac{\partial h_{j+1-k}}{\partial x_{j-k}}} + \sum_{j=i+1}^{n-1} f(x_1, \ldots, x_n) \norm{\dfrac{\partial \log p_{j+1}(y_{j+1} \, | \, \cdot)}{\partial x_j}}\prod_{k=1}^{j-i}\norm{\dfrac{\partial h_{j+1-k}}{\partial x_{j-k}}} . \stepcounter{equation}\tag{\theequation}\label{lemma2eqn2}
\end{align*}
which holds for all $i = 1, \ldots, n$. To finish the proof, combine \eqref{lemma2eqn2} and \eqref{lemma2eqn1}. The statement of the lemma follows after swapping the order of summation of $i, j$. \qed \\

Lemma \ref{integrability} breaks up the norm of the gradient estimator into individual parts so we can consider the integrability of each term seperately. If we square both sides of \eqref{integrabilitylemma}, we then also have an inequality for $\norm{\hat g}^2$ which can be used for showing that the second moment of the gradient is bounded.

Equation \eqref{integrabilitylemma} also shows the pattern of each term of the gradient estimator. One observation are the terms containing the product 
\begin{align*} 
\prod_{k=1}^{i-j}\norm{\dfrac{\partial h_{i-k+1}}{\partial x_{i-k}}} = \norm{\dfrac{\partial h_i}{\partial x_{i-1}}}\norm{\dfrac{\partial h_{i-1}}{\partial x_{i-2}}}\ldots \norm{\dfrac{\partial h_{j+1}}{\partial x_{j}}}.
\end{align*}
We of course expect such a term to arise from the chain rule. If $n$ is large and each $\norm{\partial h_{i} / \partial x_{i-1}}$ is $>>1$  (or $<<1$), the gradient can grow uncontrollably (or become close to 0). This observation is often referred to as the exploding/vanishing gradient problem. 

We now consider an alternative approach to showing unbiasedness, where we use Lipschitz continuity instead of the mean value theorem. Assuming that it is simple to show that the model $h_i$ is Lipschitz continuous or locally Lipschitz continuous, we can avoid having to verify the differentiability condition \ref{a4}. This can be useful in various neural network models, as they often have nondifferentiable points (e.g. due to ReLU nonlinearities) but are locally Lipschitz continuous. We consider this approach to mainly be useful in the case where the gradient estimator does not use score functions. Again, we give sufficient conditions for $\hat g(\theta_0)$ to be unbiased for some point $\theta_0$. 
\begin{condition}\label{a22}
Let the gradient estimator consist only of deterministic operations and pathwise derivatives so that there is no dependence on $y_1^n$. Let $\hat g(\theta_0, z_1^n)$ exist except for $z_1^n \in \mathcal{ND}(\theta_0)$ where  $\mathcal{ND}(\theta_0)$ is a subset of $Z$ with measure 0.
\end{condition}
\begin{condition} \label{a6}
\textbf{a) } \ For any $i= 1 \ldots n$, $z_1^n \in Z$, $x_{i-1} = x_{i-1}(\theta_0, z_1^{i-1})$, there exists constants $N_i(x_{i-1}, \theta_0, z_1^i)$, $\epsilon$ such that 
\begin{align*} 
\norm{h_i(x_{i-1}, \theta_1, z_i) - h_i(x_{i-1}, \theta_2, z_i)} \leq N_i(x_{i-1}, \theta_0, z_1^i) \norm{\theta_1 - \theta_2}
\end{align*}
for any $\theta_1$, $\theta_2$ such that $\norm{\theta_1 - \theta_0} \leq \epsilon, \norm{\theta_2 - \theta_0} \leq \epsilon_i$.

\noindent \textbf{b) } \ For any $i=2 \ldots n$, $z_1^n \in Z$, $x_{i-1} = x_{i-1}(\theta_0, z_1^{i-1})$, there exists a constant $K_i(x_{i-1}, \bar \theta, z_1^i)$ such that
\begin{align*} 
\norm{h_i(x_{i-1}^*, \bar \theta, z_i) - h_i(x_{i-1}^{**}, \bar \theta, z_i)} \leq K_i(x_{i-1}, \bar \theta, z_1^i) \norm{x_{i-1}^* - x_{i-1}^{**}}
\end{align*}
for any $x_{i-1}^* = x_{i-1}(\theta_1, z_1^{i-1}), \, x_{i-1}^{**} = x_{i-1}(\theta_2, z_1^{i-1}), \, \bar \theta$ where $\norm{\theta_1 - \theta_0} \leq \epsilon$, $\norm{\theta_2 - \theta_0} \leq \epsilon$, $\norm{\bar \theta - \theta_0} \leq \epsilon$. 
\end{condition}

\begin{condition} \label{a5}
For any $i= 1 \ldots n$, $z_1^n \in Z$, there exists a constant $M_i := M(x_1^*, \ldots x_{i-1}^*, x_i^{**}, \ldots x_n^{**})$ such that 
\begin{align*} 
\norm{ f(x_1^*, \ldots x_i^*, x_{i+1}^{**}, \ldots, x_n^{**}) - f(x_1^*, \ldots x_{i-1}^*, x_{i}^{**}, \ldots, x_n^{**})} \leq M_i \norm{x_i^* - x_i^{**}}
\end{align*}
for any $x_j^* = x_j(\theta_1, z_1^j)$, $j = 1, \ldots i$ where $\norm{\theta_1 - \theta_0} \leq \epsilon$ and any $x_k^{**} = x_k(\theta_2, z_1^k)$, $k = i, \ldots n$ where $\norm{\theta_2 - \theta_0} \leq \epsilon$. 
\end{condition}

Condition \ref{a6}a) states that each $h_i$ is locally Lipschitz continuous with respect to $\theta$. In the special case where there exists some Lipschitz constant that holds for any $\theta_1$, $\theta_2 \in \mathbb{R}^{\text{dim}(\theta)}$, then $h_i$ would be (globally) Lipschitz continuous. Because we only require local Lipschitz continuity, the Lipschitz constant can depend on $\theta_0$ and $x_{i-1}$. In any case, the Lipschitz constants are assumed to be random variables as they can depend on $z_1^n$. 
Condition \ref{a6}b) says that each $h_i$ ($i \geq 2$) is locally Lipschitz continuous with respect to $x_{i-1}$. Note that here we define ``local'' in the sense that the Lipschitz inequality holds for any $x_{i-1}^* = x_{i-1}(\theta_1, z_1^{i-1}), \, x_{i-1}^{**} = x_{i-1}(\theta_2, z_1^{i-1})$ where $\theta_1, \, \theta_2$ are both at most $\epsilon$ distance from $\theta_0$. This is in contrast to a more typical definition which would require $x_{i-1}^*,\, x_{i-1}^{**}$ to be within some distance to $x_{i-1}(\theta_0, z_1^{i-1})$. 
Condition \ref{a5} states that $f$ is locallly Lipschitz continuous with respect to each $x_1, \ldots x_n$.

\begin{theorem} \label{unbiasedthm2}
Let conditions \ref{a22}, \ref{a6} and \ref{a5} hold. Define 
\begin{align*} 
M_i := M(x_1(\theta^*, z_1^1), \ldots x_{i-1}(\theta^*, z_1^{i-1}), x_i(\theta_0, z_1^i), \ldots x_n(\theta_0, z_1^n)) .
\end{align*}
If the integrability condition
\begin{adjustwidth}{-2cm}{0cm} \vspace{-1.3cm}
\begin{align*} 
\underset{z_1^n \in Z \setminus \mathcal{ND}(\theta_0)}{\int}\sum_{i=1}^n M_i \sum_{j=1}^i \prod_{k=1}^{i-j}K_{i-k+1}(x_{i-k}(\theta_0, z_1^{i-k}), \theta^*, z_1^{i-k+1})N_{j}(x_{j-1}(\theta_0, z_1^{j-1}), \theta_0, z_1^j)\mu(z_1^n)dz_1^n < \infty \stepcounter{equation}\tag{\theequation}\label{thm3eqn}
\end{align*}
\end{adjustwidth}
is satisfied for any $\theta^*$ such that $\norm{\theta^* - \theta_0} \leq \epsilon$, then $\hat g(\theta_0)$ is unbiased. 
\end{theorem} 
\textit{Proof.} 
Define 
\begin{align*} 
f_i := f(x_1(\theta^*, z_1^1), \ldots, x_i(\theta^*, z_1^i), x_{i+1}(\theta_0, z_1^{i+1}), \ldots, x_n(\theta_0, z_1^n))
\end{align*}
where $\theta^*$ satisfies $\norm{\theta^* - \theta_0} \leq \epsilon$.
We have
\begin{align*} 
& \left\lvert f_n - f_0 \right\rvert = \left\lvert f_n - f_0 + \sum_{i=1}^{n-1}(f_i - f_i)\right\rvert  = \left\lvert \sum_{i=1}^n(f_i - f_{i-1})\right\rvert \leq \sum_{i=1}^n \left\lvert f_i - f_{i-1} \right\rvert . \stepcounter{equation}\tag{\theequation}\label{thm3proof0}
\end{align*}
In the rest of the proof, let us suppress the dependence of $z_1^i$ so that $x_i(\theta_0)$ refers to $x_i(\theta_0, z_1^i)$. Now using the local Lipschitz continuity per conditions \ref{a5} and \ref{a6} we have
\begin{align*} 
& \left\rvert f_i - f_{i-1} \right\rvert \leq M_i \norm{x_i(\theta^*) - x_i(\theta_0)} \\
 & = M_i \norm{h_i(x_{i-1}(\theta^*), \theta^*, z_i) -  h_i(x_{i-1}(\theta_0), \theta_0, z_i) } \\
 & = M_i \norm{h_i(x_{i-1}(\theta^*), \theta^*, z_i) -  h_i(x_{i-1}(\theta_0), \theta^*, z_i)
 + h_i(x_{i-1}(\theta_0), \theta^*, z_i) - h_i(x_{i-1}(\theta_0), \theta_0, z_i) } \\
& \leq M_i\big( K_i(x_{i-1}(\theta_0), \theta^*, z_1^i)\norm{x_{i-1}(\theta^*) - x_{i-1}(\theta_0)} +N_i(x_{i-1}(\theta_0), \theta_0, z_1^i)\norm{\theta^* - \theta_0}\big). \stepcounter{equation}\tag{\theequation}\label{thm3proof1}
\end{align*}
Thus we have the inequality 
\begin{align*} 
\norm{x_i(\theta^*) - x_i(\theta_0)} \leq K_i(x_{i-1}(\theta_0), \theta^*, z_1^i)\norm{x_{i-1}(\theta^*) - x_{i-1}(\theta_0)} + N_{i}(x_{i-1}(\theta_0), \theta_0, z_1^i)\norm{\theta^* - \theta_0}
\end{align*}
for any $i = 1, \ldots n$ (note that if $i=1$, then $\norm{x_{i-1}(\theta^*) - x_{i-1}(\theta_0)} = \norm{x_0 - x_0} = 0$). Now unpacking this recursively and continuing \eqref{thm3proof1}  we have
\begin{align*} 
 \left\rvert f_i - f_{i-1} \right\rvert \leq M_i \sum_{j=1}^{i}\prod_{k=1}^{i-j}K_{i-k+1}(x_{i-k}(\theta_0), \theta^*, z_1^{i-k+1})N_{j}(x_{j-1}(\theta_0), \theta_0, z_1^j) \norm{\theta^* - \theta_0}. 
\end{align*}
Combining the above with \eqref{thm3proof0} we have 
\begin{align*} 
& \left\lvert f_n - f_0 \right\rvert \dfrac{1}{\norm{\theta^* - \theta_0}} \leq \sum_{i=1}^n M_i \sum_{j=1}^{i}\prod_{k=1}^{i-j}K_{i-k+1}(x_{i-k}(\theta_0), \theta^*, z_1^{i-k+1})N_{j}(x_{j-1}(\theta_0), \theta_0, z_1^j) \stepcounter{equation}\tag{\theequation}\label{lipschitz-inequality}
\end{align*}
so we can conclude by \eqref{thm3eqn} that $\lvert f_n - f_0 \rvert / \norm{\theta^*-\theta_0}$ is integrable. This establishes our domating function for the dominated convergence theorem. Let us consider $\theta^* = \theta_0 + e_k \alpha$ for any $k= 1, \ldots \text{dim}(\theta)$, where $\alpha < \epsilon$ so that $\norm{\theta^* - \theta_0} = \alpha < \epsilon$. 
\pushQED{\qed}
\begin{align*} 
&\dfrac{\partial}{\partial \theta_k}\mathe_{z_1^n}[f(x_1, \ldots, x_n)] \\
 = & \underset{\alpha \rightarrow 0}{\lim}\underset{z_1^n \in Z \setminus \mathcal{ND}(\theta_0)}{\int} \frac{1}{\alpha}\big( f(x_1(\theta^*), \ldots, x_n(\theta^*)) - f(x_1(\theta_0), \ldots, x_n(\theta_0)) \big)\mu(z_1^n)dz_1^n \\
  = & \underset{z_1^n \in Z \setminus \mathcal{ND}(\theta_0)}{\int}  \underset{\alpha \rightarrow 0}{\lim} \, \frac{1}{\alpha}\big( f(x_1(\theta^*), \ldots, x_n(\theta^*)) - f(x_1(\theta_0), \ldots, x_n(\theta_0)) \big)\mu(z_1^n)dz_1^n \\
 = & \underset{z_1^n \in Z \setminus \mathcal{ND}(\theta_0)}{\int} \hat g_k(\theta_0, z_1^n) \mu(z_1^n)dz_1^n =  \mathe_{z_1^n}[\hat g_k(\theta_0)].  \qedhere
\end{align*} 
\popQED \\

The integrability condition \eqref{thm3eqn} is very similar to that given in \eqref{equnbiased1}. After using the inequality \eqref{integrabilitylemma} (of course when ignoring the terms related to score functions), we find both integrability conditions have the same structure. 

Our last result ties together unbiasedness of the gradient estimator and Lipschitz continuity of the expected objective. 
\begin{theorem}\label{lipschitzcor}
Let $\hat g(\theta^*)$ be unbiased for $\theta^*$ in some set $\Theta$. Let $\hat g(\theta^*)$ be integrable for $\theta^* \in \Theta$, i.e.
\begin{align*} 
\mathe_{z_1^n, y_1^n}[\norm{\hat g(\theta^*)}] = \underset{z_1^n, y_1^n \in Z \times Y}{\int} \norm{\hat g(\theta^*, y_1^n, z_1^n)} L(\theta^*, y_1^n, z_1^n)dz_1^ndy_1^n < \infty  .
\end{align*}
Then $\mathe_{z_1^n, y_1^n}[f(x_1, \ldots, x_n)]$ is Lipschitz continuous for $\theta \in \Theta$. 
\end{theorem}
\textit{ Proof.}
$\mathe_{z_1^n, y_1^n}[f(x_0, \ldots, x_n)]$ is differentiable on $\Theta$ because 
\begin{align*} 
\dfrac{\partial}{\partial \theta}\mathe_{z_1^n, y_1^n}[f(x_1, \ldots, x_n)] = \mathe_{z_1^n, y_1^n}[ \hat g(\theta)]
\end{align*}
which exists since $\hat g$ is integrable. Then, $\mathe_{z_1^n, y_1^n}[f(x_1, \ldots, x_n)]$ is also continuous on $\Theta$ since differentiability implies continuity. It follows that $\mathe_{z_1^n, y_1^n}[f(x_1, \ldots, x_n)]$ is Lipschitz, with Lipschitz constant
\pushQED{\qed}
\begin{align*} 
& \underset{\theta^* \in \Theta}{\sup} \norm{\mathe_{z_1^n, y_1^n}[ \hat g(\theta^*)]} \leq \underset{\theta^* \in \Theta}{\sup} \mathe_{z_1^n, y_1^n}{[ \norm{\hat g(\theta^*)}]} . \qedhere
\end{align*}
\popQED \\
We can also use theorem \ref{lipschitzcor} to show Lipschitz continuity of the expected gradient $\dfrac{\partial}{\partial \theta} \mathe_{z_1^n, y_1^n}[f(x_1, \ldots, x_n)]$. To do this, first show that $\hat g$ is unbiased. Then by applying the theorem \ref{thm1-1} estimator again on a single component of $\hat g$, we obtain an estimator for a single row of the Hessian. By showing each row of the Hessian is unbiased, it follows by theorem \ref{lipschitzcor} that $\dfrac{\partial}{\partial \theta} \mathe_{z_1^n, y_1^n}[f(x_1, \ldots, x_n)]$ is Lipschitz continuous. We give such an example in the next subsection. 

The motivation for showing Lipschitz continuity of the expected gradient would be to establish the convergence of SGD (in the sense of \eqref{bottou}) which applies when the objective is nonconvex. However, this requires global Lipschitz continuity, i.e. where $\Theta$ is all of $\mathbb{R}^{\text{dim}(\theta)}$. Otherwise, we also saw the guarentee \eqref{nemirovski} which applies when $\Theta$ is some convex set and the objective is convex. In that case we only require $\hat g$ to be unbiased on $\Theta$, and for the second moment of $\hat g$ to be bounded on $\Theta$.


\section{Application to Deep Neural Networks} \label{sec-2-4}
Let us consider a model defined as a feedforward, fully connected neural network with an arbitrary number of hidden layers.
Denote the input to model as $z$, and suppose that each dimension of $z$ is identically independently distributed as $\text{Normal}(0,1)$. Within our framework we have 
\begin{align*} 
& x_1 = \sigma(W_1 z + b_1) \\
& x_2 = \sigma(W_2 x_1 + b_2) \\
&  \qquad \vdots \\
 & x_{n-1} = \sigma(W_{n-1} x_{n-2} + b_{n-1}) \\
 & x_n = W_n x_{n-1} + b_n
\end{align*}
where $x_1$ through $x_{n-1}$ represent the hidden layers and $x_n$ is the output layer. The activation function $\sigma$ is applied element-wise and is assumed to be the same for all layers. The kernel for the $i$\textsuperscript{th} layer is $W_i \in  \mathbb{R}^{\kappa_i \times \kappa_{i-1}}$ where $\kappa_i$ is the number of neurons in the $i$\textsuperscript{th} layer (define $\kappa_0$ to be the dimension of $z$). The biases for the $i$\textsuperscript{th} layer are similarly denoted as $b_i \in \mathbb{R}^{\kappa_i}$. Following our convention of using superscripts to denote the indexing of vectors, we will denote the $(j, k)$ entry of $W_i$ as $W_i^{j,k}$, the $j$ row of $W_i$ as $W_i^j$, and the $j$ entry of $b_i$ as $b_i^j$. Similarly $z^k$ will denote the $k$\textsuperscript{th} entry of $z$. It is assumed there is a seperate parameter for each weight $W_i^{j,k}$ and bias $b_i^j$, and $\theta$ is a vector containing each weight/bias. 

Define the objective as minimizing the squared error between the model output and some ground truth
\begin{align*} 
\underset{\theta}{\min} \ \ \mathe_z \left[ f(x_1, \ldots, x_n) \right] = \underset{\theta}{\min} \ \ \mathe_z \left[ \big(x_n(\theta, z) - \hat f(z)\big)^\top\big(x_n(\theta, z) - \hat f(z)\big) \right]
\end{align*}
where the ground truth $\hat f(z) : \mathbb{R}^{\kappa_0} \rightarrow \mathbb{R}^{\kappa_n}$ is the ``true'' mapping between the input and output which we wish to learn. 

We will show how to formally prove that the gradient estimator is unbiased for $\tanh$ and $\text{ReLU}$ nonlinearities.

First let us write down the gradient estimator. We have 
\begin{align*} 
& \hat g = \sum_{i=1}^n \lambda_i^\top \dfrac{\partial h_i}{\partial \theta} \\
& \lambda_n^T = 2\big( x_n - \hat f(z)\big)^\top  \qquad \lambda_{n-1}^\top = 2\big( x_n - \hat f(z)\big)^TW_n \\
& \lambda_{i-1}^\top = \lambda_i^\top \dfrac{\partial h_i}{\partial x_{i-1}} = \lambda_i^\top\text{diag}\big(\sigma'(W_i x_{i-1} + b_i)\big)W_i \quad i = n-1, \ldots, 2  \stepcounter{equation}\tag{\theequation}\label{NNeg-adjointvariables}
\end{align*}
where $\sigma'$ is the derivative of the activation function and $\text{diag}(\cdot)$ converts a vector into a square diagonal matrix. Thus the adjoint variables have a simple form 
\begin{align*} 
&\lambda_n^T = 2\big( x_n - \hat f(z)\big)^T \\
& \lambda_i^\top = 2\big( x_n - \hat f(z)\big)^T W_n \prod_{l=1}^{n-1-i}\text{diag}\big( \sigma'(W_{n-l}x_{n-l-1} + b_{n-l})\big)W_{n-l} \quad i = 1, \ldots, n-1 \stepcounter{equation}\tag{\theequation}\label{NN-eg0}
\end{align*}

\subsubsection{Tanh activation function}
Let $\sigma( \cdot) = \tanh(\cdot)$ so that all the hidden layers use $\tanh$ nonlinearities.
\begin{theorem} \label{tanh-proof}
Let $\mathe_z\left[\shortnorm{ \hat f(z)}^2 \right] < \infty$ and for any $k =1, \ldots \kappa_0$ let $\mathe_z\left[\shortnorm{ \hat f(z)}^2\lvert z^k \rvert \right] < \infty$ and $\mathe_z\left[\shortnorm{ \hat f(z)}^2\lvert z^k \rvert^2 \right] < \infty$. Also let $\theta \in \Theta$ where $\Theta$ is a bounded subset of $\mathbb{R}^{\text{dim}(\theta)}$. Then, 
\begin{enumerate}
\item $\hat g(\theta)$ is an unbiased gradient estimator.
\item $\mathe_z[\norm{\hat g(\theta)}^2] < \infty$.
\item $\mathe_z \left[ f(x_1, \ldots, x_n) \right]$ is Lipschitz continuous on $\Theta$.
\item $\dfrac{\partial}{\partial \theta}\mathe_z \left[ f(x_1, \ldots, x_n) \right]$ is Lipschitz continuous on $\Theta$. 
\end{enumerate}
\end{theorem}

We have shown unbiasedness of the gradient in a rigorous way, instead of simply assuming that the interchange of expectation and derivative is permissable. It turns out that in this problem, we simply need the ground truth function $\hat f(z)$ to be well behaved in the sense that $\mathe_z[\,\shortnorm{\hat f(z)}\,] < \infty$ and $\mathe_z[\, \shortnorm{\hat f(z)}\, \lvert z^k \rvert \,] < \infty$ for all $k = 1, \ldots \kappa_0$. The stronger requirements given in theorem \ref{tanh-proof} are sufficient for the second moment of the gradient to be bounded. This example also shows how the same framework which we use to calculate and analyze the gradient also applies to higher order derivatives such as the Hessian. 

Perhaps the most notable result is that the expected gradient is only locally Lipschitz continuous. That is, the Hessian is bounded for any $\theta \in \Theta$ for some bounded set $\Theta$. But no bound exists for $\theta \in \mathbb{R}^{\text{dim}(\theta)}$. Without the global Lipschitz condition, we do not have the nonconvex convergence result \eqref{bottou}. This suggests the need to consider models with local Lipschitz conditions in future work in stochastic gradient methods/stochastic approximation. 

\subsubsection{ReLU activation function}
Let $\sigma(x) = \max(0, x)$, i.e. let the activations use ReLU nonlinearity. Note the definition of the derivative of ReLU, $\sigma'(x) = \mathbbm{1}( x > 0)$ where $\mathbbm{1}$ is the indicator function. 
\begin{assumption} \label{relu-asmpt}
$\hat g(\theta)$ exists with probability 1. 
\end{assumption}
The derivative of ReLU does not exist at 0, so for $\hat g(\theta)$ to exist with probability 1, there should be zero probability that any neuron has a pre-activation value of zero. In other words, we require that 
\begin{align*} 
& \mathbb{P}(W_1^j z + b_1^j = 0) = 0 \qquad \text{for all $j = 1, \ldots \kappa_1$} \stepcounter{equation}\tag{\theequation}\label{asmpt1} \\
& \mathbb{P}(W_i^jx_{i-1} + b_i^j  = 0 ) = 0 \qquad \text{for all $i=2, \ldots n-1$ and $j = 1, \ldots \kappa_i$} \stepcounter{equation}\tag{\theequation}\label{asmpt2}
\end{align*}
It follows that for the first layer, the only way \eqref{asmpt1} will not be satisfied for the $j$\textsuperscript{th} neuron is if $W_1^j$ and $b_1^j$ are both zero. For the later layers, \eqref{asmpt2} implies that $b_i^j$ should not be zero, as in general, there may be a nonzero probability that the activations of the previous layer are all zero. 

\begin{theorem} \label{relu-thm1}
Let $\theta \in \Theta$, where $\Theta$ is a bounded subset of $\mathbb{R}^{\text{dim}(\theta)}$ such that $\hat g(\theta)$ exists with probability 1 for any $\theta \in \Theta$. Let $\mathe_z\left[ \, \shortnorm{\hat f(z)}^2 \, \right]$, $\mathe_z\left[ \, \shortnorm{\hat f(z)}^2 \, \lvert z^k \rvert \, \right]$, $\mathe_z\left[ \, \shortnorm{\hat f(z)}^2 \, \lvert z^k \rvert^2 \, \right]$, $\mathe_z\left[ \, \shortnorm{\hat f(z)} \, \lvert z^k \rvert^2 \, \right]$, $\mathe_z\left[ \, \shortnorm{\hat f(z)}^2 \, \lvert z^k \rvert^3 \, \right]$ be finite for any $k = 1 \dots, \kappa_0$. Then, 
\begin{enumerate}
\item $\hat g(\theta)$ is an unbiased gradient estimator.
\item $\mathe_z[ \norm{\hat g(\theta)}^2] < \infty$.
\item $\mathe_z[ f(x_1, \ldots, x_n)]$ is Lipschitz continuous on $\Theta$. 
\end{enumerate}
\end{theorem}

Again, we see that unbiasedness and the second moment bound are achieved given some restrictions on $\hat f(z)$. Because of the ReLU nonlinearities, we cannot prove that the expected gradient is Lipschitz continuous on $\Theta$. This is because the gradient estimator is not continuous, so it follows that the Hessian estimator is biased. The simplest way to see that the gradient estimator is not continuous is to observe the adjoint variables, e.g. $\lambda_{n-2} = 2(x_n - \hat f(z))^\top W_n\text{diag}(\mathbbm{1}(W_{n-1}x_{n-2} + b_{n-1} > 0))W_{n-1}$ which is not continuous with respect to $W_{n-1}$ and $b_{n-1}$. Although we could not show it with the present analysis, it seems certain that $\frac{\partial}{\partial \theta}\mathe_z[ f(x_1, \ldots, x_n)]$ should be Lipschitz continuous on $\Theta$. For example, consider the simple neural network with a single hidden neuron and a single output, $x_1 = \max(\theta^1 z + \theta^2), \ x_2 = \theta^3 x_1 + \theta^4$, \ $f(x_1, x_2) = (x_2 - \hat f(z))(x_2 - \hat f(z))$, where $z \sim \text{Normal}(0,1)$, $x_2 \in \mathbb{R}$ is the output and $\theta^1, \theta^2, \theta^3, \theta^4$ are the parameters. For this simple example, it can be confirmed that indeed $\frac{\partial}{\partial \theta}\mathe_z[f(x_1, x_2)]$ is locally Lipschitz continuous even though $\frac{\partial}{\partial \theta}f(x_1, x_2)$ is not continuous.

We also find the seemingly benign assumption \ref{relu-asmpt} to be surprisingly restrictive, implying all the biases should be non-zero. At present it is unclear how this would change if considering a framework involving subderivatives. 

\section{The purely deterministic setting} \label{sec-2-5}
Having concluded our study in the stochastic setting, we now consider the purely deterministic setting, i.e. the problem 
\begin{align*} 
\underset{\theta}{\min} & \quad  f(x_1, \ldots, x_n)       \stepcounter{equation}\tag{\theequation}\label{deterministic}\\
\text{s.t.} & \quad x_{i} = h_i(x_{i-1}, \theta)  \quad i = 1, \ldots, n .
\end{align*}
Note that the gradient in this case was previously presented in section \ref{sec-1-2}. The deterministic setting is both simpler and of lesser interest; thus, it will be covered in a lesser amount of detail. 

First, it should be made clear that whereas an extremely simple algorithm such as stochastic gradient descent gives good results in the stochastic setting, the deterministic problem requires more sophisticated algorithms in order to achieve satisfactory results. The in-depth discussion of such algorithms is beyond the scope of this document, but it is sufficient to understand that \textit{quasi-Newton} algorithms are considered to be state of the art. These algorithms only require the user to supply the objective and gradient evaluations, but they are able to approximate the Newton search direction (i.e. a search direction which takes into account the Hessian). Slightly more detail on these algorithms is given later in \ref{algos}. For a full discussion, we offer \cite{71} as an excellent reference which is both rigorous and practically useful. 

\subsubsection{Convergence of BFGS}
The Broyden-Fletcher-Goldfarb-Shanno method (BFGS) is a well known quasi-Newton method. The variant L-BFGS-B \cite{76} is particularly widely used. Defining $\hat g(\theta)$ as the gradient, and $F(\theta) := f(h_1(x_0, \theta), h_2(h_1(x_0, \theta), \theta), \ldots)$ as the objective, we can now state a convergence result on BFGS due to \cite{bfgs-convergence}. 

If the level set 
\begin{align*} 
\Omega = \{ \theta \in \mathbb{R}^m \, : \, F(\theta) \leq F(\theta^0)\}
\end{align*}
is bounded for the initial parameter value $\theta^0$, if $\hat g(\theta)$ exists for all $\theta \in \Omega$, and $\hat g(\theta)$ is Lipschitz continuous, then BFGS will converge to a stationary point. We will assume that the level set condition holds, and again we find that we would like to be able to reason as to when the gradient is Lipschitz continuous. 

\subsubsection{Lipschitz-continuity}
Define $x_i(\theta)$ as the value of $x_i$ given $\theta$, that is
\begin{align*} 
x_i(\theta) := h_i(x_{i-1}(\theta), \theta)
\end{align*}
where $x_0(\theta)$ is defined simply as $x_0$. Let us recall the result presented in section \ref{sec-1-2} that the gradient estimator $\hat g$ in the deterministic setting is given by 
\begin{align*} 
\hat g := & \sum_{i=1}^{n}\lambda_i^\top \dfrac{\partial h_i}{\partial \theta} \\
& \lambda_{n}^\top  = \dfrac{\partial f}{\partial x_n} \\
& \lambda_{i-1}^\top = \dfrac{\partial f}{\partial x_{i-1}} + \lambda_i^\top \dfrac{\partial h_i}{\partial x_{i-1}} = 0 \quad i = n, n-1, \ldots, 2 .
\end{align*}
\begin{proposition}
For any $\theta \in \Omega$ and any $i= 1, \ldots, n$, let $h_i^*(\theta) = h_i(x_{i-1}(\theta), \theta)$ be continuous. Also let $f(\bar x_1, \ldots, \bar x_n)$ be continuous for $\{\bar x_i \in \{x_i(\theta) : \theta \in \Omega \} : i = 1, \ldots, n\}$. Then, $F(\theta)$ is continuous for any $\theta \in \Omega$.  
\end{proposition}
The proof is trivial, as it follows simply from the fact that compositions of continuous functions are continuous. 
\begin{lemma}
Let $F(\theta)$ be continuous for any $\theta \in \Omega$. For any $i=1, \ldots, n$, and any $\theta \in \Omega$, let $h_i(x_{i-1}(\theta), \theta)$ be differentiable with respect to $x_{i-1}$ and $\theta$, and let $f(x_1(\theta), \ldots, x_n(\theta))$ be differentiable with respect to $x_i$. Let the partial derivatives $\dfrac{\partial h_i}{\partial \theta}$, $\dfrac{\partial h_i}{\partial x_{i-1}}$, and $\dfrac{\partial f}{\partial x_i}$ be bounded, meaning there exists some constants $M_x$, $M_{\theta}$, $M_f$ such that 
\begin{align*} 
& \norm{\dfrac{\partial h_i}{\partial \theta}} \leq M_{\theta} \qquad \norm{\dfrac{\partial h_i}{\partial x_{i-1}}} \leq M_x \qquad \norm{\dfrac{\partial f}{\partial x_i}} \leq M_f
\end{align*}
for any $\theta in \omega$, $i= 1, \ldots, n$. Then $F(\theta)$ is Lipschitz continuous. 
\end{lemma}
\textit{Proof.}
It follows by assumption that $F(\theta)$ is both continuous and differentiable. The partial derivatives being bounded imply that $\norm{\dfrac{\partial F}{\partial \theta}} = \norm{\hat g(\theta)}$ is bounded because 
\begin{align*} 
\norm{\hat g} \leq \sum_{i=1}^n \norm{\lambda_i^\top} \norm{\dfrac{\partial h_i}{\partial \theta}} \leq \sum_{i=1}^n \norm{\lambda_i} M_{\theta} .
\end{align*}
We also have that $\norm{\lambda_i}$ is bounded because $\norm{\lambda_n^\top} \leq M_{f}$, and for any $i = n, \ldots, 2$ we have
\begin{align*} 
\norm{\lambda_{i-1}} \leq M_f + \norm{\lambda_i}M_x .
\end{align*}
It thus follows that $\norm{F(\theta)}$ is bounded. Then, by lemma \ref{lipschitz1}, $F(\theta)$ is Lipschitz continuous on $\theta \in \Omega$. \hfill $\qed$ \\

This lemma gives a way to verify Lipschitz continuity for $F$. To verify Lipschitz continuity for $\hat g$, our approach will be the same as the stochastic setting in section \ref{sec-2-4}. That is, by considering the new problem
\begin{align*} 
\underset{\theta}{\min} & \quad \sum_{i=2n+1}^{3n} x_i \\
\text{s.t.} & \quad  x_{\gamma} = h_{\gamma}(x_{\gamma-1}, \theta) \quad \gamma = 1, \ldots, n  \\
& \quad x_{n+1} = \dfrac{\partial f}{\partial x_n}^\top \\ 
& \quad  x_{n+\gamma} = \dfrac{\partial f}{\partial x_{n+1-\gamma}}^\top + \dfrac{\partial h_{n+2-\gamma}}{\partial x_{n+1-\gamma}}x_{n+\gamma -1} \quad \gamma = 2, \ldots, n \\
& \quad  x_{2n + \xi} = x_{n+\xi}^\top \dfrac{\partial h_{n+1-\xi}}{\partial \theta_j} \quad \xi = 1, \ldots, n
\end{align*}
which is equal to the $j$ component of $\hat g$.

\chapter{Selected Examples}
In this chapter we consider different examples which apply the gradient estimator discussed in chapter 1. Besides demonstrating how our optimization framework can be applied, these examples also support the discussion of section \ref{sec-1-5} and of chapter 2. The examples in section \ref{sec-3-1} and \ref{sec-3-3} explain the new capabilities of our gradient estimator and in section \ref{sec-3-4} we further discuss the implications of our methodology as it pertains to modeling.

\section{Stochastic Differential Equations} \label{sec-3-1}
Stochastic differential equations (SDE) describe the evolution of a stochastic process. They consist of both a deterministic component, the drift, and a stochastic component, the noise. In the most common case, the noise depends on a Brownian motion and is referred to as the diffusion. This standard situation is formally defined by the Ito stochastic differential equation
\begin{align*} 
dx = b(x, \theta)dt + A(x, \theta)dW  \stepcounter{equation}\tag{\theequation} \label{4.1.0}
\end{align*}
where $x(t) \in \mathbb{R}^k$ is a stochastic process, $b \in \mathbb{R}^k$ is the drift, $A \in \mathbb{R}^{k \times k}$ is the diffusion, and $W(t)$ is an $k$ dimensional Wiener process (i.e. Brownian motion). We can avoid having to deal with the complexities of the underlying continuous time process by instead working directly with a discretization of \eqref{4.1.0}. One simple and common discretization is the Euler-Maruyama discretization \cite{kloeden-platen}
\begin{align*} 
x_{i} = x_{i-1} + b(x_{i-1}, \theta)\Delta t + A(x_{i-1}, \theta)z_{i} \stepcounter{equation}\tag{\theequation}\label{4.1.1}
\end{align*}
where $x_i := x(t_i)$ for a time discretization $\{t_0, t_1, \ldots, t_n\}$ with constant intervals $\Delta t$. Each component of the vector $z_{i}$ has a $\text{Normal}(0, \Delta t)$ distribution. Equation \eqref{4.1.1} is in the form $x_{i} = h_{i}(x_{i-1}, \theta, z_{i})$, so given an arbitrary objective function $f$, applying theorem \ref{thm1-1} immediately gives the gradient estimator. In a computer implementation, it's most convenient to leave the model in the form $h_i(x_{i-1}, \theta, z_{i})$, but for exposition purposes we can write the gradient estimator as
\begin{align*} 
& \hat g(\theta) =  \sum_{i=1}^{n}\lambda_i^\top\Big( \dfrac{\partial b(x_{i-1})}{\partial \theta}\Delta t  + \dfrac{\partial A(x_{i-1}) z_i}{\partial \theta} \Big) \\
& \lambda_{n} = \dfrac{\partial f}{\partial x_n}, \qquad  \lambda_{i-1}^\top = \dfrac{\partial f}{\partial x_{i-1}} +\lambda_i^\top\Big(I + \dfrac{\partial b(x_{i-1})}{\partial x_{i-1}}\Delta t  + \dfrac{\partial A(x_{i-1}) z_i}{\partial x_{i-1}} \Big)  \stepcounter{equation}\tag{\theequation}\label{4.1.1.1}
\end{align*}
where $I$ is the $k \times k$ identity matrix. This estimator involves only pathwise derivatives. It can be applied to an arbitrary nonlinear SDE, unlike the estimators of \cite{glasserman} which only apply for linear SDEs. \eqref{4.1.1.1} gives the exact gradient of the discretization (sometimes called ``discretize first''), compared to the approach of \cite{scalable-sde} which calculates the adjoint equations for the continuous time stochastic process, and then solves the continuous time adjoint equations with some discretization. 

\subsubsection{Score Function Formulations}
By observing that $x_i$ has a multivariate normal distribution when conditioned on $x_{i-1}$, we can also consider alternative formulations of \eqref{4.1.1} using score functions. If the diffusion $A$ is lower triangular with positive diagonal entries (see online appendix), we can formulate the model as
\begin{align*} 
& x_i = y_i, \qquad p_i(y_i \, | \, x_{i-1}, \theta) = \dfrac{\exp\big( -\frac{1}{2}(y_i - \mu_i )^\top \Sigma_i^{-1}(y_i - \mu_i )\big)}{\sqrt{(2\pi)^k \text{det}(\Sigma_i)}} \stepcounter{equation}\tag{\theequation}\label{sde-lr1} \\
& \mu_i = x_{i-1} + b(x_{i-1}, \theta) \Delta t \qquad \Sigma_i = \Delta t A(x_{i-1}, \theta) A(x_{i-1}, \theta)^\top .
\end{align*}
Thus we have $h_i(x_{i-1},\theta, z_i, y_i) = y_i$. We still sample $y_i$ as $x_{i-1} + b(x_{i-1}, \theta)\Delta t + A(x_{i-1}, \theta)z_{i}$, so the forward solve is identical to \eqref{4.1.1}. The reverse solve uses only score functions, with no need to calculate the adjoint variables. It follows that \eqref{sde-lr1} can be used when the objective function is not continuous. 

We can also consider a formulation which consists of both score functions and a deterministic part. 
\begin{align*} 
x_i = x_{i-1} + b(x_{i-1}, \theta)\Delta t + y_i  \stepcounter{equation}\tag{\theequation}\label{sde-mixed} \\ 
p_i(y_i \, | \, x_{i-1}, \theta) = \dfrac{\exp\big( -\frac{1}{2}y_i^T \Sigma_i^{-1}y_i\big)}{\sqrt{(2\pi)^k \text{det}(\Sigma_i)}} 
\end{align*}
For this specific problem (i.e. noise driven by a brownian motion), this option offers no benefits. \eqref{4.1.1} should have a lower variance, and unlike \eqref{sde-lr1} the existence of $\frac{\partial f}{\partial x_i}$ is still required. But by considering different distributions for $y_i$, we find that we can easily model stochastic differential equations with different types of noise, including noise with discrete valued distributions. For example, if $y_i$ had a Poisson distribution, then \eqref{sde-mixed} corresponds to a SDE with noise driven by a Poisson point process.

Of course the mixed score function/pathwise estimators are also useful when $h_i$ is a piecewise function. To show the flexibility of our estimator, consider this stochastic process which corresponds to an ordinary differential equation (ODE) with random jumps
\begin{align*} 
& x_i = x_{i-1} + b(x_{i-1}, \theta)\Delta t + \alpha(x_{i-1}, y_i, \theta)  \qquad \alpha(\cdot) = \begin{bmatrix}\alpha^1(\cdot) & \alpha^2(\cdot) & \ldots & \alpha^k(\cdot) \end{bmatrix} \stepcounter{equation}\tag{\theequation}\label{jump-ode} \\
& \alpha^j(x_{i-1}, y_i, \theta) = \begin{cases} 
 \alpha^j_1(x_{i-1}, \theta) & \text{ if }y_i^j = 1 \\ 
  \alpha^j_2(x_{i-1}, \theta) & \text{ if } y_i^j = 2 \\ 
  \quad \vdots & \\
  \end{cases} \qquad y_i^j \in \{1, 2, \ldots, l\}.
\end{align*}
We use superscripts to indicate individual entries of a vector, and $\alpha \in \mathbb{R}^k$, $y_i \in \mathbb{R}^k$. For each of the $k$ dimensions of the ODE, there are $l$ distinct possibilities for $\alpha^j \in \mathbb{R}$, denoted as $ \alpha^j_1 \ldots \alpha^j_l$. Thus in total there are $l^k$ distinct possibilities for the functional form of the vector $\alpha$, given by combinations of the $l \cdot k$ different functions $\alpha^1_{1} \ldots \alpha^k_l$. The discrete vector $y_i$ encodes a probability distribution for which $\alpha$ to use for any given $x_{i-1}, \, \theta$. Any entry $y_i^j$ is given by a categorical distribution depending on $x_{i-1}, \, \theta$, and dictates which of the $\alpha^j_{\cdot}$ to use. We can think of each $\alpha^j$ as a discrete random variable whose support as well as pmf both depend on $x_{i-1}, \theta$.

With similar formulations, we can consider other complexities such as having noise defined by a mixture distribution, or having a piecewise drift. In any case, the key is using $y_i$ to encode the different pieces of $h_i$. As discussed in section \ref{sec-3-1}, the significance of this idea is that a) we can learn the switching condition between different branches by learning a probability distribution for $y_i$ and b) 
we avoid introducing any non-differentiable points by allowing $h_i$ to be piecewise in this way.

\subsubsection{Numerical examples}
To validate our gradient estimators, we test the standard SDE with Brownian motion \eqref{4.1.1} and our jump ODE \eqref{jump-ode}. We explicitly add dependence on the time $t_i$ and also allow dependence on the previous $\tau$ observations instead of just $x_{i-1}$, e.g. for \eqref{4.1.1}
\begin{align*} 
x_{i} = x_{i-1} + b(t_i, x_{i-\tau},\ldots, x_{i-1}, \theta)\Delta t + A(t_i, x_{i-\tau}, \ldots, x_{i-1}, \theta)z_{i} \stepcounter{equation}\tag{\theequation}\label{4.1.1-modified}
\end{align*}
and similarly for \eqref{jump-ode}. Because of this modification the adjoint variables are given by \eqref{adjointv2}. We also considered testing another stochastic process we call a piecewise ODE
\begin{align*} 
& x_i = x_{i-1} + \beta(t_i, x_{i-\tau}, \ldots, x_{i-1}, y_i, \theta)\Delta t \\
& \beta^j(t_i, x_{i-\tau}, \ldots, x_{i-1}, y_i, \theta) = 
\begin{cases} 
 \beta_1^j(t_i, x_{i-\tau}, \ldots, x_{i-1}, y_i, \theta) & \text{ if }y_i^j = 1 \\ 
  \beta_2^j(t_i, x_{i-\tau}, \ldots, x_{i-1}, y_i, \theta) & \text{ if }y_i^j = 2 \\ 
  \quad \vdots & \\
  \end{cases} \qquad y_i^j \in \{1, \ldots, l \}
\end{align*}
which is similar to the jump ODE. 

We use the electricity load dataset \cite{electricity-dataset}, which records the electricity demands of 370 customers in 15 minute intervals over a period of 3.5 years. We only used 20 of the customers for our experiments, so $x_i \in \mathbb{R}^{20}$. We take $\tau = 12$, so the models will predict the elecitricity demand of the next 15 minutes based on the previous 12 measurements (3 hours) of demand. The model predictions can then be fed back into the model so we can predict the demand arbitrarily far into the future. The models are trained by minimizing the error over a 48 hour prediction window. 

\begin{figure}[H]
\centering 
\includegraphics[ width=\textwidth]{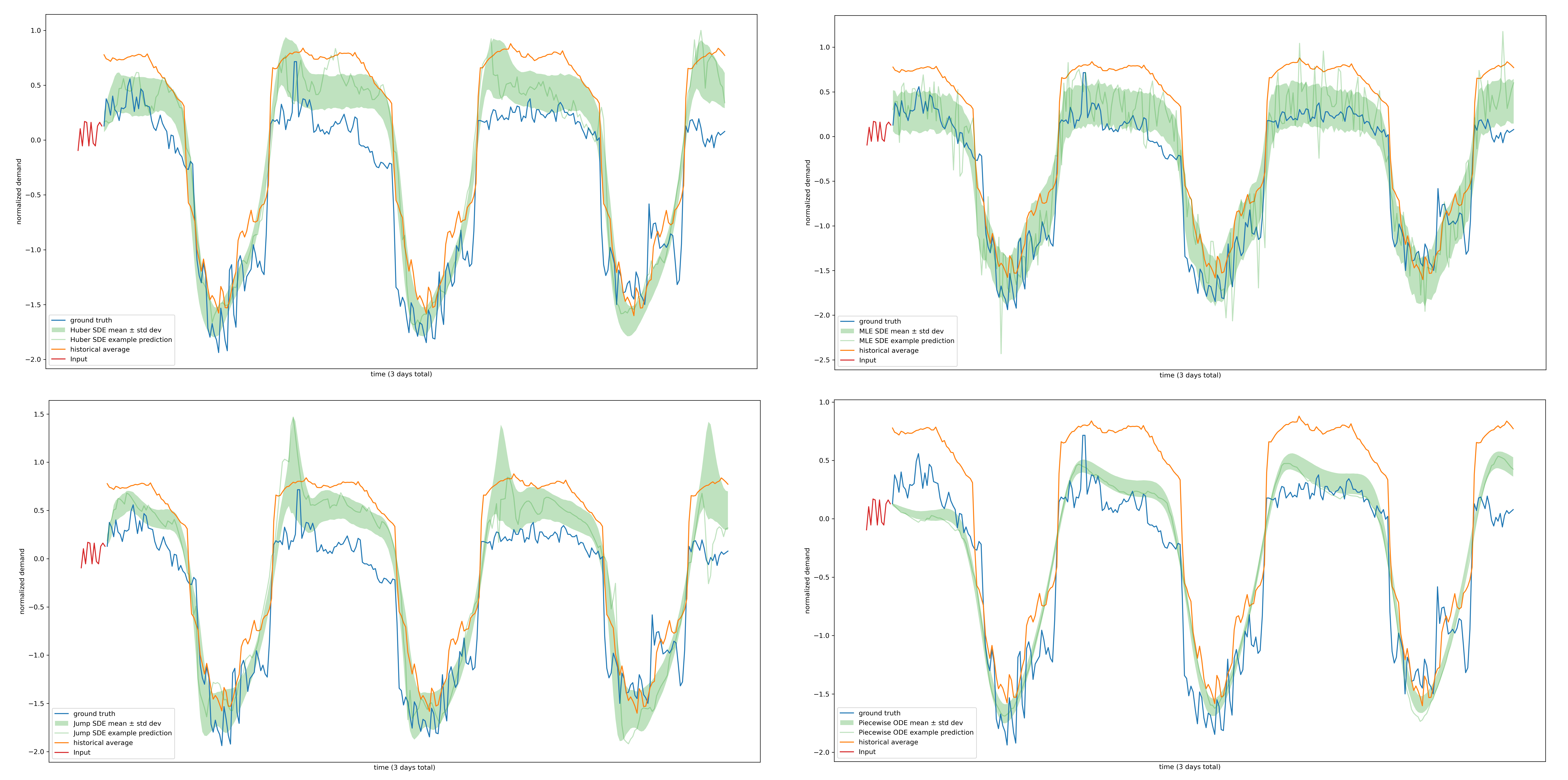}  
\caption{Predicting the next 3 days of electricity demand. Top left: Standard SDE with Huber loss. Top right: Standard SDE with MLE loss. Bottom left: Jump ODE with Huber loss. Bottom right: Piecewise ODE with Huber loss.} \label{sde-figure}
\end{figure}

The standard SDE was trained two seperate times, once with Huber loss as the objective function (Huber SDE), and also by maximizing the likelihood of the observations (MLE SDE). The jump ODE and piecewise ODE both took $l=5$ (5 possibilities in each of the 20 dimensions) and used Huber loss. All models are parametrized using deep neural networks with 3 hidden layers with a residual connection. More details are given in the appendix and the code will be made public upon publication at \url{https://github.com/ronan-keane/adjointgrad}. Figure \ref{sde-figure} shows an example prediction of 3 days for a single customer for each of the four models. Given the models only use the past 3 hours of measurements, it is probably not possible to accurately predict 3 days into the future. The point is that the models can learn long term trends as well as give reasonable short term predictions. For the Huber SDE, MLE SDE, jump ODE, and piecewise ODE, the mean squared error on the testing set is .12, .11, .14, .12 respectively, compared to a historical average which gives .23. 

More in-depth details are given in the appendix.

\section{Reinforcement Learning}
In reinforcement learning problems, an agent interacts with its environment through a sequential decision making process. In any given timestep, an agent finds itself in current environment state $s_t$ and uses its policy $\pi$ to select an action $a_t$. Based on the $(s_t, a_t)$ tuple, the environment gives a reward $r_t$ and transitions to the next state $s_{t+1}$; this process then iterates. The entire sequence of $(s_t, a_t, r_t)$ tuples is known as an episode or trajectory. The goal is to obtain a policy $\pi$ which chooses actions in a way to maximize the expected cumulative reward over the episode. Policy gradient methods directly learn a parametrization of the policy $\pi$ and are a popular method for solving reinforcement learning problems. We will see different policy gradient update rules can be understood as different choices of how to model the actions/environment.\\
In our notation, 
\begin{align*} 
x_{(i,1)} := \begin{bmatrix} s_i & a_i \end{bmatrix} \quad x_{(i,2)} := s_{i+1}
\end{align*}
so an entire trajectory containing $n$ actions and environment transitions is defined by the ordered indices \{(1, 1), (1, 2), \ldots, (n, 1), (n, 2)\} (so compared to regular indexing, $i=1 \rightarrow (1, 1)$, $2 \rightarrow (1, 2)$, $3 \rightarrow (2, 1)$ etc.). The initial state $x_{(0, 2)} := s_1$ is assumed to be the same for all episodes.

The objective is to maximize the expected value of the discounted total reward
\begin{align*} 
f(x_{(1,1)}, \ldots, x_{(n, 2)}) = \sum_{k=(1,1)}^{(n,2)} f_k(x_{k}) = \sum_{k=1}^{n} f_{(k,1)}(x_{(k,1)})  = \sum_{k=1}^n\gamma^{k-1}R(s_k, a_k)
\end{align*}
where $\gamma$ is the discount factor and the reward $R$ is assumed to be a function of the current state action pair. Policy gradient methods update the policy $\pi$ using the gradient $\partial \mathe [ f]/\partial \theta$.

\subsubsection{Reinforce }
Let us assume the actions are continuous and define the policy $\pi(a_t \, | \, s_t, \theta)$ as giving the probability density of the action $a_t$ (if the actions are discrete, then let $\pi(a_t \, | \, s_t, \theta)$ give the probability of selecting $a_t$). Similarly define the environment $env(s_{i+1} \, | \, s_i, a_i)$ as giving the probability density of transitioning to $s_{i+1}$ given the current state action pair. Within our framework the model is formulated as
\begin{align*} 
& h_{(i,1)} = (s_i, y_{(i,1)}) \qquad p_{(i,1)}(y_{(i,1)} \, | \, x_{(i-1, 2)}, \theta) = \pi(y_{(i,1)} \, | \, s_i, \theta) \\
& h_{(i, 2)} = y_{(i, 2)} \qquad p_{(i,2)}(y_{(i, 2)} \, | \, x_{(i, 1)}, \theta) = env(y_{(i,2)} \, | \, s_i, a_i)
\end{align*}
where $y_{(i,1)}$ represents $a_i$ and $y_{(i,2)}$ represents $s_{i+1}$. Since neither $h_{(i,1)}$ nor $h_{(i,2)}$ depend on $\theta$, theorem \ref{thm1-1} immediately gives
\begin{align*} 
& \hat g(\theta) = \sum_{i=1}^nf(x_{(1,1)}, \ldots, x_{(n, 2)}) \dfrac{\partial \log p_{(i,1)}(y_{(i,1)} \, | \, \cdot)}{  \partial \theta} + f(x_{(1,1)}, \ldots, x_{(n, 2)}) \dfrac{\partial \log p_{(i,2)}(y_{(i,2)} \, | \, \cdot)}{\partial \theta} \\
& \hphantom{\hat g(\theta)} = \sum_{i=1}^nf(x_{(1,1)}, \ldots, x_{(n, 2)}) \dfrac{\partial \log p_{(i,1)}(y_{(i,1)} \, | \, x_{(i-1,2)}, \theta)}{\partial \theta} \\
& \hphantom{\hat g(\theta)} = \sum_{i=1}^n \sum_{k=1}^n\gamma^{k-1}R(s_k, a_k) \dfrac{\partial \log \pi(a_i \, | \, s_i, \theta)}{\partial \theta}
\end{align*}
where the second line follows because the environment does not does not depend on $\theta$. Then applying \eqref{summable-loss} we arrive at the well known Reinforce update \cite{sutton-barto}
\begin{align*} 
& = \sum_{i=1}^n \sum_{k=i}^n\gamma^{k-1}R(s_k, a_k) \dfrac{\partial \log \pi(a_i \, | \, s_i, \theta)}{\partial \theta} \stepcounter{equation}\tag{\theequation}\label{reinforce}
\end{align*}

\subsubsection{Model based policy gradients}
If we use pathwise derivatives for the policy, we can derive policy gradients which depend on the environment model, as opposed to the model free update \eqref{reinforce}. Assume that the policy $\pi(s_i, \theta, z_{(i,1)})$ returns the action taken in state $s_i$. Similarly let $env(s_i, a_i, z_{(i,2)})$ return the next state given the current state-action pair. Here $z_{(i,1)}$, $z_{(i,2)}$ encode any randomness in the policy and environment transitions, respectively. The model is formulated as
\begin{align*} 
& h_{(i,1)} = (s_i, \pi(s_i, \theta, z_{(i,1)}) \\
& h_{(i,2)} = env(s_i, a_i, z_{(i,2)}) .
\end{align*}
The relevant partial derivatives are 
\begin{align*} 
& \dfrac{\partial h_{(i,1)}}{\partial x_{(i-1, 2)}} = \begin{bmatrix} I \\ \dfrac{\partial \pi(s_i)}{\partial s_i} \end{bmatrix} \qquad \dfrac{\partial h_{(i,1)}}{\partial \theta} = \begin{bmatrix} 0 \\ \dfrac{\partial \pi(s_i)}{\partial \theta} \end{bmatrix} \qquad \dfrac{\partial h_{(i,2)}}{\partial x_{(i,1)}} = \begin{bmatrix} \dfrac{\partial env(s_i, a_i)}{\partial s_i} & \dfrac{\partial env(s_i, a_i)}{\partial a_i} \end{bmatrix} \\
&  \dfrac{\partial h_{(i,2)}}{\partial \theta} = 0 \qquad  \dfrac{\partial f}{\partial x_{(i,1)}} = \gamma^{i-1}\begin{bmatrix} \dfrac{\partial R(s_i, a_i)}{\partial s_i} & \dfrac{\partial R(s_i, a_i)}{\partial a_i} \end{bmatrix} \qquad \dfrac{\partial f}{\partial x_{(i,2)}} = 0.
\end{align*}
Note that we assume $s_i \in \mathbb{R}^s, a_i \in \mathbb{R}^a$ so $x_{(i,1)}$, $h_{(i,1)} \in \mathbb{R}^{s+a}$ and $x_{(i,2)}, h_{(i,2)} \in \mathbb{R}^{s}$. The matrices above are interrupted as block matrices with the appropriate dimensions, 0 is a matrix of all zeros and $I$ is the identity matrix (of appropriate dimension). Recall as well that $\lambda_j$ has the dimension as $x_j$. Now plugging into theorem \ref{thm1-1} gives 
\begin{align*} 
& \hat g(\theta) = \sum_{i=1}^n \lambda_{(i,1)}^\top \dfrac{\partial h_{(i,1)}}{\partial \theta} = \sum_{i=1}^n \lambda_{(i,1)}^\top \begin{bmatrix} 0 \\ \dfrac{\partial \pi(s_i)}{\partial \theta} \end{bmatrix} \\
& \lambda_{(n,2)}^\top = 0 \qquad \ \ \ 
\lambda_{(i, 2)}^\top = \lambda_{(i+1,1)}^\top \begin{bmatrix}  I \\ \dfrac{\partial \pi(s_{i+1})}{\partial s_{i+1}} \end{bmatrix} \quad (i= n-1, \ldots, 1) \\ 
& \lambda_{(i,1)} =  \gamma^{i-1}\begin{bmatrix} \dfrac{\partial R(s_i, a_i)}{\partial s_i} & \dfrac{\partial R(s_i, a_i)}{\partial a_i} \end{bmatrix} + \lambda_{(i,2)}^\top \begin{bmatrix} \dfrac{\partial env(s_i, a_i)}{\partial s_i} & \dfrac{\partial env(s_i, a_i)}{\partial a_i} \end{bmatrix} \quad (i = n, \ldots, 1)
\end{align*}
and this is the stochastic value gradient derived in \cite{SVG}. 

The alternative choice of treating the environment, which to our knowledge has not been considered previously, is letting $env(s_{i+1} \, | \, s_i, a_i)$ return a probability density, so that 
\begin{align*} 
& h_{(i,1)} = (s_i, \pi(s_i, \theta, z_{(i,1)})) \\
& h_{(i, 2)} = y_{(i, 2)} \qquad p_{(i,2)}(y_{(i, 2)} \, | \, x_{(i, 1)} \theta) = env(y_{(i,2)} \, | \, s_i, a_i) .
\end{align*}
In this case the gradient estimator is 
\begin{align*} 
\hat g(\theta) = \sum_{i=1}^n\Big(\gamma^{i-1}\dfrac{\partial R(s_i, a_i)}{\partial a_i} + \sum_{k=i}^n\gamma^{k-1}R(s_k, a_k) \dfrac{\partial \log env(y_{(i,2)} \, | \, s_i, a_i)}{\partial a_i}  \Big)\dfrac{\partial \pi(s_i)}{\partial \theta} . \stepcounter{equation}\tag{\theequation}\label{model-based-1}
\end{align*}
\paragraph{Other formulations.}
One might wonder how the deterministic policy gradient \cite{DDPG} fits into these different formulations. Actually, that policy gradient is based on a clever rewriting of \eqref{model-based-1}. The observation is that 
\begin{align*} 
 &\mathe_{z_{(1,1)}^{(n,1)}, y_{(1,2)}^{(n,2)}}[\hat g(\theta)] \\
 & = \mathe_{z_{(1,1)}^{(n,1)}, y_{(1,2)}^{(n,2)}} \left[\sum_{i=1}^n\Big(\gamma^{i-1}\dfrac{\partial R(s_i, a_i)}{\partial a_i} + \sum_{k=i}^n\gamma^{k-1}R(s_k, a_k) \dfrac{\partial \log env(y_{(i,2)} \, | \, s_i, a_i)}{\partial a_i}  \Big)\dfrac{\partial \pi(s_i)}{\partial \theta} \right]\\
& = \mathe_{z_{(1,1)}^{(n,1)}, y_{(1,2)}^{(n,2)}}\left[ \sum_{i=1}^{n} \dfrac{\partial Q^{\pi}(s_i, a_i)}{\partial a_i} \dfrac{\partial \pi(s_i)}{\partial \theta}\right] \stepcounter{equation}\tag{\theequation}\label{ddpg-eqn}
\end{align*}
where $Q^{\pi}$ is the action value function for policy $\pi$. For the sake of completeness, the appendix shows how to derive this. 

Lastly we'll mention the alternative way of treating the reward. Instead of assuming that there is some deterministic function $R(s_i, a_i)$ which returns the reward, we assume $env(s_{i+1}, r_i \, | \, s_i, a_i)$ returns the joint probability density of receiving the reward $r_i$ and transitioning to $s_{i+1}$. Then, 
\begin{align*} 
& x_{(i,2)} = \begin{bmatrix} s_{i+1} & r_i \end{bmatrix} \\
& f(x_{(1,1), \ldots, x_{(n,2)}}) = \sum_{k=1}^n f_{(k,2)}(x_{(k,2)}) = \sum_{k=1}^n \gamma^{k-1}r_k
\end{align*}
where $h_{(i,2)} = y_{(i,2)}$. 

\section{A Piecewise ODE with Parameter Dependent Switching Condition} \label{sec-3-3}
Consider the following piecewise model
\begin{align*} 
x_{i+1} = \begin{cases} 
 x_i + \theta_1 & \text{ if  }s(x_i) < \theta_0 \\ 
  x_i + 1 & \text{ otherwise}
  \end{cases} \stepcounter{equation}\tag{\theequation}\label{piecewise-eg-1}
\end{align*}
which is a discretized ordinary differential equation (ODE) loosely based off of the famous Newell traffic flow model \cite{13}. The model has two branches, the first $x_i + \theta_1$ corresponds to the vehicle moving forward with speed $\theta_1$, and represents driving with a slow speed. The second branch $x_i + 1$ represents moving forward with the maximum speed of 1 (unitless). We only use the slow speed branch when the space headway $s(x_i)$ is less than $\theta_0$, or in other words, we use the slow speed when we are too close to the vehicle in front of us (measured as less than $\theta_0$ distance between the vehicles' bumpers). 

\begin{figure}[H] 
\centering 
\includegraphics[ width=\textwidth]{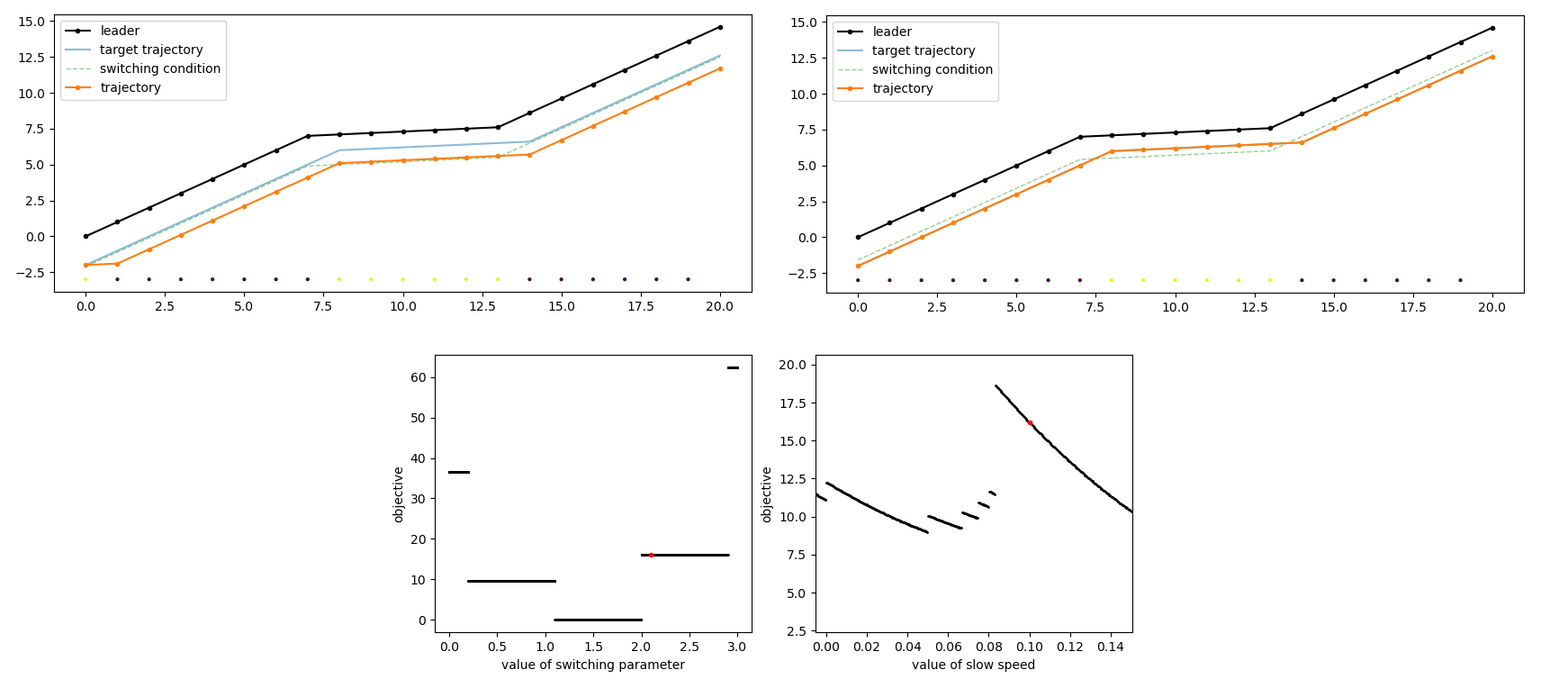}
\caption{Top: model predictions for $\theta_0 = 2.1, \theta_1 = 0.1$ (left panel) and $\theta_0 = 1.6, \theta_1 = 0.1$ (right panel). Bottom: objective values for varying values of $\theta_0$ with $\theta_1=0.1$ (left panel), objective values for varying values of $\theta_1$ with $\theta_0 = 2.1$ (right  panel).}   \label{piecewise-fig}
\end{figure}

The top two panels of figure \ref{piecewise-fig} show an example of what the model looks like for two different parameter values. The black line represents the lead vehicle which we react to. The blue line is the target for what the model should predict. The orange line is what the model actually predicts. The headway $s(x_i)$ is simply the distance between the orange and black lines at any given x coordinate. We also plot, in the semi-transparent green line, the switching condition: if we are below the green line, we use the fast speed; otherwise the speed $\theta_1$ is used. Lastly, the dots at the bottom of those panels show which branch is used, with the dark color representing the fast speed of 1. 

In this example,  any value of $\theta_0$ between roughly 1 and 2 combined with $\theta_1 = 0.1$ yields the best objective value of 0. The top left panel shows $\theta_1 = 0.1, \theta_0 = 2.1$. We see that the switching condition is not in the right place, as the first timestep is in the slow branch instead of the fast one. The top right panel shows an example of the best fit corresponding to $\theta_0 = 1.6, \theta_1 = 0.1$. Notice that in this panel, if we perturbed $\theta_0$, i.e. we changed the location of the green line, none of the timesteps would change their branch. That is the meaning of why we say that the switching condition has no sensitivity. The bottom panels show the objective values as $\theta_0$ (left) and $\theta_1$ (right) are changed. Note that objective value is always completely flat with respect to $\theta_0$, because the gradient is always 0 (or undefined, when the discontinuities occur). For $\theta_1$ (right panel), we see a striated pattern where there are regions where the objective is continuous, with discontinuities that occur whenever a timestep switches from the fast to slow branch or vice versa. This type of striated pattern is typical of the optimization landscape of piecewise models. 

Seeing as how \eqref{piecewise-eg-1} leads to a objective which is not even continuous, it's clear that giving any sort of convergence gaurentee here will be problematic. Luckily our methodology as discussed in section \ref{sec-1-5} gives a straight forward way to modify the model as follows
\begin{align*} 
& x_{i+1} =  \begin{cases} 
 x_i + \theta_1 & \text{ if } y_i = 1 \\ 
  x_i + 1 & \text{ if } y_i = 2
  \end{cases} \\
  &\mathbb{P}(y_i = 1 \, | \, x_i, z_i) = \mathbb{P}(\theta_2 z_i + s(x_i ) < \theta_0) \qquad \text{where }z_i \sim \text{Normal}(0,1) . \stepcounter{equation}\tag{\theequation}\label{piecewise-eg-2}
\end{align*}
We take the deterministic model \eqref{piecewise-eg-1} and introduce a new random variable $y_i$ which encodes whether to use the slow or fast branch. We have a great deal of flexibility in terms of how the conditional distribution of $y_i$ is defined, with really the only restriction being that we need a way to evaluate the conditional probability that $y_i$ is equal to 1. In this case we take $\mathbb{P}(y_i = 1 \, | \, \cdot) = \mathbb{P}(\theta_2 z_i + s(x_i ) < \theta_0)$, where $z_i$ has a standard normal distribution. Or in other words, the distribution of $y_i$ is parameterized using a normal distribution. There is a need to introduce a new parameter $\theta_2$ which encodes the standard deviation of the normal. 

To evaluate the model, at any given timestep we first evaluate the headway $s(x_i)$ just like in the deterministic case. But instead of simply comparing $s(x_i)$ and $\theta_0$, we first draw from a standard normal (i.e. sample a $z_i$). Then if $\theta_2 z_i + s(x_i ) < \theta_0$, we use the slow branch (i.e. $y_i = 1$), but if $\theta_2 z_i + s(x_i ) > \theta_0$ we use the fast branch ($y_i = 2$). In other words, it is as if we have added new randomness from $\theta_2 z_i$ to the old switching condition of $s(x_i)$. Then, we can calculate an unbiased gradient sample by computing the probability $\mathbb{P}(y_i)$ which will contribute score functions to the gradient according to theorem 1. 

\begin{figure}[H] 
\centering 
\includegraphics[ width=\textwidth]{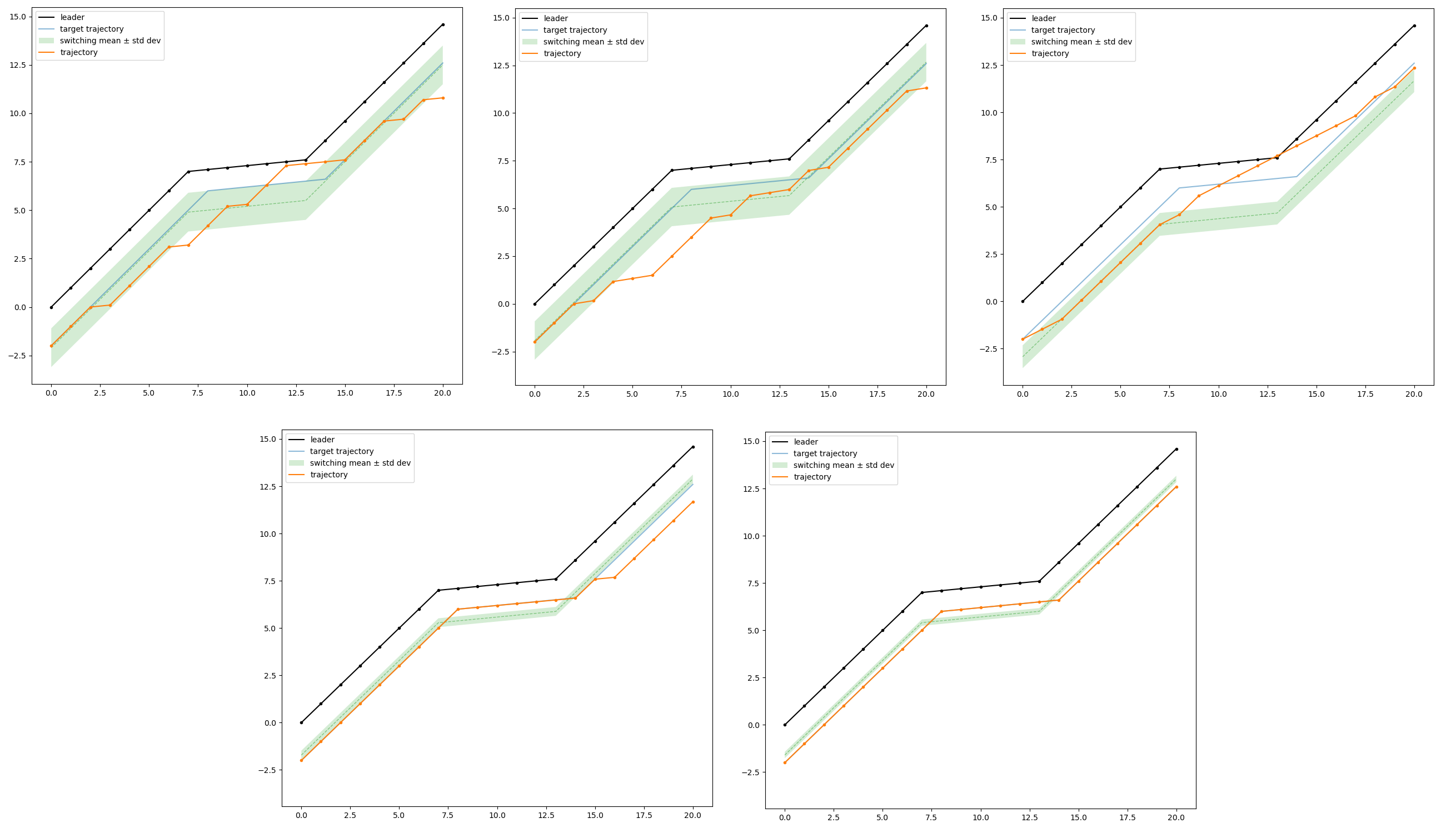} 
\caption{Using SGD to solve the piecewise model. From top left to bottom right, the plots are of iteration 1, 30, 70, 100, and 200. After roughly 200 iterations we recover the deterministic solution with an expected error close to 0.}  \label{piecewise-fig2}
\end{figure}

Figure \ref{piecewise-fig2} shows the results of using SGD to minimize the squared error between the model output and target (the blue line). The learning rate is set to 0.01, the batch size is 1, and no variance reduction is used. As before, the green line shows the headway such that $s(x_i) = \theta_0$, but now we also have the shaded green region which shows $\pm$ the standard deviation of $\theta_2$. If we are far below the shaded region, then the probability that $\theta_2 z + s(x_i ) < \theta_0$ is very small, so we will very nearly always be in the fast branch (and vice versa if we are far above the shaded region). 

In the beginning, the model output looks quite different than the original model due to the new random behavior. But as we iterate SGD, we will naturally learn the probability distribution, and eventually the standard deviation of the normal will decrease as that will lead to a lower error. After 200 iterations we have learned the correct value for $\theta_1$, and the standard deviation is small enough that the random model is almost always the same as the purely deterministic model. 

This example shows the power and the exciting possibilities for our new methodology. Not only are we now able to learn switching conditions, but also we know from the analysis in section \ref{analysis} that the expected objective will be completely smooth, so there are no more discontinuities in the problem. We also don't lose any expresstivity of the model and can easily recover the deterministic behavior if it is beneficial to the loss. 

\subsection{Implications and Future Directions}
There are many possible applications for this new methodology. It opens the door to moving beyond purely continuous models, and into the space of piecewise continuous models. This means we can work with models that use if statements. We can use models that can specialize into multiple seperate behaviors, having distinct branches and switching conditions that control which branch to use. The novelty is that this piecewise model can all be learned end to end, with an unbiased gradient estimator, without the need for any heuristics. 

To give an example of the possible applications, imagine we are developing a self driving car. There are an immense number of corner cases in driving, such as the situation when a parked vehicle (possibly with occupants still inside) has pulled over on the side of the road and is partially blocking the lane. With a purely continuous model, there is a single function which maps from the observations into the driving action. A piecewise continuous model could, for example, look at the observations, and calculate the probability that it should use the normal driving model, or a special driving model which addresses the partial lane blockage. In other words, the piecewise continuous model can learn a special behavior for dealing with such a corner case, and the conditions under which to use that behavior. Our hypothesis is that in this example, the piecewise continuous model will be easier to learn, as it more naturally fits how the car should behave: there are different, distinct behaviors corresponding to the specific driving condition. This is opposed to more monolithic, dense paradigm where there is a single continuous function which handles every single corner case as well as normal driving all at once.

Moreover, it seems there is something inherently human about piecewise functions. The idea of making distinct decisions, the idea of approaching a problem in different ways. In other words, it seems that humans thinking involves \textit{if} statements, and jumps in logic, as opposed to just being a single continuous process. Neuroscientists such as Jeff Hawkins, in works such as \cite{thousand-brains}, have extolled sparsity as a key feature to human and animal intelligence. This sparse aspect is something which is strikingly absent from current deep neural networks. By their nature, piecewise functions are inherently sparse.

\section{The Three Pitfalls of Differentiable Models} \label{sec-3-4}
Based on the gradient estimator developed in chapter 1 and the analysis in chapter 2, we have identified three problems in model formulations which we refer to as the three pitfalls. Each pitfall jeopardizes the use of gradient-based optimization. This is because they either cause discontinuites (either in the objective or the gradient), or because they may simply lead to a gradient which does not exist or is always 0. The problems are
\begin{enumerate}[label=\textbf{\arabic*})]
\item A model with multiple branches, that is a model of the form 
\begin{align*} 
h_i(\cdot) = \begin{cases} 
 h_i^1(\cdot) & \text{ if }\gamma(\cdot) > 0 \\ 
  h_i^2(\cdot) & \text{ otherwise} \\ 
  \end{cases}\, .
\end{align*}
This type of model will cause discontinuities when switching between the branches $h_i^1$ and $h_i^2$. To be more precise, based on the analysis of sections \ref{analysis} and \ref{sec-2-5}, we know that for the objective to be continuous, we require that $h_i^1  = h_i^2$ when $\gamma = 0$. Likewise, for the gradient to be continuous, the partial derivatives should be equal when $\gamma = 0$. 

It is especially problematic when the switching condition $\gamma$ depends on the parameters. Because $\gamma$ does not contribute to the gradient, there is simply no way to learn those parameters.
\item A model involving discrete variables. Discrete variables simply don't have derivatives.
\item Parameters which control how long (i.e. how many timesteps) some aspect of the model should last. Because time is discrete, this is essentially a discrete decision. Therefore we have discontinuities when the number of timesteps changes, and depending on the formulation, there may be no gradient. 
\end{enumerate}
Luckily, all of these problems can be fixed by introducing stochasticity to the model. The solutions to pitfalls 1 and 2 were already introduced in section \ref{sec-1-5}. Pitfall 3 can be addresed in a similar way, by introducing a distribution over the number of timesteps. 
\subsection{Additional Examples}
\begin{figure}[H] 
\centering 
\includegraphics[ width=\textwidth]{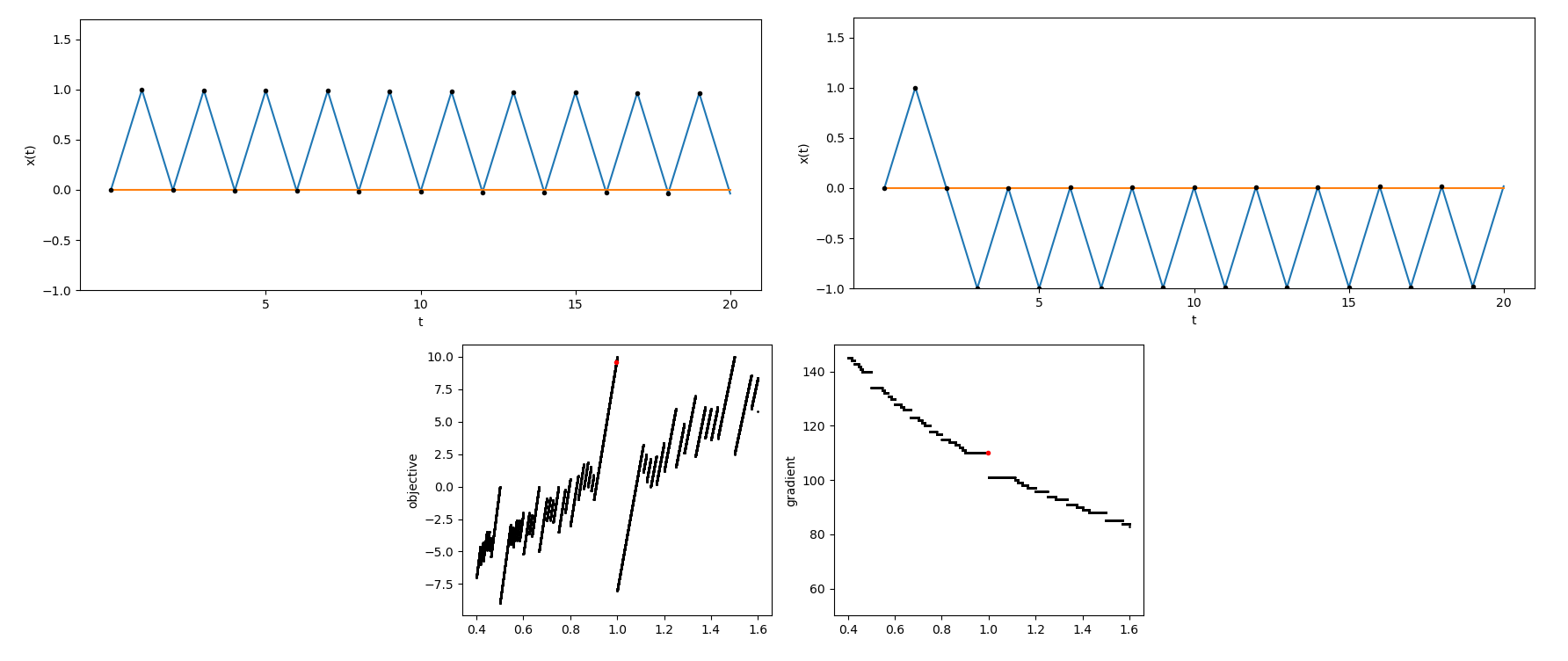} 
\caption{Plots of \eqref{switcheg1}. The more timesteps flip branches at once, the larger the resulting discontinuity. }  \label{switchfig1}
\end{figure}

We now further illustrate the pitfalls with the following examples. First consider
\begin{align*} 
x_{i+1} = \begin{cases} 
 x_i + \theta & \text{ if } x_i \leq 0 \\ 
  x_i -1 & \text{ if } x_i > 0\\ 
  \end{cases} \stepcounter{equation}\tag{\theequation}\label{switcheg1}
\end{align*}
This is a piecewise model with a single parameter that affects branch 1. Note the switching condition does not depend on the parameters. Shown in figure \ref{switchfig1}, the black dots represent each $x_i$, with blue lines connecting them. The orange line is the switching condition (and is simply the line $x=0$), so if we are above the orange line we go down by 1, and otherwise we go up by $\theta$. The interesting thing about this example is that there are certain parameter values where multiple timesteps switch branches at the same time. For example, in the top left panel, we show the model for $\theta = 0.998$. We can see that all of the black dots corresponding to indices of $2, 4, 6, \ldots$ will hit the switching condition at $\theta=1$. The resulting model output for $\theta = 1.002$ is shown in the top right panel. The observation is that the size of the discontinuities correspond to how many timesteps switch branches. The bottom panels show the objective and gradient plots, and the red dot corresponds to the values at $\theta=0.998$. The discontinuity at $\theta=1$, when 9 out of 19 timesteps switch branches, corresponds to the largest discontinuity observed in the plots. In this example, the objective function used is simply $\sum_i x_i$.

\begin{figure}[H] 
\centering 
\includegraphics[ width=\textwidth]{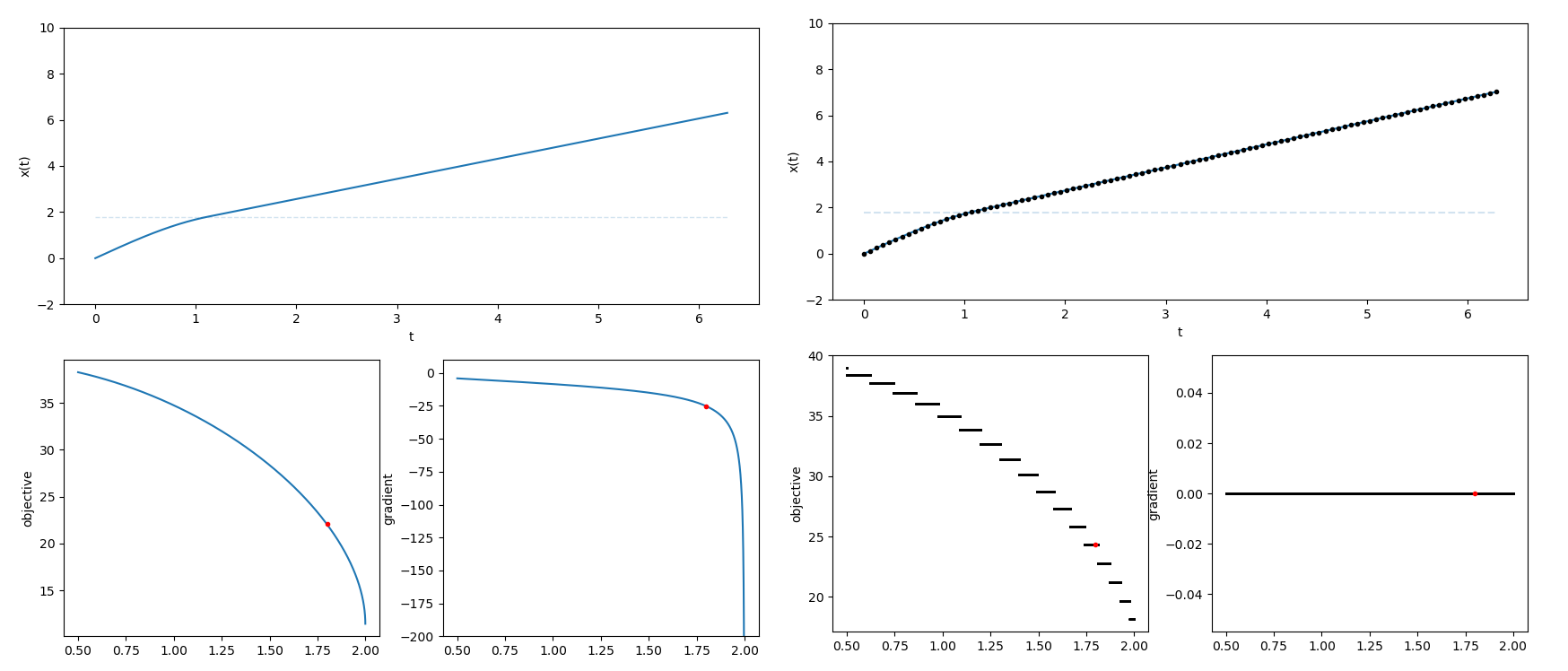}
\caption{In the left panels, ouput (top) and objective/gradient (bottom) of \eqref{switcheg2}. In the right panels, the discrete time version.}   \label{switchfig2}
\end{figure}

We now consider the piecewise ODE 
\begin{align*} 
\ddot x = \begin{cases} 
 -x & x < \theta \\ 
  0 & x \geq \theta \\ 
  \end{cases}
 \\ x(0) = 0 \ \ \ \dot x(0) = 2 \stepcounter{equation}\tag{\theequation}\label{switcheg2}
\end{align*}
Recall that if we just had the ODE $\ddot x = -x$, this would just be a simple harmonic oscillator, with $x(t)$ as just a sin wave. In \eqref{switcheg2}, when we hit the switching condition $x(t) = \theta$, we freeze the acceleration, and keep moving with whatever speed we had. The point of this example is that we can actually get a closed form solution
\begin{align*} 
x(t) = \begin{cases} 
 2 \sin(t) & t < t^* \\ 
  p + 2\cos(t^*) ( t - t^* ) & t \geq t^* \\
  \end{cases} \\
  \text{where } t^* = \arcsin(\theta/2) .
\end{align*}
In the left side panels of \ref{switchfig2}, we plot the continuous time system (again using the dummy objective of $\int_t x(t)$). The point is, when we solve for the switching time in a closed form in this way, we get a gradient with respect to the switching condition. That's because the switching time $t^* = \text{arcsin}(\theta/2)$ depends on the parameters. If we consider the discrete time version of \eqref{switcheg2}, which is shown in the right panels, we see end up in the situation where the switching condition does not contribute to the gradient. Again, this is due to the fact that perturbing $\theta$ in discrete time can not lead to a similar perturbation of the output: there is either no change, or a jump corresponding to one of the timesteps switching branches.
It is possible to get a gradient with respect to the switching condition, but this would require a perturbation to the switching parameter to perturb the time that we switch branches. In \eqref{switcheg2}, we can solve for that time in closed form. For a typical problem, solved in discrete time, it is not at all clear how to do this. For a possible solution, it is clear that we would first of all need a variable timestep length (with possibly very small timesteps in the vicinity of the switch between branches). We can then solve for $t^*$, the time that we switch branches, numerically. This would require inverting both the switching condition and the model. If this inversion is done numerically, to get a gradient we would have to propagate the gradient through the numerical inversion routine. But this then leads to new problems, as numerical solvers themselves use if statements. At the very least, any numerical solver will have an if statement for their termination conditions. In other words, we could go through this extra computation to get a gradient, but there will still be discontinuities. Because of the difficulty of implementing such a scheme, its undesirable effects with regards to the timestepping length and extra computational cost, and the fact that discontinuities will still remain in the problem, we consider this approach to be undesirable. 

Lastly, we present an example using the actual traffic model
\begin{align*} 
x_{i+1} = \min\{ x_i + \frac{1}{\theta_0}(s(x_i) + r(x_i, \theta_3)-\theta_1), x_i + \theta_2\} \stepcounter{equation}\tag{\theequation}\label{switcheg3}
\end{align*}
which is the Newell model with relaxation. The relaxation $r(x_i, \theta_3)$ is a special state that is part of lane changing. The strength of the relaxation decreases linearally at a rate of $1/\theta_3$ over $\lfloor \theta_3 \rfloor$ timesteps (more detail is given in chapter 6). The relaxation parameter $\theta_3$ therefore violates the third pitfall since $\theta_3$ controls the length of time of relaxation. However we still have sensitivity with respect to $\theta_3$ because it also controls the rate as $1/\theta_3$. The two branches (inside of the minimum) correspond to the regular driving state, and the free flow state respectively. Note that because of the $\min$, the two branches are equal when switching; thus it follows that the objective will be continuous. The gradient however, will not be continuous because the partial derivatives are not equal.

\begin{figure}[H] 
\centering 
\includegraphics[ width=\textwidth]{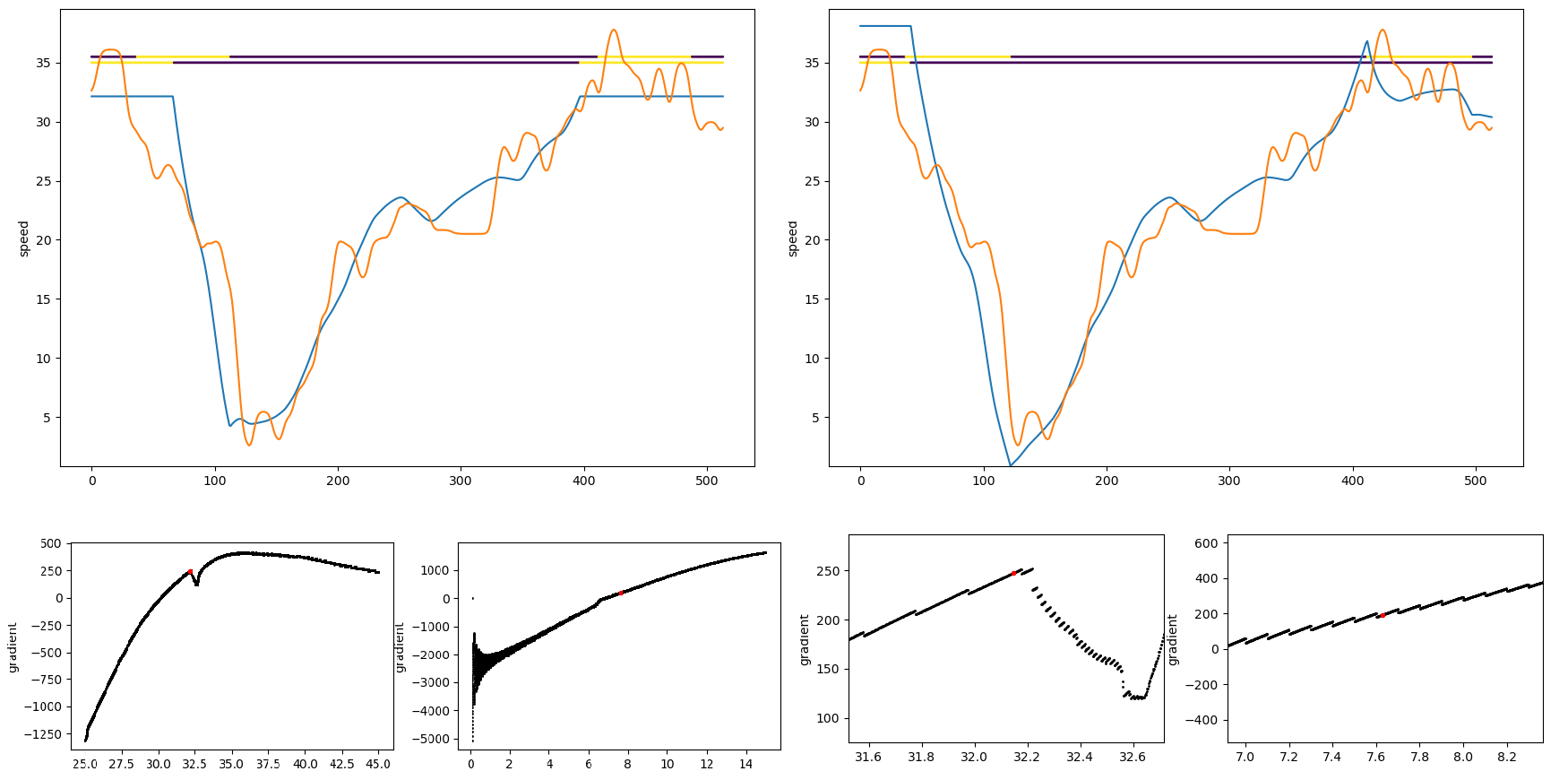}
\caption{Top panels: model outputs (blue curve) versus measurements (orange curve) for two different parameter values. Bottom panels: gradient plots with respect to free flow speed (far left, zoomed in on right), relaxation time (left, zoomed in on far right) with other parameters held fixed.}   \label{switchfig3}
\end{figure}

In figure \ref{switchfig3} the top two panels show an example of the model output, selected for a random vehicle in the NGSim \cite{6} dataset. The speed time series are plotted in blue for the model predictions, and the orange curve is the true measured speed. The two lines of color at the top of the figures represent the current branch of the model. The top line is yellow if the relaxation is active; the bottom line is yellow if the model is at the free flow speed. The bottom panels show the gradient component for the free flow speed $\theta_2$ (far left panel and right panel) and the relaxation time $\theta_3$ (left and far right panels) as the other parameters are held fixed. We can see discontinuities in the zoomed in plots (right and far right) which occur when timesteps switch their branches. 

In this example the discontinuites are less pronounced because of the large number of timesteps total (around 500). Unlike the example \eqref{switcheg1}, typically only 1 timestep will flip at a time. We can see that in the plot for the gradient with respect to $\theta_2$ (bottom far left), it appears relatively smooth with a kink around $\theta_2 = 32$. In the zoomed in plot (bottom right) we see that the ``smooth'' graph actually has persistent discontinuities. The kink is actually due to a large number of discontinuities that occur in close proximity to each other. In that region, from around $\theta_2 = 32$ to $\theta_2 = 32.6$ , the model branches change abnormally fast and irregularly. 

\begin{figure}[H] 
\centering 
\includegraphics[ width=\textwidth]{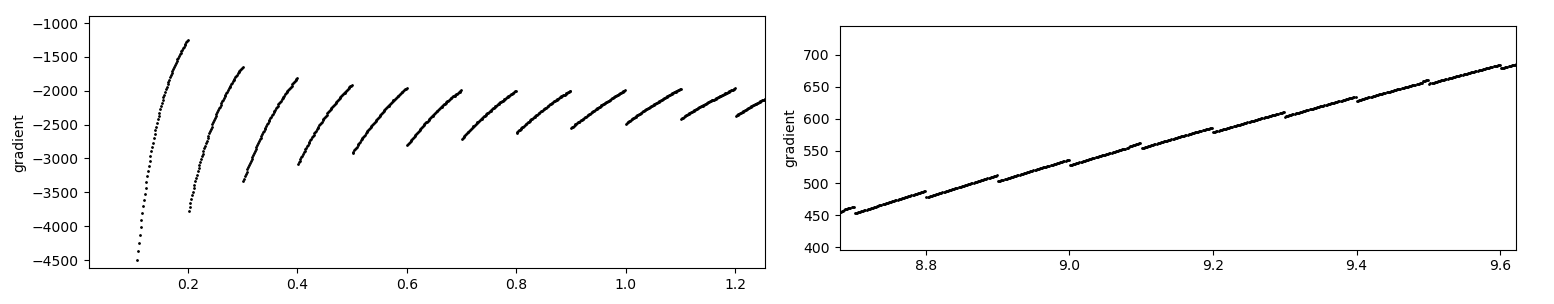} 
\caption{The size of the discontinuities can be predicted based on how many timesteps switch branches.}  \label{jump-size}
\end{figure}

We can observe that the size of the discontinuity, compared to the previous gradient value, is very strongly correlated with
\begin{align*} 
\dfrac{\text{The number of flipped timesteps}}{\text{Total number of timesteps which contribute sensitivity}} .
\end{align*}
For example, for the relaxation parameter $\theta_3$, every timestep with relaxation contributes to the gradient. The total number of such timesteps is equal to $2 \lfloor10 \theta_3\rfloor$ because there are two lane changes in this vehicle trajectory, and each lane change has $\lfloor10 \theta_3\rfloor$ timesteps of relaxation corresponding to $\lfloor \theta_3\rfloor$ seconds of relaxation (the timestep is fixed at 0.1). In figure \ref{jump-size} we can see that in the left panel, when we go from 1 timestep to 2 timesteps (which occurs at $\theta_3 = .2$), the value of the gradient component changes from about $-1500$ to $-3500$. This 133\% change is in line with the prediction of 100\% change based on having 1 flipped timestep and 1 timestep total. At $\theta_3 = .3$ we would expect roughly a 50\% discontinuity (1 flipped timestep over 2 total) and observe about 100\% (roughly -1600 to -3200). As the number of total timesteps increases, the discontinuity size decreases. In the right panel, when there are around 90 timesteps total ($\theta_3 = 9$) the discontinuities are small, as predicted. 
Interestingly, we have observed that when the discontinuites are small and only present in the gradient, they don't seem to have negative effects on gradient-based optimization. Specifically, when solving problems such as \eqref{switcheg3}, where the objective is continuous, and the majority of parameter space has small discontinuities of a regular size/frequency, a standard nonlinear optimization algorithm such as l-BFGS-b gave good results (e.g. in the experiments in section \ref{parametric-results}). The reason for this is presently unknown, but we note that there are certain gradient-based algorithms such as the gradient bundle algorithm \cite{gradient-bundle} which only requires that the objective is locally Lipschitz continuous. This lesser assumption (as opposed to the more typical, much stronger assumption of global Lipschitz continuity of the gradient) is actually satisfied in \eqref{switcheg3}.

Overall we have seen that based on the model formulation and our analysis of chapter 2, we will know ahead of time what issues to expect. We have documented and analyzed the cause of discontinuities in the objective and/or gradient, and also the problem of a lack of sensitivity which occurs in models with parameter dependent switching conditions or discrete variables. Our gradient estimator gives new methodology (section \ref{sec-1-5}) for fixing all of these problems, and this methodology was validated in the examples in \ref{sec-3-1} and \ref{sec-3-3}. 

\chapter{Variance Reduction}
We again consider the problem formulation
\begin{align*} 
\underset{\theta}{\min} & \quad  \mathe_{z_1^n,  y_1^n} [ f(x_1, \ldots, x_n) ] \\ 
\text{s.t.} & \quad x_{i} = h_i(x_{i-1}, y_{i}, z_{i}) \  \quad i = 1, \ldots, n 
\end{align*}
similar to earlier chapters. But now notice that $h_i$ has no dependence on the parameters $\theta$, so we have $x_i = h_i(x_{i-1}, y_i, z_i)$. It follows that the gradient estimator does not have any contributions from pathwise derivatives or the chain rule, so the gradient estimator consists solely of score functions
\begin{align*} 
\sum_{i=1}^n f(x_1, \ldots, x_n) \dfrac{\partial \log p_i(y_i \, | \, x_{i-1}, z_i)}{\partial \theta} . \stepcounter{equation}\tag{\theequation}\label{grad1sf}
\end{align*}
Again, the distribution of any $y_i$ explicitly depends on $\theta$, and we denote the conditional probability density of $y_i$ as $p_i(\cdot \, | \, x_{i-1}, z_i, \theta)$.

\section{Introduction to Baselines}
It is well known that score functions tend to have significantly higher variance compared to pathwise derivatives \cite{variance-of-derivative-estimators, variance-reduction-reparameterization-trick}. This large variance can potentially negatively impact the convergence of SGD. Moreover, it is not always possible to avoid SF: in chapter 3 we saw various situations where we must use SF. Baselines are one method for reducing the variance of score function estimators.

We will first consider the case where the gradient estimator consists only of score functions. In future work we will extend the results to apply to the gradient estimator of \ref{thm1-1}. In this section, we have some gradient estimator $\hat g_{\rm sf\_only}$ defined by
\begin{align*} 
\hat g_{\rm sf\_only} \, = \, \sum_{i=1}^n f(x_1, \ldots, x_n) \dfrac{\partial \log p_i(y_i \, | \, x_{i-1}, z_i)}{\partial \theta} . \stepcounter{equation}\tag{\theequation}\label{SFonlygrad}
\end{align*}
Baselines are a type of control variate with the form 
\begin{align*} 
\sum_{i=1}^{n}\beta_i(\phi) \dfrac{\partial \log p_i(y_i \,|\, x_{i-1}, z_{i})}{\partial \theta}. \stepcounter{equation}\tag{\theequation}\label{cv}
\end{align*}
The scalar functions $\{ \beta_i(\phi) : i = 1 ,\ldots, n\}$ are called baseline functions, and will be of key interest. We shall see shortly that the control variate \eqref{cv} has 0 expectation (provided the $\beta_i$ are defined appropriately), so when applied to the gradient estimator $\hat g_{\rm sf\_only}$ we get the new gradient estimator 
\begin{align*} 
\hat g \, & := \, \hat g_{\rm sf\_only} - \sum_{i=1}^{n}\beta_i(\phi) \dfrac{\partial \log p_i(y_i \,|\, x_{i-1}, z_{i})}{\partial \theta} + \mathe_{z_1^n, y_1^n} \bigg[ \sum_{i=1}^{n}\beta_i(\phi) \dfrac{\partial \log p_i(y_i \,|\, x_{i-1}, z_{i})}{\partial \theta} \bigg] \\
& = \, \sum_{i=1}^n \big( f(x_1, \ldots, x_n) - \beta_i(\phi)\big) \dfrac{\partial \log p_i(y_i \, | \, x_{i-1}, z_i)}{\partial \theta}  \stepcounter{equation}\tag{\theequation}\label{hatg}
\end{align*}
where we note that $\mathe[\hat g ] \, = \, \mathe[\hat g_{\rm sf\_only}]$. If the baselines functions  $\{ \beta_i\}$ are chosen suitably, then \eqref{cv} can be strongly correlated with $\hat g_{\rm sf\_only}$, leading to $\hat g$ having a lower variance than $\hat g_{\rm sf\_only}$. 

The parameters $\phi$ define the baseline functions. The idea of baselines is to learn $\phi$ to minimize the variance of $\hat g$ while simultaneously optimizing $\theta$. An appealing aspect of this idea is that because the baseline functions are learned from scratch, there is no need for any problem specific knowledge on how to construct a good control variate. 

\subsection{Useful Properties} \label{properties}
First, let us justify Eq. \eqref{hatg}.
\begin{assumption} \label{assumption1}
For any index $i$ and any values of $y_i$, $z_i$, $x_{i-1}$, $\theta$,
\begin{itemize}[itemsep=0pt, topsep=4pt]
\item[a)] let $p_i(y_i \, | \, z_i, x_{i-1}, \theta)$ be differentiable with respect to $\theta$
\item[b)] for some positive constant $\epsilon_{\theta}$,
\begin{align*} 
\underset{y_i}{\int} \underset{\theta^*: \norm{\theta^* - \theta} < \epsilon_{\theta} }{\sup} \norm{\dfrac{\partial p_i(y_i \, | \, z_i, x_{i-1}, \theta^*)}{\partial \theta^*}} dy_i < \infty\
\end{align*}
\item[c)] the support of $y_i$ does not depend on $\theta$.
\end{itemize}
\end{assumption}
Throughout the chapter, $\norm{\cdot}$ denotes the 2 norm (Euclidean norm). 
\begin{lemma} \label{lemma1}
Let assumption \ref{assumption1} hold. Then 
\begin{align*} 
\underset{y_i}{\int}\dfrac{\partial p_i(y_i \, | \, z_i, x_{i-1})}{\partial \theta}dy_i = 0. \stepcounter{equation}\tag{\theequation}\label{lemma1eqn}
\end{align*} 
\end{lemma}
Proofs will be given in the appendix. Assumption \ref{assumption1} gives sufficient conditions to be able to interchange the derivative and integral in \eqref{lemma1eqn}. The assumption \ref{assumption1}a) of differentiability everywhere can easily be relaxed to only require differentiability almost everywhere, and continuity everywhere (this would allow $p_i$ to be parameterized by, for example, a neural network with ReLU activation).

\begin{theorem}\label{thm1}
Let $\beta_i$ depend on any $z_1, \ldots, z_{n}$, $y_1, \ldots, y_{i-1}$, $x_1, \ldots, x_{i-1}$. Then if assumption \ref{assumption1} holds, 
\begin{align*} 
\mathe_{z_1^n, y_1^n}\left[\sum_{i=1}^{n}\beta_i(\phi) \dfrac{\partial \log p_i(y_i \,|\, x_{i-1}, z_{i})}{\partial \theta} \right] = 0.
\end{align*}
\end{theorem}
This well known result formally justifies \eqref{hatg}. Importantly, we also have that $\hat g$ is unbiased when $\beta_i$ depends on any of $x_1, \ldots, x_{i-1}$, $z_1, \ldots, z_{n}$, $y_1, \ldots, y_{i-1}$. 

For the simplest possible baseline, all $\beta_1, \ldots, \beta_n$ would take on some constant value $\beta$, and $\phi$ would be a single parameter, which gives the value of $\beta$. In the opposite extreme, $\beta_i$ would be a function (with parameters $\phi_i$) which maps from the partial sample path $x_1, \ldots, x_{i-1}$, $z_1, \ldots, z_{n}$, $y_1, \ldots, y_{i-1}$ to a constant (and $\phi$ is the concatenation of $\phi_1, \ldots, \phi_n$). A detailed discussion of how to choose/parametrize the $\beta_i$ is defered until the next section (\ref{fitting-baselines}), but it should be mentioned that a popular choice is to learn a single function which maps from $x_{i-1}$ to the value of $\beta_i$. \\

\noindent One important situation, which is frequently encountered in practice, is the objective function being \textbf{summable}.
\begin{definition}
We say the objective function is summable if it is of the form
\begin{align*} 
f(x_1, \ldots, x_n) := \sum_{k=1}^n \hat{f}_k(x_k) \stepcounter{equation}\tag{\theequation}\label{summable-defn}.
\end{align*}
for some scalar functions $\{ \hat{f}_k : k = 1, \ldots, n\}$.
\end{definition}
\begin{proposition}
For a summable objective,
\begin{align*} 
\mathe_{z_1^n, y_1^n}\bigg[ \sum_{i=1}^{n}f(x_1, \ldots, x_n) \dfrac{\partial \log p(y_i \, | \, x_{i-1}, z_i)}{\partial \theta} \bigg] = \mathe_{z_1^n, y_1^n}\bigg[ \sum_{i=1}^{n} \sum_{k=i}^n\hat{f}_k(x_k) \dfrac{\partial \log p(y_i \, | \, x_{i-1}, z_i)}{\partial \theta} \bigg] .
\end{align*}
\end{proposition}

\begin{definition}
We say the $y_i$ are conditionally independent if for any $i = 1, \ldots, n$,
\begin{align*} 
p(y_i \,|\, x_{i-1}, z_{i}, \theta) = p(y_i \, | \, z_{i}, \theta).
\end{align*}
\end{definition}
\begin{proposition}\label{proposition2}
Let the $y_i$ be conditionally independent. If any $\beta_i$ depends on $z_1, \ldots, z_n$, $y_1, \ldots, y_{i-1}, y_{i+1}, \ldots, y_n$, $x_1, \ldots, x_{i-1}, x_{i+1}, \ldots, x_n$, then $\mathe[\hat g] = \mathe[\hat{g}_{\rm sf\_only}]$.
\end{proposition}
In other words, if the $y_i$ are conditionally independent, then $\beta_i$ can depend on anything from the current sample path except for $y_i$ and $x_i$. An example of a proposed baseline which uses this property is \cite{Vimco}. That paper proposes $\beta_i = f(x_1, \ldots, x_{i-1}, \hat{x}_i, x_{i+1}, \ldots, x_n)$ as a baseline function, where $\hat{x}_i$ is defined as the average of $x_1, \ldots, x_{i-1}, x_{i+1}, \ldots, x_n$.

\subsection{Notation}
Henceforth we use the following modified notation for the gradient estimator 
\begin{align*} 
\hat g \, := \, \sum_{i=1}^n \big( f_i - \beta_i(\xi_i)\big) \dfrac{\partial \log p_i(y_i \, | \, x_{i-1}, z_i)}{\partial \theta}\, . \stepcounter{equation}\tag{\theequation}\label{hatg-actual}
\end{align*}
The notation $\xi_i$ denotes the partial sample path information that $\beta_i$ depends on; so in the default case, per theorem \ref{thm1}, $\xi_i$ is some function of $\{z_1, \ldots, z_n, y_1, \ldots, y_{i-1}\}$. We do not include $x_1, \ldots, x_{i-1}$ in the definition of $\xi_i$ since for any index $j$, $x_j$ can be determined from $z_1, \ldots z_j, y_1, \ldots y_j$. We always consider $\beta_i$ to be a function of both $\phi$ and $\xi_i$, but this dependence may be suppressed. The notation $f_i$ may mean either $f(x_1, \ldots, x_n)$ or $\sum_{k=i}^n \hat{f}_k(x_k)$, corresponding to the general case and summable objective case, respectively. Assuming that the objective is summable, it should be beneficial to use $f_i = \sum_{k=i}^n \hat{f}_k(x_k)$, as the $i$\textsuperscript{th} score function will not include any extra randomness due to $x_1, \ldots, x_{i-1}$. In section \ref{biased} we will discuss additional choices for $f_i$ which result in biased gradient estimators with lower variance.

\section{How to Fit Baselines} \label{fitting-baselines}
In each optimization step, we sample a trajectory $(z_1, y_1, x_1), (z_2, y_2, x_2), \ldots, (z_n, y_n, x_n)$ and compute $\hat g$ as well as $\nabla \phi$, where $\nabla \phi$ denotes the gradient used to update the baseline functions. The parameters are then updated as $\theta \leftarrow \theta - \alpha_{\theta} \hat g$ and $\phi \leftarrow \phi - \alpha_{\phi}\nabla \phi$ where $\alpha_{\theta}$, $\alpha_{\phi}$ are the learning rates. \\

\noindent Arguably the most popular formulation for baselines \cite{nvil, credit-assignment-techniques, Williams, sutton-88, a3c} 
is to update the $\{\beta_i\}$ according to 
\begin{align*} 
\underset{\phi}{\min} & \quad  \sum_{i=1}^n\Big( \mathe_{z_1^n, y_1^n}[\, f_i \, | \, \xi_i\, ] -  \beta_i(\xi_i)  \Big)^2 .\stepcounter{equation}\tag{\theequation}\label{RL-update}
\end{align*}
Then $\nabla \phi$ can be defined as a monte carlo sample of the gradient of \eqref{RL-update}:
\begin{align*} 
\nabla \phi = \sum_{i=1}^n  -2\big(f_i  -  \beta_i(\xi_i) \big)\dfrac{\partial \beta_i}{\partial \phi}. \stepcounter{equation}\tag{\theequation}\label{nablaphi-1}
\end{align*}
This formulation has deep connections to reinforcement learning, where baselines were first studied \cite{Williams, sutton-88}. In that context, the baselines functions are defined by $\beta_i(\xi_i) := \hat{V}(x_{i-1})$, where $\hat V$ is an approximation of the state-value function and $x_{i-1}$ represents the state in timestep $i$. Moving to a more general problem formulation, we can have some function $\hat V$ with parameters $\phi$, where $\hat V$ learns to approximate $\mathe_{z_1^n, y_1^n}[\, f_i \, | \, \xi_i\, ]$ and  $\beta_i(\xi_i) := \hat V(\xi_i)$.

In terms of variance reduction, the formulation \eqref{RL-update} and associated update \eqref{nablaphi-1} is a heuristic \cite{greensmith-2004,  weaver-tao}. Despite this, we do expect that defining the baseline functions in this way will reduce variance, as $\mathe[f_i \, | \, \xi_i]\,\frac{\partial \log p_i(y_i \, | \, x_{i-1}, z_i)}{\partial \theta}$ and $f_i \, \frac{\partial \log p_i(y_i \, | \, x_{i-1}, z_i)}{\partial \theta}$ should be correlated. 


\subsection{Definition of Optimal Baselines}\label{optimal}
The estimator $\hat g$ is a function both of the model parameters $\theta$ and the baseline function parameters $\phi$. A baseline can be called optimal if, for some given $\theta$, the parameters $\phi$ minimize the variance of $\hat g$. We define the variance of a vector-valued function as
\begin{align*} 
\text{Var}\big[ \hat g \big] \,:=\, \mathe \big[ \norm{\hat g}^2 \big] - \norm{\mathe \big[ \hat g \big]}^2 \, .\stepcounter{equation}\tag{\theequation}\label{var-defn}
\end{align*} 
Note that this defines the variance of a vector as the sum of the variance of its components (i.e. the trace of the vector's covariance matrix). 

An important observation is that if the baseline has expectation 0, then $\norm{\mathe [ \hat g ]}^2$ is constant with respect to $\phi$. This will be true even when $\hat g_{\rm sf\_only}$ is biased, assuming that the bias does not depend on $\phi$. It follows that minimizing the variance of $\hat g$ with respect to $\phi$ is equivalent to minimizing the second moment $\mathe [ \norm{\hat g}^2 ]$. Considering the aforementioned references on SGD \cite{bottou-nocedal}, \cite{nemirovski}, and \cite{almost-sure-SGD} all give convergence results which directly depend on the second moment, we have a strong theoretical justification for choosing baselines which minimize the second moment of $\hat g$. 

\subsection{Direct Variance Minimization}
One simple way to update the baseline functions is to compute a monte carlo sample of $\frac{\partial}{\partial \phi} \mathe [ \norm{\hat g}^2 ]$, giving
\begin{align*} 
\nabla \phi \, & = \, -2 \mathe_{z_1^n, y_1^n}\bigg[ \sum_{i=1}^n \hat g \dfrac{\partial \log p_i(y_i \, | \, x_{i-1}, z_i)}{\partial \theta}^\top \dfrac{\partial \beta_i}{\partial \phi} \bigg]  \stepcounter{equation}\tag{\theequation}\label{direct-baseline}
\end{align*}
To our knowledge, this approach was first proposed in \cite{relax}. Although this directly minimizes the second moment, we do not regard this approach as optimal because it gives no information on the actual values of the optimal baseline functions.

\subsection{The Optimal Baseline}
\begin{lemma}\label{lemma2}
For any two indices $i$, $j$ such that $i \neq j$, 
\begin{align*} 
\mathe_{z_1^n, y_1^n }\bigg[ \beta_i(\xi_i) \dfrac{\partial \log p_i(y_i \, | \, x_{i-1}, z_i )}{\partial \theta}\dfrac{\partial \log p_j(y_j \, | \, x_{j-1}, z_j)}{\partial \theta}^\top \ \bigg| \ \xi_j \bigg] \, = \, 0.
\end{align*}
\end{lemma}
\noindent Setting $\frac{\partial}{\partial \phi}\text{Var}(\hat g) = 0$ (i.e. setting \eqref{direct-baseline} = 0), we have 
\begin{align*} 
0 \, &  = \, \mathe_{z_1^n, y_1^n}\bigg[\sum_{j=1}^n\sum_{i=1}^n \big( f_i - \beta_i(\xi_i)\big) \dfrac{\partial \log p_i(y_i \, | \, x_{i-1}, z_i  )}{\partial \theta}\dfrac{\partial \log p_j(y_j \, | \, x_{j-1}, z_j )}{\partial \theta}^\top \dfrac{\partial \beta_j}{\partial \phi}  \bigg] \\
& = \,  \sum_{j=1}^n \mathe_{\xi_j}\bigg[\dfrac{\partial \beta_j}{\partial \phi} \mathe_{z_1^n, y_1^{n}} \bigg[ \sum_{i=1}^n \big( f_i - \beta_i(\xi_i)\big) \dfrac{\partial \log p_i(y_i \, | \, x_{i-1}, z_i )}{\partial \theta}\dfrac{\partial \log p_j(y_j \, | \, x_{j-1}, z_j )}{\partial \theta}^\top \ \bigg| \ \xi_j \bigg] \,\bigg] \\
& = \,  \sum_{j=1}^n \mathe_{\xi_j}\bigg[\dfrac{\partial \beta_j}{\partial \phi} \mathe_{z_1^n, y_1^{n}} \bigg[ \Big( \hat g_{\rm sf\_only} - \beta_j(\xi_j) \dfrac{\partial \log p_j(y_j \, | \, x_{j-1}, z_j )}{\partial \theta} \Big) \dfrac{\partial \log p_j(y_j \, | \, x_{j-1}, z_j )}{\partial \theta}^\top \ \bigg| \ \xi_j \bigg] \,\bigg]
\end{align*}
where the last line uses lemma \ref{lemma2}. Now, the optimal baseline is defined by the scalar values $\{\beta_j(\xi_j) : j = 1, \ldots, n\}$ that the baseline functions should take for any given sample path. Realizing that $\frac{\partial \beta_j}{\partial \phi}$ merely encodes the direction of steepest ascent of $\beta_j$ with respect to $\phi$, it follows that $\frac{\partial \beta_j}{\partial \phi}$ has no bearing on the actual values of the optimal baseline. Thus, for any $j$ and any partial sample path $\xi_j$, the optimal value for $\beta_j(\xi_j)$ should satisfy 
\begin{align*}
& \ \mathe_{z_1^n, y_1^{n}} \bigg[ \Big( \hat g_{\rm sf\_only} - \beta_j(\xi_j) \dfrac{\partial \log p_j(y_j \, | \, x_{j-1}, z_j )}{\partial \theta} \Big) \dfrac{\partial \log p_j(y_j \, | \, x_{j-1}, z_j )}{\partial \theta}^\top \ \bigg| \ \xi_j \bigg] = 0 \\
\Rightarrow  \ & \ \beta_j(\xi_j) = \dfrac{\mathe_{z_1^n, y_1^{n}} \bigg[\hat g_{\rm sf\_only}  \dfrac{\partial \log p_j(y_j \, | \, x_{j-1}, z_j )}{\partial \theta}^\top \ \bigg| \ \xi_j \, \bigg]}{\mathe_{z_1^n, y_1^{n}}\bigg[ \dfrac{\partial \log p_j(y_j \, | \, x_{j-1}, z_j )}{\partial \theta} \dfrac{\partial \log p_j(y_j \, | \, x_{j-1}, z_j )}{\partial \theta}^\top \ \bigg| \ \xi_j \, \bigg]}  \stepcounter{equation}\tag{\theequation}\label{optimal-baseline}
\end{align*}
Eq. \eqref{optimal-baseline} gives a closed form for the optimal value of $\beta_j$, for an arbitrary choice of $\xi_j$. However, it may be difficult to estimate these expectations if the number of possible $\xi_j$ is large. For example, if $\xi_j := x_{j-1}$ and $x_{j-1} \in \mathbb{R}$, then there are an infinite number of integrals we would have to estimate in order to calculate $\beta_j(\xi_j)$ for every possible value of $\xi_j$. \\

\noindent We will consider two different formulations for computing optimal baselines. In the first formulation, we assume that the number of distinct possibilities of $\xi_j$ is finite, with $c$ possible values total (over all indices $j$). It is further assumed that it is computationally feasible, in terms of memory usage, to use $2c$ parameters to define the baselines. We shall have $c$ baselines, one for each possible value of $\xi_j$. At index $j$, we observe a sample of $\xi_j$, which we denote as $\bar{\xi}$. The baseline function is given as $\beta_j(\bar{\xi}) := \phi_{(k,1)}/\phi_{(k,2)}$ where $k$ is the index of $\phi$ which corresponds to $\bar{\xi}$. The $\phi_{(k,1)}$ and $\phi_{(k,2)}$ are defined by
\begin{align*} 
& \underset{\phi_{(k,1)}}{\argmin}  \quad  \Big(\mathe_{z_1^n, y_1^n }\bigg[ \hat g_{\rm sf\_only} \dfrac{\partial \log p_j(y_j \, | \,  x_{j-1}, z_j )}{\partial \theta}^\top \ \bigg| \ \bar{\xi} \, \bigg] - \phi_{(k,1)} \Big)^2  \stepcounter{equation}\tag{\theequation}\label{c-constant-formulation}\\
& \underset{\phi_{(k,2)}}{\argmin} \quad  \Big(\mathe_{z_1^n, y_1^n}\bigg[ \dfrac{\partial \log p_j(y_j \, | \,  x_{j-1}, z_j )}{\partial \theta} \dfrac{\partial \log p_j(y_j \, | \,  x_{j-1}, z_j )}{\partial \theta}^\top \ \bigg| \ \bar{\xi} \, \bigg] - \phi_{(k,2)} \Big)^2 \, .  \stepcounter{equation}\tag{\theequation}\label{c-constant-formulation1}
\end{align*}
The meaning of $j$ in \eqref{c-constant-formulation}-\eqref{c-constant-formulation1} is the index that $\bar \xi$ was observed, which may be a random variable depending on the definition of the $\{\xi_j: j = 1, \ldots, n\}$. 

We call the formulation \eqref{c-constant-formulation}-\eqref{c-constant-formulation1} the \textbf{c-optimal baseline} because it directly estimates the optimal values of $c$ distinct baseline functions. To update $\phi$, we use the monte carlo gradient
\begin{align*}
& \nabla \phi_{(k,1)} = -2 \Big(\hat g_{\rm sf\_only}\dfrac{\partial \log p_j(y_j \, | \, x_{j-1}, z_j )}{\partial \theta}^\top - \phi_{(k,1)} \Big) \stepcounter{equation}\tag{\theequation}\label{per-index}\\
& \nabla \phi_{(k,2)} = -2 \Big(\dfrac{\partial \log p_j(y_j \, | \, x_{j-1}, z_j )}{\partial \theta} \dfrac{\partial \log p_j(y_j \, | \, x_{j-1}, z_j )}{\partial \theta}^\top - \phi_{(k,2)} \Big) \, . \stepcounter{equation}\tag{\theequation}\label{per-index1}
\end{align*}
Eqs. \eqref{per-index}-\eqref{per-index1} only give the gradient contribution from a single observation $\bar \xi$, so $\nabla \phi$ would consist of all the contributions from each of the observed values of $\xi_1, \ldots, \xi_n$. A possible limitation is that in a single monte carlo sample of $f(x_1, \ldots, x_n)$, we can update at most $2n$ out of the total $2c$ parameters, which may be problematic if $c >> n$.  \\

\noindent In our second proposed formulation, we assume that all possible $\xi_j$ have the same support for all $j$. We shall have two function approximators, $\beta_{\rm top}(\xi_j)$ and $\beta_{\rm bottom}(\xi_j)$, with corresponding parameters $\phi_{\rm top}$ and $\phi_{\rm bottom}$. For some $\bar \xi$, \, $\beta_{\rm top}(\bar \xi)$ should approximate 
\begin{align*} 
\mathe_{z_1^n, y_1^{n}}[\hat g_{\rm sf\_only}  \dfrac{\partial \log p_j(y_j \, | \, x_{j-1}, z_j )}{\partial \theta}^\top \, | \, \bar \xi \, ]
\end{align*} 
and $\beta_{\rm bottom}(\bar \xi)$ should approximate
\begin{align*} 
\mathe_{z_1^n, y_1^{n}}[\frac{\partial \log p_j(y_j \, | \, x_{j-1}, z_j )}{\partial \theta}\frac{\partial \log p_j(y_j \, | \, x_{j-1}, z_j )}{\partial \theta}^\top \, | \, \bar \xi \, ]
\end{align*} 
where $j$ is the index that $\bar \xi$ was observed. The baseline function is then defined as $\beta_j(\bar \xi) := \beta_{\rm top}(\bar \xi)/\beta_{\rm bottom}(\bar \xi)$. Similar to \eqref{c-constant-formulation}-\eqref{per-index1}, we can update the parameters using the monte carlo gradient
\begin{align*}
& \nabla \phi_{\rm top} =  -2\Big(\hat g_{\rm sf\_only}\dfrac{\partial \log p_j(y_j \, | \, x_{j-1}, z_j )}{\partial \theta}^\top - \beta_{\rm top}(\bar \xi) \Big) \dfrac{\partial \beta_{\rm top}(\bar \xi)}{\partial \phi}\stepcounter{equation}\tag{\theequation}\label{state-dependent} \\
& \nabla \phi_{\rm bottom} = -2\Big(\dfrac{\partial \log p_j(y_j \, | \, x_{j-1}, z_j )}{\partial \theta}\dfrac{\partial \log p_j(y_j \, | \, x_{j-1}, z_j )}{\partial \theta}^\top - \beta_{\rm bottom}(\bar \xi) \Big) \dfrac{\partial \beta_{\rm bottom}(\bar \xi)}{\partial \phi} \stepcounter{equation}\tag{\theequation}\label{state-dependent1}
\end{align*}
 We refer to this formulation as the \textbf{function approximation optimal baseline}.

\subsection{Per-Parameter Baselines}
For a per-parameter baseline, the baseline function is defined as a vector of length $m$, where $m$ is the dimension of $\theta$. The gradient estimator is then given as 
\begin{align*} 
\hat g := \sum_{i=1}^n\Big(f_i \cdot \mathbf{1} - \beta_i \Big) \odot \dfrac{\partial \log p_i(y_i \, | \, x_{i-1}, z_i )}{\partial \theta} \stepcounter{equation}\tag{\theequation}\label{per-parameter-defn}
\end{align*}
where $\odot$ denotes element-wise multiplication and $\mathbf{1}$ is a vector of all ones. Thus, each component of the baseline function affects only the corresponding component of the gradient. To our knowledge, the idea of a per-parameter baseline was first proposed in \cite{peters-2008}.

It is simple to extend the optimal baseline \eqref{optimal-baseline} to the optimal per-parameter baseline:
\begin{align*} 
\beta_{(j, l)}(\xi_j) = \dfrac{\mathe_{z_1^n, y_1^{n}} \bigg[ \big(\hat g_{\rm sf\_only}\big)_l  \dfrac{\partial \log p_j(y_j \, | \, x_{j-1}, z_j )}{\partial \theta_l}\ \bigg| \ \xi_j \, \bigg]}{\mathe_{z_1^n, y_1^{n}}\bigg[ \Big(\dfrac{\partial \log p_j(y_j \, | \, x_{j-1}, z_j )}{\partial \theta_l}\Big)^2 \ \bigg| \ \xi_j \, \bigg]} 
\end{align*}
where $\big( \hat g_{\rm sf\_only}\big)_l$ is the $l$ component of $\hat g_{\rm sf\_only}$, $\theta_l$ is the $l$ component of $\theta$, and $\beta_{(j, l)}$ is the $l$ component of the per-parameter baseline function $\beta_j(\xi_j)$. 

Similar to the c-optimal baseline, we can define the \textbf{c-optimal per-parameter baseline} by
\begin{align*} 
\beta_{j}(\bar \xi  ) := [\phi_{(k, 1, 1)}/\phi_{(k,1,2)}, \, \phi_{(k, 2, 1)}/\phi_{(k,2,2)}, \ldots, \phi_{(k, m, 1)}/\phi_{(k,m,2)}]
\end{align*}
where $k$ is the index of $\phi$ which parameterizes the baseline function for $\bar \xi $. Just like the c-optimal baseline, the per-parameter version estimates $c$ distinct baselines. The contribution to $\nabla \phi$ from $\bar \xi$ is given by the gradient
\begin{align*} 
\nabla \phi_{(k, l, 1)} = -2\Big(\big(\hat g_{\rm sf\_only}\big)_l \dfrac{\partial \log p_j(y_j \, | \, x_{j-1}, z_j)}{\partial \theta_l} - \phi_{(k,l,1)}\Big) \qquad l = 1 \ldots, m  \stepcounter{equation}\tag{\theequation}\label{per-parameter}\\
\nabla \phi_{(k, l, 2)} = -2\Big(\Big(\dfrac{\partial \log p_j(y_j \, | \, x_{j-1}, z_j)}{\partial \theta_l}\Big)^2 - \phi_{(k, l, 2)}\Big) \qquad l = 1 \ldots, m\, . \stepcounter{equation}\tag{\theequation}\label{per-parameter1}
\end{align*}
Updating the c-optimal per-parameter baseline requires approximately the same number of operations as updating the c-optimal baseline, but the per-parameter version has a memory cost which is higher by a factor of $m$ (assuming $c$ is the same in both cases).

\subsection{Other Optimal Baselines}
In \cite{peters-2008} and \cite{beyond-variance}, an alternate derivation of the optimal baseline was presented which yielded a formuli for optimal baselines specifically in the case of model free reinforcement learning. Those works proposed the constant optimal baseline
\begin{align*} 
\beta = \dfrac{\mathe_{z_1^n, y_1^n}\bigg[f(x_1, \ldots, x_n)\norm{\dfrac{\partial \log p_j(y_j \, | \, x_{j-1}, z_j )}{\partial \theta}}  \bigg] }{\mathe_{z_1^n, y_1^n}\bigg[\norm{\dfrac{\partial \log p_j(y_j \, | \, x_{j-1}, z_j )}{\partial \theta}}  \bigg]}
\end{align*}
and \cite{beyond-variance} extended this with an optimal state dependent baseline
\begin{align*} 
\beta(s) = \dfrac{\mathe_{z_1^n, y_1^n}\bigg[Q^{\pi}(s, a)\norm{\dfrac{\partial \log p_j(y_j \, | \, x_{j-1}, z_j )}{\partial \theta}}  \bigg] }{\mathe_{z_1^n, y_1^n}\bigg[\norm{\dfrac{\partial \log p_j(y_j \, | \, x_{j-1}, z_j )}{\partial \theta}}  \bigg]} \stepcounter{equation}\tag{\theequation}\label{wrong-optimal-baseline}
\end{align*}
where $Q^{\pi}(s, a)$ is the action-value function under policy $\pi$. However, we will see in section \ref{sec-4-4} that these results actually give baselines with a larger variance compared to the true optimal baselines which we have derived. We also note it is not possible to perform an unbiased update to the baseline functions using \eqref{wrong-optimal-baseline} because we do not have access to the true value of $Q^{\pi}(s, a)$ except in special problems. 

\section{Biased Gradient Estimators for Summable Objectives}\label{biased}
\subsection{Estimating the Expected Objective}\label{bootstrapping}
\begin{proposition} \label{prop3}
For a summable objective, 
\begin{align*} 
& \mathe_{z_1^n, y_1^n} \bigg[ \sum_{k=i}^n\hat f_k(x_k) \dfrac{\partial \log p_i(y_i \, | \, x_{i-1}, z_i)}{\partial \theta} \bigg]  \\
 = & \mathe_{z_1^n, y_1^i}\bigg[ \bigg(\hat f_i(x_i) + \mathe_{y_{i+1}^n} \bigg[ \sum_{k=i+1}^n \hat f_k(x_k) \ \bigg| \ z_1^n, y_1^i \, \bigg] \bigg)\dfrac{\partial \log p_i(y_i \, | \, x_{i-1}, z_i)}{\partial \theta} \, \bigg] \, . \stepcounter{equation}\tag{\theequation}\label{prop3-eqn}
\end{align*}
\end{proposition} 
Eq. \eqref{prop3-eqn} would be useful if we knew the value of $\mathe_{y_{i+1}^n}\big[ \sum_{k=i+1}^n \hat f_k(x_k) \, | \,z_1^n, y_1^i \big]$. In that case, we could take $f_i = \hat f_i + \mathe_{y_{i+1}^n}\big[ \sum_{k=i+1}^n \hat f_k(x_k) \, | \,z_1^n, y_1^i \big]$ instead of $f_i = \sum_{k=i}^n \hat f_k(x_k)$, because the former should result in lower variance. In the baseline formulation \eqref{RL-update} in section \ref{fitting-baselines}, we saw one possible way to define the baseline functions is $\beta_{i}(\xi_{i}) := \hat{V}(\xi_i)$, where $\hat V$ has parameters $\phi$ which are chosen to minimize
\begin{align*} 
\underset{\phi}{\min} & \quad  \sum_{i=1}^n\Big( \mathe_{z_1^n, y_1^n}[\, \sum_{k=i}^n\hat f_k(x_k) \, | \, \xi_i\, ] -  \hat V(\xi_i)  \Big)^2 \, . \stepcounter{equation}\tag{\theequation}\label{RL-update1}
\end{align*}
and $\xi_i$ is some subset of $\{z_1, \ldots z_n, y_1, \ldots, y_{i-1}\}$. Thus, $\hat V(\xi_{i+1})$ is an approximation to exactly the expectation we need to make \eqref{prop3-eqn} useful. We can then take
\begin{align*} 
f_i = \hat f_i(x_i) + \hat V(\xi_{i+1}) \stepcounter{equation}\tag{\theequation}\label{objective-estimate}
\end{align*}
as a biased, but lower variance alternative to $f_i = \sum_{k=i}^n \hat f_k(x_k)$. Clearly the bias will decrease to $0$ in the limit that $\hat V$ is a perfect approximation. This idea of a biased estimator was first developed in work such as [].

\subsection{Discounting} 
Discounting defines a scalar hyperparameter $\gamma \in [0, 1]$, and then modifies $f_i$ as follows
\begin{align*} 
& f_i = \sum_{k=i}^n \hat f_k(x_k)\hphantom{\Rightarrow  f_i x}  \quad \Rightarrow \quad  f_i = \sum_{k=i}^n \gamma^{k-i}\hat f_k(x_k)  \stepcounter{equation}\tag{\theequation}\label{discounting}\\
& f_i = \hat f_i(x_i) + \hat V(\xi_{i+1})   \quad \Rightarrow  \quad f_i = \hat f_i(x_i) + \gamma \hat V(\xi_{i+1}) . \stepcounter{equation}\tag{\theequation}\label{discounting1}
\end{align*}
If $\gamma = 1$ then $f_i$ is unchanged. For $\gamma < 1$ there will be `less randomness' in $f_i$ because of the weighting $\gamma^{k-i}$, so terms with an index $k$ sufficiently larger than $i$ are effectively ignored (note as well that in \eqref{discounting1} it is assumed that $\hat V$ learns to approximate the \textit{discounted} expected objective). It follows that lower values of $\gamma$ introduce more bias, but also give more variance reduction. Typical choices for $\gamma$ are fairly close to $1$, for example in the range $[0.95, 0.995]$. 

\subsection{GAE}
The Generalized Advantage Estimator (GAE) \cite{gae} is an extension of the estimator \eqref{objective-estimate}. First, realize that we can define the family of discounted, biased estimators 
\begin{align*} 
& F_i^{(1)} = \hat f_i(x_i) + \gamma \hat V(\xi_{i+1}) \\
& F_i^{(2)} = \hat f_i(x_i) + \gamma \hat f_{i+1}(x_{i+1})+ \gamma^2 \hat V(\xi_{i+2}) \\
& F_i^{(3)} = \hat f_i(x_i) + \gamma \hat f_{i+1}(x_{i+1})+ \gamma^2 \hat f_{i+2}(x_{i+2}) + \gamma^3 \hat V(\xi_{i+3})\\ 
& \hphantom{F_i^{(3)} \ \, } \vdots \\
& F_i^{(j)} =\sum_{k=i}^{n}\gamma^{k-i}\hat f_{k}(x_{k}) \qquad  \text{any } j >= n-i+1 
\end{align*} 
where $F_i^{(2)}, F_i^{(3)}, \ldots,$ can all be justified similarly to how proposition \ref{prop3} justifies $F_i^{(1)}$. The GAE is based on taking an exponential average of this family of estimators:
\begin{align*} 
 GAE_i := & (1 - \kappa) \sum_{k=1}^{\infty}F_i^{(k)}\kappa^{k-1} \\
 = & (1- \kappa) \Big(\hat f_i(x_i)[1 + \kappa + \kappa^2 + \ldots ] + \gamma \hat f_{i+1}(x_{i+1})[\kappa + \kappa^2 + \kappa^3 +\ldots ] + \\
 & \hphantom{ (1- \kappa) \Big(\hat f_i(x_i)[}\gamma^2 \hat f_{i+2}(x_{i+2})[\kappa^2 + \kappa^3 + \ldots ] + \ldots \\
 & \hphantom{(1- \kappa) \Big(}+ \gamma \hat V(\xi_{i+1}) + \gamma^2 \kappa \hat V(\xi_{i+2}) + \ldots + \gamma^{n-i}\kappa^{n-i-1}\hat V(\xi_n)  \Big) \\
 = & \sum_{k=i}^n \big(\gamma \kappa \big)^{k-i}\,\hat f_{k}(x_k) + \sum_{k=i}^{n-1} \big(\gamma \kappa\big)^{k-i}(1 - \kappa)\gamma \, \hat V(\xi_{k+1}) \stepcounter{equation}\tag{\theequation}\label{gae}
\end{align*}
The hyperparameter $\kappa \in [0, 1]$, which defines the weight of the exponential average, also controls the bias-variance trade-off, with low values of $\kappa$ corresponding to higher bias but lower variance. We have the special values $\kappa=0$ and $\kappa=1$, which correspond to $GAE_i = \hat f_i(x_i) + \gamma \hat V(\xi_{i+1})$ and $GAE_i = \sum_{k=i}^{n}\gamma^{k-i}\hat f_{k}(x_{k})$ respectively. Values of $\kappa \in (0, 1)$ can be understood as interpolating between these two extremes. 

Note that the derivation presented here has some differences compared to the original GAE in \cite{gae}. First, we considered the case where $n < \infty$, whereas the original paper only considers $n= \infty$. Second, we do not include the baseline term $-\hat V(\xi_i)$ in \eqref{gae}. For this reason, \eqref{gae} should be interpreted as an estimate of the return, and \textit{not} an estimate of the advantage. Other than those changes, \eqref{gae} is simply a rearranged form of the original GAE.

\section{Two Examples} \label{sec-4-4}
\subsection{Unfair Coin Flips}
In this game, the player flips a coin twice. They are paid \$1 for flipping two tails; \$2 for flipping two heads; or \$4 for flipping one head and one tail. The player provides their own (not necessarily fair) coin, so the question is, with what probability should the player's coin show heads to maximize their return? It's simple to show that the best probability is $3/5$, but suppose we don't know that, and wanted to solve this problem numerically. We have two parameters, $[\theta_1, \theta_2]$ which define the logits for flipping heads/tails, i.e. 
\begin{align*} 
p(\text{tails}) = \dfrac{e^{\theta_1}}{e^{\theta_1} + e^{\theta_2}} \qquad p(\text{heads}) = \dfrac{e^{\theta_2}}{e^{\theta_1} + e^{\theta_2}} \, .
\end{align*}
Then the gradient estimator for this problem is given by 
\begin{align*} 
\big(f - \beta_1\big)\dfrac{\partial \log p(y_1)}{\partial \theta} + \big(f - \beta_2(y_1)\big) \dfrac{\partial \log p(y_2)}{\partial \theta}\stepcounter{equation}\tag{\theequation}\label{coinflip1}
\end{align*}
where $f$ is the return (1, 2, or 4) and $y_1, y_2$ represent the outcome of the first and second coin flip. The first baseline $\beta_1$ is a constant by necessity, as there is nothing to condition on; $\beta_2$ can depend on $y_1$, possibly giving it two different values, but we could also choose $\beta_2$ to be some constant. We can calculate the variance of \eqref{coinflip1} in closed form. For example, if $[\theta_1, \theta_2] = [1,1]$, $\beta_1 = 0$, $\beta_2 = 0$, we have that the variance is $38/16 = 2.375$. 

Using the traditional baseline formulation \eqref{RL-update}, we have the baseline values
\begin{align*} 
\beta_1 = \mathe_{y_1, y_2}[f] \qquad \beta_2(y_1) = \mathe_{y_2}[f \, | \, y_1] \, . \stepcounter{equation}\tag{\theequation}\label{regular-baseline}
\end{align*}
If $[\theta_1, \theta_2] = [1,1]$, then $\beta_1 = 2.75$, $\beta_2(\text{tails}) = 2.5$, $\beta_2(\text{heads}) = 3$. The optimal baselines can be calculated as
\begin{align*} 
& \beta_1 =  \mathe_{y_1, y_2}\bigg[ f \Big(\dfrac{\partial \log p(y_1)}{\partial \theta} + \dfrac{\partial \log p(y_2)}{\partial \theta}\Big)\dfrac{\partial \log p(y_1)}{\partial \theta}^\top \bigg] \, \Big/ \, \mathe_{y_1}\bigg[ \dfrac{\partial \log p(y_1)}{\partial \theta}\dfrac{\partial \log p(y_1)}{\partial \theta}^\top \bigg] \\
& \beta_2(y_1) =  \mathe_{y_2}\bigg[ f \Big(\dfrac{\partial \log p(y_1)}{\partial \theta} + \dfrac{\partial \log p(y_2)}{\partial \theta}\Big)\dfrac{\partial \log p(y_2)}{\partial \theta}^\top \ \bigg| \ y_1 \, \bigg] \, \Big/ \, \mathe_{y_2}\bigg[ \dfrac{\partial \log p(y_2)}{\partial \theta}\dfrac{\partial \log p(y_2)}{\partial \theta}^\top \bigg] 
\end{align*}
which give the values $\beta_1 = 1.5$,  $\beta_2(\text{tails}) = 1$, $\beta_2(\text{heads}) = 2$ when $[\theta_1, \theta_2] = [1,1]$. As there are three distinct baseline values, we call this the c-optimal baseline with $c=3$. 

For the c-optimal baseline with $c=1$, we view $\xi_1 = \xi_2 = \{\}$ (as an empty set). Then we have just a single baseline value $\beta = \beta_1 = \beta_2$, which can be calculated as
\begin{align*} 
\beta = \mathe_{y_1, y_2}\bigg[ f \Big(\dfrac{\partial \log p(y_1)}{\partial \theta} + \dfrac{\partial \log p(y_2)}{\partial \theta}\Big)\dfrac{\partial \log p(y_j)}{\partial \theta}^\top \bigg] \, \Big/ \, \mathe_{y_1, y_2}\bigg[ \dfrac{\partial \log p(y_j)}{\partial \theta}\dfrac{\partial \log p(y_j)}{\partial \theta}^\top \bigg]
\end{align*}
where the index $j$ is viewed as a random variable which has an equal chance of being $1$ or $2$ (because half the time when we `observe' $\{\}$, it is during index 1 and the other half it is during index 2). Thus, in this simple example, the optimal constant baseline ($c=1$) is the same as $\beta_1$ for $c=3$, but this will not be the case in general. We also note that, in this particular problem, the optimal per-parameter baselines are a constant vector with the same value as the scalar optimal baseline, so the per-parameter baselines are actually equivalent to the scalar baselines in this case.

We can also test the alternative optimal baseline \eqref{wrong-optimal-baseline} that was derived in \cite{beyond-variance}, because it is possible to compute the action-value function. 

\begin{figure}[H] 
\centering 
\includegraphics[ width=\textwidth]{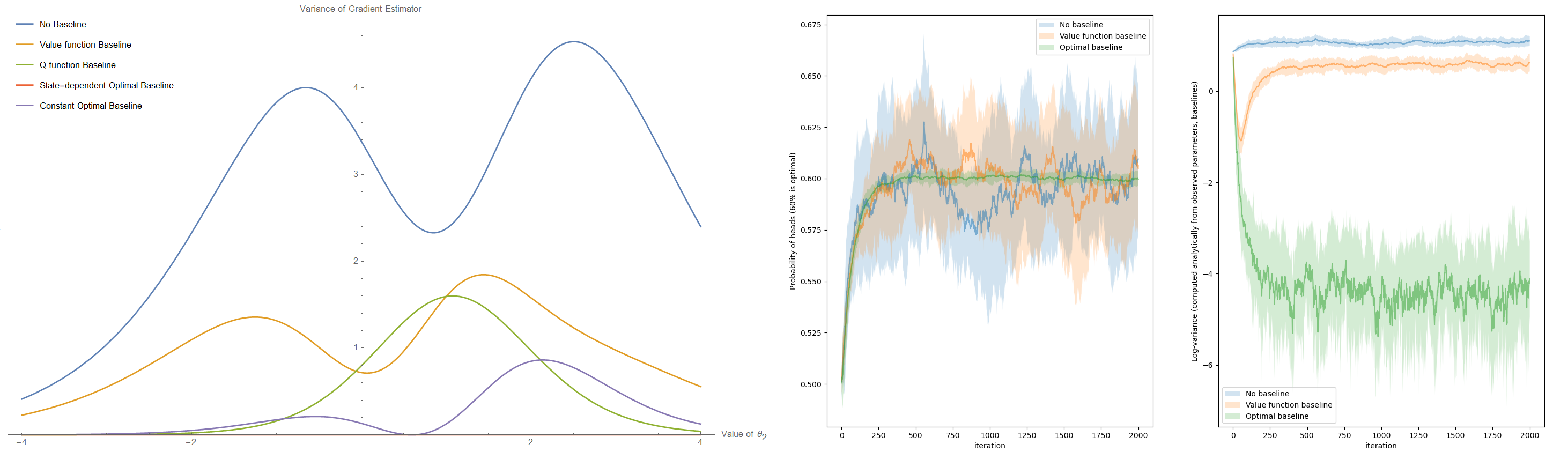}  \label{coinflipfigs}
\caption{Left panel: analytic plots of the variance of the gradient estimator \eqref{coinflip1} as a function of $\theta_2$ ($\theta_1 = 1$), for varying baseline formulations. Middle and right panel: numerical plots of the probability of heads (60\% is optimal) and variance of the gradient estimator during training.}
\end{figure}

The left panel of figure \ref{coinflipfigs} plots the variance of the gradient estimator with the different baselines, for varying values of $\theta_2$ and $\theta_1=1$. The baselines and their corresponding variance were computed analytically using computer algebra (mathematica). The regular baseline \eqref{regular-baseline} (which uses the value function), the optimal baseline \eqref{wrong-optimal-baseline} (which uses the action-value function) and our c-optimal baseline ($c=3$) are all state-dependent baselines and use 3 different baseline values. We see that only our optimal baseline always gives 0 variance, while the value function baseline or Q function optimal baseline both only reduce the variance by roughly half on average. Thus it is clear that our derivation has given the true optimal baseline. 

We also point out that this example is unusual in that it is actually possible to achieve a gradient estimator with 0 variance. This means that no matter what sample path is observed (heads-heads, heads-tail, tail-heads, or tail-tail), the gradient estimator with optimal baseline actually gives the true expected gradient. The sample paths heads-tail and tail-heads are identical, so to achieve 0 variance, the gradient estimator must give the same result (in both of its two components) for each of the 3 unique sample paths. Then, achieving 0 variance requires satisfying a system of 6 equations. In general, we would then need 6 variables; this is precisely how many unique values a state-dependent per-parameter baseline can take, which is why we can achieve 0 variance (and as mentioned previously, the per-parameter baseline is equivalent to the regular baseline in this case). More generally, because a baseline for $y_i$ can only use the partial sample path information $y_1, \ldots, y_{i-1}$, we will find that in most problems, achieving 0 variance will require more baseline values than we can possibly have, and is therefore impossible. 

The middle and right panels show the results of solving the problem numerically. We applied SGD with a learning rate of $0.01$ starting from an initial guess of $[ \theta_1, \theta_2 ] = [ 1, 1]$. Those plots show three different curves, corresponding to the case with no baseline, the value function baseline, and our state dependent optimal baseline. The baselines are solved numerically at the same time as the parameter values, so the value function baseline uses 3 additional parameters (corresponding to the baseline values), and the optimal baseline uses 6 additional parameters (corresponding to the numerator/denominator of the baseline values). The baselines also use a learning rate of $0.01$ and start from initial values of $0$. In each case, we replicate the experiment 20 times and plot the mean $\pm$ standard deviation. The right panel shows the variance of the gradient estimator, which is computed analytically based on the current values of $\theta_1, \theta_2$ and the baselines. We see that the optimal baseline gradient estimator has a variance which is roughly four orders of magnitude lower, and because of this its trajectory has far less randomness in it. 

\subsection{A 3-Arm Bandit Problem}
We now consider a bandit problem with 3 arms. Each of the arms deterministically gives a constant reward of 0, $0.7$, and $1$ respectively. Thus the optimal solution is simply to always pull the arm giving a reward of 1. The problem is parameterized with a softmax distribution over the probability of pulling the arms, so there are three parameters $\theta_1, \theta_2, \theta_3$. Because this problem only has a single action per episode, we can only have a single, constant baseline. Thus with a normal baseline, we only have 1 baseline value, or 3 baseline values by using a per-parameter baseline; it is then impossible to have 0 variance as we would require 9 baseline values to do so. 

\begin{figure}[H] 
\centering 
\includegraphics[ width=\textwidth]{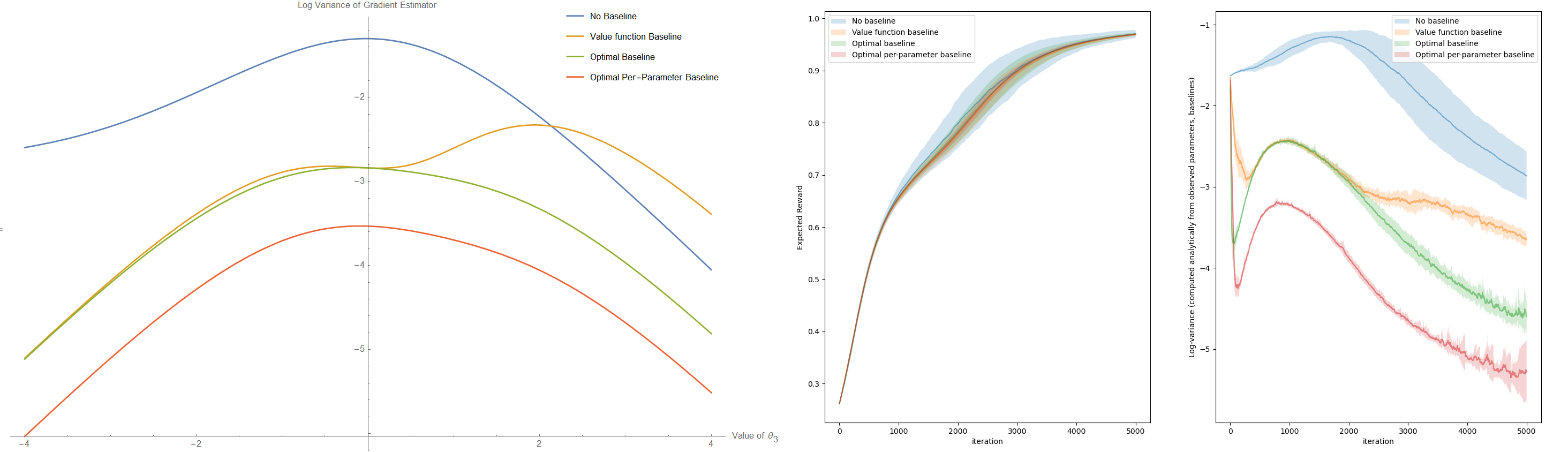} \label{mabfig} 
\caption{Left panel: analytic plots of variance of the gradient estimator as a function of $\theta_3$ ($\theta_1, \theta_2 = 1$). Middle and right panel: numerical plots of expected reward and variance of the gradient during training.} 
\end{figure}

In this example we find that the action-value function optimal baseline \eqref{wrong-optimal-baseline} actually gives the true optimal baseline, which is equivalent to our optimal baseline. We considered in total four different formulations corresponding to no baseline, the value function baseline, the optimal baseline, and optimal per-parameter baseline. The analytic plots of variance, which used closed form solutions of the baseline values, is given in the left panel of figure \ref{mabfig}. The middle and right panels show the numerical solutions when starting from an initial value of $[ \theta_1, \theta_2, \theta_3] = [3, 2, 1]$ (favoring the 0 reward and 0.7 reward arms) with learning rates of $0.01$ for the baselines and $\theta$. Again we find that the optimal baseline results in a lower variance, and also we see that the per-parameter optimal baseline improves on the optimal baseline by roughly another order of magnitude. However it is interesting to note that in this example, the lower variance doesn't actually seem to give any tangible benefit in terms of the expected reward. 

\section{Summary and Future Work}
We have introduced the notion of baselines, specialized control variates which can be applied to score functions. We derived a new optimal baseline which can achieve better variance reduction than the start of the art, and validated the methodology on some simple toy problems. Research is presently underway on several deep reinforcement learning benchmarks (e.g. lunar lander and half cheetah) to test the methodology on more complex models.

Our work on variance reduction so far has focused on score functions because of the key role they play in our gradient estimator \ref{thm1-1}. Moreover, score functions are known to have a high variance in general and often require a control variate to be used to achieve satisfactory results in practice. Based on the discussions and examples in sections \ref{sec-1-5} and \ref{sec-3-3}, which have highlighted some of the novel applications due to the inclusion of score functions (particularly with regards to learning piecewise functions, if statements, and functions using logic), it is clear that future research directions will likely involve models and gradient estimators involving score functions. 

At present, our work on variance reduction is incomplete. In the future, we will extend the methodology from the pure score function case, to the mixed score function/pathwise derivative situation.

Besides baselines, there are many more opportunities for future work in variance reduction. Or even more generally, there are many opportunities for work that modifies the gradient estimator in some way in order to achieve better convergence of SGD (because, as noted in the multi-arm bandit example and work such as \cite{beyond-variance}, lower variance does not necessarily equate to an improvement). In this area, importance sampling has long been an important topic, and will likely continue to be. There are also ideas for new control variates which can be more general than baselines. 

\chapter{Applications in Vehicular Traffic Flow: Car Following}
In chapters 5 and 6, we present two studies that were published in \cite{adjointpaper} and \cite{relaxpaper}.
\section{Introduction}\label{intro}
\subsection{The Case for Car Following and Trajectory Data}\label{intro1}

A complete traffic simulator consists of many different modules, including algorithms to determine the demand for different road sections, complex rules to govern lane changes, mergers, and diverges, a model to describe traffic flow and driving behaviors, and more. Out of all these modules, the description of traffic flow is the most fundamental. The decision of what model is used to describe traffic flow has impacts on the simulator's strengths, weaknesses, and overall performance. In particular, one important distinction is whether to simulate at the macroscopic or microscopic level. 

At the microscopic level, car following models are often used. Car following models describe how a vehicle behaves as a function of its surroundings; typically they are expressed as differential equations where any vehicle is coupled to the vehicle directly ahead (i.e. its lead vehicle). Many car following models are based on common sense rules of driving. Some simple examples are adopting a speed based on the distance between you and your leader, or accelerating based on the difference in speeds between you and your leader. 

Whereas microscopic traffic models describe the motions of individual vehicles, their macroscopic counterparts treat traffic as a fluid. At the macroscopic level, there are no vehicles, only traffic densities. Congested road sections with high densities slowly trickle forward, while uncongested traffic at low densities flow freely. This connection to fluid dynamics was originally made by \cite{39}, and the influence of the resulting traffic flow theory can hardly be overstated. In the more than six decades since, numerous works have attempted to improve upon the so-called Lighthill-Whitham-Richards (LWR) model in order to better describe various complex elements of traffic flow. Among those, \cite{37}, \cite{40}, \cite{46}, \cite{48}, \cite{53} are perhaps some of the most notable. Nonetheless, when it comes to macrosimulation, the LWR model, and its discrete numerical formulation \citep{42}, is still widely used today. 

For all its merits, the LWR model is not without its limitations. A key relationship between flow and density known as the ``fundamental diagram", must be supplied. More importantly, it is assumed vehicles are homogeneous and exhibit static behavior.
In real traffic, drivers behave differently; vehicles have varying performances; traffic demand fluctuates; controlled intersections and lane changes have complex effects. Traffic flow is incredibly dynamic and inhomogeneous. 
The shortcomings of the LWR model become apparent in practice (see \cite{46,49,41}). In the fundamental diagram, instead of equilibrium curves, one finds a wide scattering of points. The time evolution of real traffic flow shows a much richer picture than what the LWR model predicts. Examining the evolution of a vehicle trajectory in the speed-spacing plane, one sees complex hysteresis loops which are inconsistent with current macroscopic theory (\citealt{28}, \citealt{50} or Fig. \ref{fig2} are examples of such plots). Given these limitations, one would hope for some car following model that could explain all these phenomena.   

However, existing works are not prepared to offer guidelines as to which car following models are best. It is also not clear as to how much more accurate car following models are compared to macro-models (like LWR). Answering either of these pressing questions is difficult: the traffic microsimulation literature consists of literally hundreds of different models, many of which have never actually been thoroughly validated \citep{7}. Works such as \cite{9}, \cite{10}, \cite{14}, and \cite{19} have attempted to address this long-standing problem of car following models but have had varying results as to which models are best. 

Drawing conclusions on the efficacy of different car following models requires both a large vehicle trajectory dataset (e.g. \citealt{6} or \citealt{66}) to facilitate comparison, as well as a reliable methodology for model calibration and validation. The aim of this work is to develop new methodology to improve the calibration of car following models. However, one should keep in mind that this only addresses one half of the challenge; equally important is the creation of new datasets and new approaches for collecting accurate and meaningful traffic data. 

There is a need for an increased use of trajectory data to study traffic flow. This need goes far beyond the validation of existing car following models. As new technologies like autonomous vehicles, connected cities and on-demand mobility become fully realized, the LWR model assumption of homogeneous drivers exhibiting static behavior becomes increasingly unrealistic. The limitations of macrosimulation and the uncertainty of new technology are both good reasons to suspect a growing demand for microsimulation. At the same time, vehicle trajectory data (which is needed for microsimulation/car following) is easier to get than ever before because of new technologies like computer vision, lidar/radar, and GPS. Compared to the more commonly used loop detector data, trajectory data describes traffic flow at much higher temporal and spatial resolutions. These new sources of data, still largely unexploited, can help advance both micro- and macro- theory. 

\begin{figure}[h] 
\centering 
\includegraphics[ width=\textwidth]{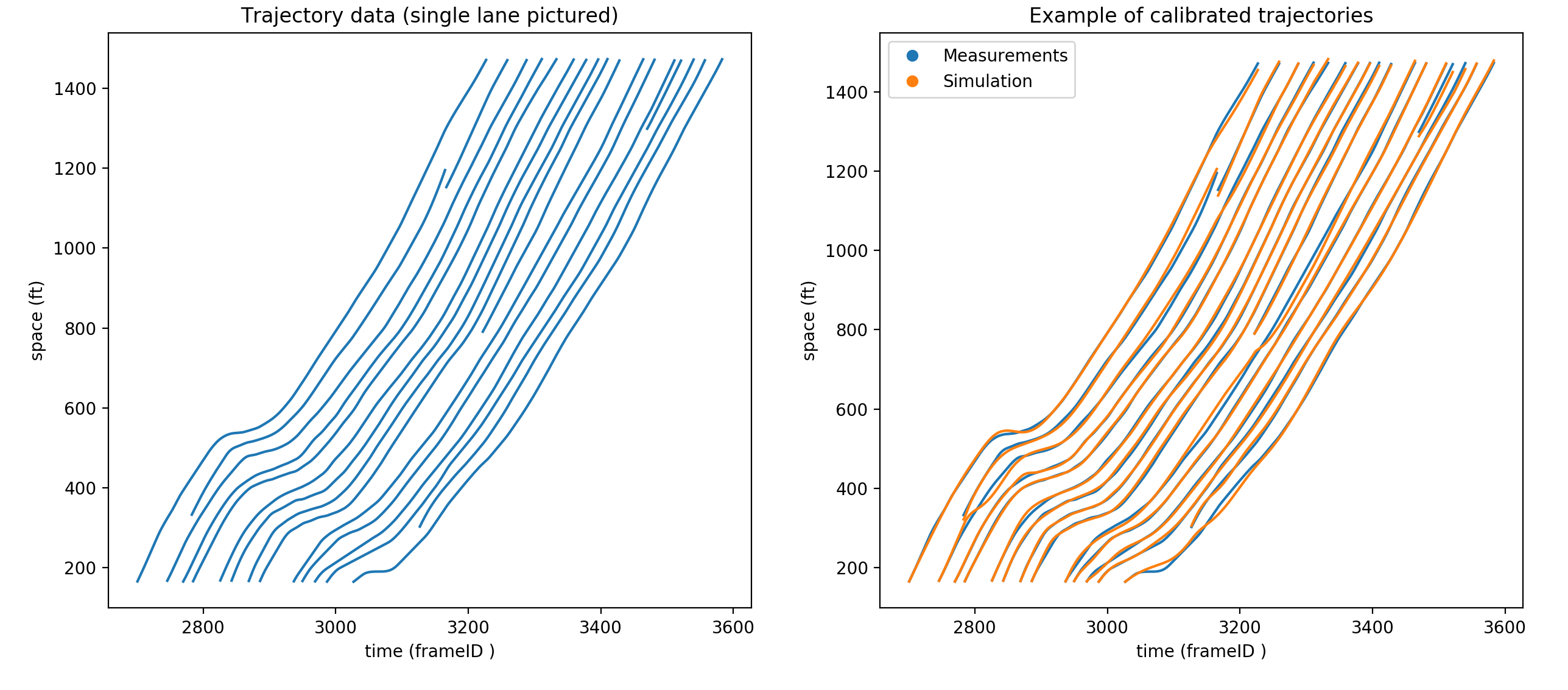} 
\caption{On the left, an excerpt of a single lane from the NGSim data. ``Incomplete" trajectories are due to vehicles changing to/out of the lane pictured. One can observe many of the important features of traffic flow such as shockwaves, lane changing events, and individual driver behaviors. Compared to macroscopic data, these features are captured at higher temporal/spatial resolutions and without aggregation. On the right, a car following model (the OVM) has been calibrated to the trajectory data.   } \label{fig1}
\end{figure}

\subsection{Calibration Using Trajectory Data}\label{intro2} 
\noindent It is relatively simple to write down some mathematical equations and call them a traffic model. Far more difficult is applying this model in practice. In order to validate a model against real data, one must find the model parameters that best explain the data: this is known as calibration. Solving the calibration problem requires both a dataset describing the quantity of interest, as well as a mathematical model that explains the data. This paper deals with car-following models that are posed as differential equations, and may include elements such as delays, bounds on acceleration/speed, and multiple driving regimes. The car-following model is to be calibrated against trajectory data, time series data that describes the motion of individual vehicles. 

Existing literature has covered various aspects of the trajectory data calibration problem: the use of meta-models, sensitivity analysis, optimization algorithms, measures of performance (e.g. speed or spacing data), and goodness-of-fit function (i.e. loss function). \cite{treibercalibration}, \cite{24}, and \cite{25} are good works focused primarily on methodology. \cite{1}, \cite{12}, \cite{14} and \cite{19} are good works focused on both calibration and the benchmarking of various models. 

In this paper calibration is posed as an optimization problem. For a wide class of models, being calibrated to trajectory data gathered from any source, the gradient of the optimization problem has been derived in an efficient way using the adjoint method. This enables the calibration problem to be solved by more efficient optimization routines which use gradient information. Several state of the art algorithms are benchmarked in order to determine both the accuracy of the fit they can achieve, as well as the computational effort required to apply the algorithm. The paper is organized as follows. Section \ref{models} introduces car following models and discusses current research directions for improving car following models. Section \ref{calibration} describes the optimization problem and discusses some issues arising in its solution. Section \ref{adjoint} applies the adjoint method to the optimization problem and compares it to finite differences. Section \ref{algos} compares different algorithms for solving the calibration problem.

\section{Car-Following Models}\label{models}

\noindent We will consider any car-following model that can be expressed as either an ordinary or delay differential equation. This describes the majority of models in the literature, as well as the majority of models used in commercial microsimulation packages \citep{47}. We deal with models of the following form 
\[ \ddot  x_n(t) = h(\dot x_{n-1}(t-\tau), x_{n-1}(t-\tau), \dot x_{n}(t-\tau), x_{n}(t-\tau), p)  \stepcounter{equation}\tag{\theequation}\label{1}\]
where $x_n(t)$ is the position of vehicle $n$ at time $t$. $h$ is the car-following model, which depends on parameters $p$, the ``following" vehicle $x_n$ and ``lead" vehicle $x_{n-1}$. $\tau$ is a time-delay parameter which represents human reaction time.  Eq. \eqref{1} states that a driver's behavior (how they choose to accelerate) is determined from the speed and position of their own car, the speed and position of the leading car, and parameters $p$. Given a lead vehicle trajectory $x_{n-1}(t)$, parameters $p$, and initial condition $x_n(t_n)$, the car following model $\eqref{1}$ will produce the follower trajectory $x_n(t)$. 
The presence of nonzero $\tau$ leads to a delay differential equation (DDE).  Many car-following models (including most of those used in commercial software) treat reaction time explicitly by using a DDE. 

Besides ODEs and DDEs, some car-following models are formulated as stochastic differential equations or derived using nonparametric techniques. We briefly discuss these types of models in section \ref{sde} but the current methodology does not address their calibration.

\subsection{Ordinary Differential Equations}\label{ODE}

\noindent Eq. $\eqref{1}$ gives a general form for car-following models. The simplest type of model is a first-order ODE where the speed of the following vehicle $n$ is completely determined by the space headway $(x_{n-1} - x_n)$. In this type of model, each driver adopts a speed based on the distance between their car and the leading car. 

This simple kind of model is important because of its connection to macroscopic theory and the LWR model. Specifically, if one assumes all vehicles are identical, and interprets the reciprocal of density as headway, then a functional relationship between traffic flow and density can be recast as a relationship between vehicle speed and headway \citep{20}. Under these assumptions, the fundamental diagram can be thought of as describing a first-order car-following model. Consider the Newell car following model an example
\[\dot x_n(t) = \min\{ v_f, \dfrac{x_{n-1}(t) - x_n(t) - s_{\text{jam}}}{\alpha}\}. \stepcounter{equation}\tag{\theequation}\label{2} \]
$v_f, \alpha$  and $s_{\text{jam}}$ are parameters describing a triangular fundamental diagram \citep{1}. Namely $v_f$ is the speed of a vehicle during free flow, $s_{\text{jam}}$ is the distance between two completely stopped vehicles, and $-s_{\text{jam}}/\alpha$ is the shockwave speed. 

The Newell model can also be written as a trajectory translation model, where the trajectory of a following vehicle is equivalent to the lead vehicle trajectory, translated over both space and time.
\[ x_n(t) = x_{n-1}(t - \alpha) - s_{\text{jam}}  \stepcounter{equation}\tag{\theequation}\label{3}\]

\begin{figure}[h] 
\centering 
\includegraphics[ width=\textwidth]{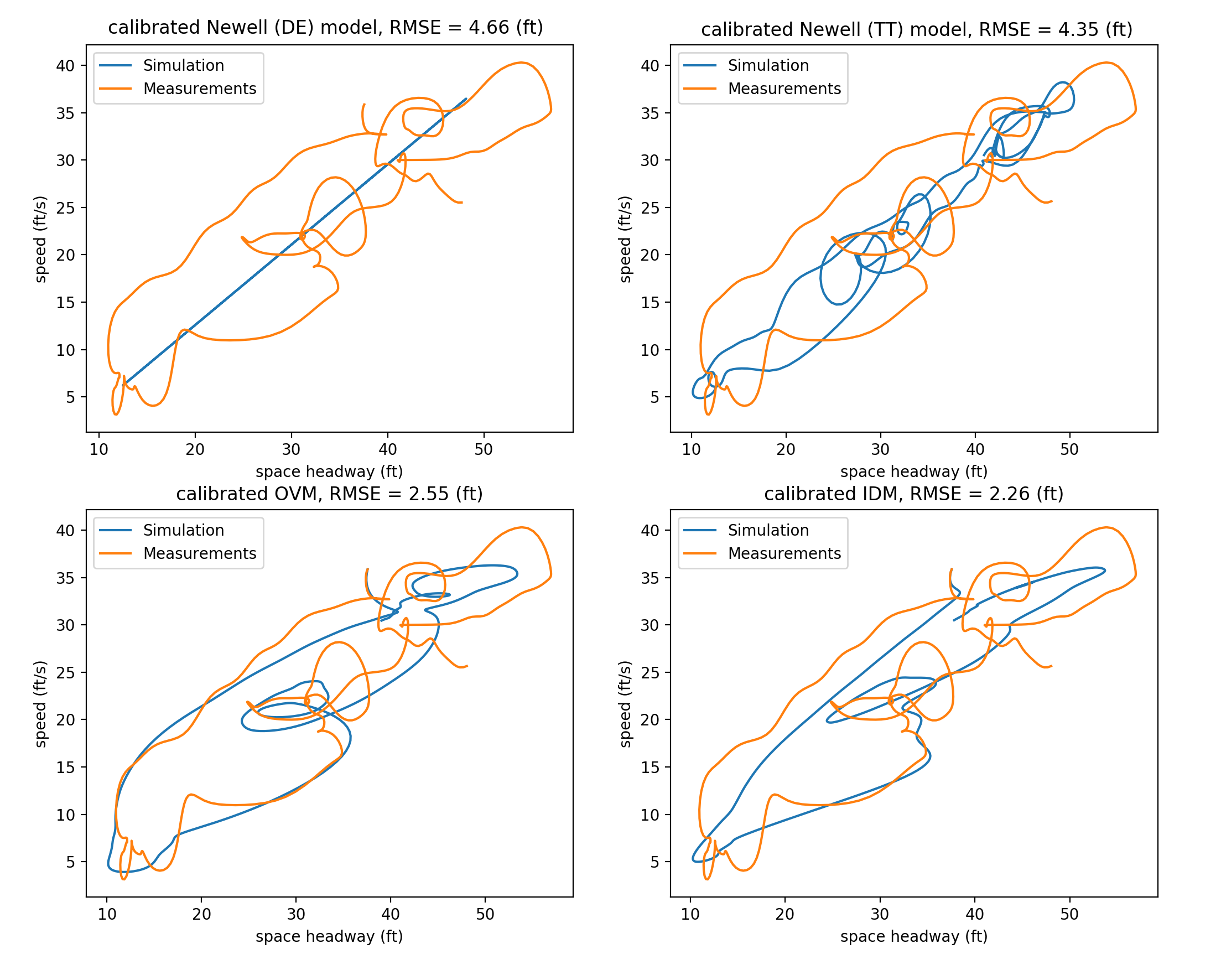} 
\caption{Four different car following models are calibrated to a single empirical vehicle trajectory by minimizing the root mean square error between simulation and measurements. The trajectories are plotted in the speed - headway plane.} \label{fig2}
\end{figure}

\noindent We refer to Eq. \eqref{3} as Newell (TT) and Eq. \eqref{2} as Newell (DE), where the acronyms stand for trajectory translation and differential equation, respectively. In the special case where the TT and DE models both use a numerical time step equal to $\alpha$, and vehicles are always in the congested regime (i.e. not moving at their free flow speeds), the two models become equivalent \citep{20}. In this special case, either model gives trajectories that exactly obey the LWR model with a triangular fundamental diagram. In general however, the two models are different (see for example Fig. \ref{fig2}). The advantages of the Newell (TT) or (DE) models are that they contain a small number of parameters, and those parameters are easily interpretable under macroscopic theory. 

More complicated car-following models (compared to those of Newell) are sometimes criticized for two main reasons: first, for not having enough empirical support for their validity, and second, for having too many parameters. Despite these criticisms, more complicated models have behaviors that attract researchers. For example, the optimal velocity model (OVM) proposed by \cite{52} exhibits a spontaneous phase transition between free flow and congested states, which may describe ``phantom" traffic jams \citep{54}. Another example is the intelligent driver model (IDM) proposed by \cite{69}, which can qualitatively reproduce a number of complex (empirical) traffic states. 
Besides more recent models like OVM or IDM, there are also a plethora of older models to complicate the mix \citep{7}. The default models in most modern traffic simulator packages (for example, VISSIM or AIMSUN) are still older car-following models \citep{47}, presumably because it is unclear if the newer car-following models offer any benefits. 

Fig. \ref{fig2} compares each of the four models just mentioned by calibrating them to a single vehicle trajectory from the reconstructed NGSim data \citep{29}. The calibration minimized the root mean square error (RMSE) between the simulation and measurements. The vehicle (ID 1013) was recorded for only 55 seconds, but even in that time many hysteresis loops were recorded in the speed-headway plane. For the NGSim data, which was recorded on the freeway in congested conditions, trajectories that qualitatively resemble this one are the rule rather than the exception. The different models all give a different caricature of the oscillatory behavior observed. In particular, the Newell (TT) model (with a simulation time step of .1 s) appears to better capture the seemingly chaotic nature of the oscillations, whereas the more complicated (and computationally expensive to calibrate) IDM or OVM models result in a lower RMSE and better capture the width of the hysteresis (the differences in speeds at a given headway). The point of this figure is not to make a case for one model over the other, but rather to point out the interesting behavior shown even in a single short trajectory. It also raises an important question which existing research seems unprepared to answer. 

With so many different models, each of varying complexities, which ones best describe traffic flow? Our work on calibration is motivated by this question.

\subsection{Delay-Differential Equations} 
\noindent Compared to ODEs, DDEs are more complicated to analyze and solve. Car-following models typically only consider the simplest case, where time delay (reaction time) is a constant. It seems the distinction between ODEs and DDEs has been given limited importance in the calibration literature, even though the addition of time delay as a parameter causes extra complications. As for the literature as a whole, the effect of nonzero time delay on car-following models is still not fully understood.  

When studying the behavior of car-following models, one common approach is to consider a line of cars with identical model parameters on a closed loop. In this special setting, one can find an ``equilibrium" solution where all vehicles have the same speed and same headway, with zero acceleration. The equilibrium solution is contrasted with an oscillatory/periodic solution, where the speed of a vehicle can oscillate in a manner reminiscent of stop-and-go or congested traffic. Since most analytic (i.e. not numerical) results come from studying the behavior around the equilibrium solution, the stability of the equilibrium solution is particularly important. Adjusting parameters in a model can be said to have a stabilizing (destabilizing) effect if the equilibrium solution will persist (degenerate) under a wider variety of conditions. Intuitively, one expects that drivers having slower reaction times will have a destabilizing effect, and this seems to be generally correct. 

 It has been shown for both the optimal velocity model (OVM) and intelligent driver model (IDM) that the inclusion of time delays has a destabilizing effect  (see \citealt{16}, \citealt{56}, \citealt{5}). But time delay is not the only model parameter that can have such effects. For example, \cite{57} showed that increasing (decreasing) the OVM's ``sensitivity parameter"  has stabilizing (destabilizing) effects (the sensitivity parameter is proportional to how strongly the driver accelerates). Time delay can also have more subtle effects: \cite{56} found the inclusion of time delays for OVM caused bistability between equilibrium and oscillatory solutions.   

 In \cite{15} the OVM is considered with different reaction times $\leq 0.3 $ seconds. In this range, no qualitative changes to the model are observed. The authors state that changing the sensitivity parameter can account for the effects of reaction time. \cite{12} and \cite{16} have done experiments on OVM with larger reaction times ($\approx 1$ seconds) and found that while it is true that reaction time is not especially important at low values, once increased past some threshold it will start to cause unstable behavior leading to vehicle collisions. A similar behavior was found in \cite{23} for the Gipps car-following model. These findings show the inclusion of reaction time will have a destabilizing effect, but large reaction times seem to have especially unstable effects, perhaps owing to some sort of phase transition. 

 In practice, one can observe that calibrated values for reaction time greatly vary across the literature, even for the same model. Consider that \cite{58},\cite{14}, and \cite{19} all calibrate the Gipps model, but all find very different means and variances for drivers' reaction times. One found the mean to be 0.57 s with a variance of 0.02 s. Another found the mean to be 1.73 s with variance 1.35 s.  
 
 Besides reaction time, another interesting extension to a model is the inclusion of ``anticipation" effects. Examples of anticipation are a driver estimating the future speed/position of a leading vehicle, or a following vehicle reacting to multiple vehicles instead of just its leader. \cite{5} showed that anticipation has a stabilizing effect for IDM, and \cite{21} showed the same for OVM. Of course, due to nonlinearities, stabilizing and destabilizing effects will not simply cancel out. 
\subsection{Stochastic and Data-Driven Models} \label{sde}
In this paper we consider the calibration of car following models formulated as ODEs or DDEs. This section discusses some recent research directions in which car following models are formulated as stochastic differential equations (SDEs), or derived from data using nonparametric models. \\
Papers using SDEs to model traffic flow typically argue that the stochastic models are better able to describe the formation and growth of congestion. Besides being able to achieve a closer fit to data, authors have concluded that deterministic models cannot reproduce some qualitative features of congestion. In particular, some recent field experiments involving 25 and 51 human driven vehicles on a straight road \citep{experimentalandempirical} showed that the standard deviation of traffic speeds initially grow in a concave way, whereas deterministic models seem to only be able to show convex growth. \\
SDEs in the literature are usually based on modifying an existing deterministic ODE model rather than created as an entirely novel model. Two common stochastic mechanisms are acceleration noise, and action points. For acceleration noise, white noise is added to the output of the deterministic model. The white noise may be sampled from either a uniform or normal distribution. Closely related to acceleration noise, is the case where the noise is added to a model parameter instead of directly to the output of the model. In \cite{laval2014} and \cite{flowarchitecture} gaussian acceleration noise was added to the Newell model and IDM respectively. In 2D-IDM from \cite{empiricalanalysis} uniform noise is added to a model parameter (in that case, the desired time headway parameter). \\
Action points are implemented by creating a model regime where the driver is insensitive to changes in the headway and/or lead vehicle velocity. Thus when inside the regime, a driver will maintain their current behavior until a sufficiently large difference in the state is achieved, at which point the driver will react to the now significantly different driving situation. The reader is referred to \cite{treiber2017} or \cite{onsomeexperimental} for specific details on implementing action points. \\ 
The motivation for including white noise is to model random human behavior and reflect the fact that humans cannot perfectly control their vehicles, whereas the motivation for action points is that drivers are not able to perceive minute changes to the traffic situation and thus a sufficiently large difference is required before they can start to react. \cite{treiber2017} shows that action points and white noise have similar effects on traffic simulations, and suggests that both mechanisms should be included. \\

\noindent Data-driven car following models are based on machine learning models and algorithms, and as a result their formulation is significantly different than the ODE, DDE, and SDE cases discussed so far. However, despite their different formulation, the inputs (current headway, follower and leader velocities) and output (current follower acceleration) are typically very similar. Papers in the literature have reported that data-driven models can achieve better fits to empirical data compared to parametric models such as OVM or IDM. The discussion of these machine learning techniques is outside the scope of this paper but we mention the different approaches which have been successfully applied. In \cite{simplenonparametric} $k$-nearest neighbor (knn) is used, and \cite{towardsdatadriven} uses locally weighted regression (loess), which can be thought of as a generalization of knn. \cite{recurrentcf} discuss using feedforward neural networks with a single hidden layer and ReLU nonlinearity. That paper also uses recurrent neural networks (RNNs) to overcome some difficulties of the feedforward neural networks. \cite{lstmcf} uses long short-term memory neural networks, a special kind of RNN. \cite{deeprlcf} use deep reinforcement learning (deep RL) and policy gradients, and do a comparison between deep RL, IDM, RNN, loess, and feedforward neural networks.

\section{Calibration}\label{calibration}
\noindent It is useful to rewrite the second order differential equation Eq. $\eqref{1}$ as a system of first order equations:
\[ \dot x_n (t) = \begin{bmatrix} \dot x_n^*(t) \\ \dot v_n(t) \end{bmatrix} = \begin{bmatrix} v_n(t) \\  h^*( x_{n-1}(t-\tau_n), x_{n}(t-\tau_n), p_n)  \end{bmatrix} = h_n(x_n(t), x_n(t-\tau_n), x_{L(n)}(t-\tau_n), p_n)  \stepcounter{equation}\tag{\theequation}\label{4}\]
\noindent where $x_n$ is now a vector consisting of the position and velocity of vehicle $n$ at time $t$. $x_n^*$ and $v_n$ refer to the position and speed for vehicle $n$. $p_n$ denotes the parameters for vehicle $n$. $L(n,t)$ refers to the leader of vehicle $n$ at time $t$, and we denote the lead trajectory as $x_{L(n)}$ (time dependence suppressed) to reflect the fact that the index corresponding to the lead vehicle may change due to lane changing. Eq. \eqref{4} is what we refer to when we say ``car-following model" and its solution $x_i(t)$ is a ``simulated trajectory".

\noindent The calibration is done for some arbitrarily sized platoon of $n$ vehicles, indexed as $1, \ldots, n$. Each vehicle has some individual parameters $p_i$ and a corresponding car following model $h_i$, so all vehicles have their own set of parameters, and $p_i$ only describes vehicle $i$. Let $p$ be a concatenated vector of all the parameters, so $p = [p_1, p_2, \ldots, p_n]$. Calibration finds the parameters $p$ which minimize the total loss between the simulated trajectories $x_i(t)$ and the corresponding measured vehicle trajectories, which are denoted $\hat x_i(t)$. The loss function is denoted as $f(x_i, \hat x_i, t)$ and represents the goodness of fit of the simulated trajectory $x_i$ at time $t$. The measurements $\hat x_i(t)$ are known in the interval $[t_i, T_i]$ for each $i$. \\
In order to compute the simulated trajectories for each vehicle $i$, we require a lead vehicle trajectory $x_{L(i)}(t)$ as well as an initial condition $x_i(t_i)$. The time interval that the lead vehicle trajectory is known is denoted $[t_i, T_{i-1}]$. 
$x_i(t_i)$ is the initial condition, which is specified as $x_i(t_i) = \hat x_i(t_i)$ for an ODE model. 

For an ODE, the optimization problem is 
\begin{align*}
\underset{p}{\min} & \quad  F = \sum_{i=1}^n\int_{t_i}^{T_{i-1}} f(x_i, \hat x_i, t) dt         \stepcounter{equation}\tag{\theequation}\label{5} \\
\text{s.t.} & \quad   \dot x_i(t)  - h_i(x_i(t),x_{L(i)}(t),p_i)  = 0, \quad t \in [t_i, T_{i-1}], \ \ \ i = 1, \ldots, n   \\
& \quad     x_i(t_{i}) - \hat x_i(t_{i})= 0 \quad  i = 1, \ldots, n    \\
& \quad b_{\rm low} \leq p \leq b_{\rm high}
\end{align*}
The objective function $F$ is defined as the integral of the loss function $f$ over time, summed over each simulated vehicle. The constraints are that the simulated trajectories $x_i$ obey the car-following model $h_i$ from time $t_i$ to $T_{i-1}$, with the initial conditions $x_i(t_{i}) = \hat x_i(t_{i})$. There may also be some box constraints $b_{\rm low}, b_{\rm high}$ which prevent parameters from reaching unrealistic values.  

The DDE case is similarly defined: 
\begin{align*}
\underset{p}{\min} & \quad  F = \sum_{i=1}^n\int_{t_i+\tau_i}^{T_{i-1}^*} f(x_i, \hat x_i, t) dt          \stepcounter{equation}\tag{\theequation}\label{6} \\
\text{s.t.} & \quad   \dot x_i(t)  - h_i(x_i(t), x_i(t-\tau_i),x_{L(i)}(t-\tau_i),p_i)  = 0, \quad t \in [t_i+\tau_i, T_{i-1}^*], \ \ \ i = 1, \ldots, n   \\
& \quad     x_i(t) -\hat x_i(t) = 0 \quad t \in [t_ i, t_i+\tau_i], \ \ \ i = 1, \ldots, n    \\
& \quad b_{\rm low} \leq p \leq b_{\rm high}
\end{align*}
where $T_{i-1}^* = \min\{T_{i-1}+\tau_{i}, T_i\}$, and the reaction time for vehicle $i$, $\tau_i$, is assumed to be a model parameter. 
$x_i(t) -\hat x_i(t) = 0$ defines the history function for $x_i$ (DDEs have history functions as their initial conditions). 

\subsection{Downstream Boundary Conditions for Car Following Models} \label{downstreamboundary}
\begin{figure}[H] 
\centering 
\includegraphics[ width=.6\textwidth]{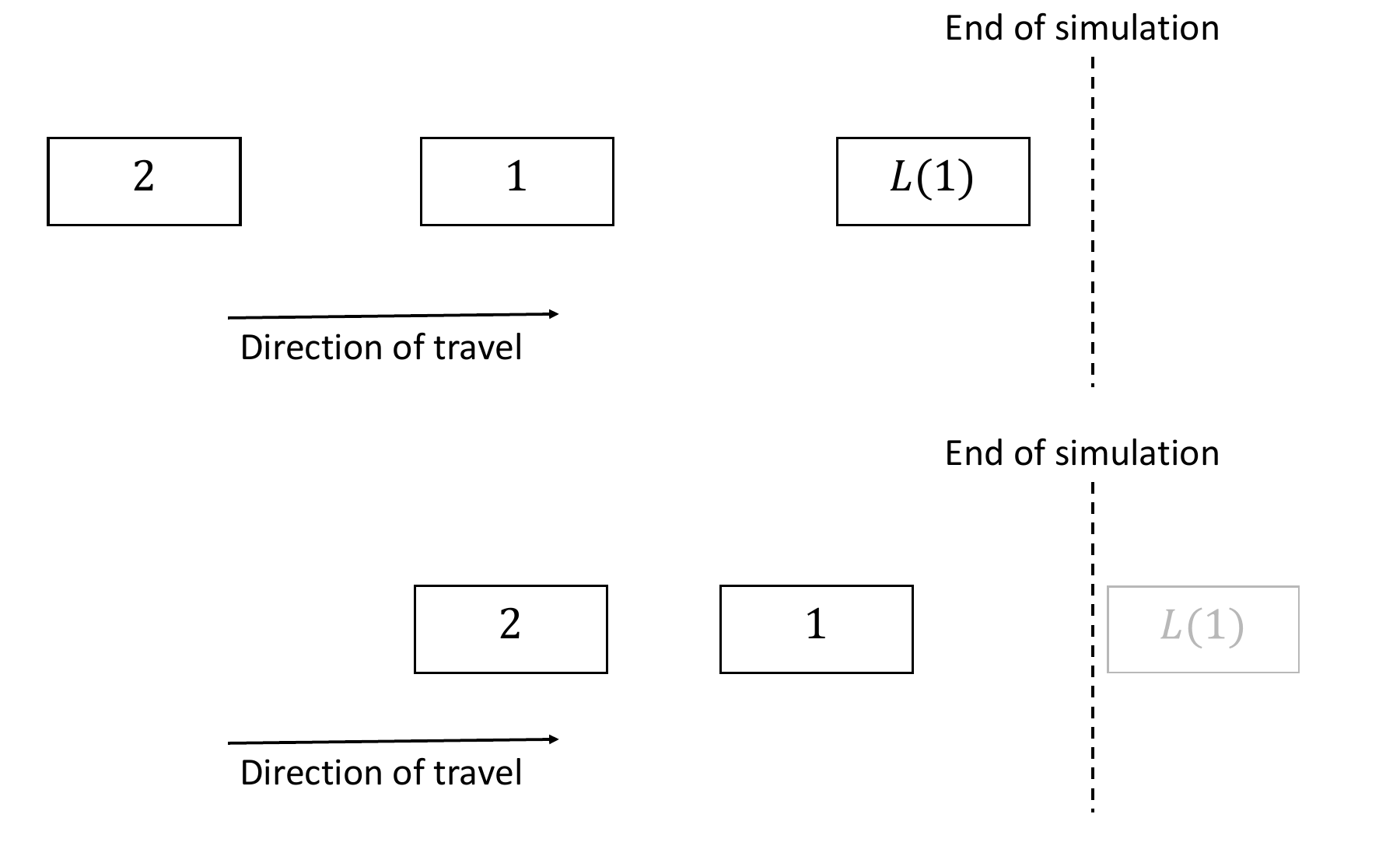} 
\caption{Boxes represent vehicles; vehicle $2$ follows $1$, and $1$ follows $L(1)$. In top panel, the leader $L(1)$ is still in the simulation and so the car following model can be used to update vehicle $1$. In bottom panel, $L(1)$ is no longer in the simulation, and so a downstream boundary condition is used to update vehicle $1$. }  \label{boundary}
\end{figure}

In Eqs. \eqref{5} and \eqref{6} we have specified the initial conditions, which give the times vehicles enter the simulation and their position/speed upon entering. These conditions might also be referred to as upstream boundary conditions since they state how vehicles enter the simulation. For video data such as NGSim, which is recorded on a fixed section of road, downstream boundary conditions also need to be specified. Video data has staggered observation times ($T_{i-1} < T_i$) meaning that a vehicle's leader leaves the study area before the vehicle itself. Figure \ref{boundary} gives a visual depiction of this. \\
Car following models require a lead vehicle trajectory as input, therefore once the lead vehicle leaves the simulation, the car following model can no longer be used. Downstream boundary conditions then give a rule for the dynamics a vehicle follows once its lead vehicle leaves the simulation (i.e. they specify the dynamics for $x_i(t), \ \ t \in [T_{i-1}, T_i]$). The the boundary conditions we implement in this paper are
\begin{align*} 
\dot x_i(t) = \dot{\hat{x}}_i(t), \ \ t \in [T_{i-1}, T_i] \stepcounter{equation}\tag{\theequation}\label{7}
\end{align*}
meaning that when a vehicle's leader is no longer available, the vehicle will move with whatever speed was recorded in the data. Under these boundary conditions, the velocity $v_i(t)$ and acceleration $\dot v_i(t)$ will be piecewise continuous (they will have a jump discontinuity at $T_{i-1}$). The objective $F$ and gradient $dF/dp$ are still continuous under this boundary condition (see sections \ref{calibration8} and \ref{continuous} for the relevant discussion on discontinuities in car following models and how they impact the calibration problem). Note that for a DDE, the time $T_{i-1}^*$ is used instead of $T_{i-1}$.  \\
It is necessary to specify the downstream boundary conditions for two reasons. First, it allows vehicles to exit the simulation area. Second, it allows congestion to propogate into the simulation area from downstream. This happens when the boundary conditions specify a reduction in speed.
\subsection{Reaction Time as a Parameter}\label{calibration6}
Simulated trajectories are computed numerically with a specified time discretization. Since we want to compare the simulated trajectories with the actual measurements, it makes sense for the simulation to use the same time discretization as the data. For example, if the data is recorded every 0.1 seconds, the simulated trajectories should use a time step of 0.1 seconds.  If the simulation used a different time step (e.g. 0.09), then the loss can only be calculated at a few points (e.g. 0.9, 1.8) unless we define some interpolation procedure. 

For a DDE, the time discretization and reaction time need to be multiples of each other. If the simulation uses a time step of 0.1 seconds, then the reaction time will have to be an integer multiple of 0.1 (e.g. 0.1, 0.2). We can either: 
\begin{enumerate} 
\item Enforce that every vehicle's reaction time is a integer multiple of 0.1 
\item Allow reaction time to be a continuous variable and define the loss function through interpolation. %
\end{enumerate}
If we choose 1, it leads to a mixed integer nonlinear programming problem (MINLP). This is a much harder problem than the original nonlinear program. One approach used in \cite{19} would be to split up all vehicles into small platoons (either pairs or triplets) and use a brute force search for reaction time. However, even when the platoons are small, using brute force vastly increases computation because the calibration has to be repeated for every possible combination of reaction times. 

The other option is to use an approach similar to \cite{12} where reaction time is a continuous variable and interpolation is used on the simulated trajectories. In general, any sort of interpolation could be used, but linear interpolation is the most common. In this approach, interpolation needs to be used on the lead trajectory and initial history function in order to define the model. Assuming the numerical time step is not equal to the delay, interpolation will need to be used on the simulated trajectory as well. 

One issue with interpolation is how to compute the loss function. Specifically, assuming the simulation is defined on a different set of times than the measurements, should the simulation be interpolated onto the measurement's times or vice versa? We suggest doing both. This avoids the possibility of error cancelation occurring during the interpolation and causing the loss function to be erroneously small. In this scenario it would be preferable to reweight the loss function as well since the loss will be computed over more points. 

\subsection{Multiple Regimes, Lane Changing, and Discontinuities in Car Following Models}\label{calibration8}
We say a model has multiple regimes when the model is expressed either as a piecewise function, or by using $\min$ and $\max$ functions. These types of modifications are frequently applied in practice. Perhaps the most common are examples like Eq. \eqref{2}, where the maximum speed is explicitly constrained by some constant $v_f$. For second order models, often the acceleration will be constrained between minimum and maximum accelerations $a_{\rm min}, a_{\rm max}$. Besides these simpler regimes which implement bounds on the acceleration and/or velocity, there are also models which use different regimes to describe different car following behavior. The Gipps model \citep{59} is one such example, which has one regime to describe normal car following behavior, and a second regime which implements an emergency braking behavior. Other examples of multi-regime models are the constant acceleration heuristic proposed in \cite{treiberbook}, the follower stopper controller proposed in \cite{FS}, or any model implementing action points. We consider 
\begin{align*} 
h_i(x_i, x_{L(i)}, p_i) = \begin{cases} 
 h_{i,2} & g_{i,2}(x_i, x_{L(i)}, p_i) \geq 0  \\ 
  h_{i,1} & \text{ otherwise} \\ 
  \end{cases} \stepcounter{equation}\tag{\theequation}\label{regime}
\end{align*}
as a general form of a two regime car following model, with the two regimes $h_{i,1}$ and $h_{i,2}$. The condition for the associated regime to become activated is $g_{i,2}$ (which is possibly a vector valued function). For example with Eq. \eqref{2}: $h_{i,1} = (x_{L(n)} - x_n - s_{\rm jam})/\alpha$, $h_{i,2} = v_f$ and $g_{i,2} = h_{i,1}   - h_{i,2}$. There must always be $m-1$ such conditions for a model with $m$ regimes, so a model with more regimes will simply have $h_{i,2}, \ldots, h_{i,m}$ and $g_{i,2}, \ldots, g_{i,m}$. \\
In the case of a first order model's velocity being bounded, or a second order model's acceleration being bounded, the resulting model $h_i$ is everywhere continuous (because both regimes will be equal when the switching condition $g_{i,2}$ is met). In general a multi-regime model need not be everywhere continuous. Gipps and follower stopper are both examples of a multi-regime model being everywhere continuous; the constant acceleration heuristic or a model implementing action points are examples not guarenteed to be continuous when switching between regimes. Of course, even for a multi-regime model which is everywhere continuous, its derivatives will have discontinuities when switching between regimes. Note that a multi-regime model written using $\min$ and $\max$ will always be continuous. The reason why we prefer using the form Eq. \eqref{regime} is because 1) it is more general, 2) we can explicitly refer to the switching condition $g_{i,2}$ and 3) the regime of the model is explicitly stated, which avoids the problem of the regime being ambiguous when the quantities inside a $\min / \max$ are equal. \\
Lane changing is another issue which causes discontinuities in car following models. Specifically, whenever a lead vehicle changes, i.e. $L(i,t)$ changes, then the lead vehicle trajectory $x_{L(i)}$ will have a jump discontinuity, and therefore $h_i$ will have a jump discontinuity as well. \\ 
Given that lane changing always causes discontinuities and multi-regime models may cause discontinuities, one may worry whether or not the objective $F$ will have discontinuities, and whether or not the gradient $dF/dp$ is even guarenteed to exist. We give sufficient conditions for the continuity of $F$ and $dF/dp$ in section \ref{continuous}.

\subsection{Algorithms for Solving the Calibration Problem}
\noindent Eqs. $\eqref{5}$ and $\eqref{6}$ are nonlinear and non-convex optimization problems that must be solved numerically. Current approaches usually rely on gradient-free optimization, with genetic algorithm (GA) and Nelder-Mead simplex algorithm (NM) being the two most commonly applied algorithms \citep{7,8,9,10,11,12,14,19}. These methods have the disadvantage of being slow to converge, and require many objective function evaluations. The objective $F$ is relatively expensive to compute in this context because we first must solve a system of differential equations (i.e. compute the simulated trajectories). As an alternative to directly solving the optimization problems $\eqref{5}$ and $\eqref{6}$, parameters can be estimated directly from data using statistics techniques \citep{2,3,4,51} but this approach is not guaranteed to find local minima. Also, models may have parameters which do not have clear physical interpretations, and therefore cannot be directly estimated. 

Gradient-based optimization algorithms can potentially solve the calibration problem faster than gradient-free approaches. However, even though most optimization algorithms use gradient information, not many gradient-based algorithms have been examined in the calibration literature. To the authors' knowledge, the only gradient-based algorithms that have been considered are simultaneous perturbation stochastic approximation (SPSA) (see \cite{70}), Lindo's Multistart (a commercial solver's global optimization routine, see \citealt{68}), and sequential quadratic programming using filtering (SQP-filter, see \citealt{72}). 
In \cite{60} SQP-filter was found to give better solutions than NM or GA, as well as solve the problem up to 10 times faster; even better results were found when it was additionally combined with gradient-free algorithm DIRECT \citep{62}. \cite{23} did not consider computational speed, but found the Multistart algorithm had a much higher chance of finding the true global minimum compared to GA or NM. 

Even if one only knows how to explicitly evaluate the objective function $F$, it is not necessary to use gradient-free optimization because the gradient (the vector $dF/dp$) can be calculated numerically. To do this, one varies the parameters $p$ one at a time, and recomputes the objective each time a parameter is varied. Then the gradient can be calculated using finite differences. We refer to this as the finite difference approach for computing the gradient, and it requires us to recompute the simulated trajectories of every single vehicle $m$ additional times, where $m$ is the number of parameters. When using a forward euler scheme with stepsize $\Delta t$ for an ODE model, the computation cost for computing the simulated trajectories once is $\mathcal{O}(T(n))$, where $T(n) = \sum_{i=1}^n(T_{i-1} - t_i)/ \Delta t$. So computing the gradient with finite differences has a complexity of $\mathcal{O}(mT(n))$. 
The adjoint method has complexity $\mathcal{O}(T(n))$, and so it is the prefered method for computing the gradient when $m$ is large \citep{22,18}.

\section{Adjoint Method and Gradient-Based Optimization}\label{adjoint}
\noindent In this paper we are primarily concerned with applying the adjoint method to Eq. $\eqref{5}$ in order to efficiently evaluate the gradient $dF/dp$ for an ODE model. The gradient can then be used with any gradient-based optimization routine. In this section we derive the gradient and compare it to using finite differences; we consider different optimization algorithms in the next section. \\
The adjoint method can also be applied to the calibration problem $\eqref{6}$ where the model includes time delay. It can also be used to calculate the hessian. Both of these extensions follow the same techniques elaborated in this section, and are included in the appendix.  \\
We first consider the calibration of a single vehicle ($n =1$) before dealing with the general case of an arbitrary number of vehicles.
\subsection{Adjoint Method for ODEs}\label{gradient ODE 1}
In our notation, $x_i$, $h_i$, and $p$ are vectors, and $f$ and $F$ are scalars. A derivative of a scalar with respect to a vector is a vector, so for example $\partial f / \partial x_i = [\partial f / \partial x^*_i, \partial f / \partial v_i]$. A derivative of a vector with respect to a vector is a matrix, where the rows correspond to the vector being differentiated, and the columns correspond to the vector being differentiated with respect to. For example, 
\begin{align*} 
\dfrac{\partial x_i}{\partial p} = \begin{bmatrix} \partial x_i^*/\partial p^1 & \partial x_i^* / \partial p^2 & \ldots & \partial x_i^* / \partial p^m \\
\partial v_i / \partial p^1 & \partial v_i / \partial p^2 & \ldots & \partial v_i / \partial p^m \end{bmatrix}
\end{align*}
where $p^j$ represents the $j{\rm th}$ parameter. Additionally, the superscript $^T$ indicates transpose.\\
\noindent Define the augmented objective function, denoted $L$
\[ L = \int_{t_n}^{T_{n-1}} dt \bigg[ f(x_n, \hat x_n, t) + \lambda^T(t)\big( \dot x_n(t) - h_n(x_n(t),x_{L(n)}(t),p_n) \big) \bigg] \stepcounter{equation}\tag{\theequation}\label{10}\]
in this equation, $\lambda(t)$ is referred to as the adjoint variable. $\lambda(t)$ has the same dimension as the state variable $x_n(t)$, so it is a length two vector for a second order model. Note that since $\dot x_n - h_n( \cdot) = 0$, we have that $L = F$. These equalities are always true regardless of the choice of $\lambda(t)$. \\
Following the observation that $L = F$, it follows that $dL/dp = dF / dp$, so differentiating Eq. \eqref{10} with respect to the parameters $p$ gives
\[ \dfrac{d}{d p} F = \dfrac{d}{d p} L = \int_{t_n}^{T_{n-1}} dt \bigg[ \dfrac{\partial f}{\partial x_n}\dfrac{\partial x_n}{\partial p} + \lambda^T(t) \bigg( \dfrac{\partial \dot x_n}{\partial p} - \dfrac{\partial h_n}{\partial x_n}\dfrac{\partial x_n}{\partial p} - \dfrac{\partial h_n}{\partial p}\bigg) \bigg] \stepcounter{equation}\tag{\theequation}\label{11}\]
where quantities are evaluated at time $t$ if not explicitly stated. Note in this case there are no terms depending on the lead trajectory $x_{L(n)}$ since the platoon consists only of $x_n$ and the lead trajectory is regarded as fixed. \\
Eq. $\eqref{11}$ cannot be computed in the current form because we have no way to calculate ${\partial x_n}/{\partial p}$. The adjoint method works by choosing the the adjoint variable $\lambda(t)$ in a way which allows us to avoid the computation of the unknown ${\partial x_n}/{\partial p}$. When doing this, one will find that the correct choice of $\lambda(t)$ results in solving a similar system to the original constraint, but ``backwards'' (e.g. when the model is an ODE, $\lambda(t)$ will end up being defined as an ODE which is solved backwards in time). The equations which define the adjoint variable are known as the adjoint system. \\
To derive the adjoint system, first apply integration by parts: 
\[  \int_{t_n}^{T_{n-1}} \lambda^T(t)\dfrac{ \partial \dot x_n}{\partial p} dt  = \lambda^T(t) \dfrac{\partial x_n}{\partial p} \bigg|^{T_{n-1}}_{t_n} - \int_{t_n}^{T_{n-1}} \dot \lambda^T(t)  \dfrac{\partial x_n}{\partial p} dt \stepcounter{equation}\tag{\theequation}\label{12}\]
The quantity ${\partial x_n(t_{n})}/{\partial p}$ is known because it simply depends on the initial conditions; based on \eqref{5} it is simply zero. We will choose $\lambda(T_{n-1}) = 0$ to eliminate the ${\partial x_n(T_{n-1})}/{\partial p}$ term. This choice becomes the initial conditions for the adjoint system; note that $L = F$ regardless of how the initial conditions for $\lambda$ are chosen. Now combining Eqs. $\eqref{11}$ and $\eqref{12}$: 
\[ \dfrac{d}{dp} F = 
 \int_{t_n}^{T_{n-1}} dt \bigg[ \left( \dfrac{\partial f}{\partial x_n} - \dot \lambda^T(t)- \lambda^T(t) \dfrac{\partial h_n}{\partial x_n} \right) \dfrac{\partial x_n}{\partial p} - \lambda^T(t)\dfrac{\partial h_n}{\partial p}\bigg]  \]
Then to avoid calculating $\partial x_n / \partial p$ we will enforce: 
\[ \dfrac{\partial f}{\partial x_n} - \dot \lambda^T(t)  - \lambda^T(t) \dfrac{\partial  h_n}{\partial x_n} = 0,  \ \ t \in [T_{n-1}, t_n]   \tag{12}\label{13} \]
Eq. $\eqref{13}$ defines a new differential equation  in $\lambda(t)$. Our choice  $\lambda(T_{n-1}) = 0$ is the corresponding initial condition. Then after solving for $\lambda(t)$ we can calculate the desired gradient: 
\[ \dfrac{d}{dp} F = -\int_{t_n}^{T_{n-1}}  \lambda^T(t) \dfrac{ \partial h_n}{\partial p} \ dt \stepcounter{equation}\tag{\theequation}\label{14}\] 
 
\subsubsection{N Car Platoon}\label{gradient ODE 2}

\noindent Now we will apply the adjoint method to a platoon of arbitrary size, and sketch a general algorithm for computing the gradient. Define the augmented objective function and proceed as before. Again, we suppress $t$ dependence for clarity. 
\[ L = \sum_{i=1}^{n}\left( \int_{t_i}^{T_{i-1}}f(x_i, \hat x_i) + \lambda_i^T\big( \dot x_i - h_i(x_i, x_{L(i)}, p_i)\big) dt + \int_{T_{i-1}}^{T_i} \lambda_i^T ( \dot x_i - \dot{\hat{x}}_i ) dt\right) \stepcounter{equation}\tag{\theequation}\label{15}\]
\begin{align*} 
& \dfrac{d}{dp} F = \dfrac{d}{d p} L  = \sum_{i=1}^{n} \left( \int_{t_i}^{T_{i-1}} \dfrac{\partial f}{\partial x_i}\dfrac{\partial x_i}{\partial p} - \lambda_i^T(t)\dfrac{\partial h_i}{\partial x_i}\dfrac{\partial x_i}{\partial p} - \lambda_i^T(t) \dfrac{\partial h_i}{\partial p}dt \right) \\ 
& +  \sum_{i=1}^{n} \left( \int_{t_{i}}^{T_{i}}\lambda_i^T(t) \dfrac{\partial \dot x_i}{\partial p} - \mathbbm{1}(G(i,t) \neq 0)\lambda^T_{G(i)}(t) \dfrac{\partial h_{G(i)}}{\partial x_i}\dfrac{\partial x_i}{\partial p}dt  \right)
\end{align*}
\noindent Here, $\lambda_n$ corresponds to the adjoint variables for vehicle $n$. Similar to how $L(n,t)$ was defined, $G(n,t)$ is defined as the index of vehicle $n$'s follower at time $t$, so that $x_{G(n)}$ is the following trajectory of vehicle $n$. If vehicle $n$ has no follower at time $t$, \textit{or if the follower is not part of the platoon being calibrated}, let $G(n,t)$ return zero. $\mathbbm{1}$ is the indicator function. When $n>1$, the downstream boundary conditions must be put into the augmented objective function, because a vehicle $i$ may affect its follower in the time $[T_{i-1}, T_i]$, even though it only contributes to the loss function during $[t_i, T_{i-1}]$. \\
Apply integration by parts, take the initial condition as $\lambda_i(T_{i}) = 0$, and gather terms with $\partial x_i / \partial p$ to form the adjoint system: 
\begin{align*} 
& - \dot \lambda_i^T(t) + \mathbbm{1}(t \leq T_{i-1}) \left( \dfrac{\partial f}{\partial x_i} - \lambda_i^T(t)\dfrac{\partial h_i}{\partial x_i} \right) - \mathbbm{1}(G(i,t) \neq 0) \lambda_{G(i)}^T(t) \dfrac{\partial h_{G(i)}}{\partial x_i} = 0, \quad t \in [T_{i}, t_i],\ \forall i \stepcounter{equation}\tag{\theequation}\label{16}
\end{align*}
Comparing this adjoint system to the one for a single vehicle \eqref{13}, the difference here is that there is an extra contribution which occurs from the coupling between vehicles. This coupling term occurs only when a vehicle acts as a leader for another vehicle in the same platoon.  
After solving the adjoint system, 
\begin{align*} 
& \dfrac{d}{dp }F  = \sum_{i=1}^n \left(  - \int_{t_i}^{T_{i-1}} \lambda_i^T(t)\dfrac{\partial h_i}{\partial p}dt  \right)\stepcounter{equation}\tag{\theequation}\label{18}
\end{align*}
All the partial derivatives in Eqs. $\eqref{16}, \eqref{18}$ are computed exactly for a given loss function and car-following model. Solving for $x_i(t), \lambda_i(t), \text{ and } d F / dp$ is done numerically with an appropriate time discretization. The same discretization should be used for all quantities, so if one uses a forward euler scheme for $x_i$, the same forward euler scheme should be used for $\lambda_i$ and the integral in $F$ should use a left Riemann sum. Note as well that for a multi-regime model, one should keep the regime for each timestep in memory, because otherwise the switching condition $g_{i,n}$ will have to be evaluated at each timestep when computing the adjoint variables. \\
We define a \textbf{general algorithm for computing $F$, $\dfrac{d}{dp} F$:} \\
\textbf{Inputs: }measured trajectory data $\hat x_1(t), \ldots, \hat x_n(t)$, any necessary lead vehicle trajectories $x_{L(i)}(t)$ for $i \in [1, \ldots, n], \ L(i, t) \notin [1, \ldots, n]$. Models $h_1, \ldots, h_n$ with parameters $p_1, \ldots, p_n$.
\begin{enumerate}
\item For each vehicle $i \in [1, \ldots, n]$:
\begin{enumerate}
\item Compute simulated trajectory $x_i(t)$, $t \in [t_i, T_{i-1}]$ by solving $\dot x_i(t) = h_i( x_i, x_{L(i)}, p_i)$ with intial condition $x_i(t_i) = \hat x_i(t_i)$. 
\item Compute $x_i(t)$, $t \in [T_{i-1}, T_i]$ using downstream boundary conditions Eq. \eqref{7}
\item Keep in memory: $x_i(t)$, $G(i,t)$, and the regime of $h_i$ at each time
\end{enumerate}
\item Compute objective $F = \sum_i^n\int_{t_i}^{T_{i-1}} f( x_i, \hat x_i, t) dt$
\item For each adjoint variable $i \in [n, \ldots, 1]:$ compute $\lambda_i(t)$, $t \in [T_i, t_i]$ by solving adjoint system Eq. $\eqref{16}$ with initial condition $\lambda_i(T_{i}) = 0$.
\item Compute gradient $dF / dp = -\sum_{i}^n\int_{t_i}^{T_{i-1}}\lambda_i^T\partial h_i / \partial p \ dt$
\end{enumerate}
\textbf{Outputs: } $F$, $d F / dp$
\subsubsection{Sufficient Conditions for a Continuous Objective and Gradient} \label{continuous}
\begin{theorem}
The following are sufficient conditions for $F$ and $dF / dp$ to be continuous:
\begin{enumerate}
\item the loss function $f(x_i, \hat x_i)$ and its partial derivative $\partial f/ \partial x_i$ are piecewise continuous on $[t_i, T_{i-1}]$ for all $i$
\item the car following model $h_i ( x_i, x_{L(i)}, p_i )$ and its partial derivatives $\partial h_i / \partial x_i$, $\ \partial h_i / \partial p$, and $\partial h_i / \partial x_{L(i)}$ are piecewise continuous on $[t_i, T_{i-1}]$ for all $i$
\item $\partial h_i / \partial x_i$ must be able to be bounded by a constant 
\item for a multi-regime model, the Eq. \ref{condition} must be satisfied
\end{enumerate}
where it is additionally assumed that $x_{L(i)}(t), L(i,t) \notin [1, \ldots, n]$ (any lead vehicle trajectories which are required but not simulated) are piecewise continuous and $\hat x_i(t), i \in [1, \ldots, n]$ are continuous.
\end{theorem}
\begin{corollary} 
If in addition to $F$ and $dF / dp$ being continuous, $\partial f / \partial x_i, \ \partial h_i / \partial x_{L(i)}$ and $\partial h_i / \partial p$ are bounded, then $F$ is Lipschitz continuous. 
\end{corollary}
We conclude that a multi-regime model need not be continuous when switching between regimes, and lane changing, which will always cause jump discontinuities in $h_i(x_i, x_{L(i)}, p_i)$, is allowed. Note that by piecewise continuity we mean a model (or loss function) can switch regimes only a finite number of times in a finite time interval (and each regime is continuous); additionally, the time interval of each regime must be continuous with respect to the model parameters. Refer to the the full proof in the appendix for further details. We also show in the appendix that the objective is Lipschitz continuous for the OVM, which was the model used for the numerical experiments in this paper.

\subsection{Implementation of Adjoint Method and Calibration Problem} \label{implementation}
To test the adjoint method and optimization algorithms, the calibration problem Eq. \eqref{5} was implemented in python for the optimal velocity model. The equations below show the functional form of the model as originally formulated in \cite{51} and \cite{52}. 
\begin{align*}
&  \ddot x_n(t) = c_4(V(s) - \dot x_n(t))  \\ 
& V(s) = c_1[ \tanh(c_2 s - c_3 - c_5) - \tanh(-c_3)] 
\end{align*}
$c_1$ through $c_5$ are the 5 parameters of the model. $s$ is the space headway, defined as $s = x_{L(i)} - x_i - l_{L(i)}$, where $l_i$ is the length of vehicle $i$. All vehicles have their own unique parameters, so the total number of parameters in the optimization problem is 5 times the number of vehicles. 

The optimal velocity model and its adjoint system were both discretized with a forward euler scheme. The data used was the reconstructed NGSim data \citep{29}, so a time step of .1 seconds was used to be consistent with this data. The loss function $f$ used was square error, which was discretized with a left Riemann sum. 
\begin{align*} 
f(x_i, \hat x_i, t) = (x_i^*(t) - \hat x_i^*(t))^2 \\
F = \sum_{i=1}^n \sum_{t = t_i}^{T_{i-1}} f(x_i, \hat x_i, t)
\end{align*}
which is equivalent to minimizing the root mean square error (RMSE)
\begin{align*} 
\text{RMSE} = \sqrt{\dfrac{F}{\sum_{i = 1}^n T_{i-1}- t_i}}
\end{align*}
Since only a car following model is being calibrated, any lane changes in the measurements will be the same lane changes in the simulation. This also means that the leader/follower relationships in the simulation are the same as they are in the data. This allows trajectories to be calibrated at the level of individual vehicles, since it does not make sense to directly compare the simulated and measured trajectories of a vehicle if the simulation and measurements have different lead vehicles. 

\subsection{Adjoint method compared to other methods for calculating the gradient}

Many optimization algorithms will automatically use finite differences to compute the gradient if an explicit gradient function is not given. Simultaneous perturbation is another option when using an optimization algorithm like the one outlined in \cite{70}. Then, assuming that one has chosen to use a gradient-based algorithm for calibration, it is of practical interest to investigate how the adjoint method compares to both of these methods. Specifically, how does the speed and accuracy of these three methods compare in practice? 
\subsubsection{Comparison of Speed} \label{adjointspeedsection}
The computational cost of an objective function evaluation is $\mathcal{O}(T(n))$, where $T(n)$ is the number of simulated timesteps, added over all $n$ vehicles. The theoretical computational complexities for the adjoint method is $\mathcal{O}(T(n))$, while forward differences has $\mathcal{O}(mT(n))$, where $m$ is the number of parameters. The simultaneous perturbation method has $\mathcal{O}(kT(n))$ complexity, where the final gradient is the average of $k$ estimates of the gradient. Note that these computational complexities are only for the gradient calculation and do not consider the complexity of whatever optimization algorithm the gradient is used with.

\begin{figure}[h] 
\centering 
\includegraphics[ width=\textwidth]{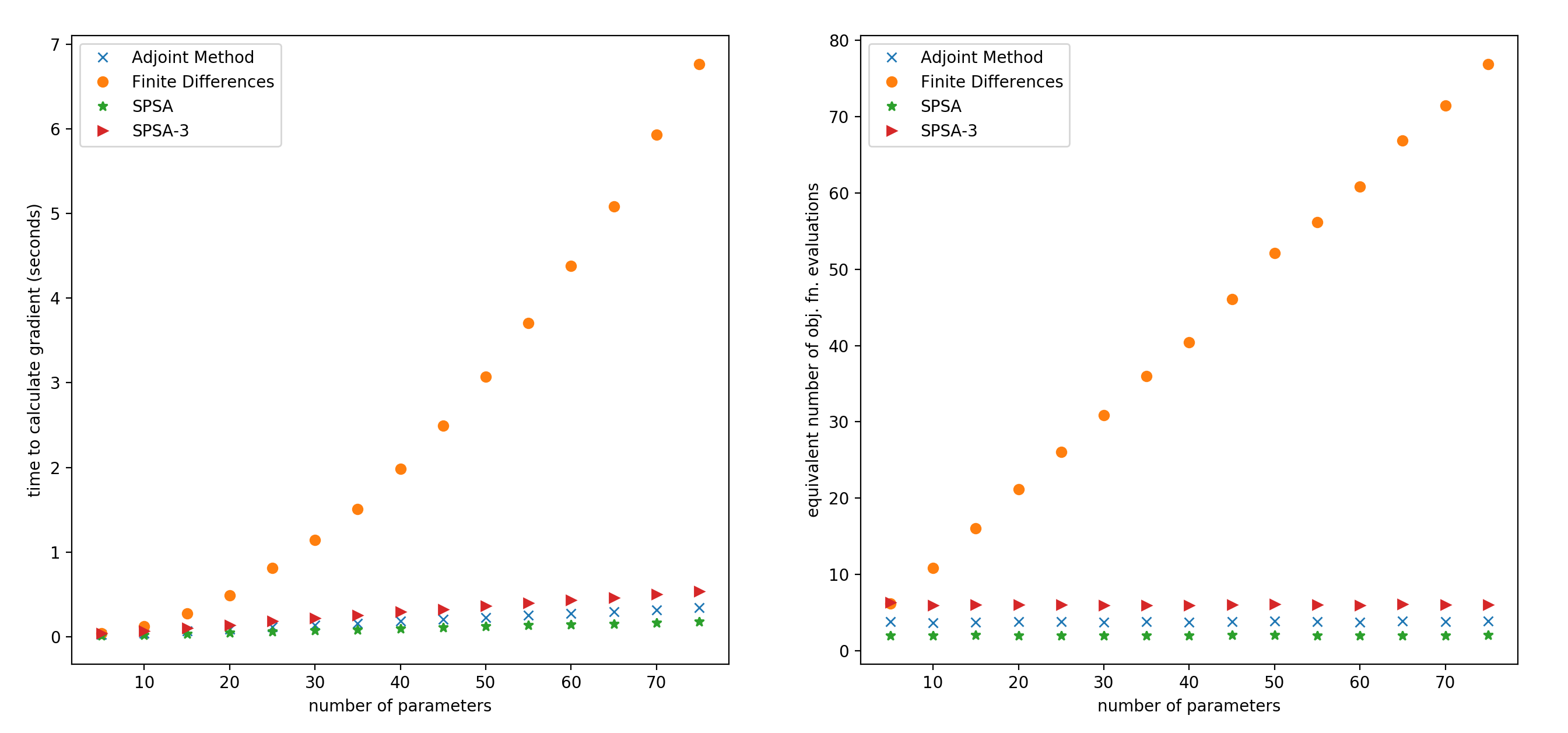} 
\caption{Time for a single evaluation of the gradient for a progressively larger calibration problem. 
The objective varied from the loss from a single vehicle (5 parameters) to 15 vehicles (75 parameters).}  \label{adjointspeed}
\end{figure}

To test these theoretical results in practice, the gradient of the calibration problem was calculated and the platoon size was varied. Since the OVM was used, the total number of parameters is 5 times the number of vehicles. We considered a platoon having between 1 and 15 vehicles. The results are shown in the figure \ref{adjointspeed}. 

\noindent Figure \ref{adjointspeed}'s left panel shows the total CPU time, measured in seconds, needed to calculate the gradient of the calibration problem using the three different methods. The simultaneous perturbation was done with both 1 and 3 trials, denoted SPSA and SPSA-3. The right panel shows the time needed to calculate the gradient divided by the time needed to calculate the objective. To minimize the randomness in the CPU times, each data point was repeated 10 times and the results were averaged. Some variation is still present because some vehicles are observed for shorter times. The same 15 vehicles were used in the experiment and always added in the same order. It should also be stated that for either a finite differencing scheme or for the adjoint method, one needs to first compute the objective before computing the gradient. Therefore, it should be implicitly understood that ``time to calculate the gradient", really refers to ``time to calculate both the objective and gradient" for those two methods. 
\noindent This shows that using a finite difference method quickly becomes intractable when the number of parameters increases. The time increases quadratically in the left panel because both $m$ and $T(n)$ are increasing. In the right panel, when the time needed to calculate the gradient is divided by the time needed to calculate the objective, one sees the complexity growing as a linear function of $m$. In that figure, we also see that the complexities for the adjoint method and simultaneous perturbation do not depend on $m$. 

\noindent The table \ref{speedtable} shows the cpu times for 5-15 parameters. 
One can observe that this implementation of the adjoint method for OVM requires the equivalent of 4 objective function solves, whereas forward differences requires $m+1$ (where there are $m$ parameters).

\begin{table}[h]
\begin{adjustwidth}{-2cm}{0cm}
\caption{Table of speeds for adjoint method compared to finite differences for computing the gradient. Relative time refers to the equivalent number of objective function evaluations.  } \label{speedtable}
\begin{tabular}{|l|l|l|l|l|l|}
\hline
Parameters & Obj. time (s) & Adjoint time (s) & Adjoint relative time  & Finite relative time \\ \hline
5                       & .0062         & .0250            & 4.03                 & 6.3         \\ \hline
10                      & .0109         & .0438            & 4.02                & 11.5        \\ \hline
15                      & .0172         & .0686            & 3.99                  & 16.3        \\ \hline
\end{tabular}
\end{adjustwidth}
\end{table}

\subsubsection{Comparison of Accuracy} \label{adjointacc}

For the calibration problem, there is no way to calculate the gradient exactly because there is no closed form solution for the car following model. To evaluate the accuracy of the adjoint method and simultaneous perturbation, we will therefore treat the result from forward differences as the true solution. In the experiment, points were evenly sampled between an initial starting guess and the global minimum of the calibration problem for a single vehicle. At each of these points in the parameter space, the relative error was calculated
$$\text{relative error} = \dfrac{||\nabla_{fd}F - \nabla F||_2}{||\nabla_{fd}F||_2}$$ 
where $\nabla_{fd}F$ is the gradient calculated with forward differences, $\nabla F$ is the gradient calculated with the adjoint method or simultaneous perturbation, and $||\cdot ||_2$ denotes the 2 norm. The results of the experiment are shown below.

\begin{figure}[h] 
\centering 
\includegraphics[ width=\textwidth]{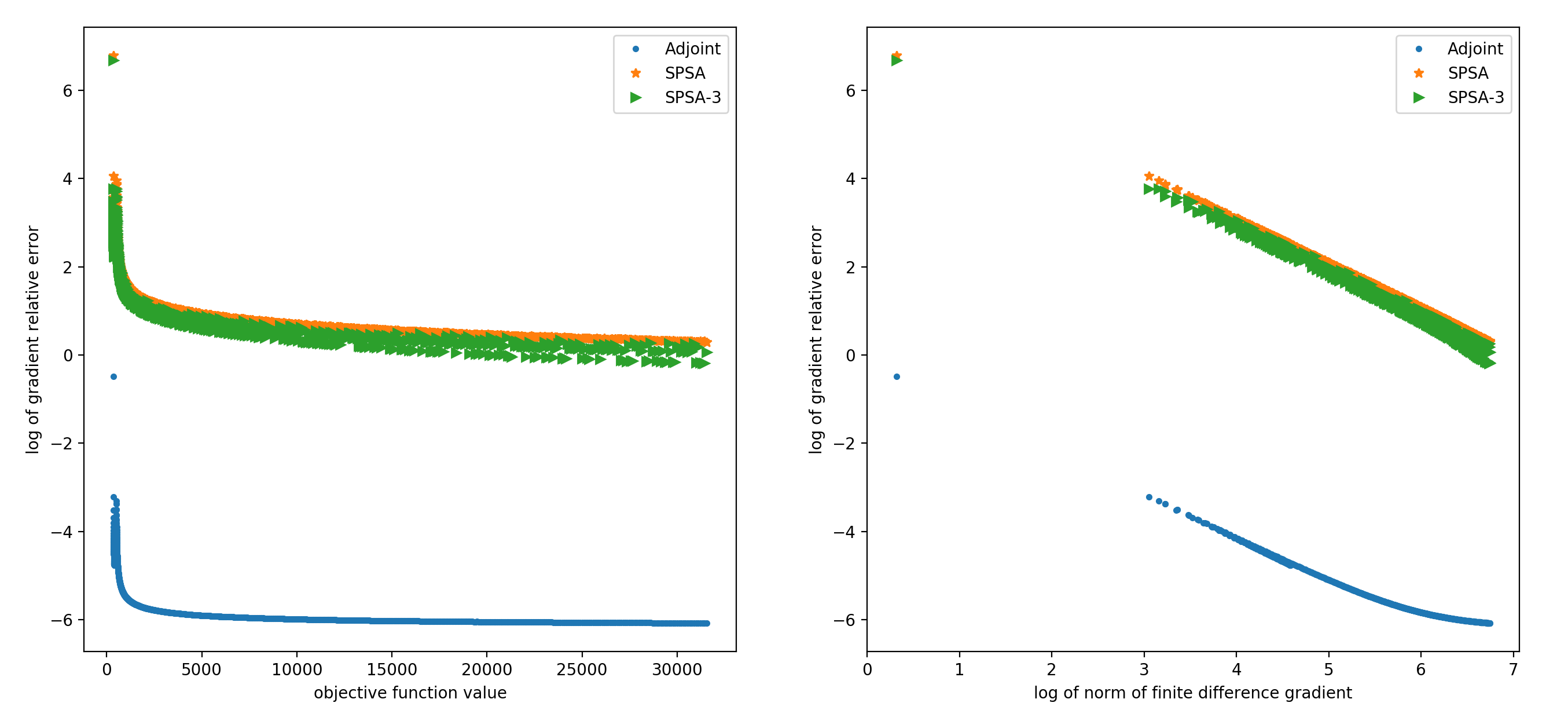} 
\caption{Compares the accuracy of the gradient for the adjoint method and simultaneous perturbation. The finite differences gradient is treated as the exact gradient. } 
\end{figure}

\noindent For the adjoint method, the relative error is usually small, on the order of $10^{-6}$. When the parameters start to become close to a local minimum, the relative error starts to rapidly increase to the order of $10^{-4}-10^{-3}$. When at the optimum, the relative error becomes as high as .3. The reason why the relative error increases is because as the parameters approach the minimum, the norm of the gradient decreases. The right panel shows the relationship between the log of the norm of the gradient calculated with finite differences, and the log of the relative error. One can see that they are inversely proportional to each other. This makes theoretical sense: we expect the residual $\nabla_{fd}F-\nabla F$ to always be on approximately the same order of magnitude since it depends mainly on truncation error which does not change with the parameters. Then, it follows that the relative error is dominated by the $||\nabla_{fd}F||$ term in the denominator, which is what we observe. 

\noindent The relative error for simultaneous perturbation follows the same shape as the adjoint method, but is always about 6 orders of magnitude larger. Of course, you would not use simultaneous perturbation as a substitute for the finite difference gradient, since it is used with its own optimization algorithm. Nonetheless, this shows that although the two methods (adjoint and simultaneous perturbation) both have a flat computational cost with the number of parameters, they have different uses. The reason why the the error for simultaneous perturbation is quite high in this context is because the gradient is scaled very poorly for the OVM model, meaning the parameters have very different sensitivities. This poor scaling is a common feature for car following models. 

\section{Benchmarking different Algorithms and the Adjoint method} \label{algos}
 The ideal optimization algorithm would be one that can not only solve the problem quickly, but also one that is able to avoid bad solutions corresponding to local extrema. Because existing literature has focused on gradient-free optimization algorithms, not much is known about using gradient/Hessian-based optimization to solve the calibration problem. The only works the authors know of in the car following calibration literature that consider gradient based optimization are \cite{60} and \cite{23}. Both those works found the gradient based algorithms to work best overall. This result is consistent with other engineering applications; for example \cite{64} and \cite{65} deal with aerodynamic shape optimization and material science respectively, and report that gradient-based optimization can offer significant speed increases.

To establish the benefits of gradient based optimization and the adjoint method, we consider 2 gradient-free and 5 gradient-based optimization algorithms. The different algorithms, their specific implementations, and abbreviations used throughout the rest of the paper, are detailed below. 
\begin{itemize}
\item Genetic Algorithm (GA) -  an initial population of trial solutions is randomly selected throughout the parameter space. In each iteration, the population is altered through stochastic processes referred to as ``mutation" and ``crossover", and then updated in a selection step. An implementation following \cite{73} is part of the scipy package in python. 
\item Nelder-Mead Simplex (NM) - a given initial guess is used to define a collection of points (a simplex). In each iteration, the corners of the simplex are updated according to deterministic rules. Implemented in scipy following \cite{74}. NM works on unconstrained problems, so a large penalty term was added to the objective if the parameters went outside the bounds.
\item Limited memory BFGS for bound constraints (BFGS) - at every iteration, gradient information from the previous iterates is used in the BFGS formula to form an approximation to the hessian. A quadratic approximation to the objective is formed, and the generalized cauchy point of this approximation is then found, which defines a set of active bound constraints. A minimization is then performed over the variables with inactive constraints, while the variables with active constraints are held fixed. The solution to this last minimization finally defines the search direction for the algorithm, and a line search is performed to determine the step length. The algorithm, referred to as l-bfgs-b, is due to \cite{75} and \cite{76} and wrapped as part of the scipy package. 
\item Truncated Newton Conjugate (TNC) - at every iteration, instead of explicitly computing the Newton direction $-\nabla^2 F^{-1} \nabla F$, the conjugate gradient method is applied to the newton system $\nabla^2 F d = -\nabla F$ (where $d$ is the search direction to be determined). Conjugate gradient is itself an iterative algorithm, and it is terminated early (truncated) to give an approximate solution to the newton system which is also guaranteed to define a direction of descent. A line search is then performed to define the step length. The conjugate gradient algorithm only needs hessian-vector products which can be computed efficiently using finite differences, so the full hessian never needs to be calculated. An implementation due to \cite{78} is wrapped as part of scipy. That implementation uses a conjugate gradient preconditioner based on the BFGS update, and the search direction is projected to enforce the bound constraints.
\item Simultaneous Perturbation Stochastic Approximation (SPSA) - At each iteration, the gradient is approximated using simultaneous perturbation. Then the algorithm moves in the direction of the negative gradient with a fixed step length, so no line search is performed. The size of the step length is gradually reduced in subsequent iterations. Implemented in python following \cite{70}; the step sizes and number of iterations were manually tuned by hand. 
\item Gradient Descent (GD) - At every iteration, the negative gradient is computed and scaled according to the spectral step length (also known as the Barzilei Borwein method). The scaled gradient is then projected onto the bound constraints to define the search direction, and a nonmonotone backtracking linesearch is used to define the step length. Implemented according to \cite{77} in python. The algorithm was tested with different line searches (backtracking, weak/strong wolfe, watchdog, see \citealt{71}) and the nonmonotone backtracking gave the best performance. 
\item Sequential quadratic programming (SQP) - At every iteration the hessian is computed and the search direction is defined by the newton direction projected onto the bound constraints. This algorithm explicitly computes the hessian instead of approximating it. Because the hessian may fail to be positive definite, the search direction is safeguarded so that if the newton direction fails to define a direction of descent, the algorithm instead searches in the direction of the negative gradient. A small regularization is also added to the hessian so that it can always be inverted. The algorithm is a simple python implementation of sequential quadratic programming based on \cite{71}. We again found the nonmonotone backtracking linesearch to be most effective. 
\end{itemize}

In summary, out of all the algorithms tested: NM and GA are gradient-free; SPSA and GA are stochastic; BFGS and TNC are quasi-newton; GD and SPSA do not use hessian information; SQP explicitly computes the hessian; GA is the only global algorithm.

For the TNC, BFGS, GD, and SQP algorithms, we considered two separate modifications for each algorithm. First, either the finite difference method or the adjoint method can be used to compute the gradient, and we denote this by either ``Fin" or "Adj" respectively. Second we considered starting from multiple initial guesses. Multiple initial guesses are specified, and after each calibration, if the RMSE was above a specified threshold, the next initial guess would be used. The same 3 initial guesses were used in the same order for all algorithms requiring an initial guess. So with a threshold of 0, 3 guesses are always used, and with an arbitrarily large threshold, only a single guess will be used. The thresholds of 0, 7.5 and $\infty$ were used, and denoted in reference to the algorithm. The initial guesses used are based off of the calibrated parameters found from \cite{51}.

\subsection{Evaluating Algorithm Performance}
As detailed in section \ref{implementation}, we tested the calibration problem on the OVM car following model using squared error in distance as the loss function (equivalent to minimizing the RMSE). In this experiment, for each algorithm, every vehicle in the reconstructed NGSim dataset was calibrated using the true measurements of its lead trajectory. In total, there were 7 unique algorithms and 23 total algorithms tested when including the variants described above. 

The calibration results are evaluated primarily based on three different measures: the average calibration time for a single vehicle, the \% of the time the algorithm found the global minimum, and the average RMSE of a calibrated vehicle. Since there is no way to know the actual global minimum for the calibration problem, we regard the global minimum for a vehicle as being the best result out of all the algorithms, and we regard an algorithm as having found the global minimum if its RMSE is within 1/12 ft (1 inch) of this best result. 

\begin{figure}[h] 
\centering 
\includegraphics[ width=\textwidth]{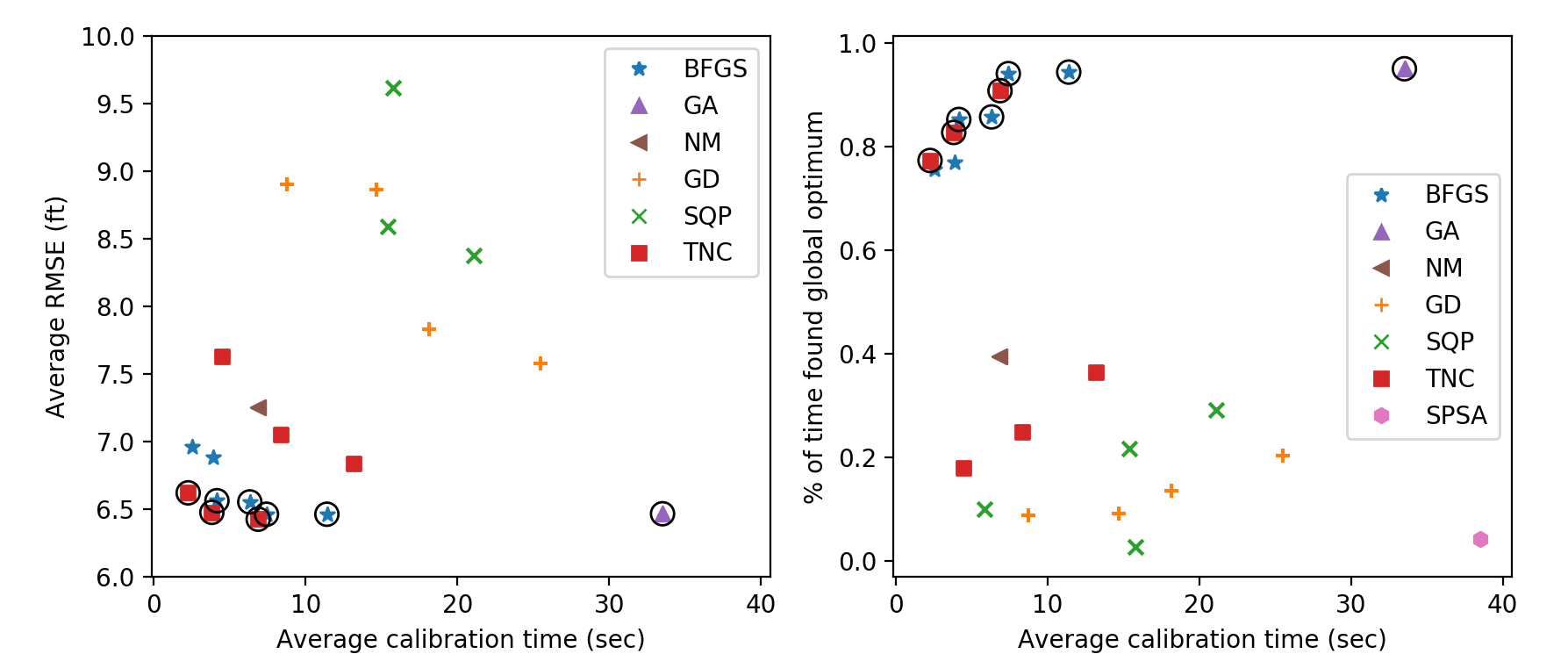} 
\caption{All algorithms plotted by the three metrics time, RMSE, and \% found the global optimum. Algorithms on the pareto front with respet to these three metrics circled in black.}  \label{paretofig}
\end{figure}

Based on these three metrics, all algorithms on the pareto front were identified and are shown in the table \ref{table2}. 
In figure \ref{paretofig} every algorithm is plotted in the RMSE - time and \% found global optimum - time plane. The algorithms on the pareto front are circled in black. 

\begin{table}[h]
\begin{adjustwidth}{-1cm}{0cm}
\caption{Algorithms on the Pareto front. All algorithms tested shown in appendix.} \label{table2}
\begin{tabular}{|l|l|l|l|l|l|l|l|l|}
\hline
Algorithm & \begin{tabular}{@{}c@{}}\% found  \\global opt\end{tabular} &  \begin{tabular}{@{}c@{}} Avg.  \\ RMSE (ft) \end{tabular} &  \begin{tabular}{@{}c@{}} Avg. \\  Time (s)\end{tabular} & \begin{tabular}{@{}c@{}} Avg. RMSE / \% \\  over global opt\end{tabular} &\begin{tabular}{@{}c@{}}Initial  \\ Guesses\end{tabular} & \# Obj. Evals \\ \hline
Adj BFGS-0 & 94.0  & 6.46& 7.42 & .10 / 2.0\% &3 &307.1 \\ \hline
Adj BFGS-7.5 &85.2 &6.56 &4.16   & .20/6.8\%& 1.67 & 165.0\\ \hline
GA & 95.0& 6.47 &33.5  & .10/2.2\% & -  & 5391 \\ \hline
Fin BFGS-0& 94.3 & 6.46 & 11.4  & .10/2.2\% &3 & 311.5   \\ \hline
Fin BFGS-7.5&85.7 & 6.55 & 6.32   & .19/6.3\%&1.66 & 164.8\\ \hline
Adj TNC-0& 90.7  & 6.43 & 6.87 & .05/2.4\% &3 & 285.7  \\ \hline
Adj TNC-7.5& 82.7  & 6.48 & 3.82  & .11/5.0\%& 1.63 & 152.3  \\ \hline
Adj TNC-$\infty$& 77.3 &  6.62& 2.26 & .25/7.7\%& 1 & 94.1  \\ \hline
\end{tabular}
\end{adjustwidth}
\end{table}

The only three algorithms on the pareto front were the GA, TNC, and BFGS. GA gave the best \% of finding the global optimum. TNC with the adjoint method and three guesses gave the best overall RMSE, and was the fastest algorithm overall when used with the adjoint method and a single guess. BFGS was slightly slower than TNC and has a slightly higher RMSE, but it has a higher chance of finding the global optimum. 

\begin{figure}[h] 
\centering 
\includegraphics[ width=\textwidth]{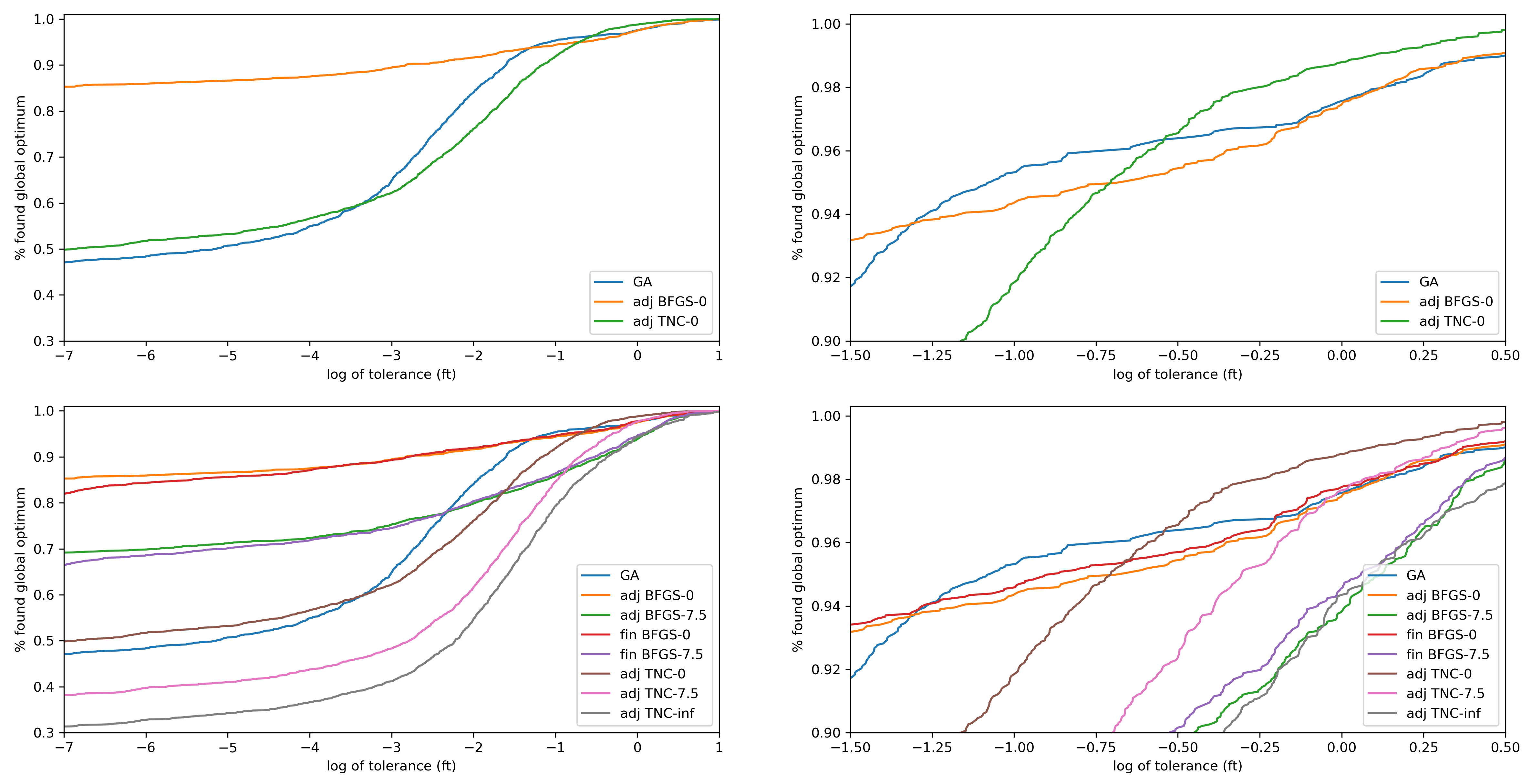} 
\caption{Shows the distribution of RMSE over the global optimum for GA, Adj TNC-0, Adj BFGS-0 in top two panels. Bottom two panels show the distributions over all algorithms on the pareto front (with a tolerance of 1/12 used for obtaining the pareto front). Right panels are zoomed in windows of left panels.} \label{rmsedist}
\end{figure}

The conclusion of what algorithms are on the pareto front depends strongly on how ``finding the global optimum" is defined. Here, we treat an algorithm as having found the global optimum if it gives a result within an inch (1/12) of the best RMSE. Let us refer to this value of 1/12 as the ``tolerance" for finding the global optimum. If a smaller tolerance is used, then GA will no longer be on the pareto front. If a larger tolerance is used, GA and BFGS will no longer be on the pareto front. 
For example, with a tolerance of 0, the pareto front consists of only the adj TNC and adj BFGS variants, and adj BFGS-0 gives the highest chance of finding the global optimum. With a tolerance of 1/2, the adj TNC variants are the only algorithms on the pareto front. The effects of changing the tolerance are shown in the table \ref{table3}. Adj TNC will always be on the pareto front regardless of the tolerance because it gives the best average RMSE for any given speed. No algorithms other than adj TNC, BFGS, and GA are ever on the pareto front. 
\begin{table}[h]
\centering 
\caption{Algorithms on the Pareto front depend on how ``finding the global optimum" is defined (without that metric, TNC with the adjoint method would be the only algorithm on the pareto front). 
Set notation indicates the strategies used for initial guesses.} \label{table3}
\begin{tabular}{|l|l|}
\hline
 \begin{tabular}{@{}c@{}}Tolerance  for  \\global opt (ft) \end{tabular}  &  Algorithms on the pareto front\\ \hline
0 & Adj TNC-\{0, 7.5, Inf\}, Adj BFGS-\{0, 7.5, Inf\} \\ \hline
1/24 & Adj TNC-\{0, 7.5, Inf\}, Adj BFGS-\{0, 7.5, Inf\}, Fin BFGS-\{0, 7.5\} \\ \hline
1/12 & Adj TNC-\{0, 7.5, Inf\}, Adj BFGS-\{0, 7.5\}, Fin BFGS-\{0, 7.5\}, GA \\ \hline
1/4 & Adj TNC-\{0, 7.5, Inf\}, GA \\ \hline
1/2 & Adj TNC-\{0, 7.5, Inf\}\\ \hline
\end{tabular}
\end{table}
The behavior shown in table \ref{table3} is explained by figure \ref{rmsedist}. That figure shows how the \% found global optimum metric changes with tolerance. For example, a tolerance of 1/12 corresponds to -1.08 in a log scale, and the top right panel shows that at that point, the GA has the highest chance of finding the global optimum. The same panel shows that if the tolerance increases, adj TNC-0 will have the highest chance, and that adj BFGS-0 will have the highest chance if the tolerance decreases. 

BFGS typically found the best solution out of all the algorithms tested, but its distribution of RMSE has a heavier tail in the sense that there were also some vehicles for which BFGS couldn't find a good solution. TNC is most consistent because its distribution shows less of a heavy tail (it is the first to reach 100\% found global optimum as the tolerance increases). The GA has a tail similar to BFGS and performs well when only a moderate tolerance is needed. 

The appendix includes a plot that shows the distributions of RMSE (on a non-log scale), and the differences between the three algorithms are essentially indistinguishable at that resolution. 

Overall, these differences between the three algorithms (GA, adj TNC-0, adj BFGS-0) are relatively minor, with all three giving essentially the same average RMSE (GA had a slightly worse RMSE than the other two). It's not clear to the authors whether the subtle differences between them would cause any differences in practice. However, the BFGS and TNC algorithms both give a significant speed increase; the adjoint method with BFGS or TNC is about 5 times faster than the GA on average. Even when using finite differences instead of the adjoint method, BFGS is still about 3 times faster than the GA (finite differences with TNC did not work well). A much larger speed increase can be realized with only a small trade off in accuracy. Adj TNC-Inf had an average RMSE of 6.62 compared to 6.47 for the GA, but is 15 times faster. Adj TNC-7.5 had an average RMSE of 6.48 while being 9 times faster than the GA.

The comparison between the different strategies for initial guesses (0, 7.5, Inf) shows the trade off between speed and accuracy when starting a local search based algorithm multiple times. For car following models, it seems that even a suboptimal local minimum still gives a reasonable accuracy. The difference in accuracy made by starting a algorithm multiple times is smaller than the difference from using another algorithm, so regardless of the initial guess, the local search seems to converge to similar parameters. We also saw that a good local search algorithm (in this case, the quasi-newton algorithms) can give results equivalent to a global search algorithm (the GA) when a small number of initial guesses are used. Of course, each initial guess used greatly increases the time needed for the algorithm, so a hybrid strategy similar to the one proposed in \cite{60} can be considered if a large number of initial guesses are needed to achieve good performance. 

Comparing the results of using the adjoint method to finite differences, the differences between the two methods depends heavily on the algorithm used. For BFGS, using finite differences instead of the adjoint method gives essentially the same result, in the same number of objective/gradient evaluations, and the difference is speeds is consistent with the speed increase found in section \ref{adjointspeed}. For TNC, the algorithm does not perform well when finite differences is used. 

All the attention so far has been concentrated on the TNC, BFGS, and GA algorithms since those gave good results. None of the other algorithms performed well. NM is designed for unconstrained problems, so it is unsurprising that it did not perform as well as an algorithm designed specifically to deal with box constraints (recall that here the box bounds were enforced by adding a penalty term). GD suffered from the problem widely reported in the literature where it takes a very large number of iterations to fully converge. The majority of the time, the algorithm terminates due to exceeding the maximum number of evaluations allowed. For the SQP (recall in this case we explicitly computed the Hessian), we found that the newton direction often fails to define a direction of descent, and the algorithm simply searches in the direction of the gradient instead. Then, the algorithm spends a lot of computational time computing the Hessian but ends up not using it for anything, so explicitly computing the hessian seems to not work poorly for car following calibration; BFGS updating to estimate the Hessian, or a truncated newton method for estimating the newton direction should be used instead. SPSA performed worst out of all algorithms tested because of the vastly different sensitivities for car following models, as well as the rapidly changing magnitude of the gradient. A variant of SPSA such as c-SPSA \citep{cspsa} or W-SPSA \citep{wspsa} may have overcome these issues. 

The analysis in this section has shown the value of gradient based algorithms and the adjoint method; compared to gradient free methods, they can give the same overall performance while offering a significant speed increase. We found that for the calibration of a single vehicle, the truncated newton (TNC) algorithm with the adjoint method gives a slightly more accurate calibration than a genetic algorithm, while also being 5 times faster. TNC is 15 times faster than the genetic algorithm if $\approx$ 2\%  decrease in accuracy is acceptable (when only a single initial guess is used). Actually, as we will see in the next subsection, when the calibration problem becomes larger, the benefit of gradient based algorithms and the adjoint method becomes even larger.

\subsection{Calibration of larger platoons}\label{platoonsize}
\noindent Define platoon size as the number of vehicles calibrated at the same time ($n$ in Eq. \eqref{5}). The simplest case is when $n=1$: a single vehicle is calibrated at a time (this was the strategy used in the above section). Taking $n=1$ is how most of the literature treats car following calibration. Moreover, trajectories are typically calibrated to the measured, as opposed to simulated, lead trajectory. To the authors' knowledge, \cite{19} is the only paper which has taken a platoon size of larger than 1 ($n=2$ in that paper) and \cite{recurrentcf} is the only paper which considers the issue of predicting on the simulated vs. measured lead trajectory. In actual traffic simulations, all vehicles are simulated at the same time, and measured trajectories aren't assumed to be available. In that case, errors from one vehicle can accumulate in the following vehicles and produce unrealistic effects. This motivates the question of calibrating platoons of vehicles, where each vehicle uses the simulated lead trajectory.

In this section, we are interested in two questions regarding the calibration for a platoon size of $n>1$. First, for the different methods considered so far (gradient free, gradient based with finite differences, adjoint method) how does the time needed to solve the calibration problem scale with the number of parameters. Second, what is the benefit (in terms of RMSE) from considering a larger platoon of vehicles.

A platoon of 100 vehicles was formed from the NGSim data. The platoon was formed by randomly selecting the first vehicle to be calibrated, and then repeatedly adding vehicles which have their leaders in the platoon. We considered calibrating these 100 vehicles with a platoon size $n = 1, 2, \ldots 10$. When the platoon size doesn't divide 100, the last platoon used will be the remainder (e.g. for $n = 3$, the last platoon is of size 1). Here the calibration is done sequentially, so each vehicle is calibrated to the simulated trajectory of its leader. 

\begin{figure}[h] 
\centering 
\includegraphics[ width=.65\textwidth]{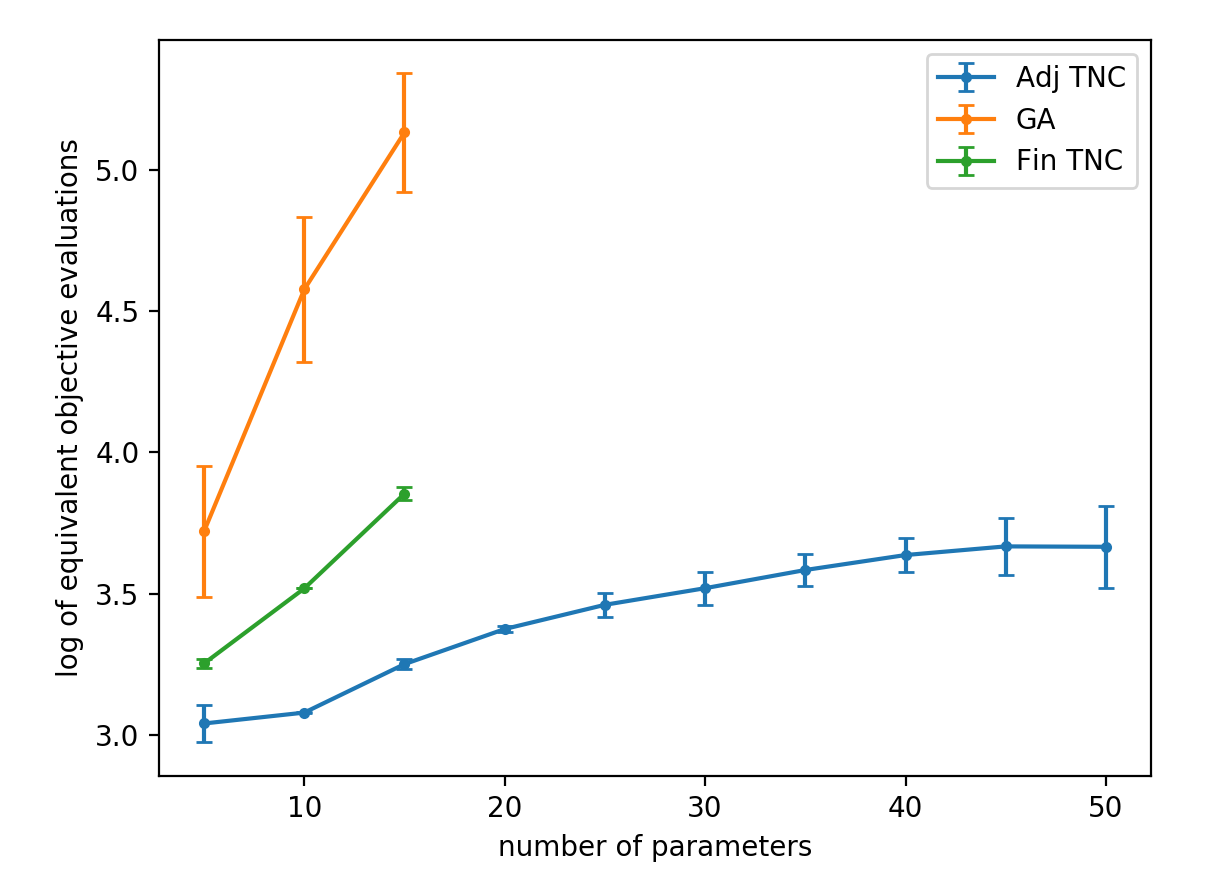}  
\caption{Log (base 10) of equivalent number of objective evaluations required to solve the calibration problem for an increasing platoon size ($n)$. The number of parameters in each platoon is $5n$. The brackets show the standard deviation for the data points. The work required to solve the calibration problem increases at a much faster rate when using gradient free optimization or finite differences compared to the adjoint method.  } \label{speed2}
\end{figure}

To compare how different algorithms scale, we performed calibration using the GA, Adj TNC, and Fin TNC algorithms, as shown in figure \ref{speed2}. For each platoon of a specific size, the number of equivalent objective evaluations is recorded. Gradient evaluations are converted into objective evaluations following the results of \ref{adjointspeedsection} (adjoint method can compute the gradient and objective in the equivalent of 4 objective evaluations, finite differences computes the gradient and objective in the equivalent of $m+1$ objective evaluations, where $m$ is the number of parameters). The results are shown in figure \ref{speed2}.

\begin{table}[h]
\centering
\caption{Compares the number of equivalent objective evaluations for the calibration problem for an increasing number of parameters, relative to the Adj TNC-0 algorithm. }
\begin{tabular}{|l|l|l|l|l|l|}
\hline
\# parameters & Adj TNC & Fin TNC &  \begin{tabular}{@{}c@{}}Fin TNC  \\relative evals. \end{tabular} & GA & \begin{tabular}{@{}c@{}}GA \\relative evals. \end{tabular}     \\ \hline
5             & 1098    & 1790 & 1.63   & 5256  & 4.79  \\ \hline
10            & 1200    & 3300  & 2.75  & 37715 & 31.4\\ \hline
15            & 1780    & 7139 & 4.01  & 135485 & 76.1\\ \hline
\end{tabular}
\end{table}

Assuming that a gradient based algorithm converges in approximately the same number of gradient/objective evaluations for either using finite differences or the adjoint method, the equivalent number of objective evaluations can be easily converted between the two methods. Namely, if using the adjoint method requires $g(m)$ total computational effort for computing the gradient, for some arbitrary function $g$, then finite differences will require $\frac{m+1}{4} g(m)$. So using finite differences will always scale an order of $m$ worse than the adjoint method. 
For the GA, it is clear that its cost is growing extremely fast with the number of parameters, and we only considered a platoon up to size 3 because of this reason. It was about 75 times as expensive as using TNC with the adjoint method for a problem size of only 15 parameters. 
We conclude that when the number of parameters becomes large, using the adjoint method with gradient based optimization can offer an arbitrarily large speed increase compared to either gradient free optimization or finite differences. 

\begin{figure}[h] 
\centering 
\includegraphics[ width=\textwidth]{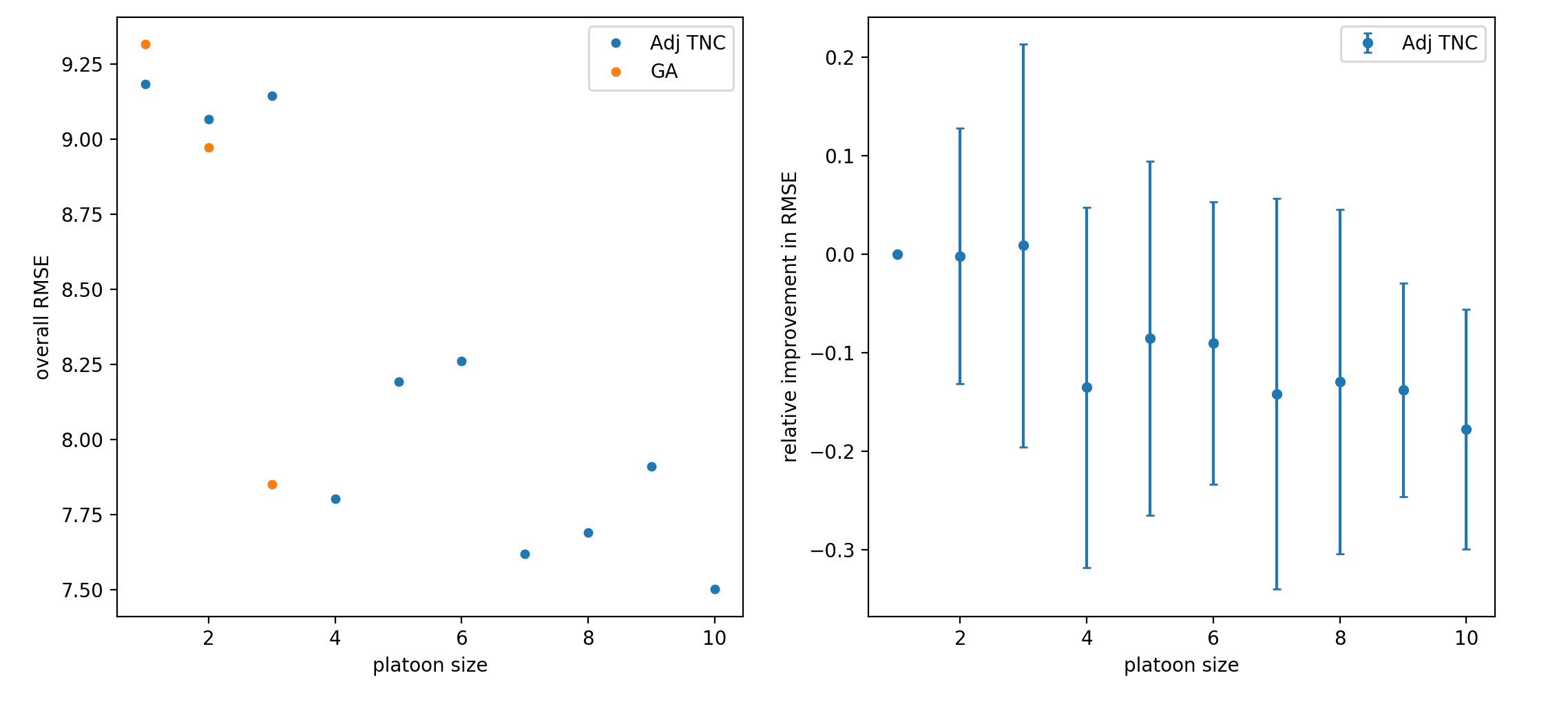} 
\caption{Left panel shows overall RMSE when using a varying platoon size for the calibration problem. Right panel shows the percentage improvement from using the larger platoon sizes compared to calibrating a single vehicle at a time. The bars show the standard deviation for each point. } \label{platacc}
\end{figure}
As for the effect of platoon size on the overall RMSE of all 100 vehicles, the results are shown in figure \ref{platacc}. Note overall RMSE is different than average RMSE because vehicles are weighted according to how many observations they have (see section \ref{implementation}). Also, the RMSE is expected to be higher here compared to the previous section since the calibration is being done sequentially using the leader's simulated trajectory instead of the leader's measured trajectory. The left panel shows the overall RMSE versus platoon size. The right panel shows the relative improvement in RMSE. To calculate the relative improvement, the percent change in RMSE compared to a platoon size of 1 is calculated for a specific platoon. Then the average and standard deviation of the percent change is calculated over all platoons of the specific size (e.g. 33 size 3 platoons or 10 size 10 platoons). The standard deviations are quite large: for any specific platoon, the improvement could be as large as 50\%, or the RMSE could sometimes be significantly worse. For a platoon size $n = 2$ or $3$, we saw a neglible overall improvement. In both those cases, we saw that the majority of platoons had an improvement of around $8\%-9\%$, but there were some platoons which achieved a significantly worse fit (20\%-40\%+ worse). These significantly worse fits suggest that in those cases, the algorithm was converging to bad local minima which apparently were avoided when calibrating vehicles one at a time. As the platoon size increased, the improvement became more consistent, and for platoon size $n = 9$ and $10$ we saw an overall $13.8\%$ and $17.8\%$ improvement respectively. The authors stress that this improvement was achieved without changing the algorithm or the model at all. The only thing that was changed was the platoon size ($n$ in Eq. \eqref{5}). 

Overall, these results suggest that considering a larger platoon size can improve calibration results, and further research on larger platoon sizes should be considered. In figure \ref{platacc} we saw that the GA starts to perform better than TNC when the platoon size increases. We also saw that for larger platoon sizes, there were some platoons which ended up with a significantly worse fit compared to the case of $n=1$. This suggests that the local search algorithm (TNC) was having problems converging to bad local minima for $n>1$. We therefore conclude that while using multiple guesses was adequate for the calibration of a single vehicle, a more sophisticated strategy is needed when multiple vehicles are being calibrated.

\section{Conclusion}\label{conclusion}
We considered an optimization based formulation of the calibration problem for car following models being calibrated to trajectory data. It was shown how to apply the adjoint method to derive the gradient or hessian for an arbitrary car following model formulated as either an ordinary or delay differential equation. Several algorithms for solving the calibration problem were compared, and it was found that the best overall algorithms were the genetic algorithm, l-bfgs-b, and truncated newton conjugate. 

For the calibration of a single vehicle at a time, it was found that using the adjoint method and a quasi-newton method gives slightly better performance than a genetic algorithm, and is 5 times faster. As the number of parameters increases, the speed increase offered by the adjoint method and gradient based optimization can become arbitrarily large (for 15 parameters, a truncated newton algorithm using the adjoint method was 75 times faster than a genetic algorithm). The adjoint method will always perform better than finite differences as the number of parameters becomes large, because computing the gradient with the adjoint method has a flat cost with respect to the number of parameters, whereas using finite differences has a cost which increases linearally with the number of parameters. Numerical experiments used the reconstructed NGsim data and the optimal velocity model.  

There are two main directions for future research. In this paper we consider only the calibration of a car following model, while a full microsimulator has several other important modules, such as route choice or lane changing decision models. The question of how to apply the adjoint method to those other components can be considered in the future so that the methodology could be applied to a full microsimulation model. There are also more questions regarding the calibration of car following models. Future research can consider the effect of a larger platoon size on the calibration problem, so that multiple vehicles (which share leaders) are calibrated at a time. Another interesting question is using the new highly accurate vehicle trajectory data \citep{66} to compare and validate different car following models.

\chapter{Applications in Vehicular Traffic Flow: Lane Changing}
\section{Introduction}
Lane changing (LC) is a vital component of traffic dynamics which plays a role in many observed macroscopic phenomena. Recent works have confirmed that shockwaves on the highway are often initiated by lane changing maneuvers. \cite{105} and \cite{106} both examined the same traffic data from Interstate I-80 in the U.S., with the former concluding that all shockwaves observed in that data originated from disturbances caused by lane changing. The latter concluded that 16 out of 18 traffic oscillations in the data were caused by lane changing. An example of lane changing causing a shockwave is shown in figure \ref{lcjamfig}. \cite{106} also looked at another study site, and found 12 out of 35 traffic oscillations were caused by lane changing. These results indicate that understanding lane changing dynamics is necessary for understanding congestion. Another example of the importance of lane changing is \cite{40}, where incorporating lane changing dynamics into a kinematic wave model was found to explain the drop in the discharge rate (i.e. capacity drop) frequently observed at the onset of congestion.  \\
Despite evidence indicating that lane changing plays a key role in the formation and propagation of congestion, it seems that most current works emphasize only the role of car following behavior (e.g. from string instability, see figure \ref{phantomjamfig}), and lane changing models remain a niche area which has received only limited rigorous academic study. \\
One well known feature of lane changing dynamics is the so-called relaxation phenomenon. This refers to the observation that drivers are willing to accept abnormally short spacings at the onset of lane changes, and that these short spacings gradually transition (``relax") back to a normal spacing. Car following models react to this short spacing by decelerating too strongly, causing unrealistic behavior. Fig. \ref{fig1} shows some empirical examples of the relaxation phenomenon, and Fig. \ref{fig2-2} shows an example of the unrealistic behavior which can occur in car following models because of lane changing. \\
This paper proposes a model for the relaxation phenomenon which addresses this unrealistic behavior for an arbitrary model. The relaxation model is meant to augment an existing traffic microsimulation model consisting of both a car following model (which determines the speeds/accelerations of vehicles) and a lane changing model (to determine how/when vehicles should initiate lane changes).
\begin{figure}[H] 
\centering 
\includegraphics[ width=\textwidth]{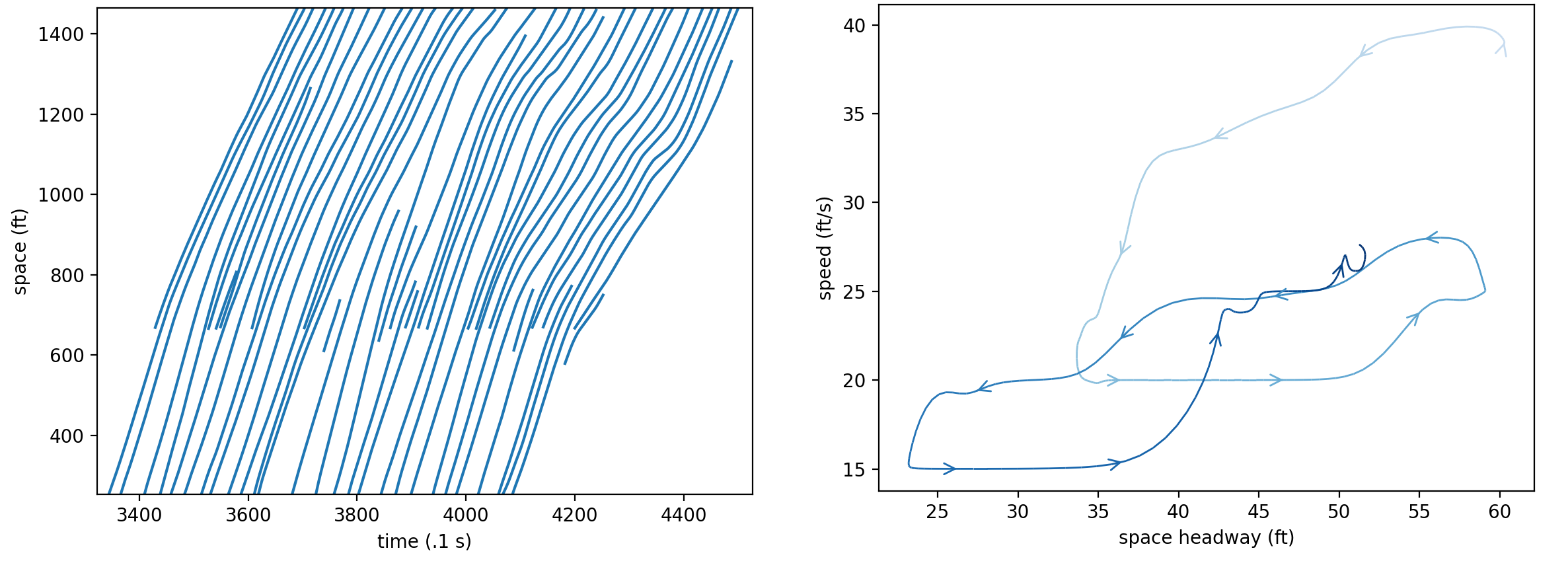} 
\caption{Left panel: space-time plot of vehicle trajectories from lane 6 in the NGSIM I-80 dataset, recorded in congested conditions. There is a merge bottleneck at space = 670 ft. owing to an on-ramp. But despite the abnormally short headways due to merging vehicles, there are no immediate drops in speeds; rather, vehicles gradually relax to a normal spacing (there are also several vehicles which move over to the less congested lanes shortly after merging). \newline
Right panel: speed-headway plot of a typical merging vehicle (id 1336). Darker colors indicate later times, and arrows point in the direction of the trajectory. The vehicle initially merges at a comparatively high speed before transitioning to equilibrium. }  \label{fig1}
\end{figure}
\begin{figure}[H] 
\centering 
\includegraphics[ width=\textwidth]{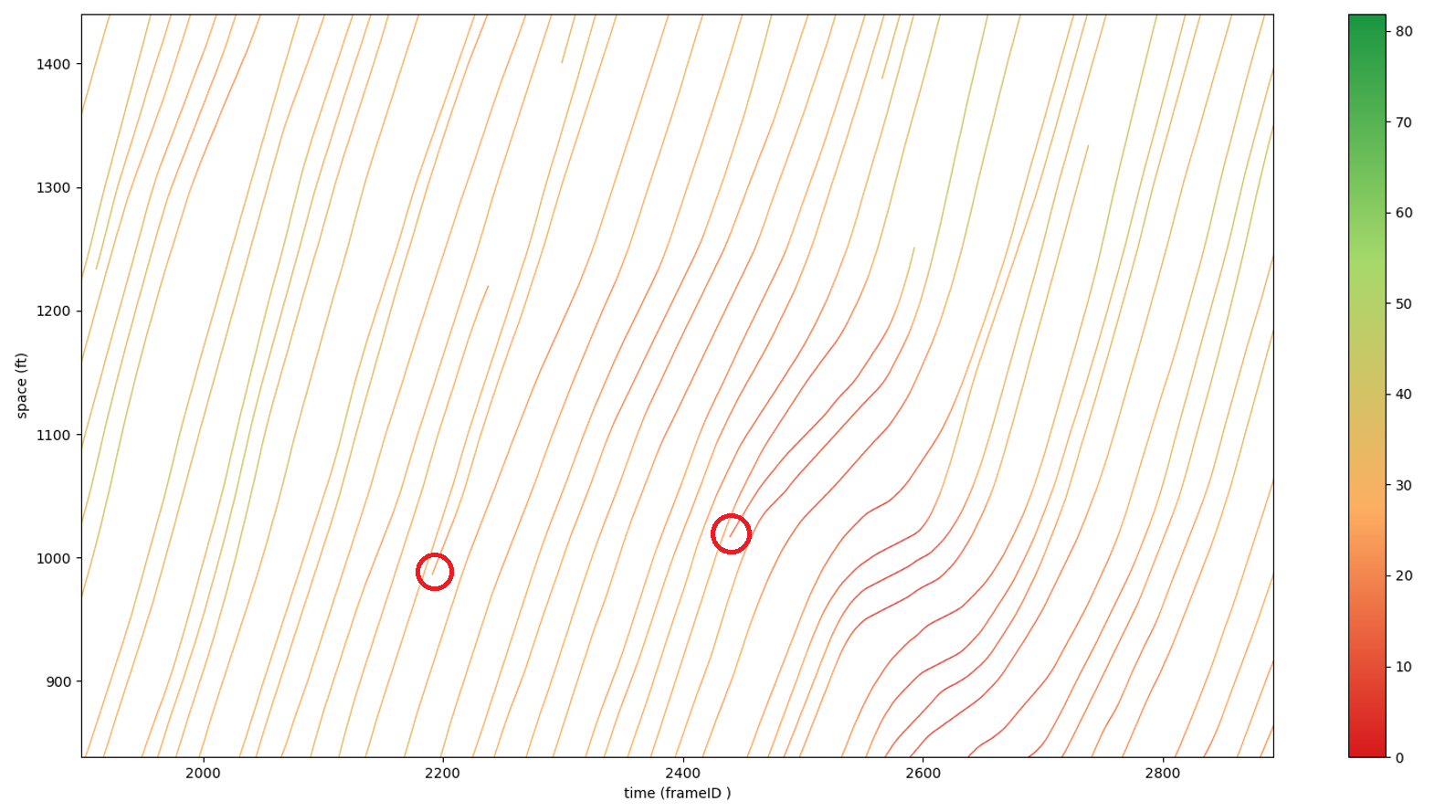}
\caption{An example of lane changing (circled) directly causing the formation of a shockwave in the NGSIM dataset.}  \label{lcjamfig}
\end{figure}
\begin{figure}[H] 
\centering 
\includegraphics[ width=\textwidth]{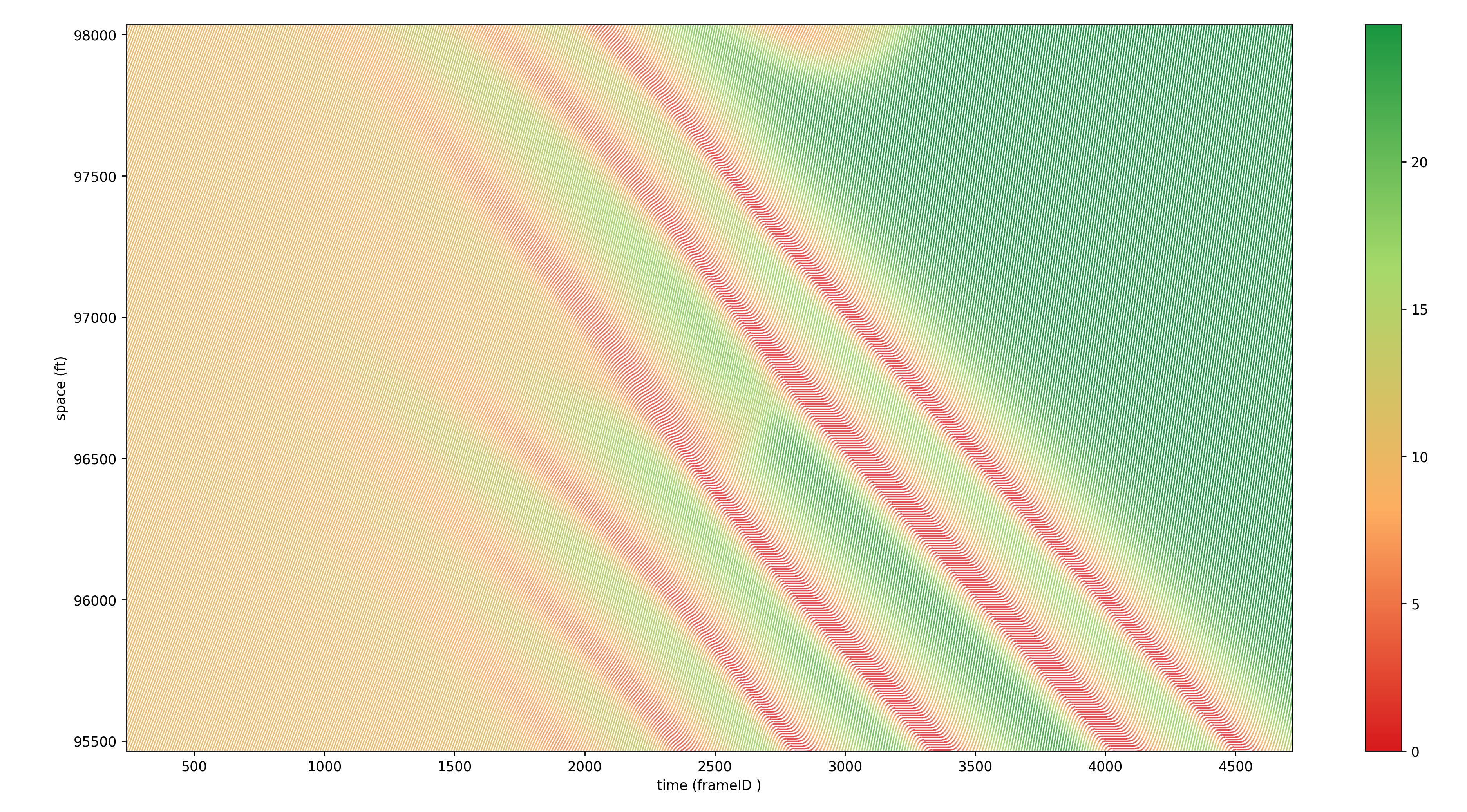} 
\caption{In a simulation run in havsim, vehicles were simulated on a straight road with random noise added to a small number of the leading vehicles. After roughly 6 minutes, this random noise causes the transition from equilibrium flow (left side of figure) to an oscillatory flow with shockwaves (right side of figure).} \label{phantomjamfig}
\end{figure}
\begin{figure}[H] 
\centering 
\includegraphics[ width=\textwidth]{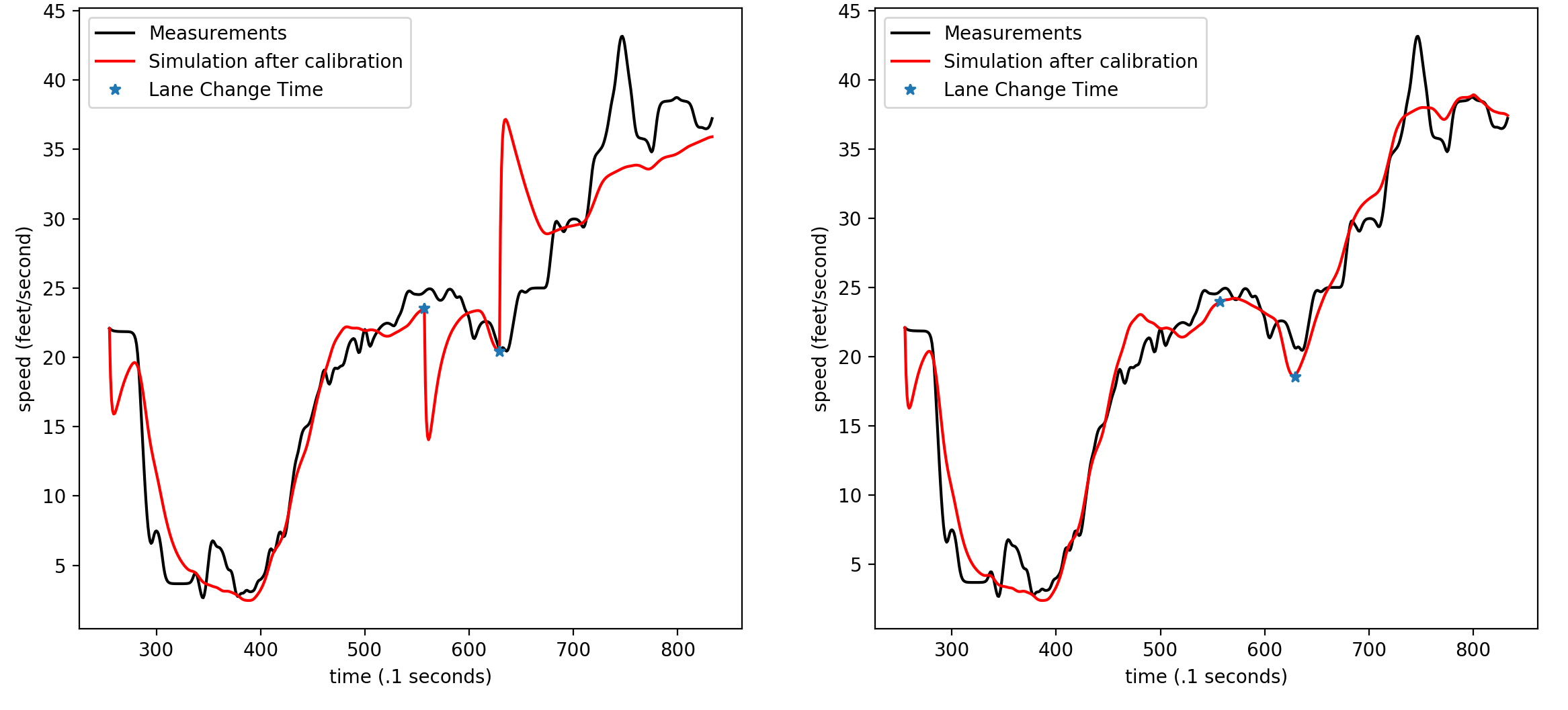} 
\caption{The optimal velocity model (OVM) was calibrated to the trajectory of vehicle 31 in the NGSIM I-80 dataset. Stars indicate the times of the lane changes. The left shows the calibration result when there is no relaxation model--- the model overreacts to the sudden change in headway after a lane change. On the right, the calibration result for OVM with the proposed relaxation model added.} \label{fig2-2}
\end{figure}
\subsection{Lane changing decisions and lane changing dynamics} \label{sec12}
\begin{figure}[H] 
\centering 
\includegraphics[ width=.9\textwidth]{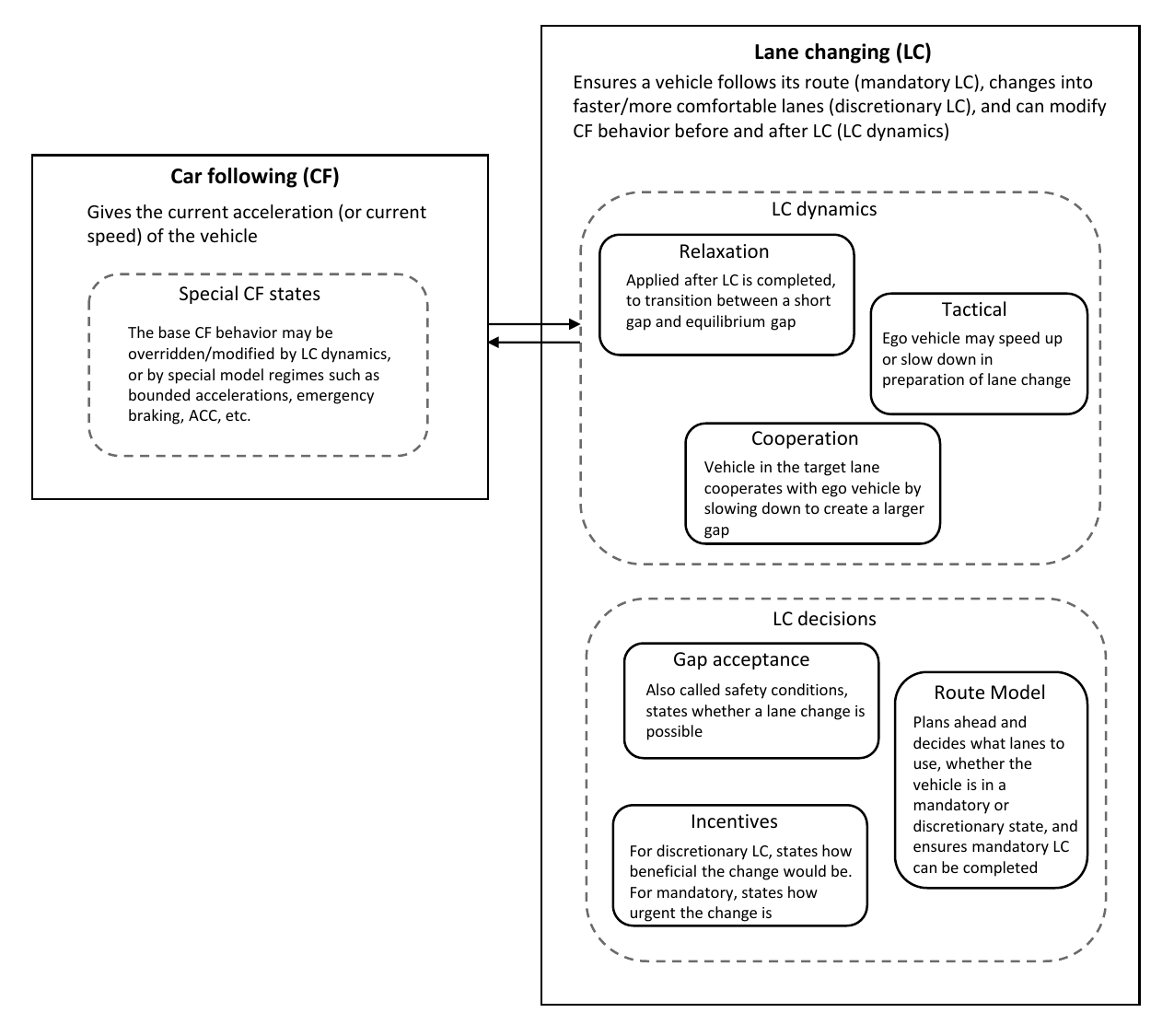}
\caption{High level summary of the different components of a typical lane changing model used in microsimulation.}  \label{cflc}
\end{figure}

A complete lane changing (LC) model consists of several interacting parts (see Fig. \ref{cflc}). One important distinction is between lane changing decisions (the decision processes behind lane changing, which control whether a vehicle will change lanes), and lane changing dynamics (changes to the driving dynamics applied as a result of lane changing). A typical lane changing model (for example, \cite{34}, \cite{1}, \cite{sumolc}, \cite{toledolc}, \cite{gippslc}, \cite{hidaslc}) can be summarized as consisting of the following parts
\begin{enumerate}
\item A gap acceptance model or `safety condition', which determines whether there is sufficient space in the target lane to initiate the LC. The gap required usually depends on both the speeds and positions of the leading and following vehicles in the target lane (the preceding and proceeding vehicles if one was to change lanes). 
\item Rules for both discretionary and mandatory LC. In a discretionary LC, the ego vehicle changes lanes to obtain a more favorable traffic condition. Typically there is some `incentive' which promotes changing into faster lanes; passing rules should also be taken into consideration. In  mandatory LC, the ego vehicle must change lanes to stay on their desired route. This includes planning to decide which lanes the vehicle should use, and rules for handling LC which need to be completed urgently.
\item Lane changing dynamics, which modify the vehicle's CF model in various ways. The most common example is cooperation, where a vehicle in the target lane slows down with the intention of creating a gap for the ego vehicle to use. For an urgent mandatory LC, cooperation may be forced in the target vehicle, simulating aggression of the ego vehicle. There may also be rules (sometimes referred to as a tactical model) for the LC vehicle to either accelerate or decelerate in order to find a suitable gap in the target lane to use. Whereas a tactical/cooperation model is applied before an LC, a relaxation model is applied after the LC, when vehicles must adjust to their new driving situations.
\end{enumerate}
Much of the existing literature concentrates on LC decisions, particularly for discretionary LC. But with a discretionary LC model, it is only possible to model a straight highway section. For an arbitrary highway network, consisting of on/off ramps, bottlenecks, and weaving sections, both mandatory LC and LC dynamics need to be considered. \\
Examining the literature describing complete LC models, we found that although LC dynamics typically include cooperation, a formulation of relaxation is often lacking. Relaxation is critical for modeling short gaps. With relaxation, short gaps can be accepted, because vehicles will adjust to equilibrium gradually without overreacting; without a relaxation model, vehicles would have to wait until a larger gap occurs naturally or is created by the cooperation/tactical model. In \cite{mergingbehavior}, it was found that existing lane changing models were not able to describe many of the LC observed in real data because the models lacked an implementation of the relaxation phenomenon. \\
In a recent LC model literature review, \cite{107} identified one literature gap as a general model of the relaxation phenomenon which can be combined with an arbitrary car following/microscopic model. For existing relaxation models, they are either formulated for macrosimulation or are formulated for specific microsimulation models. \cite{31}, \cite{41}, and \cite{108} all develop formulations of the relaxation phenomenon, but these works are only interpretable in the setting of kinematic wave theory/the LWR model. \cite{82} developed a relaxation model which only applies to the Newell car following model. \cite{81}, \cite{80}, \cite{1}, \cite{32} also developed formulations of relaxation for car following models. Those works incorporated relaxation by changing some parameter in the model in order to temporarily accept the short spacing. The present work differs because instead of changing model parameters, the input to the model (i.e. the headways, speeds) is altered. This creates a formulation of the relaxation phenomenon which can be easily applied to an arbitrary car following model.
\section{The Relaxation Model} \label{sec2}
\begin{figure}[H] 
\centering 
\includegraphics[ width=.5\textwidth]{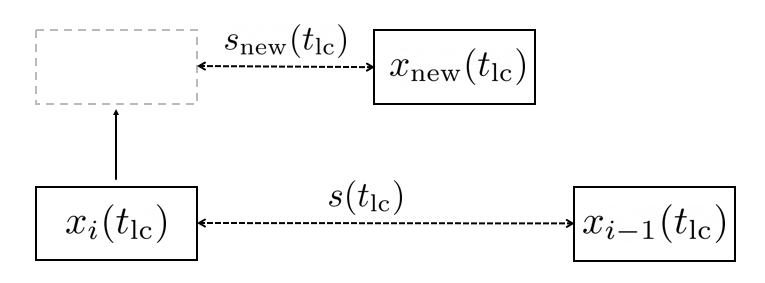} 
\caption{At time $t_{\rm lc}$, vehicle $x_i$ initiates a lane change and begins following $x_{\rm new}$ instead of $x_{i-1}(t_{\rm lc})$.}  \label{fig3}
\end{figure}

\begin{figure}[H] 
\centering 
\includegraphics[ width=\textwidth]{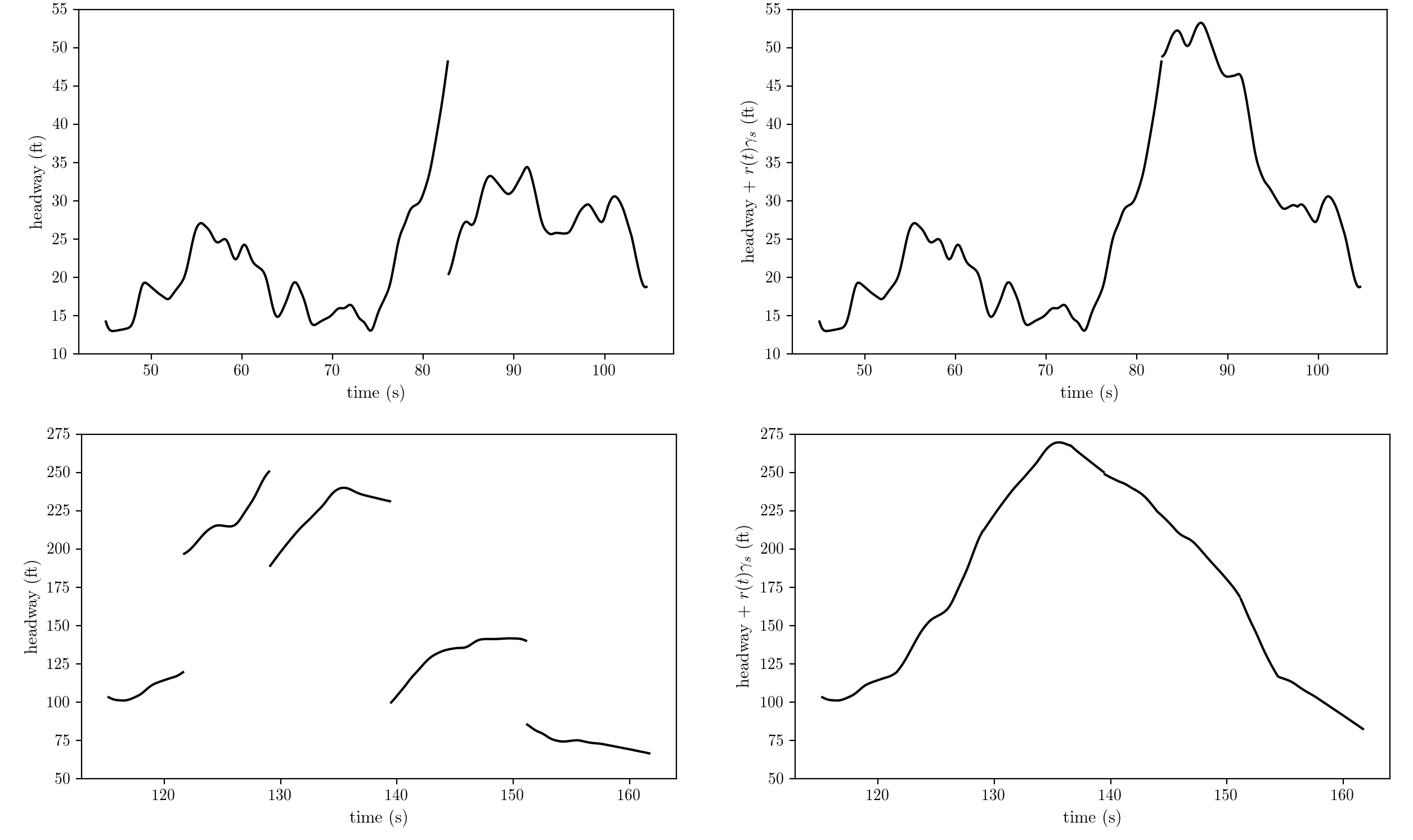} 
\caption{Top left shows the recorded headway for vehicle 93, and bottom left shows the recorded headway for vehicle 320. The gaps in headway are caused by lane changing maneuvers. The right shows the relaxed headway for their respective vehicles, using a value of $c = 15$ seconds. The relaxed headway is used in place of the normal headway as an input to the car following model. } \label{fig4}
\end{figure}

Consider the second order ODE car following model $h$ 
\begin{align*} 
\ddot x_i(t) = h(s(t), \ \dot x_{i-1}(t), \ \dot x_i(t)) \stepcounter{equation}\tag{\theequation}\label{1}
\end{align*}
where $s(t)$ is the (space) headway at time $t$, $\dot x_{i-1}(t)$ is the speed of the lead vehicle at time $t$, and $\dot x_{i}(t)$ is the speed of following vehicle at time $t$. The model with relaxation added is 
\begin{align*} 
\ddot x_i(t) = h(s(t) + r(t)\gamma_s, \ \dot x_{i-1}(t) + r(t)\gamma_v, \ \dot x_i(t)) \stepcounter{equation}\tag{\theequation}\label{2}
\end{align*}
where $r(t)$ is the relaxation and $\gamma_s, \gamma_v$ are the relaxation amounts for headway and speed.
\begin{align*} 
& r(t) = \begin{cases} 
1 - \dfrac{1}{c}(t - t_{\rm lc}) & t_{\rm lc} < t < t_{\rm lc} + c \\ 
  0 & \text{otherwise} \\ 
  \end{cases} \\
& \gamma_s =  s(t_{\rm lc}) - s_{\rm new}(t_{\rm lc}) \\
& \gamma_v =  \dot x_{i-1}(t_{\rm lc})  - \dot x_{\rm new}(t_{\rm lc}) \stepcounter{equation}\tag{\theequation}\label{3}
\end{align*}
Where $t_{\rm lc}$ is the time of the lane change, $s_{\rm new}$ is the headway with respect to the new leader, and $\dot x_{\rm new}$ is the speed of the new leader. This formulation introduces one parameter, $c$, which defines the relaxation time. See Fig. \ref{fig3} for a visual explanation. Note we treat $t_{\rm lc}$ as the last timestep the vehicle follows the old leader.
As opposed to existing relaxation models which temporarily modify model parameters, our relaxation model is based on modifying the headway/lead vehicle speed. This formulation avoids the sudden jumps in those quantities which occur when $x_{i-1}$ changes due to a lane change. See fig. \ref{fig4} for some examples of the jumps in unmodified headway and what the relaxed headway looks like. There are two main benefits to our formulation. First, because we change the inputs and not a model parameter, the relaxation can be applied to an arbitrary model, including different classes of car following models and also nonparametric models. Second, the model $h$ will be continuous even during lane changes. This helps to ensure smooth transitions without overreactions to lane changes.
\subsection{Mathematical analysis of the relaxation model}\label{math}
\begin{figure}[H] 
\centering 
\includegraphics[ width=\textwidth]{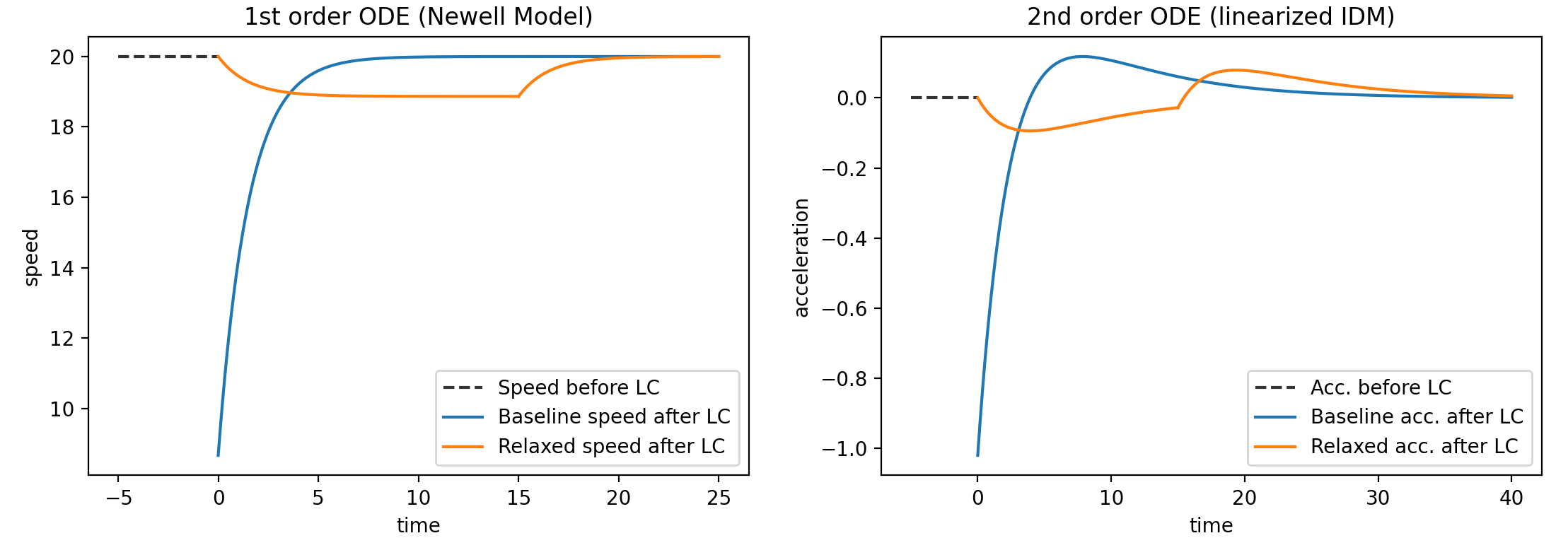} 
\caption{The following vehicle was initially at equilibrium and changed leaders at $t=0$, resulting in a relaxation amount of $\gamma_s = 17$ ($\gamma_v = 0$). For the baseline model with no relaxation, there are strong over-reactions to the lane change.}  \label{fig5}
\end{figure}

We can obtain closed form solutions for linear models in order to analyze the effect of relaxation for a vehicle adjusting to a non equilibrium spacing. Consider the following first order and second order linear car following models 
\begin{align*} 
& \dot x_i(t) = \beta_1 ( s(t) - \beta_2 ) \stepcounter{equation}\tag{\theequation}\label{4} \\
& \ddot x_i(t) = \beta_1 s(t) + \beta_2 \dot x_i(t) + \beta_3 \dot x_{i-1}(t) + \beta_4 \stepcounter{equation}\tag{\theequation}\label{5}
\end{align*}
where the $\beta$ are model parameters. The Eq. \eqref{4} is equivalent to the Newell model, where $\beta_2$ is the jam spacing and $\beta_1\beta_2$ is the wave speed. We used the parameters $\beta_1 = 2/3$ and $\beta_2 = 2$ (with units of 1/seconds and meters respectively). Eq. \eqref{5} is a generic linear second order car following model, and we obtained the parameters $\beta_1 = .06$, $\beta_2 = -.55$, $\beta_3 = .45$, $\beta_4 = .14$ by linearizing the IDM. \\
We solve for the trajectory of both models experiencing a lane change at $t_{\rm lc} = 0$ where the previous leader $x_{i-1}$  and new leader $x_{\rm new}$ both have constant speeds of 20 m/s. The initial position of the vehicles was chosen so that $\gamma_s = 17$ m for both models, and the follower was assumed to be in equilibrium prior to the change, so that $\dot x_i(t) = 20$ before the change. Fig. \ref{fig5} shows the resulting speed profiles (for Newell) and acceleration profiles (for linearized IDM) using a relaxation time of $c = 15$.
Turning our attention to the Newell model, the speed profile for the baseline is 
\begin{align*} 
\dot x_i(t) = v - \gamma_s \beta_1 \exp( -\beta_1 t)
\end{align*}
where $v$ is the (constant) speed of the previous/new leader. This means that the follower is initially $\gamma_s \beta_1$ m/s away from the equilibrium speed of $v$, and that this difference decreases at an exponential rate of $\beta_1$. Note that for the baseline model, the time needed to reach equilibrium is proportional to $1/\beta_1$. The speed profile for the relaxed model is 
\begin{align*} 
\dot x_i(t) = \begin{cases} 
  -\dfrac{\gamma_s-cv}{c} + \dfrac{\gamma_s}{c}\exp(-\beta_1 t)  &  0 <= t < c \\ 
  v - \alpha_1 \exp(-\beta_1 t) & c <= t \\ 
  \end{cases} 
\end{align*}
where $\alpha_1$ is a constant. Assuming that the relaxation time $c$ is significantly longer than $1/\beta_1$, then the speed will adjust to $-(\gamma_s - cv)/c$ and hold that constant speed for the duration of the relaxation. When the relaxation ends, the model will behave like the baseline model, but the initial speed difference will be $\gamma_s/c$ instead of $\gamma_s \beta_1$. \\
Define the time to equilibrium (TTE) as the length of time needed to adjust within $\delta$ of the equilibrium speed.
\begin{align*} 
\rm TTE = \dfrac{1}{\beta_1} \log \left( \dfrac{\beta_1 \gamma_s}{\delta} \right)  \quad \text{ for baseline }\\
\rm TTE = c + \dfrac{1}{\beta_1} \log \left( \dfrac{\gamma_s}{\delta c} \right) \quad \text{ for relaxation}
\end{align*}
Note that the TTE is not the same as the relaxation time parameter $c$--- it also depends on the car following parameter $\beta_1$ (or more generally, for a higher order car following model the TTE will be inversely proportional to the largest root of the characteristic equation). The relaxed TTE is also not exactly equal to the baseline TTE $+ \ c$: it is shorter because when the relaxation ends the vehicle is already closer to equilibrium. \\
We also define the deceleration time (DT) as the length of time that the vehicle spends decelerating after the LC. For the Newell model, DT is simply equal to $c$ and the unrelaxed model has DT$=0$. For second order models, the unrelaxed model has some DT $> 0$, and the relaxed DT is approximately $c$ greater than the unrelaxed DT. \\
Overall this analysis shows that for a leader with constant speed, the relaxed Newell follower will essentially adjust to the new equilibrium at a constant speed. It's not true that the model will reach equilibrium after $c$ time; to characterize a vehicle's adjustment towards equilibrium, we suggest calculating/estimating the TTE and DT values.
One important finding is that different models have different baseline TTEs/DTs, meaning it is not appropriate to give different models the same $c$ values. For example, the Newell model with realistic parameters has a TTE $\approx$2--4 seconds and DT$=0$, whereas the IDM has a TTE $\approx$10--20 seconds and DT$\approx$1-2 seconds. The $c$ value should ideally be calibrated with the rest of the parameters. Alternatively, if one has observed (i.e. from traffic data) approximate values for TTE or DT, it is possible to calculate $c$ analytically.

\subsection{Extra details}
Calculating the relaxation amounts $\gamma_s, \gamma_v$ in Eq. \eqref{3} requires both the old leader $x_{i-1}(  t_{\rm lc})$ and new leader $x_{\rm new}$. But for merging vehicles, there is typically no old leader as vehicles are on the end of the on-ramp/lane. In that case, we calculate the relaxation amounts as 
\begin{align*} 
& \gamma_s = s_{\rm eql}(\dot x_i(t_{\rm lc}) ) - s_{new}(t_{\rm lc}) \\
& \gamma_v = \dot x_i(t_{\rm lc}) - \dot x_{\rm new}(t_{\rm lc})
\end{align*}
where $s_{\rm eql}(v)$ is the equilibrium headway of vehicle $i$ corresponding to speed $v$. \\
Some vehicles may change lanes again before their current relaxation ends. In these cases, the relaxations are added together (and each lane change has its own $\gamma, r(t)$ associated with it). \\
Since positive values of $\gamma_s$ increase the headway, it is possible that the relaxation model can cause collisions in a model that is otherwise accident free. In our experiments, we found that this occurred rarely, only in cases where the new leader suddenly slows down shortly after the change (e.g. due to a shockwave that arrives in the new lane). To prevent this we safeguarded the relaxation in the following way
\begin{align*} 
\rm{z} = \dfrac{\max(s(t) - s_j - \alpha \dot x_i(t), \epsilon)}{\dot x_i(t) - \dot x_{i-1}(t) } \\
r(t) = \begin{cases} 
 r(t)\left( \frac{z}{\beta} \right) &  0 < \rm{z} < \beta \\ 
 r(t) & \rm otherwise \\ 
  \end{cases}
\end{align*}
Where $s_j$ is the jam spacing of the CF model, $\epsilon$ is some small positive constant, and $\alpha, \beta$ are two new safeguard parameters. We took $\alpha = .6$ and $\beta = 1.5$ seconds and did not calibrate those values.  \\
Lastly, we apply the relaxation to all vehicles which experience a leader change. This means that a single lane change results in relaxation being applied to 3 vehicles. There is the lane changing vehicle $x_i$, the old follower (the vehicle following $x_i$ at $t_{\rm lc}$) and the new follower (vehicle following $x_{\rm new}$  at $t_{\rm lc}$). This is slightly different from existing models such as \cite{80} or \cite{41} which do not apply relaxation to the old follower. Also, we apply the relaxation whether the $\gamma_s$ is positive or negative. In those previous papers, they only consider relaxation for $\gamma_s$ being positive (i.e. headway decreases after lane change). 

\section{Microscopic Validation}
\subsection{Models}
To test that the relaxation model can be used with an arbitrary parametric model, we applied it to the optimal velocity model (OVM) \cite{52}, intelligent driver model (IDM) \cite{12}, and Newell model \cite{13}. As a benchmark for existing relaxation models, we considered \cite{41} which formulates a 1 parameter relaxation model for Newell, and \cite{80} which formulates a 2 parameter relaxation model for IDM. \\
The parametrization for OVM is 
\begin{align*} 
& \ddot x_i(t) = c_4(V(s(t)) - \dot x_i(t)) \\
 & V(s) = c_1[ \tanh(c_2 s - c_3 - c_5) - \tanh(-c_3)]
\end{align*}
where $V(s)$ is the optimal velocity function with shape parameters $c_1, c_2, c_3, c_5$. The maximum speed is $c_1(1-\tanh(-c_3))$, $c_5/c_2$ is the jam spacing, and $c_4$ controls the strength of acceleration. \\
The parametrization for IDM is 
\begin{align*} 
\ddot x_i(t) = c_4\left( 1 - \left( \dfrac{\dot x_i(t)}{c_1} \right)^4  - \left( \dfrac{s^*}{s(t)} \right)^2\right) \stepcounter{equation}\tag{\theequation}\label{6} \\
s^*  = c_3+c_2\dot x_i(t) + \dfrac{\dot x_i(t) (\dot x_i(t) - \dot x_{i-1}(t))}{2 \sqrt{c_4 c_5}}  
\end{align*}
where $s^*$ is the desired headway. $c_1$ is the maximum speed, $c_2$ is the desired time headway, $c_3$ is the jam spacing, $c_4$ is the acceleration and $c_5$ is the comfortable deceleration. \\
In \cite{80} their relaxation model for IDM works by modifying the time headway parameter $c_2$. Let $T^*(t)$ be the relaxed time headway. In that paper, the initial value for time headway is set to $\min(c_2, \ d \cdot T_{\rm min} + (1-d)c_2 )$ where $d$ is the `desire' which changes based on the current lane changing model incentives, and $T_{\rm min}$ is the minimum possible time headway. The time headway is then updated as
\begin{align*} 
T^*(t + \Delta t) = T^*(t) + (c_2 - T^*(t)) \dfrac{\Delta t}{\tau}
\end{align*}
where $\tau$ is the relaxation time parameter, and $\Delta t$ is the simulation time step. Thus their formulation had 2 relaxation parameters ($T_{\rm min}, \tau$), and is also integrated with the parameters of the lane changing model through the desire. Since we only want to look at car following dynamics in this section, we have set the initial value for time headway to $\min(c_2, T_0)$ where $T_0$ is the new relaxation parameter. This retains a similar degree of freedom while removing the need for jointly calibrating the specific lane changing model used in that paper. \\
The parametrization for Newell is 
\begin{align*} 
x_i(t + \tau) = \min(x_i(t) + v_f \tau, x_{i-1}(t) -  l_{i-1} - \delta)
\end{align*}
where $\delta$ and $\tau$ are the space and time shift parameters, and $v_f$ is the free flow speed (also a parameter). $l_{i-1}$ is the length of the lead vehicle. \\
In \cite{41} their relaxation model for Newell is formulated as 
\begin{adjustwidth}{-2cm}{0cm} \vspace{-1.3cm}
\begin{align*} 
& x_i(t) = \min( x_i(t - \Delta t) + \min(v_f, \dot x_i(t - \Delta t) + a \Delta t ) \Delta t, \ x_i(t - \Delta t) + \dot x_{i-1}(t)\Delta t - \dfrac{\Delta N(t)}{K(\dot x_{i-1}(t))} ) \\
& K(v) = \dfrac{\omega \kappa}{v + \omega}
\end{align*}
\end{adjustwidth}
where $\omega, \kappa, v_f$ are the parameters, and the relaxation is controlled by $\Delta N(t)$ and its single parameter $\epsilon$ (refer to the paper for the equations used to initialize and update $\Delta N$). Note that this formulation of Newell is equivalent to the previous one when $\Delta N$ is 1, $\omega = \delta/\tau$, $\kappa = 1/\delta$, and $\tau = \Delta t$. The only exception is there is also a maximum acceleration $a$, which was set to 3.4 $m/s^2$ and not calibrated.
\subsection{Methodology}
In this section we only consider calibrating the car following and relaxation model parameters to vehicle trajectory data. There is no lane changing model, so the lane changes and leader/follower relationships are taken from the data. This allows the simulated trajectories to be directly compared to the measured trajectories at an individual vehicle level. Each vehicle has its own individual parameters which represent only that particular vehicle. \\
For any given vehicle to be calibrated, the lead vehicle trajectories and initial condition of the vehicle are taken from the data. Then the simulated vehicle trajectory is generated using given the car following/relaxation parameters, and the mean squared error (MSE) in position between the simulation and measurements is calculated. A genetic algorithm \cite{73} then finds the parameters which minimize the MSE. \\
We used the reconstructed NGSim data \cite{29} as the source of data, which contains 2037 vehicle trajectories total. Of those, 4 vehicles do not have their lead vehicles in the dataset and cannot be calibrated. Of the remaining 2033, 1101 vehicles change lanes themselves, or have a leader which changes lanes. There are also 158 vehicles which merge from the on-ramp onto the highway.
\subsection{Results} \label{parametric-results}
\begin{figure}[H] 
\centering 
\includegraphics[ width=\textwidth]{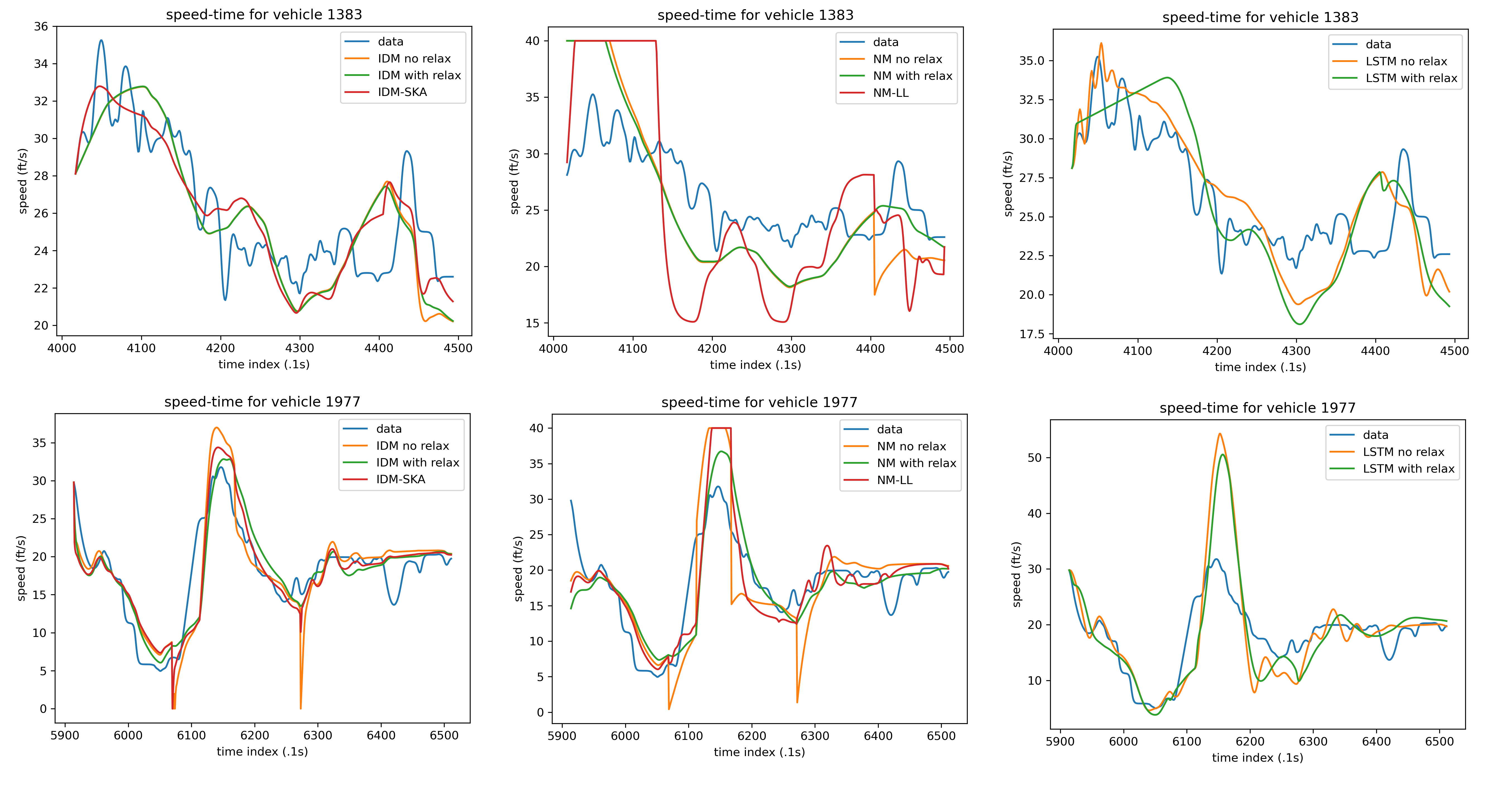} 
\caption{Speed time plots of calibrated trajectories (NM = Newell Model). Many of the Newell trajectories (e.g. middle top/bottom panels) appear unrealistic because of strong accelerations after lane changes. The bottom left panel shows an example of this same effect for the IDM no relax and IDM SKA.} \label{speedfig}
\end{figure}

`Relax' refers to the calibrated results for the car following model with relaxation applied to all the lane changing and merging vehicles. `No relax' are the calibrated results for all the lane changing and merging vehicles, but with no relaxation added. `No LC' are the results for only the vehicles with no lane changing. IDM SKA is the relaxation model due to \cite{80} and Newell LL is the relaxation model due to \cite{41}. Since SKA uses two relaxation parameters, we also tested using our relaxation model with 2 parameters, which is labeled as `IDM relax 2p'. In that formulation, there are two separate relaxation times for the headway/speed relaxation amounts $\gamma_s, \gamma_v$. \\
For the metrics, MSE is the mean squared error averaged over all vehicles, `MSE many LC' is the MSE averaged only over vehicles with 3 or more lane changes, and `MSE merges' is the MSE averaged over merging vehicles. `MSE near LC' is the MSE of trajectories in the 10 seconds after a lane change, averaged over all lane changing events. For all those metrics, we report the mean/median/standard deviation of the distributions of MSE.  \\
Realistic acc. (acceleration) is a measure of whether the calibrated trajectory has any large accelerations/decelerations which were not recorded in the data. This is meant to be a measure of whether the model tends to overreact to lane changes. For each vehicle, we calculate the maximum allowed acceleration/deceleration as $\max ( 4 \rm{m/s}^2, 1.1\max( \hat{\ddot{x}}_i)\ )$, $\min ( -6 \rm{m/s}^2, 1.1\min(\hat{\ddot{x}}_i) \ )$ respectively, where $\max( \hat{\ddot{x_i }})$ is the empirical observed maximum acceleration. For virtually all vehicles, this results in the acceleration bounds of $[4, -6]$ which confirms the plausibility of those bounds. Then the calibrated trajectory is either classified as ``realistic'' or ``unrealistic'' if the simulation has any timesteps with accelerations which are outside those bounds.
\begin{table}[H]
\begin{adjustwidth}{-3cm}{0cm}
\begin{tabular}{|l|l|l|l|l|l|l|}
\hline
                    & MSE ($m^2$) & MSE near LC & Realistic Acc.  & MSE many LC & MSE merges \\ \hline
IDM relax   & 3.33/1.93/7.0   & 5.34/1.86/14.4  & 77\% & 5.84/2.93/14.3 & 1.39/.76/2.0 \\ \hline
IDM no relax    & 4.41/2.50/9.6   & 7.45/2.66/24.2  & 57\% & 8.20/3.45/20.0 & 2.00/.99/3.5  \\ \hline
IDM no LC    & 1.87/1.22/2.0   & -  & 94\% & - & -  \\ \hline
IDM SKA    & 3.07/1.71/6.6   & 4.50/1.55/13.4  & 62\% & 5.30/2.43/13.5 & 1.16/.52/2.0  \\ \hline
IDM relax 2p   & 2.89/1.61/6.7  & 4.23/1.49/12.3  & 77\% & 4.90/2.44/13.7 & .92/.56/1.0  \\ \hline
OVM relax & 4.68/2.83/8.6   & 7.10/2.88/18.5  & 58\% & 7.82/4.64/16.3 & 1.73/.91/3.2  \\ \hline
OVM no relax    & 7.98/4.90/14.2   & 13.98/6.34/32.8  & 39\% & 14.73/7.32/29.1 & 4.08/2.23/5.3  \\ \hline
OVM no LC    & 2.90/1.80/3.7   & -  & 74\% & - & -  \\ \hline
Newell relax   & 8.47/4.57/17.2   & 11.37/4.44/29.3  & 64\% & 13.17/6.38/20.6 & 3.11/1.77/4.4\\ \hline
Newell no relax    & 16.03/9.09/26.2   & 27.74/13.75/50.7  & 7\% & 25.90/14.56/37.4 & 10.32/5.49/14.1  \\ \hline
Newell no LC   & 5.73/3.35/8.7   & -  & 84\% & - & -  \\ \hline
Newell LL  & 11.39/6.27/21.3   & 18.68/7.07/58.5  & 39\% & 18.80/9.54/38.6 & 5.66/2.99/9.3  \\ \hline
\end{tabular}
\caption{Calibration results of the unrelaxed/relaxed parametric models.} \label{table 1} 
\end{adjustwidth}
\end{table}
Out of the models tested, IDM consistently performed the best over all metrics, which is a result other papers have also had when comparing different car following models. By contrast, Newell performed the worst, except for the realistic acceleration metric, for which OVM performed the worst. However it is also true that Newell has 3 parameters (+1 with relaxation) compared to IDM/OVM which have 5 (+1 with relaxation, or +2 for SKA and relax 2p). \\
From the mean/median/standard deviation, it is clear that all the distributions are skewed right, with certain vehicles which have high MSEs. These vehicles which contribute the high MSEs tend to be the same across all the different models. Often, these are vehicles which have many lane changing events over their trajectory--- consider that the `MSE many LC' is significantly higher than the MSE for all models. In general, the MSE for lane changing vehicles is significantly higher than vehicles with no lane changing, even after adding relaxation. The MSE near LC metric confirms that this is due to the lane changing itself, since that metric is significantly higher than the average MSE. \\
As for merging vehicles, we found their average MSE were significantly lower than the average lane changing vehicle. This is because the merging vehicles in the dataset tended to only experience one lane change (the merge itself), and then stayed in the most congested rightmost lane. Thus, merging vehicles tended to have few lane changes, and stay at relatively constant speeds compared to the average lane changing vehicle, resulting in their lower MSE. \\
For all models tested, the `realistic acceleration' for lane changing vehicles is lower than for non lane changing vehicles. Note though that even though the metric is called ``realistic'', a trajectory can still have accelerations which fall outside the [4, -6] m/s$^2$ bounds and still be realistic. In general we found that nearly all of the trajectories for IDM using either SKA or the relaxation model appeared qualitatively realistic, but there they have some short time periods with acceleration just outside the [4, -6] bounds and therefore contribute against the metric. On the flip side, there were many trajectories for OVM and Newell that still had severe and clearly unphysical acceleration spikes, even after adding relaxation. \\
Comparing Newell relax to the existing relaxation model Newell LL, our formulation had significantly better results in every category. IDM relax (+1 parameter) did slightly worse than IDM SKA (+2 parameters) in every category except realistic acceleration, where our model did significantly better. But when we also used 2 relaxation parameters, our model is again outperforming the existing model in every category. As for average relaxation values, we found an average/median of 21.1/8.7 for IDM, and 26.5/15.3 for Newell. The larger relaxation parameters for Newell are to be expected and confirm the analysis of section \ref{math}. \\ 
Some randomly chosen examples of speed time series are shown in Fig. \ref{speedfig}.
Overall, we summarize the results for this section as follows
\begin{itemize}
\item The MSE for lane changing vehicles is significantly higher than for non lane changing vehicles, and base car following models with no relaxation tend to overreact to lane changes. 
\item Our relaxation model is comparable or better than existing model specific relaxation models, in terms of level of fit and ability to prevent overreaction.
\item Even after adding relaxation, MSE for lane changing vehicles is still significantly higher than non lane changing vehicles. This points to a current literature gap which is how to properly model and calibrate lane changing at a microscopic level. Future research should consider models which use leaders/followers in neighboring lanes, and develop methodology for calibrating not only relaxation but also tactical/cooperative lane changing models at a microscopic level.
\end{itemize}
\subsection{Validation with a neural network model}
\begin{figure}[H]
\centering 
\includegraphics[ width=.85\textwidth]{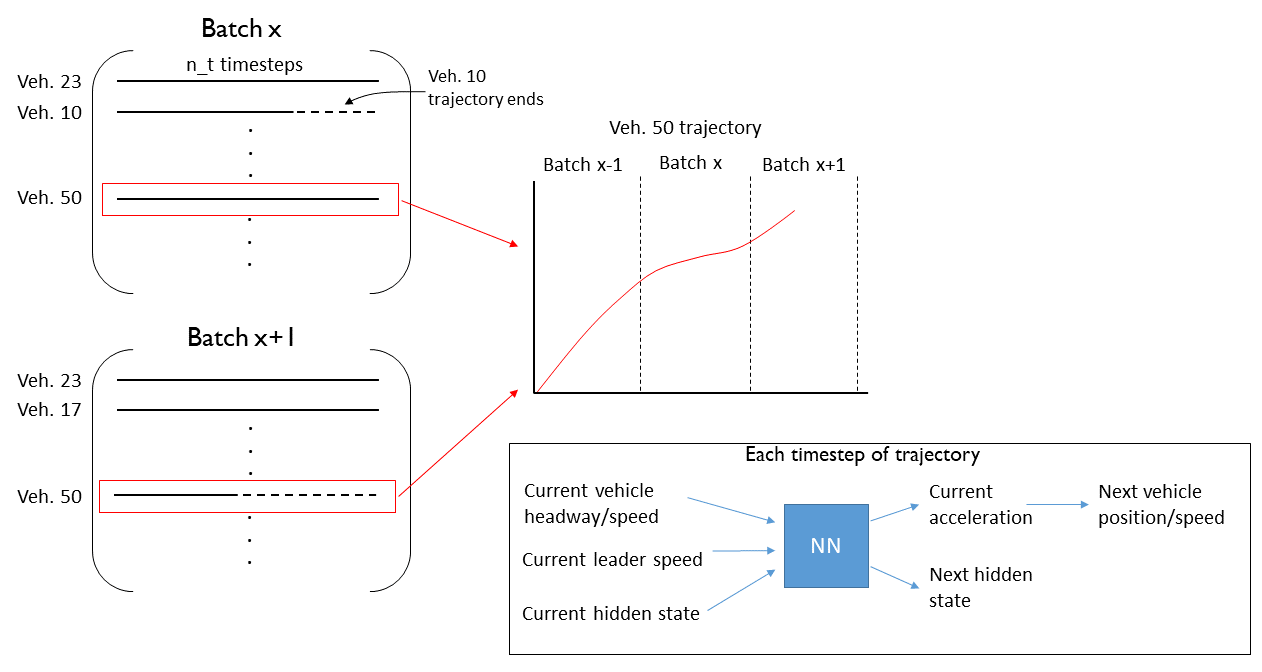} 
\caption{Illustration of batching and inputs/outputs of the LSTM model.}  \label{NNfig}
\end{figure}

To test if the relaxation model can be applied to nonparametric models, we tested a long short-term memory (LSTM) neural network based car following model. LSTM is a common type of recurrent neural network architecture which is designed to process time series data. Our methodology is primarily based off of \cite{zhou-rnn} and is also similar to other works such as \cite{huang-rnn, zhang-rnn}. \\
At every timestep, the current headway, vehicle's speed, and leader's speed are normalized and fed as inputs. The current hidden state of the LSTM, which is initialized as all zeros, is also fed as input. The neural network consists of one layer with 60 LSTM units connected to a second dense layer with 10 neurons. The LSTM produces a new hidden state, and the dense layer then connects to the output, which is a single float value representing the current acceleration of the vehicle. The outputted acceleration then updates the vehicle's position/speed, and the new current hidden state is recorded. \\
Note that this follows the approach used in \cite{zhou-rnn} where the same hidden state is constantly updated along the entire length of the vehicle trajectory, which typically lasts around 500-700 timesteps (50-70 seconds). In this way, the LSTM layer only has to process one timestep worth of data per prediction. In contrast, \cite{huang-rnn} generates a new hidden state for each timestep by feeding the past $M$ timesteps as an input (where $M$ is a hyperparameter taken as 50 timesteps). \\
The main difference in our neural network is how we handled the batching. In existing papers using deep learning for traffic modeling, they typically just state the batch size without elaborating further. This leads to confusion: are entire trajectories being simulated during training, do they just predict the next timestep given the past empirical trajectory? We found that the latter typically lead to bad testing results. If the former is used, do the batches consist of sequential timesteps for the same vehicle, or are multiple vehicles considered in each batch? We formed batches by selecting $n\_veh$ vehicles randomly. Each vehicle in the batch will have up to a maximum of $n\_t$ timesteps simulated, starting with the initial condition. 
In the next batch, the next $n\_t$ timesteps will be simulated for each of the $n\_veh$ vehicles, starting from where the previous batch's simulation ended. Once a vehicle's entire trajectory has been simulated, it is replaced with a new random vehicle in the next batch. This is explained visually in the Fig. \ref{NNfig}. Compared to having a fixed batch size, the final testing error was significantly improved by starting with a small value of $n\_t$ (50) before increasing it to around 500, so entire trajectories will be simulated in each batch. \\
Other minor details are as follows. We didn't clean or filter the data ourselves, and we used every vehicle possible in the reconstructed NGSim dataset. Vehicles were put into either the training (85\%) or testing (15\%) set, and the loss of the testing set is computed after training to ensure there is no overfitting. A dropout rate of .3 was applied to the LSTM layer, and l2 regularization with constant .02 was applied to the kernels of the dense and LSTM layers. The LSTM layer used a tanh activation and sigmoid recurrent activation, and the dense layer used ReLU activation. Adam with a learning rate of .0008 was used as the optimizer, using MSE in position as the loss function.
\subsubsection{Results of the LSTM model}
Applying relaxation to the LSTM model is the same as applying it to a parametric model, with one exception: when a lane change occurs, we pass the relaxation amounts as an extra input. Otherwise, the extra input is zero. We did this so there is a way for the neural net to `see' when a lane change occurs. The relaxation parameter was not learned during training, rather we treated it as a hyperparameter which was set as 12 manually.
\begin{table}[H]
\begin{adjustwidth}{-3cm}{0cm}
\begin{tabular}{|l|l|l|l|l|l|l|}
\hline
                    & MSE ($m^2$) & MSE near LC & Realistic Acc.  & MSE many LC & MSE merges \\ \hline
LSTM relax   & 52.0/24.1/120   & 99.4/26.8/343  & 85\% & 88.7/34.5/232 & 37.4/15.8/71.9 \\ \hline
LSTM relax no LC    & 39.0/18.9/64.8  & -  & 100\% & - & -  \\ \hline
LSTM no relax   & 77.5/28.0/189   & 132/29.3/458  & 99\% & 144/48.8/344 & 42.6/13.3/82.3\\ \hline
LSTM no relax no LC   & 32.4/15.1/54.9  & -  & 100\% & - & - \\ \hline
\end{tabular} 
\caption{Calibration results of the unrelaxed/relaxed neural network (LSTM) model.}\label{table 2}
\end{adjustwidth}
\end{table}
First, it should be stated that these MSE are much higher than those reported for the parametric models - this is not a surprise, as here we are learning a single set of parameters to predict the trajectory of any vehicle in the NGSim data. That is, the LSTM model encodes the behavior for the average driver, as opposed to having vehicle specific parameters. \\
Regarding the distributions of MSE, they all follow the similar patterns observed for the parametric models. The MSE is skewed right, with the error primarily dominated by vehicles which change lanes several times. We see that when training with relaxation, the prediction error on lane changing vehicles is significantly reduced (from 77.5 to 52. on average), but even with the relaxation added the error for lane changing vehicles is still significantly higher than for vehicles with no lane changing. For the LSTM with relax, the error on vehicles with no LC is slightly higher than the LSTM with no relaxation, however the LSTM with relax has a significantly lower error when considering the prediction over all vehicles (46.0 v 56.8). Thus, the relaxed LSTM model is more successful at learning the behavior of the average driver. LSTM with relax had a lower realistic acceleration metric, but upon visual inspection of the trajectories, we found this was caused by having accelerations just outside the [4, -6] bounds, as opposed to the trajectories being appearing qualitatively unrealistic.\\
Overall we found that the relaxation model is general enough to be easily applied to a neural network model, and that it had a significant improvement on the prediction of car following trajectories with lane changing.

\section{Macroscopic Experiments and Validation}

\begin{figure}[H] 
\centering 
\includegraphics[ width=.6\textwidth]{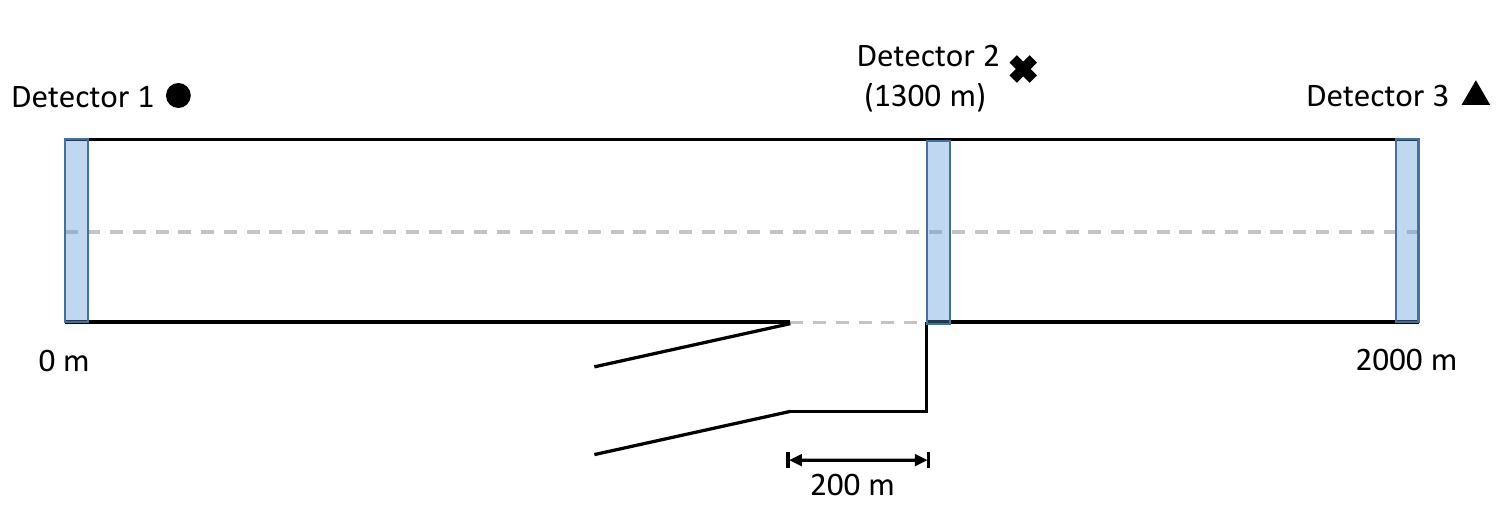} 
\caption{Configuration of road network used for macroscopic experiments.} \label{roadnetwork}
\end{figure}

\begin{figure}[H] 
\centering 
\includegraphics[ width=\textwidth]{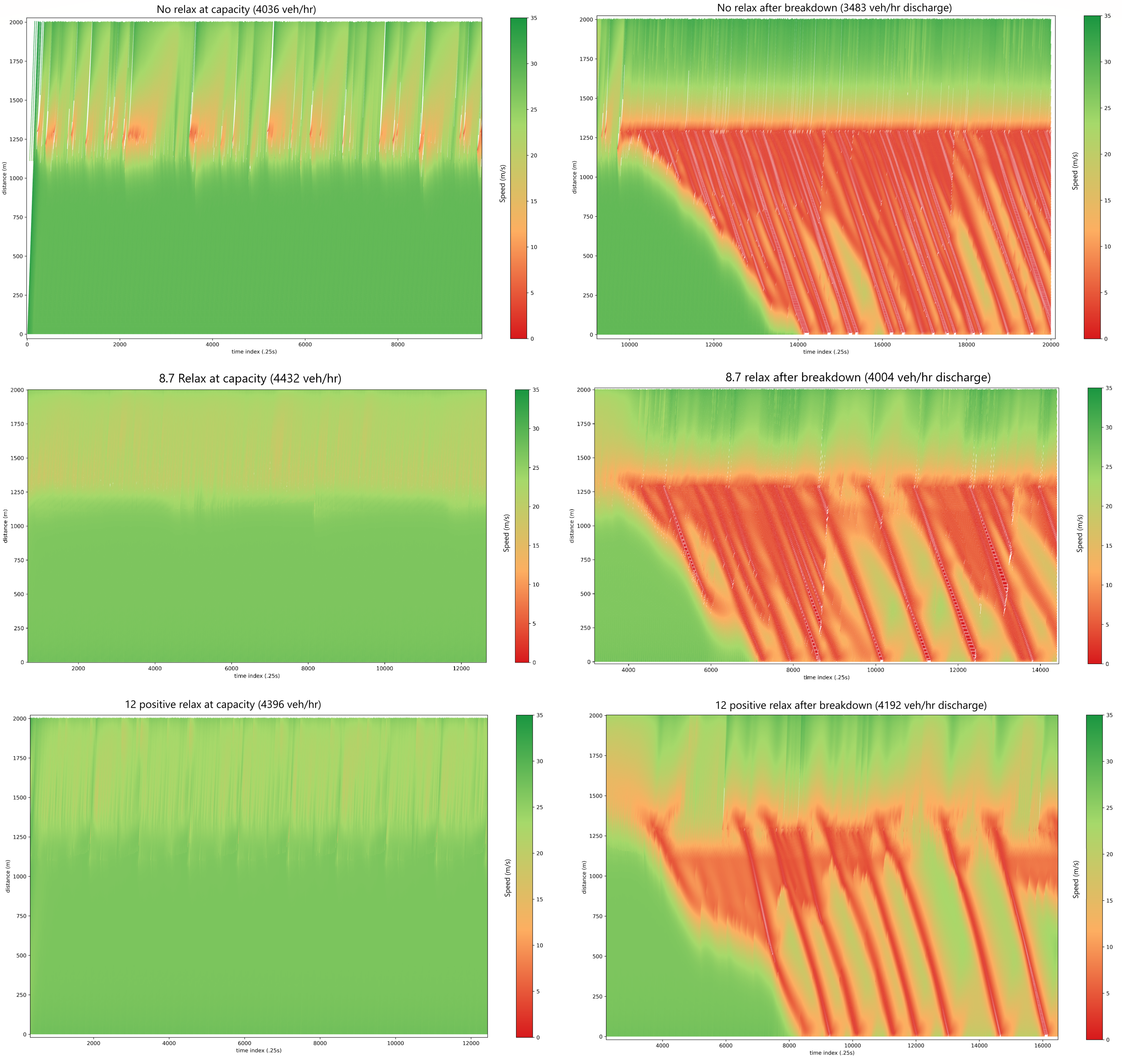} 
\caption{Space-time trajectories of the right lane (connected to the on-ramp). Trajectories are colored according to their speeds. 
} \label{macrofig1}
\end{figure}

To study the effects of relaxation on capacity drop, we considered the 2 lane highway with an on-ramp bottleneck shown in Fig. \ref{roadnetwork}. Simulations were done using our open source simulator havsim. The current havsim model is most similar to the open source simulator traffic-simulation-de \cite{TreiberKestingBook}, and it is explained fully in the appendix. Note that our model is completely deterministic, except for discretionary LC, and that we used identical vehicles in the simulations.
The reason for capacity drop at a merge bottleneck can be explained intuitively in the context of microsimulation. While traffic is uncongested and below capacity, vehicle trajectories immediately downstream of the bottleneck will be close to equilibrium (i.e. vehicles tend to have a fairly constant speed when leaving the bottleneck). When the capacity drop occurs, it is because merging vehicles cause disturbances on the main road--- this forms waves which propagate upstream on the main road, and the downstream trajectories will no longer be in equilibrium. It is well known that the equilibrium solution corresponds to a higher flow state than an oscillatory solution, and so because of the oscillations induced by merging vehicles, the discharge rate is reduced. Note that this intuitive explanation is in general agreement with work such as \cite{leclerqcapacitydrop} and \cite{duretcapacitydrop}, which have analyzed capacity drop in the Newell model with bounded acceleration. However, the key difference is that whereas the reduced flow in the Newell trajectories is due to empty space (voids) in the downstream trajectories, for a generic second order car following model the reduced flow is due to the oscillatory downstream trajectories. \\
In our experiments, we found that the relaxation time parameter had a strong effect on the oscillations created by merging vehicles. The Fig. \ref{macrofig1} shows the contrast between none, moderate, and strong relaxation. To produce those figures, we kept a constant inflow of 400 veh/hr on the on-ramp, and found the maximum constant inflow we could send on the main road before traffic breakdown would occur. The top left shows the maximum flow state for the model without relaxation. Even though the equilibrium solution corresponding to the maximum flow gives 4420 veh/hr, the model is only able to produce a maximum of 4036 veh/hr discharge. The relaxed model (bottom/middle left), are able to reach 4420 veh/hr, and its downstream trajectories show significantly less oscillation. The right figures show the unrelaxed/relaxed models after traffic breakdown. The unrelaxed model has more frequent and severe waves emerging from the merge, resulting in a significantly lower discharge rate. As the relaxation increases, oscillations tend to decrease, leading to higher discharge rates and lower values of capacity drop. 

One important aspect of capacity drop is that its magnitude depends on the amount of flow on the merging lane. This result has been found in \cite{leclerqcapacitydrop} for an analytic model as well as in \cite{duretmetering} and \cite{integrationdrop} for the microsimulators Aimsun and Integration respectively. Intuitively this is not too surprising: if there are more merging vehicles, there will tend to be more disturbances/oscillations, so the capacity and discharge rate both decrease (the capacity drop will increase). However we were not able to find any empirical studies which considered the amount of on-ramp flow as a key exogenous factor in capacity drop. \\
The effects of changing the relaxation time and inflow amounts are summarized in the two tables \ref{table 4} and \ref{table 5}. In those tables, period refers to the average time between shockwaves. We calculated the period by recording the arrival times of waves at the detector 1 (upstream). 
\begin{table}[H]
\centering
\begin{tabular}{|l|l|l|l|l|l|}
\hline
   Relax   (s)         & capacity & discharge (left/right) & drop \% (stdev)& period (stdev) (min.)\\ \hline
   0 & 4036 & 3483 (1851/1632) & 13.7 (2.9) & 1.73 (.82)  \\ \hline
    2 & 4432 & 3567 (1895/1672) & 19.5 (3.3) & 1.73 (.69)  \\ \hline
     4 & 4432 & 3710 (1924/1786) & 16.3 (3.1) & 1.89 (.96) \\ \hline
      7 & 4432 & 3859 (2001/1859) & 12.9 (3.1) & 2.88 (.86)  \\ \hline
      10 & 4432 & 4015 (2054/1961) & 9.4 (3.3) & 3.15 (1.39)  \\ \hline
        15 & 4396 & 3985 (2039/1947) & 9.3 (2.8) & 3.2 (1.43)  \\ \hline
        $10^*$ & 4432 & 4119 (2121/1998) & 7.1 (1.3) & 3.71 (1.02)  \\ \hline
\end{tabular}  
\caption{Results of changing relaxation time with constant on-ramp inflow of 400 veh/hr}\label{table 4}
\end{table}

\begin{figure}[H] 
\centering 
\includegraphics[ width=\textwidth]{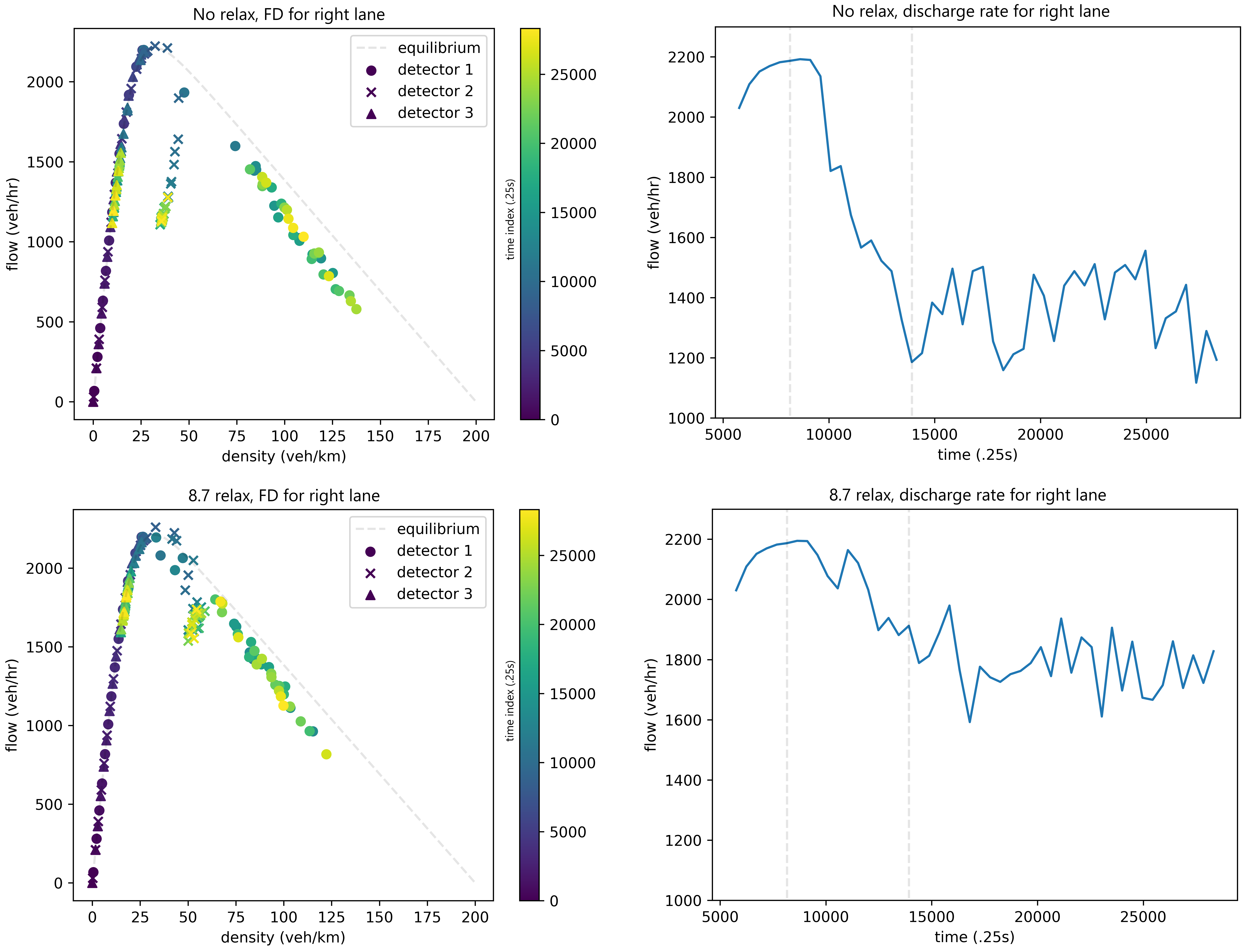} 
\caption{In left plots, the dashed line corresponds to the equilibrium solution of the car following model; the colors represent the time the measurement was taken. In right plots, the two dashed lines are at 34 and 58 minutes (during this period, the on-ramp inflow was gradually increased). }  \label{macrofig2}
\end{figure}

\begin{table}[H]
\centering
\begin{tabular}{|l|l|l|l|l|}
\hline
   Relax    (s)        & capacity & discharge (left/right) & drop \% (stdev)& period (stdev) (min.)\\ \hline
   0 & 3608 & 3110 (1720/1390) & 13.8 (4.5) & .88 (.40)  \\ \hline
    2 & 4040 & 3211 (1735/1476) & 20.5 (4.0) & 1.02 (.43)  \\ \hline
     4 & 4112 & 3305 (1723/1583) & 19.6 (4.3) & .98 (.40)  \\ \hline
      7 & 4148 & 3412 (1765/1647) & 17.7 (3.8) & 1.21 (.56) \\ \hline
      10 & 4256 & 3718 (1889/1829) & 12.6 (3.8) &  1.56 (.77)  \\ \hline
        15 & 4256 & 3676 (1881/1795) & 13.6 (4.1) & 1.62 (.55)  \\ \hline
        $10^*$ & 4184 & 3947 (2006/1941) & 5.7 (3.6) & 2.19 (1.28) \\ \hline
\end{tabular}  
\caption{Results of changing relaxation time with constant on-ramp inflow of 800 veh/hr}\label{table 5}
\end{table}
Note that there is some ambiguity in defining the capacity drop. As is pointed out in \cite{coifmanrelax} (and observed in our simulations), the discharge rate will increase above capacity shortly before breakdown. Further, since the capacity depends on the on-ramp inflow, the observed capacity drop will be different if the on-ramp inflow varies. The capacity and discharge rates were calculated as in the Fig. \ref{macrofig1}, where we keep the on-ramp inflow constant and vary the mainroad inflow until traffic breakdown; the capacity is then defined as the maximum inflow we could send before breakdown. To measure the discharge, we record the flow after breakdown at detector 3 in 2 minute intervals, averaged over $\approx$45 minutes. \\
We used our vanilla formulation which applies relaxation to both positive and negative $\gamma$. We also tried only applying positive relaxation, i.e. $\gamma >0$, as is done in other relaxation formulations, and those results are marked by an asterisk. The table \ref{table 6} shows the corresponding TTE and DT, which we estimated for a follower with an initial headway of 15 m and speed of 29 m/s, whose leader has a constant speed of 29 m/s (equilibrium headway is 55 m in this case). Those values were chosen to represent a merge in uncongested conditions; in a different merging situation, the TTE and DT would be different.
\begin{table}[H]
\centering
\begin{tabular}{|l|l|l|l|l|l|l|}
\hline
   Relax  (s) & 0 & 2 & 4 & 7 & 10 & 15  \\ \hline
 TTE/DT (s) & 24.3/1.8 & 25.5/3.5 & 26.7/5.4 & 28.5/8.2 & 30.3/10.9 & 33.4/15.4 \\ \hline
\end{tabular}  
\caption{Estimates of time to equilibrium (TTE) and deceleration time (DT) for IDM with varying relaxation time.}\label{table 6}
\end{table}
Overall this shows that increasing the relaxation time leads to higher discharge rates and longer periods between shockwaves. The capacity tends to remain fairly constant, meaning the capacity drop decreases as relaxation increases. Decreasing the negative relaxation (i.e. relaxation for larger gaps after LC) seems to have a similar effect as increasing the positive relaxation. Changing only the relaxation parameter, we produced capacity drop for a 2 lane highway in the range of $\approx$10-20\%, with shockwave periods between $\approx$1-5 minutes, which is in good agreement with empirical data \cite{capacitydropreview}.\\
Lastly, we show some examples of the FD and discharge rate in Fig. \ref{macrofig2}. In those experiments, we first generated the uncongested branch of the FD by keeping the on-ramp inflow at 0, and linearally increasing the main road inflow from 0 to 2196 veh/hr/lane over 24 minutes. The 2196 veh/hr/lane inflow is then kept constant for the rest of the experiment. Then from 34-58 minutes, the on-ramp inflow is increased linearally from 0 to 800 veh/hr, and kept constant for the rest of the 2 hour simulation. Points in the FD are aggregated over 2 minute time intervals, and measured over 100m distances using Edie's generalized definitions of flow/density; the discharge rate shows the flows of the downstream detector (3). We used a value of 8.7 for the relaxation time, which was the median value calibrated for IDM in the previous section. The fundamental diagrams show that 700m downstream of the merge, trajectories have returned to equilibrium. Immediately downstream of the merge, trajectories can be far from equilibrium, with flow values approximately equal to or slightly lower than the discharge rate, but with higher densities.
\section{Future outlook} \label{conclusion}
This work shows the importance of considering lane changing (LC) dynamics when formulating and and analyzing microsimulation models. LC dynamics are necessary to be able to understand and explain the trajectories of lane changing vehicles. More importantly, LC dynamics such as relaxation play a significant role in key traffic phenomena such as the formation of waves and capacity drop. We believe traffic flow researchers need to move towards a more holistic view of traffic microsimulation, which considers not only car following models, but also LC dynamics and mandatory/discretionary LC. To this end, we have proposed a relaxation model that can be applied to an arbitrary car following model, and made our complete microsimulation model available online at \url{https://github.com/ronan-keane/havsim}. That repository also includes the trained LSTM model and the code to generate the other results of the paper. \\
In the future, we hope that a general tactical/cooperation model can be developed, with the goal of obtaining a complete LC dynamics model which can be applied to an arbitrary car following rule. Since many traffic waves seem to originate from large perturbations caused by lane changing, as opposed to phantom traffic jams, a better understanding of LC dynamics would likely lead to better control strategies for improving traffic flow.

\chapter{Conclusion and Future Research Directions}
In recent years, gradient-based optimization has seen a marked uptick in popularity, especially with regards to its use in machine learning and deep learning. Our point of view is that this is simply due to the ability of machine learning to capitalize on two pertinent trends: the ever-increasing power of computers, and the emergence of new sources of high quality data. In the future, it will probably be the case that prosperous methodologies will be the ones which are best able to scale up to large models and best handle large, real world datasets. For these reasons, we expect that machine learning will only continue to become more dominant in all of engineering, and likewise that various disciplines will all move towards emphasizing larger, more complex models.

Gradient-based optimization represents one of the most powerful tools currently available for dealing with large models. We have seen how reverse differentiation can be used to get an estimate of the gradient in the same asymptotic time complexity as an objective evaluation, which makes gradient-based optimization uniquely attractive when considering problems with extremely large numbers of parameters. Indeed, we have shown that this can be done for problems involving stochastic elements (i.e. involving random variables), for problems with nondifferentiable objectives, for problems that involve discrete variables, and for problems involving piecewise functions (if-else statements). As far as the author is aware, there is no such thing as a problem that cannot be formulated as an unconstrained nonlinear optimization problem, and then solved using reverse differentiation and a gradient-based optimization algorithm. This dissertation has attempted to provide a comprehensive framework for doing so, with an emphasis on the gradient itself.

\noindent There are many promising avenues for future research: \\
\noindent \textit{\textbf{Variance Reduction.}} Variance reduction remains a fruitful area of research in machine learning and stochastic optimization. Following the same line of research as the optimal baseline discussed in chapter 4, there are opportunities for the applications of baselines to the mixed score function/pathwise derivative case. This would rigorous justify the application of baselines to the theorem 1.1 estimator. It is clear that the optimal baseline has a different derivation/interpretation in the mixed score function/pathwise case compared to the pure score function case which we discussed in this dissertation. Another good question is whether optimal baselines would benefit reinforcement learning, and how much extra variance reduction a (possibly per-parameter) optimal baseline can add to the variance reduction given by GAE alone.

There are many additional opportunities outside of baselines. Based on the new understanding of reverse differentiation, an obvious idea is designing control variates for the reverse solve, i.e. control variates which address the variance of the adjoint variables $\lambda$. This is attractive because it is difficult to design a control variate for the gradient when the number of parameters grows large. By focusing on the adjoint variables, we can potentially learn a control variate with a much smaller dimension, which would make the problem tractable. This idea would specifically target pathwise derivatives (seeing as score functions by themselves do not lead to a reverse solve). Another promising area is in importance sampling. Similar to gradient estimation, importance sampling is an old idea which has continued to found new uses and appeal. There are still more opportunities for using importance sampling for variance reduction, and more generally using importance sampling to promote more `well-behaved' SGD iterates.

\noindent \textbf{Other Applications for Score Functions.} In chapter 3.3 we argued for the importance of piecewise functions. Truly we believe that this idea of learning logic holds a great deal of promise for future modeling applications. Although still in an early stage of conception/development, this new idea for working with piecewise continuous functions could lead to a departure from the space of continuous functions (which is where modeling currently resides).

Another potential idea is the application of score functions in mixed integer nonlinear programming. In chapter 1.5 we discussed how to reformulate discrete variables using score functions, and this is an idea which transforms a MINLP into a regular, unconstrained nonlinear programming problem. The idea of relaxing integer programs is obviously not new, but the exciting aspect is that that our formulation would give an \textit{upper bound} as opposed to a continuous relaxation which gives a \textit{lower bound}.

\noindent \textbf{Differentiable Traffic Simulators.} Havsim, the first differentiable traffic simulator, is still under active development. The point of havsim is not to show that we can use gradient based optimization in traffic applications. The point of havsim is to build a new software tool for transportation researchers that can enable new developments such as \textbf{a)} a rigorous treatment of stochastic models and stochastic behaviors \textbf{b)} enabling realistic driver heterogeneity by making it tractable to consider different sets of parameters for individual drivers \textbf{c)} being able to calibrate lane changing dynamics end-to-end, without the use of heuristics, by learning piecewise functions \textbf{d)} the ability to easily utilize different models (such as neural network based approaches) and different types of big data (including individual vehicle trajectories but also space mean speeds/densities). Havsim is a python package that can be used to manipulate and visualize trajectory data, easily implement customizable traffic models, build simulations on arbitrary highway networks, and solve optimization problems which depend on those simulations. It can be used both for control and calibration problems, and API is being developed to handle both of those use cases. The ultimate goal of havsim is to serve as a tool for transportation researchers who rely on traffic modeling, especially those working on Intelligent Transportation Systems (ITS) or connected and autonomous vehicles. The project can be followed at \url{www.github.com/ronan-keane/havsim} and will be updated as the official release approaches.

\appendix 
\chapter{Chapter 1 Appendix}
\section*{Adjoint Method Tutorial} \label{adjoint-method-tutorial}
To derive the gradient for
\begin{align*}
\underset{\theta}{\min} & \quad  f(x_1, \ldots, x_n)       \stepcounter{equation}\tag{\theequation}\label{tut0}\\
\text{s.t.} & \quad x_{i} = h_i(x_{i-1}, \theta)  \quad i = 1, \ldots, n 
\end{align*}
begin by differentiating the objective
\begin{align*} 
\dfrac{\partial f(x_1, \ldots, x_n)}{\partial \theta} = \sum_{i=1}^{n} \dfrac{\partial f ( \cdot)}{\partial x_i}\dfrac{\partial x_i}{\partial \theta} \label{tut1}\stepcounter{equation}\tag{\theequation} .
\end{align*}
The key to the adjoint method is to avoid calculating the quantities $\partial x_i / \partial \theta$. To do this, first take the gradient of the model $x_{i} = h_i(x_{i-1}, \theta)$ to yield
\begin{align*} 
\dfrac{\partial x_{i}}{\partial \theta} - \dfrac{\partial h_i}{\partial x_{i-1}}\dfrac{\partial x_{i-1}}{\partial \theta} - \dfrac{\partial h_i}{\partial \theta} = 0 \quad \forall  \ i \in 1, \ldots, n. 
\end{align*}
Since the above is equal to zero, we can add some multiple $\lambda_i$ of it to Eq. $\ref{tut1}$, yielding
\begin{align*} 
\dfrac{\partial f(x_1, \ldots, x_n)}{\partial \theta} =  \sum_{i=1}^{n} \dfrac{\partial f}{\partial x_i}\dfrac{\partial x_i}{\partial \theta} + \lambda_i^\top\left(\dfrac{\partial x_{i}}{\partial \theta} - \dfrac{\partial h_i}{\partial x_{i-1}}\dfrac{\partial x_{i-1}}{\partial \theta} - \dfrac{\partial h_i}{\partial \theta} \right) \label{tut2}\stepcounter{equation}\tag{\theequation}
\end{align*}
where $\lambda_i$ has the dimension as $x_i$. Note that equality holds for any possible value of $\lambda_1, \ldots, \lambda_n$. Now, to avoid calculating any of the $\partial x_i / \partial \theta$, we will collect all the terms containing $\partial x_i / \partial \theta$ for each $i$. 
\begin{align*} 
=&\sum_{i=1}^n\left( \dfrac{\partial f}{\partial x_i}\dfrac{\partial x_i}{\partial \theta} + \lambda_i^\top\dfrac{\partial x_i}{\partial \theta} \right) - \sum_{i=1}^{n-1}\left( \lambda_{i+1}^\top\dfrac{\partial h_{i+1}}{\partial x_i}\dfrac{\partial x_i}{\partial \theta}  \right) - \sum_{i=1}^n \lambda_i^\top \dfrac{\partial h_i}{\partial \theta} \qquad \text{(note } \frac{\partial x_0}{\partial \theta} = 0) \\
= &\left(\dfrac{\partial f}{\partial x_n} + \lambda_n^T \right)\dfrac{\partial x_n}{\partial \theta } + \sum_{i=1}^{n-1}\left( \dfrac{\partial f}{\partial x_i} + \lambda_i^\top - \lambda_{i+1}\dfrac{\partial h_{i+1}}{\partial x_i} \right)\dfrac{\partial x_i}{\partial \theta} - \sum_{i=1}^n \lambda_i^\top \dfrac{\partial h_i}{\partial \theta} \stepcounter{equation}\tag{\theequation}\label{tut3}
\end{align*}
Since the equality holds for any value of $\lambda_i$, we can choose $\lambda_i$ to satisy the equations
\begin{align*} 
& \dfrac{\partial f}{\partial x_n} + \lambda_n^T = 0 \\
& \dfrac{\partial f}{\partial x_i} + \lambda_i^\top - \lambda_{i+1}\dfrac{\partial h_{i+1}}{\partial x_i} = 0 \quad i = n-1, \ldots, 1 .\label{tut4}\stepcounter{equation}\tag{\theequation}
\end{align*}
Thus we have derived the adjoint equations, which wipe out all $\partial x_i / \partial \theta$ terms for $i = 1, \ldots, n$. Whereas the $x$ are solved in the order $x_1 \ldots x_n$, the $\lambda$ are solved in the reverse order $\lambda_n \ldots \lambda_1$, hence \eqref{tut4} can be referred to as the reverse pass. The remaining terms of \eqref{tut3} then gives us the gradient 
\begin{align*} 
& \dfrac{\partial f(x_1, \ldots, x_n)}{\partial \theta} = - \sum_{i=1}^n \lambda_i^\top \dfrac{\partial h_i}{\partial \theta} .
\end{align*}
In the equations presented in the main text, we use also the substitution $\lambda_i^* = -\lambda_i$ to wipe out all the minus signs. Lastly, we note that this derivation requires existence of the jacobians $\partial f/\partial x_i$, $\partial h_i/\partial x_i$, $\partial h_i/\partial \theta$.
\subsubsection*{Benefit of the Adjoint method}
To better understand exactly what the reverse pass in reverse differentiation affords us, it is instructive to try explicitly calculating the gradient of \eqref{eqn1}
\begin{align*} 
\dfrac{\partial f}{\partial \theta} = \sum_{i=1}^n \dfrac{\partial f}{\partial x_i}\dfrac{\partial x_i}{\partial \theta}  . \stepcounter{equation}\tag{\theequation}\label{RD0}
\end{align*}
To calculate $\partial x_i/ \partial \theta$, we take the gradients of both sides of the equality $x_i = h_i(x_{i-1}, \theta)$ to yield 
\begin{align*} 
\dfrac{\partial x_{i}}{\partial \theta} = \dfrac{\partial h_i}{\partial x_{i-1}}\dfrac{\partial x_{i-1}}{\partial \theta} + \dfrac{\partial h_i}{\partial \theta}. \stepcounter{equation}\tag{\theequation}\label{RD1}
\end{align*}
The issue with \eqref{RD1} is that it does not contain vector-Jacobian or Jacobian-vector products; rather, it requires Jacobian-matrix products and full Jacobians, so there is no way to efficiently compute $\partial x_i/\partial \theta$ given $\partial x_{i-1}/\partial \theta$. But looking at \eqref{RD0}, $\partial f/ \partial x_i$ is a vector, so can we somehow convert \eqref{RD1} into vector-Jacobian products? Writing out \eqref{RD0}, \eqref{RD1} for $n=3$:
\begin{align*} 
\dfrac{\partial f}{\partial \theta} = \dfrac{\partial f}{\partial x_1}\dfrac{\partial h_1}{\partial \theta} + \dfrac{\partial f}{\partial x_2}\left( \dfrac{\partial h_2}{\partial x_1}\dfrac{\partial h_1}{\partial \theta} + \dfrac{\partial h_2}{\partial \theta} \right) + \dfrac{\partial f}{\partial x_3}\left( \dfrac{\partial h_3}{\partial x_2}\left( \dfrac{\partial h_2}{\partial x_1}\dfrac{\partial h_1}{\partial \theta} + \dfrac{\partial h_2}{\partial \theta} \right) + \dfrac{\partial h_3}{\partial \theta}\right) . \stepcounter{equation}\tag{\theequation}\label{RD2}
\end{align*}
Note that if one directly applyies the chain rule to calculate the gradient, one ends up with \eqref{RD2}. Our point here is that this form is not convenient to work with. Although we can represent each of the six terms in \eqref{RD2} as nested vjp, for example 
\begin{align*} 
\dfrac{\partial f}{\partial x_2}\dfrac{\partial h_2}{\partial x_1}\dfrac{\partial h_1}{\partial \theta} = \text{vjp}(\text{vjp}(\text{vjp}(1, f, x_2)^\top, h_2, x_1)^\top, h_1, \theta)
\end{align*}
it's clear that evaluating each of these six terms independently would be highly inefficient because of the repeated calculations. So it is logical to then grouping common factors, 
\begin{align*} 
\left(\dfrac{\partial f}{\partial x_1}+ \left(\dfrac{\partial f}{\partial x_2} + \dfrac{\partial f}{\partial x_3}\dfrac{\partial h_3}{\partial x_2}\right)\dfrac{\partial h_2}{\partial x_1}\right)\dfrac{\partial h_1}{\partial \theta} + \left(\dfrac{\partial f}{\partial x_2} +  \dfrac{\partial f}{\partial x_3}\dfrac{\partial h_3}{\partial x_2}\right)\dfrac{\partial h_2}{\partial \theta} + \dfrac{\partial f}{\partial x_3}\dfrac{\partial h_3}{\partial \theta}
\end{align*}
and we arrive at an expanded form of \eqref{2.2} (each $\lambda_i$ is expanded according to \eqref{2.3}). 

The point of this is to illustrate that a reverse differentiation technique like the adjoint method is neither mysterious nor ``just the chain rule''. Rather, the adjoint method gives a mechanical procedure for calculating the chain rule in the ``correct'' order, an order that can be computed efficiently by vector-Jacobian products. It is clumsy to try to calculate the gradient directly using the chain rule, and one ends up having to manipulate the expressions into the form which the adjoint method provides by default.
\chapter{Chapter 2 Appendix}
\section{Proof of Theorem 2.4}
Because $\tanh$ is continuous and differentiable, it follows that conditions \ref{a2}, \ref{a4}, \ref{a3} are trivially satisfied. From theorem \ref{unbiased1}, to show unbiasedness of $\hat g(\theta)$ it is sufficient to show that 
\begin{align*} 
\underset{z \in \mathbb{R}^{\kappa_0}}{\int} \underset{\bar \theta \in [\theta, \theta + \alpha e_k]}{\sup}\left\lvert \hat g^k(\bar \theta) \right\rvert \mu(z) dz < \infty \stepcounter{equation}\tag{\theequation}\label{tanh-proof0}
\end{align*}
for each component $k = 1, \ldots, \text{dim}(\theta)$ of the gradient. In \eqref{tanh-proof0}, $\alpha$ is some positive constant (which is nonzero but can be made arbitrary small) and $\mu(z)$ is the joint probability density of $z$. 

The reason why we will consider only a single component of the gradient at a time is because of this problem's structure, where a given weight/bias affects only a single $h_i$. This means that $\hat g^k := \sum_{i=1}^{n} \lambda_i^\top \frac{\partial h_i}{\partial \theta^k} = \lambda_i^\top \frac{\partial h_i}{\partial \theta^k}$ where the parameter $\theta^k$ affects $h_i$ only (that is, $\theta^k$ is one of the entries of either $W_i$ or $b_i$). By considering a single component of the gradient at a time, we also avoid having to deal with the complications of how $\{W_i, b_i : i = 1, \ldots n\}$ should be reshaped into a single parameter vector. 

First let us consider $\theta^k$ belonging to either $W_i$ or $b_i$ for an $i \in \{1, \ldots n-1\}$. Recalling \eqref{NN-eg0} we have
\begin{align*}
\left\lvert \hat g^k \right\rvert & \leq \norm{\lambda_i} \norm{\dfrac{\partial h_i}{\partial \theta^k}} \\
& \leq 2 \shortnorm{x_n - \hat f(z)} \norm{W_n} \prod_{l=1}^{n-1-i}\norm{\text{diag}\big( \sech^2 (W_{n-l}x_{n-l-1} + b_{n-l})\big)}\norm{W_{n-l}} \norm{\dfrac{\partial h_i}{\partial \theta^k}} \\
& \leq 2 \shortnorm{x_n - \hat f(z)} \norm{W_n} \prod_{l=1}^{n-1-i}\norm{W_{n-l}} \norm{\dfrac{\partial h_i}{\partial \theta^k}} \\
& \leq 2 \big(\norm{W_nx_{n-1} + b_n} + \shortnorm{\hat f(z)}\big) \norm{W_n} \prod_{l=1}^{n-1-i}\norm{W_{n-l}} \norm{\dfrac{\partial h_i}{\partial \theta^k}} \\
&  \leq 2 \big(\norm{W_n}\sqrt{\kappa_{n-1}} + \norm{b_n} + \shortnorm{\hat f(z)}\big) \norm{W_n} \prod_{l=1}^{n-1-i}\norm{W_{n-l}} \norm{\dfrac{\partial h_i}{\partial \theta^k}}. \stepcounter{equation}\tag{\theequation}\label{tanh-proof1}
\end{align*}
Now let us establish bounds for the norm of $\partial h_i / \partial \theta^k$. If $i \in \{2, \ldots n-1\}$ then we have the bounds
\begin{align*} 
& \norm{\dfrac{\partial h_i}{\partial W_i^{j,k}}} = \norm{e_j \sech^2\big(W_i^j x_{i-1} + b_i^j\big)x_{i-1}^k } \leq \norm{x_{i-1}^k} \leq 1 \\
& \norm{\dfrac{\partial h_i}{\partial b_i^{j}}} =  \norm{e_j \sech^2\big(W_i^j x_{i-1} + b_i^j\big)} \leq 1
\end{align*}
for $i =1$ we have 
\begin{align*} 
& \norm{\dfrac{\partial h_1}{\partial W_1^{j,k}}} = \norm{e_j \sech^2\big( W_1^jz + b_1^j\big)z^k} \leq \lvert z^k \rvert \qquad \norm{\dfrac{\partial h_1}{\partial b_1^{j}}} = \norm{e_j \sech^2\big(W_1^j z + b_1^j\big)} \leq 1
\end{align*}
and lastly when $i=n$
\begin{align*} 
& \norm{\dfrac{\partial h_n}{\partial W_n^{j,k}}} = \norm{e_j x_{n-1}^k} \leq 1 \qquad \norm{\dfrac{\partial h_n}{\partial b_n^{j}}} = \norm{e_j} \leq 1 .
\end{align*}
If $\theta^k$ belongs to $b_i$ for $i \in \{1, \ldots n-1\}$, or if $\theta^k$ belongs to $W_i$ for $i \in \{2, \ldots n-1\}$ we have from these inequalities and \eqref{tanh-proof1}
\begin{align*} 
\left\lvert \hat g^k \right\rvert \leq 2 \big(\norm{W_n}\sqrt{\kappa_{n-1}} + \norm{b_n} + \shortnorm{\hat f(z)}\big) \norm{W_n} \prod_{l=1}^{n-1-i}\norm{W_{n-l}} \stepcounter{equation}\tag{\theequation}\label{tanh-proof2}
\end{align*}
Now owing to the assumption that each entry of $\theta$ is bounded, it follows that there exists some constants $M_w$, $M_b$ such that $\norm{W_i} < M_w$, $\norm{b_i} < M_b$ for any $i = 1, \ldots n$. Additionally, for any $i = 1, \ldots n$, and any $k = 1, \ldots \text{dim}(\theta)$ we have the inequalities 
\begin{align*}
\underset{\bar \theta \in [\theta, \theta + \alpha e_k]}{\sup} \norm{W_i} \leq M_w + \alpha, \qquad \underset{\bar \theta \in [\theta, \theta + \alpha e_k]}{\sup} \norm{b_i} \leq M_b + \alpha  \stepcounter{equation}\tag{\theequation}\label{tanh-proof22}
\end{align*}
which are derived from the triangle inequality.
From the assumption $\mathe_z\left[\shortnorm{ \hat f(z)}^2 \right] < \infty$ it follows that $\mathe_z\left[\shortnorm{ \hat f(z)} \right] < \infty$ and so it is clear from \eqref{tanh-proof2} that
\begin{align*} 
& \underset{z \in \mathbb{R}^{\kappa_0}}{\int} \underset{\bar \theta \in [\theta, \theta + \alpha e_k]}{\sup}\left\lvert \hat g^k(\bar \theta) \right\rvert \mu(z) dz  \\
\leq &  \underset{z \in \mathbb{R}^{\kappa_0}}{\int} 2\big((M_w + \alpha)\sqrt{\kappa_{n-1}} + M_b+\alpha + \shortnorm{\hat f(z)} \big)\big(M_w + \alpha\big)^{n-i}\mu(z) dz  < \infty \stepcounter{equation}\tag{\theequation}\label{tanh-proof3}
\end{align*}
which shows the integrability condition is satisfied for $\theta^k$ belonging to $b_1, \ldots b_{n-1}$, or $W_2, \ldots W_{n-1}$. Showing the integrability condition is satisfied for the other $\theta^k$ follows similarly. For $\theta^k$ belonging to $W_1$:
\begin{align*} 
& \underset{z \in \mathbb{R}^{\kappa_0}}{\int} \underset{\bar \theta \in [\theta, \theta + \alpha e_k]}{\sup}\left\lvert \hat g^k(\bar \theta) \right\rvert \mu(z) dz \\
\leq & \underset{z \in \mathbb{R}^{\kappa_0}}{\int} 2\big((M_w + \alpha)\sqrt{\kappa_{n-1}} + M_b+\alpha + \shortnorm{\hat f(z)} \big)\big(M_w + \alpha\big)^{n-i}\lvert z^k \rvert \mu(z) dz < \infty \stepcounter{equation}\tag{\theequation}\label{tanh-proof4}
\end{align*}
where the last inequality follows since $\mathe_z\left[\shortnorm{ \hat f(z)}^2\lvert z^k \rvert^2 \right] < \infty$ implies that\\
 $\mathe_z\left[\shortnorm{ \hat f(z)}\lvert z^k \rvert \right] < \infty$. 
Lastly for a $\theta^k$ belonging to either $W_n$ or $b_n$ 
\begin{align*} 
& \underset{z \in \mathbb{R}^{\kappa_0}}{\int} \underset{\bar \theta \in [\theta, \theta + \alpha e_k]}{\sup}\left\lvert \hat g^k(\bar \theta) \right\rvert \mu(z) dz \ \,
\leq  \underset{z \in \mathbb{R}^{\kappa_0}}{\int} 2\big((M_w + \alpha)\sqrt{\kappa_{n-1}} + M_b+\alpha + \shortnorm{\hat f(z)} \big)\mu(z) dz < \infty. \stepcounter{equation}\tag{\theequation}\label{tanh-proof5}
\end{align*}
Having shown the integrability condition \eqref{tanh-proof0} is satisfied for all $\theta^k$, it follows from theorem \ref{unbiased1} that $\hat g(\theta)$ is an unbiased estimator for any $\theta \in \Theta$. Additionally, since \eqref{tanh-proof0} being satisfied also implies 
\begin{align*} 
\underset{z \in \mathbb{R}^{\kappa_0}}{\int} \left\lvert \hat g^k(\theta) \right\rvert \mu(z) dz < \infty
\end{align*}
it follows immediately from theorem \ref{lipschitzcor} that $\mathe_z[f(x_1, \ldots x_n)]$ is Lipschitz continuous for any $\theta \in \Theta$.

Showing that the second moment of $\hat g(\theta)$ is bounded is also trivial after having shown \eqref{tanh-proof3}, \eqref{tanh-proof4}, \eqref{tanh-proof5}. Note that our asumptions  $\mathe_z\left[\shortnorm{ \hat f(z)}^2\lvert z^k \rvert^2 \right] < \infty$, $\mathe_z\left[\shortnorm{ \hat f(z)}\lvert z^k \rvert^2 \right] < \infty$ and $\mathe_z\left[\shortnorm{ \hat f(z)}^2 \right] < \infty$ are what guarentees the second moment bound. 

\subsection*{Showing that the expected gradient is Lipschitz continuous}
We can use the theorem \ref{lipschitzcor} in order to show that the expected gradient, $\frac{\partial}{\partial \theta}\mathe[f]$, is Lipschitz continuous. To do this, we consider a new optimization problem where the forward pass computes a single component of the original problem's gradient. We can then calculate the gradient of this new optimization problem in order to derive an estimator for a single row of the Hessian of the original problem. If, for each row of the Hessian, we derive an unbiased estimator and show that the estimator is integrable, then it follows that we have an unbiased, integrable estimator for the entire Hessian. We can then conclude that the expected gradient is Lipschitz continuous.

Let $\theta^k$ belong to the $i$\textsuperscript{th} layer ($i \in \{1, \ldots n\}$). Now we define the following model, which calculates $\hat g^k(\theta)$.
\begin{align*} 
& x_1 = \tanh(W_1 z + b_1) \\
& \qquad \ \ \ \vdots \\
& x_\gamma = \tanh(W_{\gamma} x_{\gamma-1} + b_{\gamma}) \qquad \gamma = 2, \ldots n-1 \\
& \qquad \ \ \ \vdots \\
& x_n = W_n x_{n-1} + b_n \\
& x_{n+1} = 2\big(x_n - \hat f(z)\big) \\
& x_{n+2} = W_n^\top x_{n+1} \qquad \text{if $i < n$} \\
& \qquad \ \ \ \vdots \\
& x_{n + \gamma} = W_{n+2-\gamma}^\top \text{diag}(\sech^2(W_{n+2-\gamma}x_{n+1-\gamma} + b_{n+2-\gamma}))x_{n+\gamma-1} \qquad \gamma = 3, 4, \ldots n+1-i \\
& \qquad \ \ \ \vdots \\
& x_{2n+2-i} = \dfrac{\partial h_i}{\partial \theta^k}^\top x_{2n+1-i} \stepcounter{equation}\tag{\theequation}\label{tanh-proof6}
\end{align*}
In this model, $x_1$ through $x_n$ calculate the forward pass of the neural network as normal. Then, $x_{n+1}$ through $x_{2n+1-i}$ calculate the adjoint variables for the reverse pass of the network, so that $x_{n+\gamma}$ corresponds to $\lambda_{n+1-\gamma}$ defined by the reverse pass \eqref{NNeg-adjointvariables}. Because $\theta^k$ belongs to the $i$\textsuperscript{th} layer, we compute up to $\lambda_i$ (i.e. up to $x_{2n+1-i}$) and then lastly $x_{2n+2-i}$ gives $\hat g^k(\theta)$.

The loss function used is $f(x_1, \ldots x_{2n+2-i}) = x_{2n+2-i}$ so that the loss function returns $\hat g^k$ (note that this is a scalar quantity). Then by computing the gradient estimator for \eqref{tanh-proof6}, we obtain an estimator for the $k$\textsuperscript{th} row of the Hessian matrix. The adjoint variables for this calculation are given by the following equations.
\begin{align*} 
& \lambda_{2n+2-i}^\top = 1 \\
& \lambda_{2n+1-i}^\top = \dfrac{\partial h_i}{\partial \theta^k}^\top \\
& \qquad \ \ \ \vdots \\
& \lambda_{n+\gamma}^\top = \dfrac{\partial h_i}{\partial \theta^k}^\top \,\prod_{l=1}^{n-i-\gamma+1}W_{i+l}^\top\text{diag}(\sech^2(W_{i+l}x_{i+l-1} + b_{i+l})) \qquad \gamma = n-i, n-i-1, \ldots 2 \\
& \qquad \ \ \ \vdots \\
& \lambda_{n+1}^\top = \lambda_{n+2}^\top W_n^\top \qquad \text{(if $i < n$)} \\
& \lambda_n^T = 2 \lambda_{n+1}^\top \\
& \lambda_{n-1}^\top = \lambda_n^TW_n \\
& \qquad \ \ \ \vdots \\
& \lambda_{\gamma}^\top = \lambda_{\gamma+1}^\top\text{diag}(\sech^2(W_{\gamma+1}x_{\gamma} +b_{\gamma+1}))W_{\gamma+1} + \lambda_{2n+1-\gamma}^\top \big[ W_{\gamma+1}^\top \\
& \hphantom{\lambda_{\gamma}^\top = \quad} \text{diag}(-2\tanh(W_{\gamma+1}x_{\gamma} +b_{\gamma+1}))\text{diag}(\sech^2(W_{\gamma+1}x_{\gamma} +b_{\gamma+1}))\text{diag}(x_{2n-\gamma})W_{\gamma+1}\big] \\
& \hphantom{\lambda_{\gamma}^\top = \quad \text{diag}(-2\tanh(W_{\gamma+1}x_{\gamma} +b_{\gamma+1}))\text{diag}(\sech^2(W_{\gamma+1}x_{\gamma} +b_{\gamma+1}))} \gamma =n-2, n-3, \ldots i \\
& \qquad \ \ \ \vdots \\
& \lambda_{i-1}^\top = \lambda_i^\top \text{diag}(\sech^2(W_i x_{i-1}+ b_i))W_i + \dfrac{\partial h_{2n+2-i}}{\partial x_{i-1}} \qquad (\text{if } 1 < i < n) \\
& \lambda_{\gamma}^\top = \lambda_{\gamma+1}^\top \text{diag}(\sech^2(W_{\gamma+1}x_{\gamma}+b_{\gamma+1}))W_{\gamma+1} \qquad \gamma = i-2,\ldots 1 \stepcounter{equation}\tag{\theequation}\label{tanh-proof7}
\end{align*}
The extension \eqref{adjointv2} is used to derive these equations, as $h_{n+\gamma}$ depends on both $x_{n+1-\gamma}$ and $x_{n+\gamma-1}$ (where $\gamma \in \{3, \ldots n+2-i\}$).

As already explained, the forward pass \eqref{tanh-proof6} corresponds to the computation of $\hat g^k$. We will use the notation $\hat{\mathcal{H}}^{k,m}$ to refer to the $m$\textsuperscript{th} component of the $k$\textsuperscript{th} row of the Hessian, i.e. 
\begin{align*} 
\hat{\mathcal{H}}^{k,m}(\theta) := \dfrac{\partial^2}{\partial \theta^k \partial \theta^m} \bigg(\big(x_n(\theta) - \hat f(z)\big)^\top\big(x_n(\theta) - \hat f(z)\big)\bigg) .
\end{align*}
We can compute any $(k,m)$ entry of the Hessian as 
\begin{align*} 
\hat{\mathcal{H}}^{k,m}(\theta) = \begin{dcases}  \lambda_j^T\dfrac{\partial h_j}{\partial \theta^m} +  \lambda_{2n+2-j}^\top\dfrac{\partial h_{2n+2-j}}{\partial \theta^m} &  j >= i \\
 \lambda_j^T\dfrac{\partial h_j}{\partial \theta^m} & j < i \\
\end{dcases}
\stepcounter{equation}\tag{\theequation}\label{tanh-proof8}
\end{align*}
where the parameter $\theta^m$ is part of the $j$\textsuperscript{th} layer (that is, it belongs to either $W_j$ or $b_j$), $\theta^k$ is part of the $i$\textsuperscript{th} layer, the model is defined per \eqref{tanh-proof6} and adjoint variables defined per \eqref{tanh-proof7}. As a sanity check, it can be verified that indeed $\hat{\mathcal{H}}^{k,m} =  \hat{\mathcal{H}}^{m,k}$. 

When evaluating \eqref{tanh-proof8}, we already have given the derivatives $\partial h_j / \partial W_j^{r,c}$ and $\partial h_j /\partial b_j^{r}$ for $j \in \{1, \ldots n\}$ which were derived in the first part of proof for showing $\hat g$ is unbiased. We still need to give the following derivatives
\begin{adjustwidth}{-3cm}{0cm} \vspace{-1.3cm}\begin{align*} 
& \dfrac{\partial h_{n+2}}{\partial W_n^{r,c}} = e_r x_{n+1}^c  \qquad \ \dfrac{\partial h_{n+2}}{\partial b_n^{r,c}} = 0 \qquad (i < n, j=n) \\
& \dfrac{\partial h_{n+\gamma}}{\partial W_{n+2-\gamma}^{r,c}} = \big(W_{n+2-\gamma}^r\big)^\top\big(-2\tanh(W_{n+2-\gamma}^rx_{n+1-\gamma} + b_{n+2-\gamma}^r)\big)\sech^2(W_{n+2-\gamma}^rx_{n+1-\gamma} + b_{n+2-\gamma}^r)x_{n+\gamma-1}^rx_{n+1-\gamma}^c \\
& \hphantom{\dfrac{\partial h_{n+\gamma}}{\partial W_{n+2-\gamma}^{r,c}} = \ \ } + e_c\sech^2(W_{n+2-\gamma}^rx_{n+1-\gamma} + b_{n+2-\gamma}^r)x_{n+\gamma-1}^r \qquad (i < n-1, j = {n+2 - \gamma}, \gamma \in \{3, \ldots n+1-i\})  \\
 & \dfrac{\partial h_{n+\gamma}}{\partial b_{n+2-\gamma}^r} =  \big(W_{n+2-\gamma}^r\big)^\top\big(-2\tanh(W_{n+2-\gamma}^rx_{n+1-\gamma} + b_{n+2-\gamma}^r)\big)\sech^2(W_{n+2-\gamma}^rx_{n+1-\gamma} + b_{n+2-\gamma}^r)x_{n+\gamma-1}^r  \\
 & \hphantom{\dfrac{\partial h_{n+\gamma}}{\partial b_{n+2-\gamma}^r} =  \big(W_{n+2-\gamma}^r\big)^\top\big(-2\tanh(W_{n+2-\gamma}^rx_{n+1-\gamma}} \qquad (i < n-1, j = {n+2 - \gamma}, \gamma \in \{3, \ldots n+1-i\}) \\
 & \dfrac{\partial h_{2n+2-j}}{\partial \theta^m} = \dfrac{\partial^2 h_i}{\partial \theta^k \partial \theta^m}^\top x_{2n+1-i} \qquad (i = j) .
\end{align*}
\end{adjustwidth}
We will also make use of this expanded form of $\lambda_{\gamma}$
\begin{adjustwidth}{-3cm}{0cm} \vspace{-1.3cm} \begin{align*} 
 \lambda_{\gamma}^\top = & \lambda_{n-1}^\top \prod_{l=1}^{n-\gamma-1} \text{diag}\big(\sech^2(W_{n-l}x_{n-l-1} + b_{n-l})\big)W_{n-l} + \sum_{s=1}^{n-\gamma-1}\Big[ \lambda_{n+2+s}^\top W_{n-s}^\top \text{diag}\big(-2 \tanh(W_{n-s}x_{n-1-s} + b_{n-s})\big) \\
& \text{diag}\big( \sech^2(W_{n-s}x_{n-1-s} + b_{n-s})\big) \text{diag}(x_{n+1+s}) W_{n-s} \prod_{l=1}^{n-s-\gamma-1}\text{diag}\big( \sech^2(W_{n-s-l}x_{n-s-l-1} + b_{n-s-l}) \big)W_{n-s-l} \Big]
\end{align*}
\end{adjustwidth}
which is valid for $\gamma \in \{n-2, \ldots, i\}$. 

At this point we have given closed form expressions for the $(k,m)$ entry of the Hessian estimator, for any $(k, m)$ such that $m>=k$. Since the Hessian is symmetric, this is sufficient to evaluate every entry of the Hessian. Now to show unbiasedness we need to verify that the integrability condition
\begin{align*} 
\underset{z \in \mathbb{R}^{\kappa_0}}{\int} \underset{\bar \theta \in [\theta, \theta + \alpha e_m ]}{\sup}\left\lvert \hat{\mathcal{H}}^{k,m}(\bar \theta)\right\rvert \mu(z) dz < \infty \stepcounter{equation}\tag{\theequation}\label{tanh-proof9}
\end{align*}
is satisfied for every entry of the Hessian. This follows using the same arguments we used to show that $\hat g$ is unbiased; in particular since $\theta$ is assumed to be bounded we have the inequalities \eqref{tanh-proof22}. To give an example, let us consider $\theta^k = W_3^{r,c}$ and $\theta^m = W_5^{q,d}$. 
\begin{adjustwidth}{-3cm}{0cm} \vspace{-1.3cm} \begin{align*} 
& \hat{\mathcal{H}}^{k,m} = \lambda_5^\top \dfrac{\partial h_5}{\partial W_5^{q,d}} + \lambda_{2n-3}^\top \dfrac{\partial h_{2n-3}}{\partial W_5^{q,d}} \\
& = \lambda_{n-1}^\top \prod_{l=1}^{n-6}\Big[ \text{diag}\big(\sech^2(W_{n-l}x_{n-l-1} + b_{n-l})\big)W_{n-l}\Big]\dfrac{\partial h_5}{\partial W_5^{q,d}} + \sum_{s=1}^{n-6}\Big[ \lambda_{n+2+s}^\top W_{n-s}^\top \text{diag}\big(-2 \tanh(W_{n-s}x_{n-1-s} + b_{n-s})\big) \\
& \text{diag}\big( \sech^2(W_{n-s}x_{n-1-s} + b_{n-s})\big) \text{diag}(x_{n+1+s}) W_{n-s} \prod_{l=1}^{n-s-6}\text{diag}\big( \sech^2(W_{n-s-l}x_{n-s-l-1} + b_{n-s-l}) \big)W_{n-s-l} \Big]\dfrac{\partial h_5}{\partial W_5^{q,d}}\\
& + \sech^2(W_3^rx_2 + b_3^r)x_2^c e_r^\top W_4^\top \text{diag}\big(\sech^2(W_4x_3 + b_4)\big)\Big((W_5^r)^\top (-2\tanh(W_5^r x_4 + b_5^r))\sech^2(W_5^rx_4 + b_5^r)x_{2n-4}^rx_{4}^c \\
& + e_c\sech^2(W_5^rx_4 + b_5^r)x_{2n-4}^r\Big)
\end{align*}
\end{adjustwidth}
This estimator is easily bounded
\begin{adjustwidth}{-3cm}{0cm} \vspace{-1.3cm}
\begin{align*} 
& \underset{\bar \theta \in [\theta, \theta + \alpha e_m ]}{\sup} \left\lvert \hat{\mathcal{H}}^{k,m}(\bar \theta) \right\rvert \\
& \leq \underset{\bar \theta \in [\theta, \theta + \alpha e_m ]}{\sup} \norm{\lambda_{n-1}}(M_w + \alpha)^{n-6} + \sum_{s=1}^{n-6}\norm{\lambda_{n+2+s}}\norm{x_{n+1+s}}(M_w+\alpha)^{n-s-4} + 2(M_w+\alpha)^2\norm{x_{2n-4}} \\
& \leq  \underset{\bar \theta \in [\theta, \theta + \alpha e_m ]}{\sup} (M_w + \alpha)^{2n-8} + \sum_{s=1}^{n-6}2\shortnorm{x_n(\bar \theta) - \hat{f}(z)}(M_w+\alpha)^{2n-s-8} + 4\shortnorm{x_n(\bar \theta) - \hat f(z)} (M_w + \alpha)^{n-3} \stepcounter{equation}\tag{\theequation}\label{tanh-proof10}
\end{align*}
\end{adjustwidth}
and \eqref{tanh-proof10} is clearly integrable. Since all entries of the Hessian are integrable, and conditions \ref{a2}, \ref{a4} and \ref{a3} are satisfied because $\tanh$ and $\sech$ are continuous and differentiable, it follows from theorem \ref{unbiased1} that $\hat{\mathcal{H}}(\theta)$ is an unbiased Hessian estimator. Therefore from theorem \eqref{lipschitzcor}, the expected gradient of the original problem, $\dfrac{\partial}{\partial \theta} \mathe_z[f(x_1, \ldots, x_n) ]$ is Lipschitz continuous.  \qedhere

\section{Proof of Theorem 2.5}
In this case, it will not be so straightforward to verify condition \ref{a4}. 
Thus, we will elect to use the theorem \ref{unbiasedthm2}. Since $ReLU$ is a Lipschitz function and $Wx + b$ is locally Lipschitz, it follows that
each layer will be locally Lipschitz, so condition \ref{a6} is satisfied. Quadratic functions are also locally Lipschitz, so condition \ref{a5} will be satisfied as well. To apply the theorem we need to identify the Lipschitz constants so we can then verify the integrability condition. 
It is straightforward to verify 
\begin{align*} 
& \norm{f(x_n^*) - f(x_n^{**})} \leq \Big(\underset{\bar \theta \in S(\theta)}{\sup}\norm{x_n(\bar \theta, z)} + \shortnorm{\hat f(z)}\Big) \norm{x_n^* - x_n^{**}} \\
& \norm{h_i(x_{i-1}^*, \theta) - h_i(x_{i-1}^{**}, \theta)} \leq \norm{W_i}\norm{x_{i-1}^* - x_{i-1}^{**}} \qquad (i>1) \\
& \norm{h_i(x_{i-1}, \theta_1) - h_i(x_{i-1}, \theta_2)} \leq \kappa_i (\norm{x_{i-1}} + 1) \norm{\theta_1 - \theta_2} \qquad( i>1) \\
& \norm{h_1(\theta_1, z) - h_1(\theta_2, z)} \leq \kappa_0 (\norm{z} + 1) \norm{\theta_1 - \theta_2}
\end{align*}
where $x_i^* = x_i(\theta_1, z), x_i^{**} = x_i(\theta_2, z), \theta_1, \theta_2$ satisfy $\norm{\theta_1 - \theta} \leq \epsilon$, $\norm{\theta_2 - \theta} \leq \epsilon$, and the set $S(\theta)$ is defined as $\{\bar \theta: \norm{\bar \theta - \theta} < \epsilon\}$. The constant $\epsilon >0$ can be chosen arbitrarily. Let $\mathcal{ND}(\theta)$ denote the subset of $\mathbb{R}^{\kappa_0}$ for which $\hat g(\theta, z)$ does not exist when $z \in \mathcal{ND}(\theta)$. By theorem \ref{unbiasedthm2}, $\hat g(\theta)$ is unbiased if
\begin{align*} 
& \underset{z \in \mathbb{R}^{\kappa_0} \setminus \mathcal{ND}(\theta)}{\int} \Big(\underset{\bar \theta \in S(\theta)}{\sup}\norm{x_n(\bar \theta, z)} + \shortnorm{\hat f(z)}\Big)\Big( \sum_{j=2}^n \prod_{k=1}^{n-j}\norm{W_{n+1-k}}\kappa_{j} (\norm{x_{j-1}(\theta^*, z)} + 1) \\
& \hphantom{\underset{z \in \mathbb{R}^{\kappa_0} \setminus \mathcal{ND}(\theta)}{\int}} + \prod_{k=1}^{n-1}\norm{W_{n-k+1}}\kappa_1(\norm{z} + 1)\Big)\mu(z) dz < \infty \stepcounter{equation}\tag{\theequation}\label{relu-proof0}
\end{align*}
for any $\theta^*$ such that $\norm{\theta^* - \theta} \leq \epsilon$. For \eqref{relu-proof0} to be true, clearly every weight matrix $W_i$ needs to have bounded norm, so it follows that each parameter must be bounded (i.e. let $\theta \in \Theta$). Moreover the output $x_n(\theta, z)$ is just a composition of affine transformations, so again assuming that each parameter is bounded, there exists some $A, b \in \mathbb{R}$ such that $\norm{x_n(\theta, z)} \leq A \norm{z} + b$. A similar argument holds for the other $x_i$, for $i \in \{1, \ldots, n-1\}$. It follows that \eqref{relu-proof0} is integrable provided that $\mathe_z\left[\shortnorm{\hat f(z)}\right] < \infty$ and $\mathe_z\left[\shortnorm{\hat f(z)}\lvert z^k \rvert\right] < \infty$ for any $k = 1, \ldots \kappa_0$. 
Then we have shown that $\hat g(\theta)$ is unbiased provided that $\theta \in \Theta$. From the inequality \eqref{lipschitz-inequality} and \eqref{relu-proof0} we have
\begin{adjustwidth}{-2cm}{0cm} \vspace{-1.3cm}
\begin{align*} 
& \lvert \hat g^k(\theta) \rvert = \left\lvert \dfrac{\partial}{\partial \theta^k}f(x_1, \ldots x_n)\right\rvert \leq  \Big(\underset{\bar \theta \in S(\theta)}{\sup}\norm{x_n(\bar \theta, z)} + \shortnorm{\hat f(z)}\Big)\Big( \sum_{j=2}^n \prod_{k=1}^{n-j}\norm{W_{n+1-k}}\kappa_{j} (\norm{x_{j-1}(\theta^*, z)} + 1) \\
& \hphantom{\lvert \hat g^k(\theta) \rvert = \left\lvert \dfrac{\partial}{\partial \theta^k}f(x_1, \ldots x_n)\right\rvert \leq} + \prod_{k=1}^{n-1}\norm{W_{n-k+1}}\kappa_1(\norm{z} + 1)\Big)  \stepcounter{equation}\tag{\theequation}\label{relu-proof1}
\end{align*}
\end{adjustwidth}
and so we can immediately conclude that $\mathe_z[f(x_1, \ldots x_n)]$ is Lipschitz continuous on $\Theta$.

Squaring both sides of \eqref{relu-proof1} we find that the second moment will be bounded if all of $\mathe_z\left[\shortnorm{\hat f(z)}^2\right]$, $\mathe_z\left[\shortnorm{\hat f(z)}^2 \, \lvert z^k \rvert \right]$, $\mathe_z\left[\shortnorm{\hat f(z)}^2 \, \lvert z^k \rvert^2 \right]$, $\mathe_z\left[\shortnorm{\hat f(z)} \, \lvert z^k \rvert^2 \right]$, $\mathe_z\left[\shortnorm{\hat f(z)} \, \lvert z^k \rvert^3 \right]$ are finite. \qedhere

\subsection*{Definitions and Properties of Lipschitz Continuous Functions} \label{lipschitz-appendix}
\begin{definition}
A function $g(x) : \mathbb{R}^m \rightarrow \mathbb{R}^n$ is Lipschitz continuous on $X$ if 
\begin{align*} 
\norm{g(x^*) - g(x)} \leq K \norm{x^* - x} \quad \forall x^*, x \in X .
\end{align*}
for some Lipschitz constant $K$. If $X$ is not specified it is assumed to be all of $\mathbb{R}^m$.
\end{definition}

\begin{definition}
A function $g(x) : \mathbb{R}^m \rightarrow \mathbb{R}^n$ is locally Lipschitz continuous at $x_0$ if 
\begin{align*} 
\norm{g(x^*) - g(x)} \leq K(x_0) \norm{x^* - x} \quad \forall x^*, x \text{ such that }\norm{x^* - x_0} < \delta, \norm{x - x_0} < \delta.
\end{align*}
for some Lipschitz constant $K(x_0)$. If $x_0$ is not specified $g(x)$ is assumed to be locally Lipschitz at any $x_0 \in \mathbb{R}^m$. 
\end{definition}

\begin{lemma} \label{lipschitz1}
Let $g(x) : \mathbb{R}^m \rightarrow \mathbb{R}^n$ be Lipschitz continuous. Then, $g(x)$ is almost-everywhere differentiable. At every point $x \in \mathbb{R}^m$ where $g(x)$ is differentiable,
\begin{align*} 
\norm{ \dfrac{d g}{d x} } \leq M 
\end{align*}
where $M$ is the Lipschitz constant and $\norm{\cdot}$ denotes the induced matrix norm.
\end{lemma}
The fact that $g(x)$ is almost-everywhere differentiable is due to Radermacher's Theorem \cite{radermacher-thm}. \\
To show the bound on the Jacobian, consider some point $x$ where $g(x)$ is differentiable, let $t$ be some positive scalar constant, and $v$ some vector in $\mathbb{R}^m$. By definition, 
\begin{align*} 
 & M \geq  \dfrac{\norm{g(x + tv)  - g(x)}}{\norm{tv}}  \\
 =& \dfrac{\norm{\frac{1}{t}\big[\,g(x + tv)  - g(x)\, \big]}}{\norm{v}} .
\end{align*}
Now consider the limit $t \rightarrow 0$:
\begin{align*} 
= \dfrac{\norm{\frac{d g(x)}{dx}v}}{\norm{v}} = \norm{\dfrac{d g(x)}{d x} \dfrac{v}{\norm{v}}} .
\end{align*}
As this holds for any $v$, $x$, it follows that $\norm{dg / dx} \leq M$. \qed 

\begin{lemma} \label{lipschitz2}
Let $g(x) : \mathbb{R}^m \rightarrow \mathbb{R}^n$ be continuous for any $x \in X$. For any two points $x, x^* \in X$, consider the interval $I := \{x + t(x^* - x) : t \in [0, 1]\}$. Let $g(x_0)$ be differentiable for $x_0 \in I \setminus \mathcal{B}$ where $\mathcal{B}$ is a finite set. Lastly, let the Jacobian $d g / d x$ be bounded for every point at which $g(x)$ is differentiable. Then $g$ is Lipschitz continuous on $X$, with lipschitz constant 
\begin{align*} 
M = \underset{x \in X}{\sup} \norm{ \dfrac{d g}{d x} }.
\end{align*}
\end{lemma}
Let $x, x^*$ be two arbitrary points in $X$. Define 
\begin{align*} 
\bar g(t) := g\big(x + t(x^* - x)\big) \quad \forall t \in [0, 1]
\end{align*}
where $t$ is a scalar. Let $\{\alpha_i : i = 1, \ldots, k-1\}$ be the set of all points in $(0, 1)$ where $\bar g(t)$ is not differentiable, sorted in descending order. Also let $\alpha_0 = 1$ and $\alpha_k = 0$. 
\begin{align*} 
& \norm{ \bar g(1) - \bar g(0) }\\
= & \norm{ \bar g(\alpha_0) - \bar g(\alpha_k) + \sum_{i=1}^{k-1} \big(\bar g(\alpha_i) - \bar g(\alpha_i)\big)  } \\
\leq & \sum_{i=0}^{k-1} \norm{ \bar g(\alpha_i) - \bar g(\alpha_{i+1}) } \stepcounter{equation}\tag{\theequation}\label{lipschitz2proof0}
\end{align*}
Now since by construction $\bar g$ is differentiable on each $(\alpha_i, \alpha_{i+1})$, and $\bar g$ is continuous by assumption, we can use Rudin's mean value theorem \cite{mvt-thm} (theorem 5.19) which states that 
\begin{align*} 
\norm{ \bar g(\alpha_i) - \bar g(\alpha_{i+1}) } \leq (\alpha_i  - \alpha_{i+1}) \norm{ \dfrac{d \bar g}{d t}\big|_{c_i} }
\end{align*}
for some $c_i \in (\alpha_i, \alpha_{i+1})$. Applying this to \eqref{lipschitz2proof0}, 
\begin{align*} 
\leq &\sum_{i=0}^{k-1} (\alpha_i  - \alpha_{i+1}) \norm{ \dfrac{d \bar g}{d t}\big|_{c_i} } \\
= &  \sum_{i=0}^{k-1} (\alpha_i  - \alpha_{i+1}) \norm{\dfrac{d g}{d x}\big|_{x + c_i(x^* - x)} (x^* - x) } \\
= & \sum_{i=0}^{k-1} (\alpha_i  - \alpha_{i+1}) \norm{ \dfrac{d g}{d x}\big|_{x + c_i(x^* - x)} \dfrac{x^* - x}{\norm{x^* - x}}  } \norm{x^* - x} \\
\leq & \sum_{i=0}^{k-1} (\alpha_i  - \alpha_{i+1}) \underset{x_0 \in X}{\sup}\norm{ \dfrac{d g}{d x}\big|_{x_0} \dfrac{x^* - x}{\norm{x^* - x}}  } \norm{x^* - x} \\
\leq & \sum_{i=0}^{k-1} (\alpha_i  - \alpha_{i+1}) M \norm{ x^* - x} \\
= & M \norm{ x^* - x } .  \qedhere
\end{align*}
\chapter{Chapter 3 Appendix}
\subsection*{SDE Experiment Architecture}
All models have two seperate neural networks. All neural networks use the same architecture with the same inputs ($x_{i-12}, \ldots, x_{i-1}, t_i, \theta$) with the only exception being different shapes for the output layers. There are three fully connected hidden layers with 100 units each followed by the output layer. The first two fully connected layers use $\tanh$ activation and the last uses ReLU activation. There is also a skip connection between the first and third hidden layer. The time $t_i$ is encoded by a 4 dimensional vector, two dimensions each for the time of day and time of year.

For the standard SDEs (both the Huber and MLE versions), the two neural networks represent the drift and diffusion. The drift output is in $\mathbb{R}^{20}$. The diffusion matrix $A$ is parametrized as the cholesky factor of the covariance matrix (as opposed to assuming a diagonal covariance matrix), so it has $20*(20+1)/2 = 210$ outputs. The jump ODE neural networks represent the drift and jump magnitudes/probabilities. The latter outputs 100 probabilities and 80 different jump magnitudes, as we assume one of the ``jumps'' corresponds to 0 magnitude (i.e. no jump) and $l = 5$ total jump possibilities. The piecewise ODE neural networks output the drifts (of which there $l = 5$ in each dimension) and the probabilities, so there are 100 outputs for both networks.

We use ADAM as the optimizer, and gradually increase the prediction length of the model during training, from 3 hours up to the chosen target of 2 days. The learning rate is also gradually reduced. All models have L2 regularization added, and the Huber SDE additionally has a entropy maximization term added as regularization. Some minor hyperparameter optimization was done manually.

\subsection*{Modeling using the Multivariate Normal Distribution} \label{appendix-multivariate-normal}
In $\mathbb{R}^k$, we parametrize the distribution as
\begin{align*} 
\mu(\theta) + A(\theta)z \stepcounter{equation}\tag{\theequation}\label{multivariate-normal}
\end{align*}
where $\mu : \theta \rightarrow \mathbb{R}^k$ is the mean, $z \in \mathbb{R}^k$ is a vector of standard normals, and $A : \theta \rightarrow \mathbb{R}^{k\times k}$ is the cholesky factor of the covariance matrix. That is, $A$ is a nonsingular, lower triangular matrix such that $\Sigma := A A^T$ is the positive definite covariance matrix. To ensure $A$ is nonsingular (and thus, that the covariance is positive definite), it is sufficient that the diagonal of $A$ contains only positive entries. We can ensure this by exponentiating the diagonal entries of some arbitrary lower triangular matrix. 

To compute the probability density, we require both $\Sigma^{-1}$ and $\text{det}(\Sigma)$. $\Sigma^{-1} = A^{-T}A^{-1}$ can be computed efficiently by using forward substitution on $A$. $\text{det}(\Sigma)^{1/2} = \text{det}(A)$ is given by taking the product of the diagonal entries of $A$. To sample from the distribution, generate $z$ and use \eqref{multivariate-normal}.

If the dimension $k$ is large, we may elect to consider $A$ as a diagonal matrix, or impose some other special structure on $A$.

\subsection*{The Deterministic Policy Gradient}
With \eqref{model-based-1} in mind, observe that
\begin{adjustwidth}{-3cm}{0cm} \vspace{-1.3cm}
\begin{align*} 
& \mathe_{y_{(i,2)}^{(n,2)}, z_{(i+1,1)}^{(n,1)},} \left[\Big(\gamma^{i-1}\dfrac{\partial R(s_i, a_i)}{\partial a_i} + \sum_{k=i}^n\gamma^{k-1}R(s_k, a_k) \dfrac{\partial \log env(y_{(i,2)} \, | \, s_i, a_i)}{\partial a_i} \Big)\dfrac{\partial \pi(s_i)}{\partial \theta}  \ \bigg| \ z_{(1,1)}^{(i,1)}, y_{(1,2)}^{(i-1,2)}\right] \\
= & \underset{y_{(i,2)}}{\int}\underset{z_{(i+1,1)}}{\int}\ldots \underset{z_{(n,1)}}{\int}\underset{y_{(n,2)}}{\int}\Big[\Big(\gamma^{i-1}\dfrac{\partial R(s_i, a_i)}{\partial a_i} + \sum_{k=i}^n\gamma^{k-1}R(s_k, a_k) \dfrac{\partial \log env(y_{(i,2)} \, | \, s_i, a_i)}{\partial a_i} \Big)\dfrac{\partial \pi(s_i)}{\partial \theta} \\
& \hphantom{\underset{z_{(i,1)}}{\int}\underset{y_{(i,2)}}{\int}\ldots \underset{z_{(n,1)}}{\int}\underset{y_{(n,2)}}{\int}\Big[} env(y_{(i,2)}\, | \, s_i, a_i)\ldots env(y_{(n,2)}\, | \, s_n, a_n) \mu(z_{(i,1)}^{(n,1)})\Big] dy_{(i,2)}^{(n,2)} dz_{(i+1,1)}^{(n,1)} \\
= & \underset{y_{(i,2)}}{\int} \Big( \gamma^{i-1}\dfrac{\partial R_i }{\partial a_i}\dfrac{\partial \pi(s_i)}{\partial \theta}env(y_{(i,2)} \, | \, \cdot) + \dfrac{\partial env(y_{(i,2) \, | \, \cdot})}{\partial a_i}\dfrac{\partial \pi(s_i)}{\partial \theta}\Big(\gamma^{i-1}R_i + \underset{z_{(i+1,1)}}{\int}\underset{y_{(i+1,2)}}{\int}\mu(z_{(i+1, 1)})env(y_{(i+1, 2)} \, | \, \cdot)\Big(\gamma^i R_{i+1} \\
& + \underset{z_{(i+2,1)}}{\int}\underset{y_{(i+2,2)}}{\int}\ldots \underset{z_{(n,1)}}{\int}\underset{y_{(n,2)}}{\int}\gamma^{n-1}R_n\, \mu(z_{(n,1)}) env(y_{(n,2)} \, | \, \cdot)\Big)\,dy_{(i,2)}^{(n,2)} dz_{(i+1,1)}^{(n,1)} \\
= & \dfrac{\partial}{\partial a_i}\underset{y_{(i,2)}}{\int} env(y_{(i,2)} \, | \, \cdot) \Big( \gamma^{i-1}R_i  + \underset{z_{(i+1,1)}}{\int}\underset{y_{(i+1,2)}}{\int}\mu(z_{(i+1, 1)})env(y_{(i+1, 2)} \, | \, \cdot)\Big(\gamma^i R_{i+1} \\
& + \underset{z_{(i+2,1)}}{\int}\underset{y_{(i+2,2)}}{\int}\ldots \underset{z_{(n,1)}}{\int}\underset{y_{(n,2)}}{\int}\gamma^{n-1}R_n\, \mu(z_{(n,1)}) env(y_{(n,2)} \, | \, \cdot)\, \Big) dy_{(i,2)}^{(n,2)} dz_{(i+1,1)}^{(n,1)}  \dfrac{\partial \pi(s_i)}{\partial \theta} \\
& = \dfrac{\partial Q^{\pi}(s_i, a_i)}{\partial a_i}\dfrac{\partial \pi(s_i)}{\partial \theta} \stepcounter{equation}\tag{\theequation}\label{ddpg-eqn2}
\end{align*}
\end{adjustwidth}

where we have defined $R_{i} := R(s_i, a_i)$. The third equality rearranges terms, and the fourth equality pulls the partial derivative with respect to $a_i$, and also $\partial \pi / \partial \theta$, outside the integral. We then see that the integral is simply equal to $Q^{\pi}(s_i, a_i)$. The ddpg update \eqref{ddpg-eqn} then follows from applying \eqref{ddpg-eqn2} to \eqref{model-based-1}.

\chapter{Chapter 4 Appendix}
\section*{Proofs}
\subsection{Lemma 4.1}
\begin{align*}
&\dfrac{\partial}{\partial \theta_k} \underset{y_i}{\int} p_i(y_i \, | \, z_i, x_{i-1}, \theta) dy_i = \underset{\alpha \rightarrow 0}{\lim}  \bigg(\underset{y_i}{\int} p_i(y_i \, | \, z_i, x_{i-1}, \theta + \alpha e_k) dy_i - \underset{y_i}{\int} p_i(y_i \, | \, z_i, x_{i-1},  \theta) dy_i \bigg)/\alpha \\
= & \underset{\alpha \rightarrow 0}{\lim}  \underset{y_i}{\int} \dfrac{ p_i(y_i \, | \, z_i, x_{i-1}, \theta + \alpha e_k) -  p_i(y_i \, | \, z_i, x_{i-1},\theta)}{\alpha}dy_i  \stepcounter{equation}\tag{\theequation}\label{lemma1proof1}
\end{align*}
Where we define $e_k$ as the $k$ unit vector, i.e. a vector with 1 in index $k$ and 0's elsewhere. The  $k$\textsuperscript{th} component of $\theta$ is denoted as $\theta_k$. \\
By assumption, $p_i(\cdot)$ is differentiable with respect to $\theta$, so by the mean value theorem \cite{dieudonne-mvt-thm} (Theorem 8.5.2) we have 
\begin{align*}
\left \lvert \dfrac{p_i(y_i \, | \, z_i, x_{i-1}, \theta + \alpha e_k) - p_i(y_i \, | \, z_i, x_{i-1}, \theta)}{\alpha} \right\rvert \leq \underset{\bar \theta \in [\theta, \theta + \alpha e_k]}{\sup}\left \lvert \dfrac{p_i(y_i \, | \, z_i, x_{i-1}, \bar \theta_k)}{\partial \bar \theta_k} \right \rvert \\
\leq \underset{\bar \theta \in [\theta, \theta + \alpha e_k]}{\sup} \norm{\dfrac{p_i(y_i \, | \, z_i, x_{i-1}, \bar \theta)}{\partial \bar \theta}} \, .\stepcounter{equation}\tag{\theequation}\label{lemma1proof2}
\end{align*}
where $\alpha < \epsilon_{\theta}$. As \eqref{lemma1proof2} is integrable by assumption \ref{assumption1}b), this establishes the dominating function for \eqref{lemma1proof1}; we can use Lebesgue's dominated convergence theorem to conclude that \eqref{lemma1proof1} is equal to 
\begin{align*} 
 \underset{y_i}{\int} \underset{\alpha \rightarrow 0}{\lim} \dfrac{ p_i(y_i \, | \, z_i, x_{i-1}, \theta + \alpha e_k) -  p_i(y_i \, | \, z_i, x_{i-1},\theta)}{\alpha}dy_i  =  \underset{y_i}{\int} \dfrac{\partial p_i(y_i \, | \, z_i, x_{i-1},\theta)}{\partial \theta_k}dy_i .
\end{align*}
From the definition of probability density, we also have that 
\begin{align*} 
\dfrac{\partial}{\partial \theta_k} \underset{y_i}{\int} p_i(y_i \, | \, z_i, x_{i-1}, \theta) dy_i = \dfrac{\partial}{\partial \theta_k} 1 = 0 .
\end{align*}
Because we can make this argument for an arbitrary value of $k$, it follows that 
\pushQED{\qed}
\begin{align*} 
& \underset{y_i}{\int} \dfrac{\partial p_i(y_i \, | \, z_i, x_{i-1},\theta)}{\partial \theta}dy_i = 0.  \qedhere
\end{align*}
\popQED
\subsection{Theorem 4.1}
For any index $i \in 1, \ldots, n$:
\begin{adjustwidth}{-3cm}{0cm}\vspace{-1.3cm}
\begin{align*} 
& \mathe_{z_1^n, y_1^n}\bigg[ \beta_i(\phi) \dfrac{\partial \log p_i(y_i \, | \, x_{i-1}, z_i)}{\partial \theta} \bigg] \\
= & \underset{z_1}{\int} \mu(z_1)\ldots \underset{z_n}{\int}\mu(z_n)\underset{y_1}{\int}p_1(y_1 \, | \, z_1, x_0) \ldots \underset{y_{i-1}}{\int}p_{i-1}(y_{i-1} \, | \, z_{i-1}, x_{i-2}) \underset{y_i}{\int}\beta_i(\phi)\dfrac{\partial p_i(y_i \, | \, z_i, x_{i-1})}{\partial \theta} \underset{y_{i+1}}{\int}p_{i+1}(y_{i+1} \, | \, z_{i+1}, x_{i})\ldots \\
& \underset{y_n}{\int} p_n(y_n \, | \, z_n, x_{n-1}) dz_1\ldots dz_n dy_1 \ldots dy_n \\
= & \underset{z_1}{\int} \mu(z_1)\ldots \underset{z_n}{\int}\mu(z_n)\underset{y_1}{\int}p_1(y_1 \, | \, z_1, x_0) \ldots \underset{y_{i-1}}{\int}p_{i-1}(y_{i-1} \, | \, z_{i-1}, x_{i-2}) \beta_i(\phi) \bigg( \underset{y_i}{\int}\dfrac{\partial p_i(y_i \, | \, z_i, x_{i-1})}{\partial \theta} dy_i\bigg) dz_1\ldots dz_n dy_1 \ldots dy_{i-1}  \\
= & \underset{z_1}{\int} \mu(z_1)\ldots \underset{z_n}{\int}\mu(z_n)\underset{y_1}{\int}p_1(y_1 \, | \, z_1, x_0) \ldots \underset{y_{i-1}}{\int}p_{i-1}(y_{i-1} \, | \, z_{i-1}, x_{i-2}) \beta_i(\phi) \big( {0} \big) dz_1\ldots dz_n dy_1 \ldots dy_{i-1} = 0 .
\end{align*}
\end{adjustwidth}
The second equality uses that $\beta_i$ does not depend on $y_i$ through $y_n$; the third equality uses lemma \ref{lemma1}. $\hfill \qed$
\subsection{Proposition 4.1}
\begin{adjustwidth}{-2cm}{0cm} \vspace{-1.3cm}
\begin{align*} 
& \mathe_{z_1^n, y_1^n}\bigg[ \sum_{i=1}^{n}f(x_1, \ldots, x_n) \dfrac{\partial \log p(y_i \, | \, z_i, x_{i-1})}{\partial \theta} \bigg] \ = \   \mathe_{z_1^n}\bigg[ \mathe_{y_1^n}\bigg[ \sum_{i=1}^{n}f(x_1, \ldots, x_n) \dfrac{\partial \log p(y_i \, | \, z_i, x_{i-1})}{\partial \theta} \ \bigg| \ z_1^n \ \bigg] \bigg] \\
\end{align*}
\end{adjustwidth}
Focusing on the inner expectation:
\begin{align*}
& \mathe_{y_1^n}\bigg[ \sum_{i=1}^{n}f(x_1, \ldots, x_n) \dfrac{\partial \log p(y_i \, | \, z_i, x_{i-1})}{\partial \theta} \ \bigg| \ z_1^n  \bigg] \\
& = \underset{y_1, \ldots, y_n}{\int} \sum_{i=1}^{n}\sum_{k=1}^n \hat{f}(x_k) \prod_{\substack{j=1 \\ j \neq i}}^{n} p_j(y_j \, |\, \cdot ) \dfrac{\partial p_i(y_i \, | \, \cdot)}{\partial \theta}dy_1, \ldots, dy_n \\
& = \sum_{i=1}^{n}\sum_{k=1}^n \underset{y_1, \ldots, y_v}{\int}  \hat{f}(x_k) \prod_{\substack{j=1 \\ j \neq i}}^{v} p_j(y_j \, |\, \cdot ) \dfrac{\partial p_i(y_i \, | \, \cdot)}{\partial \theta}dy_1, \ldots, dy_v . \stepcounter{equation}\tag{\theequation}\label{prop1eqn1}
\end{align*}
where $v = \max\{{i, k}\}$. All terms with $i > k$ drop out due to lemma \ref{lemma1} since
\begin{align*} 
& \underset{y_1, \ldots, y_i}{\int}  \hat{f}(x_k) \prod_{j=1}^{i-1} p_j(y_j \, |\, \cdot ) \dfrac{\partial p_i(y_i \, | \, \cdot)}{\partial \theta}dy_1, \ldots, dy_i \\
& = \underset{y_1, \ldots, y_{i-1}}{\int}  \hat{f}(x_k) \prod_{j=1}^{i-1} p_j(y_j \, |\, \cdot ) \bigg(\underset{y_i}{\int}\dfrac{\partial p_i(y_i \, | \, \cdot)}{\partial \theta} dy_i \bigg) dy_1, \ldots, dy_{i-1} =  0.
\end{align*}
Eq. \ref{prop1eqn1} then becomes
\begin{align*} 
& = \sum_{i=1}^{n}\sum_{k=i}^n \underset{y_1, \ldots, y_k}{\int}  \hat{f}(x_k) \prod_{\substack{j=1 \\ j \neq i}}^{k} p_j(y_j \, |\, \cdot ) \dfrac{\partial p_i(y_i \, | \, \cdot)}{\partial \theta}dy_1, \ldots, dy_k \\
& = \mathe_{y_1^n}\bigg[\sum_{i=1}^{n}\sum_{k=i}^n \hat{f}(x_k) \dfrac{\partial \log p(y_i \, | \, z_i, x_{i-1})}{\partial \theta} \, \bigg| \, z_1^n \bigg]
\end{align*}
and the result follows trivially. $\hfill \qed$
\subsection{Proposition 4.2}
\pushQED{\qed}
\begin{align*} 
& \mathe_{z_1^n, y_1^n}\bigg[ \beta_i(\phi) \dfrac{\partial \log p_i(y_i \, | \, z_i)}{\partial \theta} \bigg]  = \mathe_{z_1^n, y_1^{i-1}, y_{i+1}^n}\bigg[ \mathe_{y_i}\bigg[ \beta_i(\phi) \dfrac{\partial \log p_i(y_i \, | \, z_i)}{\partial \theta} \ \bigg| \ z_1^n, y_1^{i-1}, y_{i+1}^n \bigg] \bigg] \\
& = \mathe_{z_1^n, y_1^{i-1}, y_{i+1}^n}\bigg[ \beta_i(\phi) \bigg(\underset{y_i}{\int}\dfrac{\partial p_i(y_i \, | \, z_i)}{\partial \theta}dy_i \bigg) \bigg] = 0. \qedhere
\end{align*}
\popQED
\subsection{Lemma 4.2}
If $j > i$, then 
\begin{align*} 
& \mathe_{z_1^n, y_1^n}\bigg[ \beta_i(\xi_i) \dfrac{\partial \log p_i(y_i \, | \, x_{i-1}, z_i )}{\partial \theta}\dfrac{\partial \log p_j(y_j \, | \, x_{j-1}, z_j)}{\partial \theta}^\top \ \bigg| \ \xi_j \bigg] \\
= & \mathe_{z_1^n, y_1^{j-1} }\bigg[ \mathe_{z_1^n, y_1^n} \bigg[ \beta_i(\xi_i) \dfrac{\partial \log p_i(y_i \, | \, x_{i-1}, z_i )}{\partial \theta}\dfrac{\partial \log p_j(y_j \, | \, x_{j-1}, z_j)}{\partial \theta}^\top \ \bigg| \ z_1^n, y_1^{j-1}, \xi_j \bigg]\bigg] \\ 
= & \mathe_{z_1^n, y_1^{j-1}}\bigg[ \beta_i(\xi_i) \dfrac{\partial \log p_i(y_i \, | \, x_{i-1}, z_i )}{\partial \theta} \mathe_{y_j^n} \bigg[ \dfrac{\partial \log p_j(y_j \, | \, x_{j-1}, z_j)}{\partial \theta}^\top \ \bigg| \ z_1^n, y_1^{j-1}\bigg]\bigg] = 0
\end{align*}
as the inner expectation is zero due to lemma \ref{lemma1}. Otherwise if $i > j$, then we have essentially the same argument
\pushQED{\qed}
\begin{align*} 
& \mathe_{z_1^n, y_1^n}\bigg[ \beta_i(\xi_i) \dfrac{\partial \log p_i(y_i \, | \, x_{i-1}, z_i )}{\partial \theta}\dfrac{\partial \log p_j(y_j \, | \, x_{j-1}, z_j)}{\partial \theta}^\top \ \bigg| \ \xi_j \bigg] \\
= & \mathe_{z_1^n, y_1^{i-1}}\bigg[ \beta_i(\xi_i) \dfrac{\partial \log p_j(y_j \, | \, x_{j-1}, z_j )}{\partial \theta}\mathe_{y_i^n}\bigg[\dfrac{\partial \log p_i(y_i \, | \, x_{i-1}, z_i)}{\partial \theta}^\top \ \bigg| \ z_1^n, y_1^{i-1} \bigg] \bigg] = 0. \qedhere
\end{align*}
\popQED
\subsection{Proposition 4.3}
\pushQED{\qed}
\begin{align*} 
& \mathe_{z_1^n, y_1^n} \bigg[ \sum_{k=i}^n\hat f_k(x_k) \dfrac{\partial \log p_i(y_i \, | \, x_{i-1}, z_i)}{\partial \theta} \bigg] \\
= & \mathe_{z_1^n, y_1^i} \bigg[  \mathe_{y_{i+1}^n}\bigg[ \sum_{k=i}^n\hat f_k(x_k) \dfrac{\partial \log p_i(y_i \, | \, x_{i-1}, z_i)}{\partial \theta} \  \bigg| \ z_1^n, y_1^i \, \bigg] \bigg] \\
= & \mathe_{z_1^n, y_1^i}\bigg[ \bigg(\hat f_i(x_i) + \mathe_{y_{i+1}^n} \bigg[ \sum_{k=i+1}^n \hat f_k(x_k) \ \bigg| \ z_1^n, y_1^i \, \bigg] \bigg)\dfrac{\partial \log p_i(y_i \, | \, x_{i-1}, z_i)}{\partial \theta} \, \bigg] \qedhere
\end{align*}
\popQED

\chapter{Chapter 5 Appendix}
\section{Proof of Theorem 5.1 and Corollary 5.1}
  \begin{definition}
We say that the function $h(y(t), z(t))$ is piecewise continuous on an interval $\mathcal{I}$ if there are a finite number of points $t \in \mathcal{I}$ where $h(y(t), z(t))$ is not continuous.
\end{definition}
\begin{definition}
(Existence and Uniqueness Theorem.) Define the initial value problem 
\begin{align*} 
\dot x(t) = h(t, x(t)), \quad x(t_0) = x_0
\end{align*}
where $h(t,y(t))$ is a continuous function on the region $\mathcal{R} = \{ (t,x) :  | t- t_0 | < s, |x - x_0 | < y \}$, 
and $h$ is uniformly Lipschitz in $x$, meaning 
\begin{align*} 
\left| \dfrac{\partial h}{\partial x} \right| \leq C
\end{align*}
for some constant $C$ in the region $\mathcal{R}$. Then the solution $x(t)$ exists and is unique on the interval $|t - t_0| < s*$ for some $0 < s^* < s$.
\end{definition}
 \begin{prf}[Proof of Theorem 1.]
We say that there is a discontinuity for vehicle $i$, either in its car following model or its adjoint variable, if at time $t$: a) its lead vehicle changes, b) its model regime changes, c) the vehicle changes from the car following model to the downstream boundary condition, or d) the loss function $f$ is not continuous at time $t$. We will refer to these different causes of discontinuities as types a) - d) respectively. Define the times $\theta_j$ for $j = 0, \ldots, m$ as the set of all times when when any vehicle $x_i, i = 1, \ldots, n$ experiences a discontinuity. The set $\{ \theta_j \}$ is of measure zero because each vehicle can contribute only a finite number of discontinuities. The assumption that $f$ and $h_i$ are piecewise continuous and switch regimes only a finite number of times means a vehicle can contribute only a finite number of type b) or d) discontinuities. A vehicle switches to the downstream boundary only once, contributing a single type c) discontinuity. And based on the physical nature of the problem it can be assumed that a vehicle can change its lead vehicle only a finite number of times in a finite interval. \\
Some of the times $\theta_j$ may depend on the switching condition $g$ in Eq. \eqref{regime}. These times may therefore depend on the parameters $p$, and we use the notation $\theta_j(p)$ to refer to the switching times for parameters $p$. Besides the set $\{ \theta_j \}$ being finite, we also require that 
\begin{align*} 
\underset{\epsilon \rightarrow 0}{\lim} \{\theta_j(p+\epsilon)\} = \{ \theta_j (p)\} \tag{17}\label{condition}
\end{align*}
meaning that the times of discontinuities must be continuous with respect to the parameters. \\
Let the set $\{ \theta_j \}$ be ordered in increasing time,and let it include the first and last times of the simulation $\theta_0 = \min \{t_1, \ldots, t_n\}$ and $\theta_m = \max \{T_{1}, \ldots, T_{n} \}$. Since all vehicles have the same lead vehicle on each time interval $[\theta_j, \theta_{j+1}]$, on that interval there exists some ordered set of vehicles $\chi_j$ which contains all vehicles which are simulated in that time interval, ordered such that the first vehicle has no simulated vehicles as leaders (i.e. the vehicle farthest downstream) and the last vehicle has no simulated vehicles as followers (i.e. the vehicle farthest upstream). Following the definition it is true that for $i \in \chi_j$ and $t \leq T_{i-1}$ all the partial derivatives $\partial f / \partial x_i$, $\partial h_i / \partial x_i$, $\partial h_i / \partial p$, $\partial h_i/ \partial x_{L(i)}$ as well as $f$ and $h_i$ are continuous with respect to their arguments, except possibly at the endpoints. \\
\begin{figure}[H] 
\centering 
\includegraphics[ width=.8\textwidth]{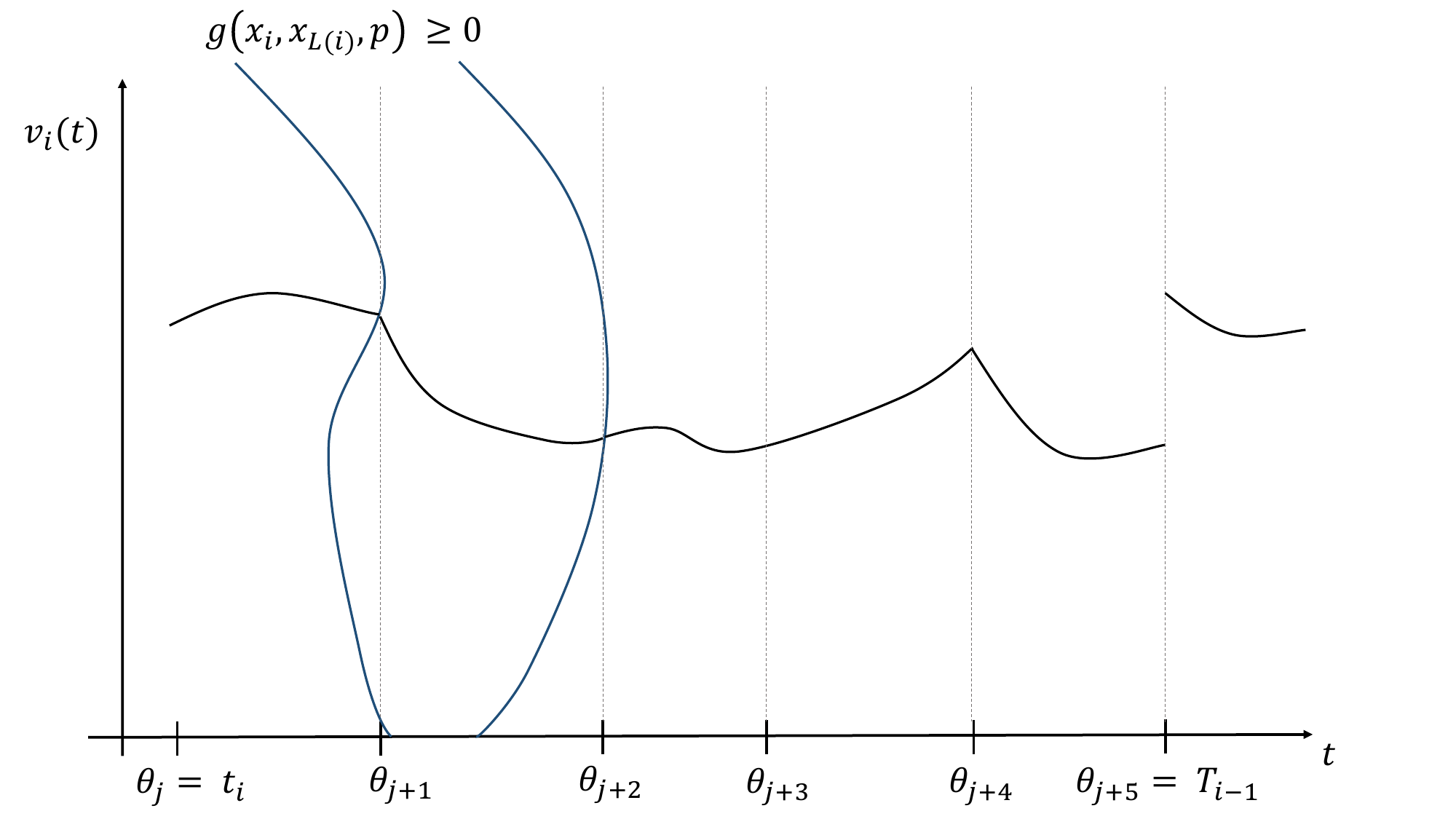} 
\caption{Illustration of the velocity $v_i(t)$ when $h_i(\cdot)$ may have discontinuities. The times $\theta_j$, where discontinuities can occur are marked as dashed lines. The region inside the two blue lines is the region where the switching condition $g(x_i, x_{L(i)}, p) \geq 0$, so the model changes regimes at times $\theta_{j+1}$ and $\theta_{j+2}$. The model is discontinuous at $\theta_{j+1}$ and $\theta_{j+2}$ but $v_i(t)$ is still continuous. At $\theta_{j+3}$ some other vehicle in the platoon experiences a discontinuity, but $i$ is unaffected. At $\theta_{j+4}$, the lead vehicle changes, causing a discontinuity in the model (but $v_i$ is still continuous at that time). At $\theta_{j+5}$, the vehicle switches to the downstream boundary condition which causes a jump discontinuity in $v_i$. } 
\end{figure}
First we show that $x_i(t), i \in \chi_0$ is continuous on the interval $[\theta_0, \theta_1]$. This is guaranteed by applying the existence and uniqueness theorem in order to each vehicle. Since by construction $h_i(x_i, x_{L(i)},p_i)$ is continuous in this interval excluding the right endpoint, and $\partial h_i / \partial x_i$ is bounded, then the initial value problem $\dot x_i = h_i$ with initial condition $x_i(t_i) = \hat x_i(t_i)$ has a solution, namely
\begin{align*} 
x_i(t) = \hat x_i(t_i) + \int_{t_i}^{\theta_1} h_i(x_i, x_{L(i)}, p_i) dt, \ \ t_i \leq t < \theta_1
\end{align*}
which is valid for $t_i \leq t < \theta_1$. Any discontinuity which may exist at $\theta_1$ is not felt by the integral since the point has measure 0. Since we solve for each vehicle following the order $\chi_0$, the continuity of $x_{L(i)}$ for $L(i) \in \chi_0$ is already obtained by the time it is needed to compute its follower. \\
For some arbitrary interval $[\theta_j, \theta_{j+1}]$, where the previous interval is already solved, take the left sided limit of the previous interval as the initial condition, i.e. $x_i(\theta_j) = \underset{\epsilon \rightarrow 0}{\lim} \ x_i(\theta_j-\epsilon)$. This point will always be defined by the previous interval. Then the solution on the new interval is defined as
\begin{align*} 
& x_i(t) = x_i(\theta_j) + \int_{\theta_j}^{\theta_{j+1}}h_i(x_i, x_{L(i)}, p_i) dt, \ \ \theta_j \leq t < \theta_{j+1} \\
& x_i(\theta_j) = \underset{\epsilon \rightarrow 0}{\lim} \ x_i(\theta_j-\epsilon)
\end{align*}
Note that any possible discontinuities in $h_i$ at times $\theta_j$ and $\theta_{j+1}$ do not affect the integral because they are of measure 0. Applying the existence and uniqueness theorem again, we have that $x_i(t)$ is continuous on $[\theta_j, \theta_{j+1}]$. Our choice of initial conditions guarentees continuity on the larger interval $[\theta_{j-1}, \theta_{j+1}]$. Thus it is clear from an induction argument that $x_i(t)$ will be continuous on the interval $[t_i, \theta_{j+1}]$, despite any possible discontinuities in $h_i$. \\
The exception to the continuity of $x_i$ is at time $T_{i-1}$, where the downstream boundary conditions will cause a jump discontinuity. Upon reaching the $\theta_j = T_{i-1}$, the remaining trajectory of $x_i$ is to be taken as 
\begin{align*} 
& x_i(t) = x_i(T_{i-1}) + \int_{T_{i-1}}^{T_i} \dot{\hat{x}}_i(t) dt \\
& x_i(T_{i-1}) = (\underset{\epsilon \rightarrow 0}{\lim} \ x_i^*(T_{i-1}-\epsilon) , \  \hat v_i(T_{i-1}) )^T
\end{align*} 
which guarantees continuity only in position (i.e. $x_i^*$, the first component of vector $x_i$) and the speed $v_i$ will have a jump discontinuity. \\
Thus we proven that the simulated trajectories $x_i$ are continuous on the interval $[t_i, T_{i-1}]$ (excluding $T_{i-1}$) by repeatedly applying the existence and uniqueness theorem on the intervals $[\theta_j, \theta_{j+1}]$ for an arbitrary $j$. \\
Then for the objective 
\begin{align*} 
F = \sum_{i=1}^n \int_{t_i}^{T_{i-1}}f(x_i, \hat x_i, t) dt
\end{align*}
since $f$ is assumed to have a finite number of discontinuities in $t$, $\hat x_i$ is assumed to be continuous, and we have shown $x_i(t)$ is continuous on the interval $t_i \leq t < T_{i-1}$, it follows that $F$ is continuous. \\
To show that $\lambda_i(t)$ is continuous, apply the existence and uniqueness theorem starting with the last interval $[\theta_{m}, \theta_{m-1}]$, going backwards over vehicles on $\chi_{m-1}$. Define 
\begin{align*} 
& \gamma_i = \mathbbm{1}(t \leq T_{i-1}) \left( \dfrac{\partial f}{\partial x_i} - \lambda_i^T(t)\dfrac{\partial h_i}{\partial x_i} \right) - \mathbbm{1}(G(i,t) \neq 0) \lambda_{G(i)}^T(t) \dfrac{\partial h_{G(i)}}{\partial x_i}
\end{align*}
and for an arbitrary interval $[\theta_{j}, \theta_{j-1}]$ define 
\begin{align*} 
& \lambda_i(t) = \lambda_i(\theta_j) + \int_{\theta_j}^{\theta_{j-1}} \gamma_i dt, \ \ \theta_j \geq t > \theta_{j-1} \\
& \lambda_i(\theta_j) = \underset{\epsilon \rightarrow 0}{\lim} \ \lambda_i(\theta_j + \epsilon)
\end{align*}
where additionally for the first interval for $\lambda_i$ (the interval where $\theta_j = T_i$), the initial condition should be $\lambda_i(\theta_j) = 0$. The justification for why $\lambda_i(t)$ is continuous follows the same argument as for $x_i$. The Lipschitz condition for $\lambda_i$ necessary to apply the existence and uniqueness theorem for $\lambda_i(t)$ is the same as the Lipschitz condition for $x_i$ ($\partial h_i/ \partial x_i$ is bounded). Additionally, by assumption we have all the partial derivatives of $h_i$ are continuous on each interval excluding the endpoints. Since the adjoint system contains the term $\mathbbm{1}(G(i,t) \neq 0) \lambda_{G(i)}^T(t) \dfrac{\partial h_{G(i)}}{\partial x_i}$ the $\lambda_i$ need to be solved backwards in the order $\chi_j$ so that the follower's adjoint variable is already proven as continuous by the time it is needed. Also note that $\dfrac{\partial h_{G(i)}}{\partial x_i}$ is simply $\partial h_k / \partial x_{L(k)}$ where $k = G(i)$.  \\
Because $\lambda_i$ is continuous on $[t_i, T_i]$, $x_i$ is continuous on $[t_i, T_{i-1})$, and $\partial h_i / \partial p$ contains a finite number of discontinuities, it follows that $\lambda_i^T(t) \partial h_i / \partial p$ is a piecewise continuous function with a finite number of discontinuities. Then the gradient
\begin{align*} 
\dfrac{d}{dp} F = \sum_{i=1}^n\left( -\int_{t_i}^{T_{i-1}} \lambda_i^T(t) \dfrac{\partial h_i}{\partial p} dt \right)
\end{align*}
is continuous since the number of discontinuities over the integral is of measure 0. \qed \\
\end{prf}
\begin{prf}[Proof of Corrolary 1.]
Showing $F$ is lipschitz continuous is equivalent to showing that $d F / dp$ is bounded. To do this define 
\begin{align*} 
& \alpha(t) = \int_{T_i}^{t} \left| \mathbbm{1}(t \leq T_{i-1}) \dfrac{\partial f}{\partial x_i} \right| +\left| \mathbbm{1}(G(i,t) \neq 0) \lambda_{G(i)}^T(t) \dfrac{\partial h_{G(i)}}{\partial x_i} \right| dt \\
& \beta(t) = \dfrac{\partial h_i}{\partial x_i}
\end{align*}
Then by definition we have
\begin{align*} 
\lambda_i(t) \leq \alpha(t) + \int_{T_i}^t \beta(s) \lambda_i(s) ds
\end{align*}
and by Gronwall's inequality in integral form
\begin{align*} 
\lambda_i(t) \leq \alpha(t) \exp \left( \int_{T_i}^t\beta(s) ds \right) \tag{18}\label{gronwall}
\end{align*}
To show that all $\lambda_i(t)$ are bounded, prove that each $\lambda_i , i \in \chi_j$ are bounded in the interval $[\theta_j, \theta_{j-1}]$ going backwards over the order specified by $\chi_{j-1}$, starting with $j=m$. Since we assumed that $\partial f / \partial x_i$ and $\partial h_i / \partial x_i$ are both bounded, and for the last vehicle $k$ in $\chi_j$, $G(i,t) = 0$, then it follows from Eq. \eqref{gronwall} that $\lambda_k(t)$ is bounded in the interval $[\theta_j, \theta_{j-1}]$. Then after showing this for $k$, it can be shown for the other vehicles in $\chi_j$. Repeating this process for each interval shows that each $\lambda_i(t)$ is bounded on $[T_i, t_i]$. \\
Then for the gradient Eq. \eqref{16} since we showed $\lambda_i(t)$ is bounded, and $\partial h_i / \partial p_i$ is bounded by assumption, then it follows that $d F / dp$ is also bounded, and hence $F$ is Lipschitz continuous. \qed
\end{prf} 
 \subsection*{Example with OVM}
 Written as a first order model OVM is 
 \begin{adjustwidth}{-2cm}{0cm} \vspace{-1.3cm} \begin{align*} 
 \dot x_i(t) = \dfrac{d}{dt} \begin{bmatrix} x_i^*(t) \\ v_i(t) \end{bmatrix} = \begin{bmatrix} v \\ c_4c_1 \tanh ( c_2 s_i(t) - c_3 - c_5) - c_4c_1 \tanh ( - c_3) - c_4 v_i(t) \end{bmatrix} = h_i(x_i, x_{L(i)}, p_i) 
 \end{align*}
 \end{adjustwidth}
Where $p_i = (c_1, c_2, c_3, c_4, c_5) > \textbf{0}$ and $s_i(t) = x_i^*(t) - x_{L(i)}(t) - l$ (where $l$ is the length of the lead vehicle). Since it is a single regime model, it will be everwhere continuous. The partial derivative 
\begin{align*} 
\dfrac{\partial h_i}{\partial x_i} = \begin{bmatrix} 0 & 1 \\ -c_1c_2c_4 \sech ( c_3 + c_5 - c_2(s_i))^2 & -c_4 \end{bmatrix} \Rightarrow \left| \dfrac{\partial h_i}{\partial x_i} \right| \leq \begin{bmatrix} 0 & 1 \\ c_1c_2c_4 & c_4 \end{bmatrix} 
\end{align*}
so it can be bounded by a constant for an arbitrary vector space. The other partial derivatives can be bounded as
\begin{align*} 
\left| \dfrac{\partial h_i}{\partial x_{L(i)}} \right|  \leq  \begin{bmatrix} 0 & 0 \\ c_1c_2c_4 & 0 \end{bmatrix} \\
\left| \dfrac{\partial h_i}{\partial p_i} \right| \leq \begin{bmatrix} 0, & 0, & 0, & 0, & 0 \\ 2c_4, & c_1 c_4 s_i, & c_1c_4, & v_i + 2 c_1, & c_1c_4  \end{bmatrix}
\end{align*}
And so the objective will be Lipschitz continuous assuming that the headway $s_i(t)$ and velocity $v_i(t)$ are finite.
 \section*{Supplemental figures/tables for comparison of algorithms}
 
\begin{table}[H] 
\begin{adjustwidth}{-3.5cm}{0cm}
\centering
\caption{The full table showing all variants of the 7 algorithms considered.}
\begin{tabular}{|l|l|l|l|l|l|l|l|l|}
\hline
Algorithm & \begin{tabular}{@{}c@{}}\% found  \\global opt\end{tabular} & \begin{tabular}{@{}c@{}} Avg.  \\ RMSE (ft) \end{tabular} &  \begin{tabular}{@{}c@{}} Avg. \\  Time (s)\end{tabular} & \begin{tabular}{@{}c@{}} Avg. RMSE / \% \\  over global opt\end{tabular} &\begin{tabular}{@{}c@{}}Initial  \\ Guesses\end{tabular} & \# Obj. & \# Grad.& \# Hess. \\ \hline
Adj BFGS-0 & 94.0  & 6.46& 7.42 & .10 / 2.0\% &3 &307.1 & 307.1 & 0 \\ \hline
Adj BFGS-7.5 &85.2  &6.56 &4.16 & .20/6.8\% & 1.67 & 165.0&  165.0& 0 \\ \hline
Adj BFGS-$\infty$& 75.6& 6.96& 2.55 & .59 / 15.4\% & 1 & 107.3 &107.3 & 0\\ \hline
GA & 95.0 & 6.47 &33.5 & .10/2.2\%& -  & 5391 & 0& 0\\ \hline
NM & 39.4 & 7.25 & 6.85   & .88/21.0\%& 1 & 1037 & 0 &0 \\ \hline
Fin BFGS-0& 94.3 & 6.46 & 11.4  & .10/2.2\%&3 & 311.5 &311.5 &0 \\ \hline
Fin BFGS-7.5&85.7 & 6.55 & 6.32  & .19/6.3\%&1.66 & 164.8& 164.8& 0\\ \hline
Fin BFGS-$\infty$& 77.0& 6.88 & 3.91 & .52/13.8\%& 1& 107.3& 107.3& 0\\ \hline
Adj GD-0&20.2 & 7.58 & 25.5  & 1.21/28.2\%& 3 & 1374 & 542.5 & 0\\ \hline
Adj GD-7.5& 13.4 & 7.83 & 18.1   & 1.46/40.8\%& 1.93 & 1094 & 436.6 &0 \\ \hline
Adj GD-Inf& 8.8& 8.90 & 8.75 & 2.53/67.2\% & 1 & 719.3 &282.6 &0 \\ \hline
Adj SQP-0& 29.1 & 8.37 &21.1   & 2.01/61.7\%&3 &287.6 & 83.8 & 80.8\\ \hline
Adj SQP-7.5&21.5  & 8.59 & 15.4 & 2.22/71.4\%& 2.17 & 244.8 & 71.3 & 69.1 \\ \hline
Adj SQP-$\infty$& 10.0  &11.1 & 5.87 & 4.70/140.7\% & 1 & 141.6 & 39.4 &  38.4 \\ \hline
SPSA& 4.2  & 53.2  & 38.5  & 46.8/786\%& 1 & 6001 & 0 &0  \\ \hline
Fin GD-$\infty$& 9.1 & 8.86 & 14.7  & 2.50/66.2\%& 1 & 657.3 & 258.5 &0 \\ \hline
Fin SQP-$\infty$& 2.6 & 9.61 & 15.8   & 3.25/81.8\%&1 & 179.4 & 67.2 & 66.2 \\ \hline
Adj TNC-0& 90.7 & 6.43 & 6.87  & .05/2.4\%&3 & 285.7 & 285.7& 0 \\ \hline
Adj TNC-7.5& 82.7  & 6.48 & 3.82 & .11/5.0\%& 1.63 & 152.3 &152.3 &0 \\ \hline
Adj TNC-$\infty$& 77.3 &  6.62& 2.26 & .25/7.7\%& 1 & 94.1 & 94.1 & 0 \\ \hline
Fin TNC-0& 36.3 & 6.84 & 13.2 & .47/11.0\% & 3 & 302.4& 302.4& 0 \\ \hline
Fin TNC-7.5& 24.9  & 7.05 & 8.35 & .69/20.8\% & 1.82 & 183.2 & 183.2 & 0 \\ \hline
Fin TNC-$\infty$& 18.0 & 7.63 & 4.49  & 1.27/33.2\%& 1 & 100.8& 100.8& 0\\ \hline
\end{tabular}
\end{adjustwidth}
\end{table}

\begin{figure}[H] 
\centering 
\includegraphics[ width=\textwidth]{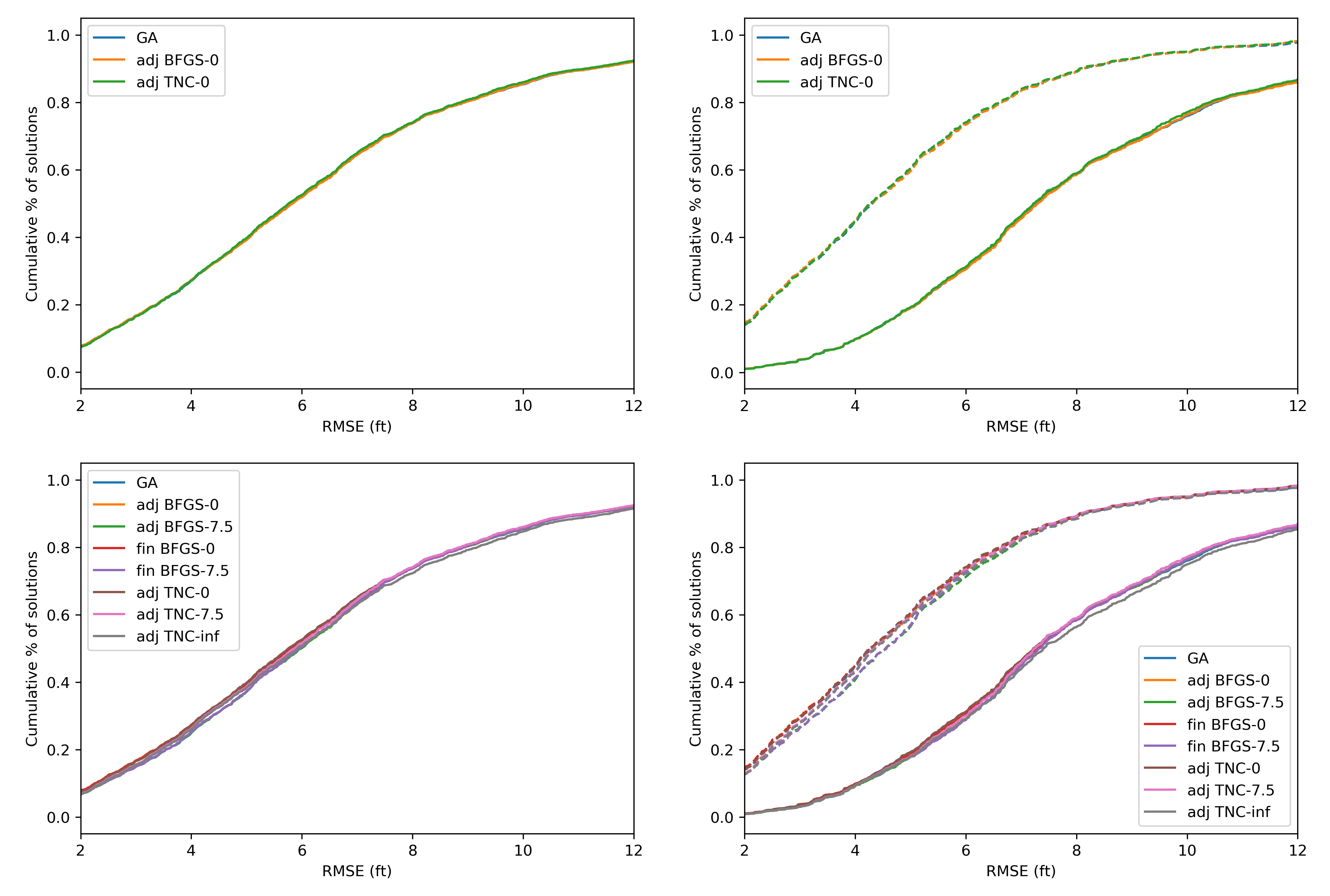} 
\caption{Shows the distribution of RMSE for different algorithms tested. Left panels show overall RMSE, right panels show the RMSE for vehicles that experience lane changes (solid) and those that don't (dashed). The RMSE for lane changing vehicles is much higher because car following models don't describe lane changes well.  For example, for the genetic algorithm, the average RMSE for vehicles which do not experience lane changes is 4.75, whereas it is 8.17 for vehicles which do. The other algorithms have very similar numbers. } 
\end{figure}

\section*{Adjoint method for gradient with DDE model}
\noindent We will now deal with calculating the gradient for Eq. \eqref{6}. As before, if not explicitly stated, quantities can be assumed to be evaluated at time $t$. Define the augmented objective function: 
\begin{align*} 
& L = \sum_{i=1}^n \bigg( \int_{t_i + \tau_i}^{T_{i-1}^*} f(x_i, \hat x_i) + \lambda_i^T(t) \bigg( \dot x_i(t) - h_i(x_i(t), x_i(t-\tau_i), x_{L(i)}(t-\tau_i), p_i) \bigg) dt  + \\
& \qquad  \int_{T_{i-1}^*}^{T_i} \lambda_i^T ( \dot x_i - \dot{\hat{x}}_i ) dt \bigg)   \tag{19}\label{27}
\end{align*}
and since $F = L$
\begin{adjustwidth}{-2.5cm}{0cm} \vspace{-1.3cm} \begin{align*}
&\dfrac{d}{dp} F = \sum_{i=1}^n \bigg[ \int_{t_i + \tau_i }^{T_{i-1}^* } \dfrac{\partial f}{\partial x_i}\dfrac{\partial x_i}{\partial p} - \lambda_i^T(t)\dfrac{\partial h_i}{\partial x_i}\dfrac{\partial x_i}{\partial p} - \lambda_i^T(t)\dfrac{\partial h_i}{\partial x_i(t-\tau_i)}\dfrac{\partial x_i(t- \tau_i)}{\partial p} 
- \lambda_i^T(t)\dfrac{\partial h_i}{\partial x_i(t-\tau_i)}\dot x_i(t-\tau_i) \dfrac{\partial (t-\tau_i)}{\partial p} \\ 
& - \lambda_i^T(t)\dfrac{\partial h_i}{\partial p} dt  - \int_{t_i + \tau_i}^{T_i} \dot \lambda_i^T(t)\dfrac{\partial x_i}{\partial p} dt \bigg] 
+ \sum_{i=1}^{n} \sum_{j \in S(i)}\int_{t_{i}+\tau_j}^{T_{i}+\tau_j}\mathbbm{1}(G(i,t-\tau_j) = j)\bigg[ + \ldots \\ & 
 -\lambda_{j}^T(t) \dfrac{ \partial h_{j}}{\partial x_i(t-\tau_{j})}\dfrac{\partial x_i(t-\tau_{j}) }{\partial p} - 
\lambda_{j}^T(t)\dfrac{\partial h_{j}}{\partial x_i(t - \tau_{j})}\dot{x}_i(t-\tau_{j})\dfrac{\partial(t - \tau_{j})}{\partial p} dt\bigg] \tag{20}\label{28} 
\end{align*}
\end{adjustwidth}
where $S(i)$ is defined as the set of all followers for vehicle $i$ and we have chosen $\lambda_i^T(T_{i-1}^*) = 0$. For a DDE, use a change of coordinates on terms containing $\partial x_i(t - \tau) / \partial p $. 
\begin{align*}
&-\int_{t_i+\tau_i}^{T_{i-1}^*}  \lambda_i^T(t)\dfrac{\partial h_i}{\partial x_i(t-\tau_i)}\dfrac{\partial x_i(t- \tau_i)}{\partial p_i}dt 
= -\int_{t_i+\tau_i}^{T_{i-1}^* - \tau_i} \lambda_i^T(t+\tau_i) \dfrac{\partial h_i(t + \tau_i)}{\partial x_i(t- \tau_i)}\dfrac{ \partial x_i(t)}{\partial p_i}dt  \\
& \sum_{i=1}^{n} \sum_{j \in S(i)}\int_{t_{i}+\tau_j}^{T_{i}+\tau_j}-\mathbbm{1}(G(i,t-\tau_j) = j) \lambda_{j}^T(t) \dfrac{ \partial h_{j}}{\partial x_i(t-\tau_{j})}\dfrac{\partial x_i(t-\tau_{j}) }{\partial p}dt \\ 
& = \sum_{i=1}^{n} \sum_{j \in S(i)}\int_{t_{i}+\tau_{i}}^{T_{i}}-\mathbbm{1}(G(i,t) = j) \lambda_{j}^T(t+\tau_j) \dfrac{ \partial h_{j}(t+\tau_j)}{\partial x_i(t-\tau_{j})}\dfrac{\partial x_i(t) }{\partial p}dt  \tag{21}\label{29}
\end{align*}
Where we make use of the fact that $\partial x_i (t) / \partial p  = 0$ for $t \in [t_i, t_i+\tau_i]$ since the initial history function doesn't depend on the parameters. Now substituting Eq. $\eqref{29}$ into Eq. $\eqref{28}$ and grouping terms, one obtains the adjoint system: 
\begin{align*}
& - \dot \lambda_i^T(t) + \mathbbm{1}(t \leq T_{i-1}^*)\left( \dfrac{\partial f}{\partial x_i} - \lambda_i^T(t)\dfrac{\partial h_i}{\partial x_i} \right) - \mathbbm{1}(t \leq T_{i-1}^* - \tau_i) \lambda_i^T(t+\tau_i)\dfrac{\partial h_i(t + \tau_i)}{\partial x_i(t-\tau_i)} \ldots \\ 
&  - \sum_{j \in S(i)}\mathbbm{1}( G(i,t) = j)\lambda_{j}^T(t+\tau_{j})\dfrac{ \partial h_{j}(t+\tau_{j})}{\partial x_i(t - \tau_{j})}        = 0, \quad  t \in [T_i, t_i + \tau_i], \ \ \ \forall i    \tag{22}\label{32}
\end{align*}   
\noindent with initial conditions $\lambda_i(T_{i}) = 0$. For a DDE model, the adjoint system also becomes a DDE. After solving for the adjoint variables, one can compute the gradient: 
\begin{align*} 
&\dfrac{d}{dp} F = \sum_{i=1}^n \bigg[ \int_{t_i+\tau_i}^{T_{i-1}^*} - \lambda_i^T(t)\dfrac{\partial h_i}{\partial x_i(t - \tau_i)} \dot x_i(t - \tau_i)\dfrac{\partial (t - \tau_i)}{\partial p} -\lambda_i^T\dfrac{\partial h_i}{\partial p}dt \bigg] \\
& + \sum_{i=1}^{n} \sum_{j \in S(i)}\int_{t_{i}+\tau_j}^{T_{i}+\tau_j}-\mathbbm{1}(G(i,t-\tau_j) = j)\lambda_{j}^T(t)\dfrac{\partial h_{j}}{\partial x_i(t - \tau_{j})}\dot{x}_i(t-\tau_{j})\dfrac{\partial(t - \tau_{j})}{\partial p} dt 
 \tag{23}\label{31}
\end{align*} 

 \section*{Extension of adjoint method for Hessian}
  In this section, for a scalar being differentiated with respect to two vectors, the result is a matrix where the rows correspond to the first vector, and the columns correspond to the second vector. For example, 
 \begin{align*} 
\dfrac{ \partial f }{\partial x_i \partial p} = \begin{bmatrix} \partial f / \partial x_i^* \partial p^1 & \partial f / \partial x_i^* \partial p^2 & \ldots & \partial f / \partial x_i^* \partial p^m \\ 
 \partial f / \partial v_i \partial p^1 & \partial f / \partial v_i \partial p^2 & \ldots & \partial f / \partial v_i \partial p^m \end{bmatrix} 
 \end{align*}
 For a matrix being differentiated to two vectors, the result is a 3-dimensional array which should be interpreted as a vector of matrices. For example 
 \begin{align*} 
 \dfrac{\partial h_i}{\partial p \partial x_i } = \begin{bmatrix} \partial h_i^1 / \partial p \partial x_i \\ \partial h_i^2 / \partial  p \partial x_i \end{bmatrix}
 \end{align*}
 where $h_i^1$ and $h_i^2$ refer to the two components of $h_i$. So the normal rules of linear algebra can be applied, e.g. 
 \begin{align*} 
& \dfrac{\partial x_i}{\partial p}^T \dfrac{\partial h_i}{\partial p \partial x_i}^T \lambda_i = \dfrac{\partial x_i}{\partial p}^T \begin{bmatrix} \dfrac{h_i^1}{\partial p \partial x_i}^T,  & \dfrac{h_i^2}{\partial p \partial x_i}^T \end{bmatrix} \lambda_i = \begin{bmatrix} \dfrac{\partial x_i}{\partial p}^T \dfrac{\partial h_i^1}{\partial x_i \partial p},  & \dfrac{\partial x_i}{\partial p}^T \dfrac{\partial h_i^2}{\partial x_i \partial p} \end{bmatrix} \begin{bmatrix} \lambda_i^1 \\ \lambda_i^2 \end{bmatrix} \\
& =  \dfrac{\partial x_i}{\partial p}^T \dfrac{\partial h_i^1}{\partial x_i \partial p}\lambda_i^1 + \dfrac{\partial x_i}{\partial p}^T \dfrac{\partial h_i^2}{\partial x_i \partial p} \lambda_i^2
 \end{align*}
 where $\lambda_i^1$ and $\lambda_i^2$ are the two components of the adjoint variable. 
 
 \noindent The adjoint method extends in a straightforward way for computing the Hessian. Using finite differences would have complexity $\mathcal{O}(m^2 T(n))$, whereas the adjoint method can be shown to have complexity $\mathcal{O}(m T(n))$ \citep{30}. \\
 When calculating the gradient, the benefit of the adjoint method is that you avoid calculating $\partial x_i / \partial p$ terms. When calculating the Hessian, the adjoint method will be used to avoid calculating $\partial ^2 x_i/ \partial p^2$ terms. We will still be left with the unknowns $\partial x_i / \partial p$, which need to be calculated using the ``variational" approach, also known as the ``forward sensitivity" approach (see \cite{18, 22}).  

To obtain the variational system, differentiate the model Eq. \eqref{4} with respect to the parameters 
\[ \dfrac{d}{dp} \dot x_i = \dfrac{d}{dp} h_i(x_i, x_{L(i)}, p_i)  \]
and then switch the order of differentiation on the left hand side to obtain the variational system 
\begin{align*} 
& \dfrac{d}{dt} \dfrac{\partial x_i}{\partial p} = \dfrac{\partial h_i}{\partial x_i}\dfrac{\partial x_i}{\partial p} + \dfrac{ \partial h_i}{\partial p_i} + \mathbbm{1}(L(i,t) \in [1, n]) \dfrac{\partial h_i}{\partial x_{L(i)}}\dfrac{\partial x_{L(i)}}{\partial p} , \quad t \in [t_i, T_{i-1}] \ \ \ \forall i . \tag{24}\label{24}
\end{align*}
Eq. $\eqref{24}$ gives ${\partial x_i(t)}/{\partial p}$ starting from the initial conditions ${\partial x_i(t_i)}/{\partial p}$. Note that the variational system for $x_i$ is only defined up to time $T_{i-1}$ because $\partial x_i (t)/ \partial p = \partial x_i (T_{i-1})/ \partial p$ for $t \in [T_{i-1}, T_i]$. Now starting from Eq. \eqref{10}, differentiate twice instead of once
\begin{adjustwidth}{-2cm}{0cm} \vspace{-1.3cm} \begin{align*} 
&\dfrac{d^2}{d p^2} F = \dfrac{d^2}{d p^2} L = \sum_{i=1}^n\bigg[ \int_{t_i}^{T_{i-1}}\dfrac{\partial x_i}{\partial p}^T \dfrac{\partial ^2 f}{\partial x_i^2}^T\dfrac{\partial x_i}{\partial p} + \dfrac{\partial f}{\partial x_i}\dfrac{\partial^2 x_i}{\partial p^2} - \dfrac{\partial x_i}{\partial p}^T\dfrac{\partial^2 h_i}{\partial x_i^2}^T \lambda_i \dfrac{\partial x_i}{\partial p} - \lambda_i^T\dfrac{\partial h_i}{\partial x_i}\dfrac{\partial ^2 x_i}{\partial p^2} + \ldots \\
 &- \dfrac{\partial x_i}{\partial p}^T \dfrac{\partial^2 h_i}{\partial p \partial x_i}^T\lambda_i - \lambda_i^T \dfrac{\partial^2 h_i}{\partial p^2}  - \dfrac{\partial^2 h_i}{\partial x_i \partial p}^T\lambda_i \dfrac{\partial x_i}{\partial p} dt + \int _{t_i}^{T_i} \lambda_i^T\dfrac{\partial^2 \dot{x}_i}{\partial p^2}dt \bigg] +\ldots \\
&+ \sum_{i=1}^{n}\int_{t_{i}}^{T_i}  \mathbbm{1}(G(i,t) \neq 0)\bigg[ - \dfrac{\partial x_i}{\partial p}^T\dfrac{\partial^2 h_{G(i)}}{\partial x_i^2}^T\lambda_{G(i)} \dfrac{\partial x_i}{\partial p} - \dfrac{\partial x_i}{\partial p}^T\dfrac{\partial^2 h_{G(i)}}{\partial p \partial x_i}^T\lambda_{G(i)} - \dfrac{\partial x_i}{\partial p}^T\dfrac{\partial^2 h_{G(i)}}{\partial x_{G(i)}\partial x_i}^T \lambda_{G(i)}\dfrac{\partial x_{G(i)}}{\partial p} + \ldots \\ 
&- \lambda_{G(i)}^T\dfrac{\partial h_{G(i)}}{\partial x_i}\dfrac{\partial^2 x_i}{\partial p^2} - \dfrac{\partial^2 h_{G(i)}}{\partial x_i \partial p}^T \lambda_{G(i)}\dfrac{\partial x_i}{\partial p} - \dfrac{\partial x_{G(i)}}{\partial p}^T\dfrac{\partial^2 h_{G(i)}}{\partial x_i \partial x_{G(i)}}^T \lambda_{G(i)}\dfrac{\partial x_i}{\partial p} dt \bigg]
\end{align*}\end{adjustwidth}

The adjoint variables are defined the same way whether the gradient or Hessian is being computed. Eq. \eqref{16} gives the adjoint system for the unmodified objective $F$.  
After solving for all the adjoint variables, in addition to solving the variational system Eq. \eqref{24}, one can calculate the Hessian:\begin{adjustwidth}{-2cm}{0cm} \vspace{-1.3cm} \begin{align*} 
& \dfrac{d^2}{d p^2} F = \sum_{i=1}^n\bigg[ \int_{t_i}^{T_{i-1}}\dfrac{\partial x_i}{\partial p}^T \dfrac{\partial ^2 f}{\partial x_i^2}^T\dfrac{\partial x_i}{\partial p}  - \dfrac{\partial x_i}{\partial p}^T\dfrac{\partial^2 h_i}{\partial x_i^2}^T \lambda_i \dfrac{\partial x_i}{\partial p}  - \dfrac{\partial x_i}{\partial p}^T \dfrac{\partial^2 h_i}{\partial p \partial x_i}\lambda_i - \lambda_i^T \dfrac{\partial^2 h_i}{\partial p^2}  +\ldots \\ 
&- \dfrac{\partial^2 h_i}{\partial x_i \partial p}^T\lambda_i \dfrac{\partial x_i}{\partial p} dt \bigg] 
+ \sum_{i=1}^{n}\int_{t_{i}}^{T_i}  \mathbbm{1}(G(i,t) \neq 0) \bigg[  - \dfrac{\partial x_i}{\partial p}^T\dfrac{\partial^2 h_{G(i)}}{\partial x_i^2}^T\lambda_{G(i)} \dfrac{\partial x_i}{\partial p}  -\dfrac{\partial x_i}{\partial p}^T\dfrac{\partial^2 h_{G(i)}}{\partial p \partial x_i}^T\lambda_{G(i)} + \ldots \\
&- \dfrac{\partial x_i}{\partial p}^T\dfrac{\partial^2 h_{G(i)}}{\partial x_{G(i)}\partial x_i}^T \lambda_{G(i)}\dfrac{\partial x_{G(i)}}{\partial p} - \dfrac{\partial^2 h_{G(i)}}{\partial x_i \partial p}^T \lambda_{G(i)}\dfrac{\partial x_i}{\partial p} - \dfrac{\partial x_{G(i)}}{\partial p}^T\dfrac{\partial^2 h_{G(i)}}{\partial x_i \partial x_{G(i)}}^T \lambda_{G(i)}\dfrac{\partial x_i}{\partial p} dt \bigg] + \ldots  \tag{25}\label{25}
\end{align*} \end{adjustwidth}

\chapter{Chapter 6 Appendix}
\section{Appendix - havsim microsimulation model for highways}
Note that our model is almost entirely deterministic, as the only source of randomness is from discretionary LC. The car following, safety condition, and discretionary model follow traffic-simulation-de \cite{TreiberKestingBook}. This is combined with our relaxation and tactical/cooperation model, with modifications to the discretionary/mandatory LC model and upstream boundary conditions of traffic-simulation-de. 
\subsubsection*{Car following}
The IDM (Eq. \eqref{5}) is used with parameters $[c_1, c_2, c_3, c_4, c_5] = [35, \  1.3, \ 2, \ 1.1, \ 1.5]$
\subsubsection*{Lane changing model notation}
The ego vehicle $x_i$ is the vehicle for which the lane changing model is being evaluated. $x_i$ has the current leader, follower vehicles $x_{i-1}, x_{i+1}$ respectively. If $x_i$ were to change lanes, it would have the vehicles $x_{i-1}^*, x_{i+1}^*$ as the new leader, follower; we call $x_{i-1}^*, x_{i+1}^*$ the LC side leader, follower, respectively. LC side refers to either the left or right lane, depending on which side the ego vehicle is evaluating the change for. The current leader/follower relationships are refered to as the vehicle order, and if this order were to change due to lane changing, we call the new leader/follower relationships the new vehicle order.
Lastly, define $s(x_{i}, x_{i-1})$ as the space headway between vehicles $x_i$ and $x_{i-1}$, and define $h(x_i, x_{i-1})$ as the car following model evaluated for vehicle $x_i$ with $x_{i-1}$ as the leader.
\subsubsection*{Gap acceptance/safety condition}
The safety of the LC is evaluated by calling the car following model for the ego vehicle and LC side follower, under the new potential vehicle order. Thus the LC side follower safety is defined as $h(x_{i+1}^*, x_i)$ and ego vehicle safety is $h(x_i, x_{i-1}^*)$. For the potential change to be evaluated as safe, both safety conditions must be greater than the threshold value 
$$h(x_{i+1}^*, x_i) \text{ and } h(x_i, x_{i-1}^*) > d_1\frac{\dot x_i}{v_{\rm max}} + d_2(1 - \frac{\dot x_i}{v_{\rm max}})$$
where $v_{\rm max}$ is the maximum possible speed of $x_i$ (for the IDM, the $v_{\rm max} = c_1$), and $d_1, d_2$ are parameters taken as $-8, -20$. 
\subsubsection*{Discretionary LC and incentive}
For a discretionary change to be accepted, the safety condition and incentive condition must both be met. The incentive condition follows the MOBIL, defined as 
$$ h(x_i, x_{i-1}^*) - h(x_{i}, x_{i-1}) + d_4\left(h(x_{i+1}, x_{i-1}) - h(x_{i+1}, x_i) + h(x_{i+1}^*, x_i) - h(x_{i+1}^*, x_{i-1}^*)\right) + d_5 > d_3$$
where $d_3$ is the incentive threshold, $d_4$ is the politeness, and $d_5$ is the bias for the left side ($d_6$ is the bias for right side). Those parameters are taken as $.6,\  .1,\  0,\  .2$ respectively. \\
While in a discretionary state, there is a $d_7 = .1$ probability to check the incentive and safety conditions in any given timestep. If a discretionary change is completed, the discretionary model will not be checked for the next $d_9 = 20$ timesteps.  
\subsubsection*{Mandatory LC}
When vehicles on the on-ramp are able to merge onto the main road, they enter a mandatory LC state. In this state they will check the safety condition every timestep, and change as soon as the safety conditions are met. 
\subsubsection*{Tactical/Cooperation model}
The tactical and cooperation model is applied a) when the vehicle is in a mandatory state and one or both of the safety conditions are not met or b) the vehicle is in a discretionary state and the incentive is met but not the safety conditions. In case b), the discretionary state becomes ``activated'' so that the LC model will be continuous evaluated for the next $d_8=20$ timesteps (as opposed to the normal case, when there is only a $d_7$ probability to check the model). \\
Our tactical and cooperation model is very simple and based on either adding acceleration or deceleration to the baseline car following acceleration. A vehicle can either have $a2=2$ acceleration added or $a_3 =  -2$ deceleration added. Note that this has the effect of shifting the equilibrium of the baseline model. \\
In the cooperative model, the ego vehicle seeks cooperation from either the LC side follower or the LC side follower's follower. The headway between the potentially cooperating vehicle and the ego vehicle must be greater than the jam spacing of the cooperative vehicle. In the discretionary state, the cooperative vehicle will only choose to accept the cooperation with probability $a_1 = .2$. In the mandatory state, vehicles will always be willing to cooperate with the ego vehicle. If the LC side follower's safety is violated, the cooperating vehicle will have deceleration added to its baseline car following acceleration, so that it will give a greater amount of space. \\
In the tactical model, the ego vehicle adjusts its acceleration so that it will position itself better. If the LC side follower's safety is violated, the ego vehicle always has acceleration added; otherwise, only the ego vehicle's safety is violated, so deceleration is added to the ego vehicle.
\subsubsection*{Relaxation}
Follows the formulation of section 2.0 and 2.2.
\subsubsection*{Boundary Conditions}
The downstream boundary condition is applied whenever a vehicle lacks a leader. We used a free boundary, which is equivalent to taking the limit of the car following model when the leader position goes to infinity and leader speed goes to $v_{\rm max}$. For the IDM, this yields
$$\ddot{x}_i = c_4\left( 1 - \left( \dfrac{\dot x_i(t)}{c_1} \right)^4\right)$$
For the upstream boundaries, each lane with inflow has an ``inflow buffer'' which is incremented according to the instantaneous flow rate at any given timestep. When the inflow buffer $>=1$, we attempt to add a vehicle to the simulation at the start of the corresponding lane. The vehicle will potentially be added with an initial speed $v_0$
$$v_0 = \max\{\dot x_{i-1}, v_{\rm eql}^i(s(x_i, x_{i-1})) \} $$
provided that the headway satisfies
$$s(x_i, x_{i-1}) >= b^* s_{\rm eql}^i(v_0)$$
$$b^* = \begin{cases} 
 b_1 & v_0 > b_2 \\ 
  1 & \text{o.w.}
  \end{cases}$$
 Where we use the notation $v_{\rm eql}^i$ and $s_{\rm eql}^i$ to refer to the equilibrium solution for vehicle $i$. 
 $b_1$ and $b_2$ are parameters controlling the boundary condition. $b1 < 1$ allows vehicles to be aggresively added to the simulation. Under congested conditions, vehicles tend to be added with an initial speed of $b2$. We took $b_1=.8$ and $b_2=18.85$ (18.85 is the speed giving maximal flow for the chosen parameters of IDM).

\bibliography{keane_sources}

\end{document}